\documentclass[10pt,twocolumn,letterpaper]{article}

\usepackage[final]{cvpr}      %

\usepackage{times}
\usepackage{helvet}  %
\usepackage{courier}  %
\usepackage[hyphens]{url}  %
\usepackage{graphicx} %
\usepackage{natbib}  %
\usepackage{caption} %
\usepackage{xspace}
\usepackage{graphicx}
\usepackage{amsmath,amssymb}
\usepackage{multirow}
\usepackage{xspace}
\usepackage{makecell}
\usepackage{amssymb}
\usepackage{algorithm}
\usepackage{algpseudocode}
\usepackage{booktabs} 
\usepackage{amsfonts} 
\usepackage{bbold}

\usepackage{adjustbox}
\usepackage{makecell}

\usepackage{algorithm}

\usepackage{newfloat}
\usepackage{listings}
\DeclareCaptionStyle{ruled}{labelfont=normalfont,labelsep=colon,strut=off} %
\floatstyle{ruled}
\newfloat{listing}{tb}{lst}{}
\floatname{listing}{Listing}
\newcommand{\modelname}{\textsc{TempoControl}\xspace}

\usepackage{siunitx}
\usepackage{tikz}
\usepackage{pgfplots}
\pgfplotsset{compat=1.18}

\definecolor{cvprblue}{rgb}{0.21,0.49,0.74}
\usepackage[pagebackref,breaklinks,colorlinks,allcolors=cvprblue]{hyperref}

\def\confName{CVPR}
\def\confYear{2026}

\title{\modelname: Temporal Attention Guidance for Text-to-Video Models}
\author{
Shira Schiber \qquad Ofir Lindenbaum \qquad Idan Schwartz\\
Bar-Ilan University\\
{\tt\small \{shira-dvora.schiber, ofir.lindenbaum, idan.schwartz\}@biu.ac.il}
}

\begin{document}
\maketitle
\begin{abstract}
Recent advances in generative video models have enabled the creation of high-quality videos based on natural language prompts. However, these models frequently lack fine-grained temporal control, meaning they do not allow users to specify when particular visual elements should appear within a generated sequence. In this work, we introduce \modelname, a method that allows for temporal alignment of visual concepts during inference, without requiring retraining or additional supervision. \modelname utilizes cross-attention maps, a key component of text-to-video diffusion models, to guide the timing of concepts through a novel optimization approach. Our method steers attention using three complementary principles: aligning its temporal pattern with a control signal (correlation), adjusting its strength where visibility is required (magnitude), and preserving semantic consistency (entropy). \modelname provides precise temporal control while maintaining high video quality and diversity. We demonstrate its effectiveness across various applications, including temporal reordering of single and multiple objects, action timing, and audio-aligned video generation. Project page: \url{https://shira-schiber.github.io/TempoControl/}.
\end{abstract}

\begin{figure*}[hbt]
    \centering
    \includegraphics[width=0.95\linewidth]{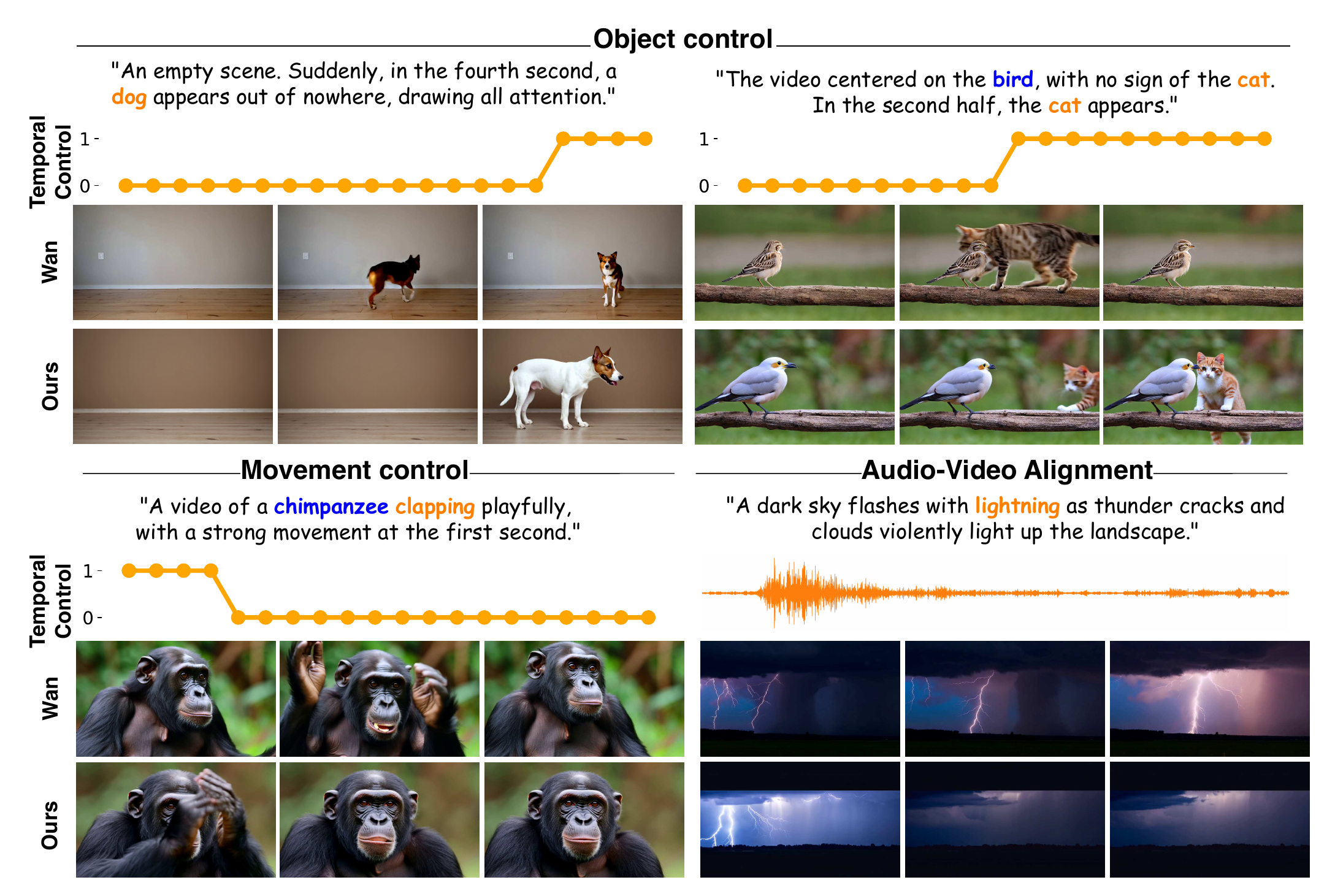}
    \caption{Applications of our inference-time temporal control method (\modelname), which enables multi-object and single-object control, motion control, and audio alignment. Words that are controlled to appear throughout the entire video are shown in blue. Words that appear according to a temporally varying control signal are shown in orange.}\vspace{-10pt}
    \label{fig:teaser}
\end{figure*}

\section{Introduction}

Generative video models have improved dramatically, enabling the synthesis of high-quality, temporally coherent videos from natural language descriptions~\cite{singer2022make, blattmann2023align, yang2024cogvideox, bar2024lumiere, wan2025wan}. These models have enabled creative applications in animation, design, and virtual content creation by generating visually compelling sequences that align with textual descriptions. However, while current models offer strong spatial consistency and global semantics, they lack fine-grained \emph{temporal control}. Specifically, guiding the model to generate visual elements at specific time points is particularly challenging. This includes making an object appear in the middle of a scene or aligning visual aspects of lighting with the sound of thunder (see Figure~\ref{fig:teaser}).

Recent work on generative video modeling has proposed various spatial and motion control mechanisms, including object masking, layout conditioning, or added camera controls~\cite{gupta2024photorealistic, wang2024boximator, feng2025blobgen, he2024cameractrl}. In contrast, temporal control has received significantly less attention. Incorporating such control typically requires additional supervision in the form of temporally annotated datasets. Since video data is often scarce, extending annotations to capture temporal variation is expensive and usually impractical at scale. Synthetic data could be used; however, generating diverse and temporally accurate video-text pairs, especially for abstract concepts such as movement, is a highly non-trivial challenge.

In this work, we propose a lightweight yet effective inference-time control method, \modelname, which introduces temporal control for generated video. Instead of retraining the generative model, \modelname leverages a property already embedded in these systems: the text-vision cross-attention mechanism.

Cross-attention layers inherently encode strong signals about \emph{which} specific words in the prompt are realized in each generated frame. By directly influencing these attention maps during the denoising process, we can shift the temporal alignment of visual concepts without sacrificing quality or requiring additional data. The approach is data-efficient and preserves both the fidelity and diversity of the original generation process.

Attention control is achieved by carefully steering the latent variables during the diffusion process. At each denoising step, we apply a few stochastic gradient descent iterations \citep{chefer2023attend,battash2024revisiting} until a satisfactory level of temporal alignment is achieved, notably, without updating the model parameters.

To guide temporal alignment, \modelname introduces a novel loss with three complementary components, based on Pearson correlation, magnitude, and entropy. The Pearson term encourages the temporal shape of attention to match that of an external control signal, operating on the normalized attention vector to align the presence of a concept with the desired timing. However, since correlation is scale-invariant, it may yield high scores even when attention values are too low to render the object visible. To mitigate this, we incorporate a magnitude term that directly promotes stronger attention in frames where the temporal signal is high, and suppresses it elsewhere. Additionally, we introduce an entropy regularization term to maintain spatial focus, ensuring that when attention is activated, it remains coherent and not diffusely spread across the frame.

Beyond single-object scenarios, as shown in Figure~\ref{fig:teaser}, our method generalizes well to multi-object scenes. In such cases, it is crucial to maintain the visual grounding of non-target entities. We find that preserving one object (e.g., the bird) while selectively manipulating another (e.g., the cat) helps retain the overall scene structure. In addition to nouns, our method can also manipulate verbs (e.g., clapping), effectively modifying the motion depicted in the video. Finally, we explore the potential of aligning external audio cues with video generation. Although we focus on simple cases involving a single audio event (e.g., a lightning strike), we observe promising results, suggesting that our approach may naturally extend to zero-shot multimodal control without additional training.

Our contributions are threefold: 
(i) We introduce the first inference-time method for temporally controlling the appearance of specific words in text-to-video diffusion models, without additional supervision or model finetuning, allowing temporal control while preserving the fidelity and diversity of existing models. 
(ii) We propose a set of spatiotemporal attention losses that shape both the timing and strength of word grounding, together with an entropy-based regularization term that maintains semantic consistency. 
(iii) We demonstrate broad applicability across multi-object control, motion timing, and audio-to-video alignment, and introduce a new metric for temporal alignment, a dimension not covered by current benchmarks.

\begin{figure*}
    \centering
    \includegraphics[width=1\linewidth]{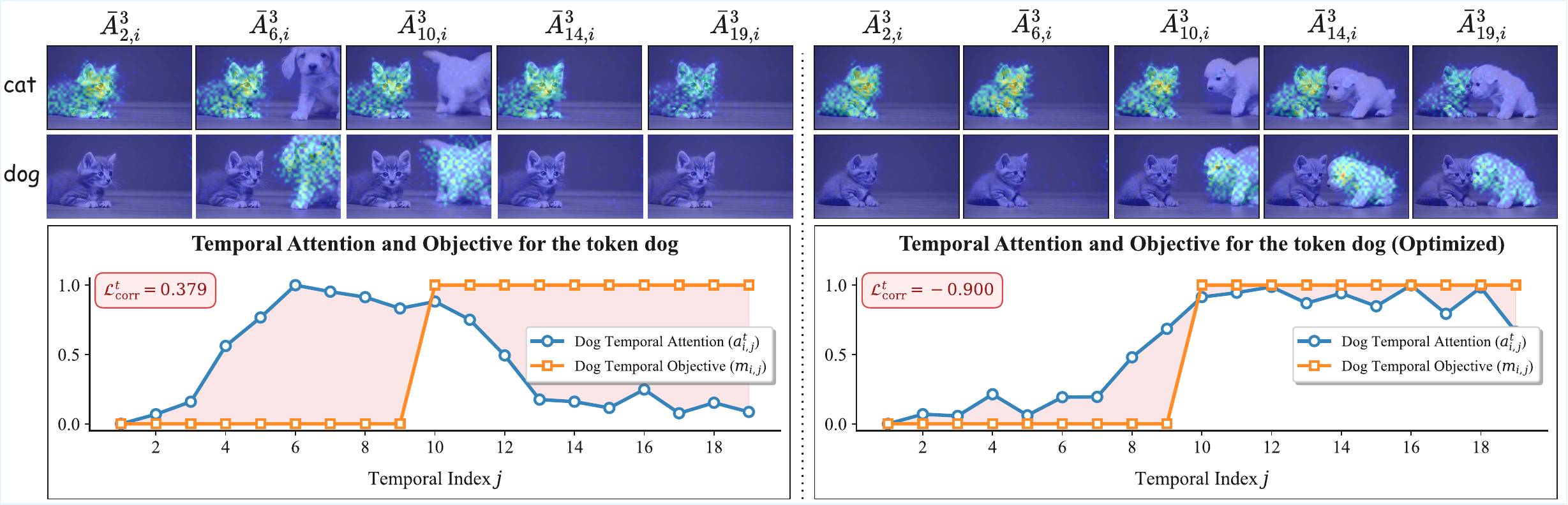}
\caption{Motivation for our approach. \textbf{Top:} We show a video generated for the prompt ``The video begins with a serene view centered on the cat, with no sign of the dog. In the second half, the dog unexpectedly appears, altering the dynamic of the scene.'' The top row displays the attention maps for the tokens \textit{cat} and \textit{dog}, extracted from the denoising step $t{=}3$, for frames $j{=}2,6,10,14,19$. On the left, the video is generated without our optimization. Despite the prompt specifying that the dog should appear in the second half, it appears early. This behavior is common, as Wan~2.1 often fails to depict objects or movements according to temporal cues in the prompt. On the right, after applying our conditioning method, the dog correctly appears in the second half of the video. \textbf{Bottom:} Temporal attention $a_{i,j}^t$ (blue) versus target mask $m_{i,j}$ (orange), with the corresponding Pearson correlation loss shown.}
    \label{fig:atten}
\end{figure*}

\section{Related Work}

\noindent\textbf{Attention-based Control.} Attention manipulation has become a powerful tool for controlling semantics in text-to-image generation. Methods such as Attend-and-Excite~\cite{chefer2023attend, li2023divide, battash2024obtaining, dahary2024yourself} perform inference-time optimization over cross-attention maps to ensure that all textual entities are faithfully represented in the generated image. Stable Flow~\cite{avrahami2025stable} further introduces a training-free editing framework that selectively injects attention features into key layers of a diffusion transformer. Additional approaches, including Prompt-to-Prompt~\cite{hertz2022prompt}, leverage attention restructuring for fine-grained, text-driven edits.

\noindent\textbf{Inference time optimization.} Inference-time optimization for text-to-video diffusion has only recently attracted attention. \cite{li2025training} pre-compute a spatio-temporal sketch and inject plan-consistent noise during sampling to tighten prompt alignment. \cite{wei20253dv} converts a single keyframe into a textured, deformable 3D mesh that is re-rendered at every denoising step to maintain consistency in the garment's appearance. \cite{shaulov2025flowmo} derive variance-based optical-flow cues from the model’s latent predictions and add a flow-coherence loss on-the-fly, smoothing motion without retraining. \cite{lian2023llm} let a large language model generate scene graphs and temporal layouts that act as attention masks, improving complex spatio-temporal grounding. While these methods improve fidelity and consistency without necessitating additional training, none offer explicit, user-defined control over the timing of textual element unfolding.

\noindent\textbf{Controllable Video Generation}
Controllable video generation has advanced rapidly, with methods exploring a broad set of control modalities. Early multimodal systems, such as VideoComposer \cite{wang2023videocomposer}, incorporate depth, sketch, and motion vector guidance to steer video synthesis. More recent approaches introduce explicit camera and object motion control on top of text-to-video generation \cite{wang2024motionctrl, he2024cameractrl, kuang2024collaborative, yang2024direct, feng2025blobgen, bahmani2025ac3d}. Another line of work investigates the controllability of image-to-video object motion~\cite{wang2024boximator, xu2024camco}. While effective, these training approaches rely on large paired datasets and extensive supervision, which can be costly and limit generalization.

A parallel line of work studies training-free controllability, including CamTrol~\cite{hou2024training} for camera motion, MotionClone and DiTFlow~\cite{ling2024motionclone, pondaven2025video} for transferring motion from reference videos, and MOFT~\cite{xiao2024video} for editing motion at inference. These rely on dense motion trajectories, but cannot specify when a concept should appear or disappear. 

The most related work, MinT~\cite{wu2025mind}, arranges multiple textual instructions in temporal order by learning when each should begin. However, MinT provides only coarse event sequencing, whereas our method offers fine-grained timing for individual concepts. Moreover, MinT requires fine-tuning on temporally annotated caption datasets centered on human activities, which limits its generality across diverse prompts. Our approach operates entirely at inference time and generalizes across settings without any training or domain-specific supervision.

\section{Preliminaries}
We build \modelname on top of Wan~2.1~\cite{wan2025wan}, a state-of-the-art diffusion model that serves as our backbone for text-to-video generation. Let $v \in \mathbb{R}^{T_v \times H_v \times W_v \times 3}$ represent the input video, where $T_v$ is the number of frames, $H_v \times W_v$ denotes the spatial resolution of each frame, and the final dimension corresponds to the three RGB color channels. A 3D causal variational autoencoder first encodes the video into a latent tensor
\begin{equation}
z \in \mathbb{R}^{T' \times H' \times W' \times C},
\end{equation}
where $T'$, $H'$, and $W'$ denote the temporal and spatial dimensions of the latent representation, which are reduced relative to the input video.
Flattening the spatial and temporal dimensions, $z$ is reshaped into a sequence of $n_v$ video tokens, where $n_v = T' \cdot H' \cdot W'$, and each token is a $C$-dimensional feature vector representing a specific spatiotemporal region of the video.

The text prompt $\mathcal{P} = [p_1, \dots, p_{n_p}]$ is embedded using a pretrained text encoder. For brevity, we refer to tokens as words, although a word may consist of multiple subtokens. All operations described for words are applied to the average of their corresponding subtokens.
Video synthesis is then performed as a denoising diffusion process in the latent space, where the model iteratively refines latent variables $\{z_t\}_{t=T}^{0}$, gradually denoising $z_T$ into $z_0$. At each timestep $t$, a transformer-based network $\operatorname{DiT}(z_t, \mathcal{P}, t)$ predicts the noise in $z_t$, conditioned on the prompt $\mathcal{P}$ and the timestep $t$.

\subsection{Cross-attention in video diffusion models}
\label{sec:atten}

A key challenge in text-to-video generation is maintaining semantic and temporal alignment with the input prompt. In diffusion-based architectures, this alignment is achieved through cross-attention layers within the transformer blocks at each denoising step~$t$.

During cross-attention, video tokens act as queries that attend to text tokens, which serve as keys and values. This mechanism enables the model to inject contextually relevant linguistic information into each video token representation. The attention weights are computed via scaled dot products between queries and keys, producing a set of attention tensors across layers, $A^t \in \mathbb{R}^{L \times h \times n_v \times n_p}$, where $L$ is the number of cross-attention layers, $h$ is the number of attention heads, $n_v$ is the number of video tokens, and $n_p$ is the number of text tokens. A softmax is applied along the text-token dimension, yielding a probability distribution over prompt words for each video token, per head and per layer.

To obtain a global view of token-level influence, we average attention maps across heads and cross-attention layers, producing an aggregate attention matrix $\bar{A}^t \in \mathbb{R}^{n_v \times n_p}$. This matrix quantifies the influence of each text token on each video token. 

The effect of a specific word~$i$ is measured by extracting the corresponding column $\bar{A}^t_i \in \mathbb{R}^{n_v}$. The vector $\bar{A}^t_i$ can be reshaped at temporal index $j$, yielding $\bar{A}^{t}_{j,i} \in \mathbb{R}^{H' \times W'}$, which represents the spatial influence of word~$i$ at temporal position~$j$ and denoising step~$t$. Figure~\ref{fig:atten} illustrates this for the words \textit{cat} and \textit{dog} in the prompt ``A cat and a dog. In the second half of the video, the dog appears.'' Attention maps are upsampled to frame resolution for display.

The attention map captures the spatiotemporal grounding of words in the generated video; \modelname is specifically designed to steer the \textit{temporal} influence of a token. Thus, we focus on the scalar attention value $\hat{A}^{t}_{j,i} = \langle \bar{A}^{t}_{j,i} \rangle_{x,y} \in \mathbb{R}$, where $\langle \cdot \rangle_{x,y}$ denotes summation over the spatial coordinates $x \in [1, H']$ and $y \in [1, W']$. Intuitively, $\hat{A}^{t}_{j,i}$ measures the influence of word~$i$ on the latent representation at temporal position~$j$ during denoising step~$t$.

To aggregate the temporal attention pattern of word~$i$ at denoising step~$t$, we collect the scalar values across frames into a vector:
\begin{equation}
a_i^t = [\hat{A}^{t}_{1,i}, \hat{A}^{t}_{2,i}, \ldots, \hat{A}^{t}_{T',i}] \in \mathbb{R}^{T'}.
\end{equation}

We omit the first frame due to unstable attention values that do not reliably reflect word grounding. This vector serves as the target for our steering method, described next.

\begin{figure}[t]
    \centering
    \includegraphics[width=0.90\linewidth]{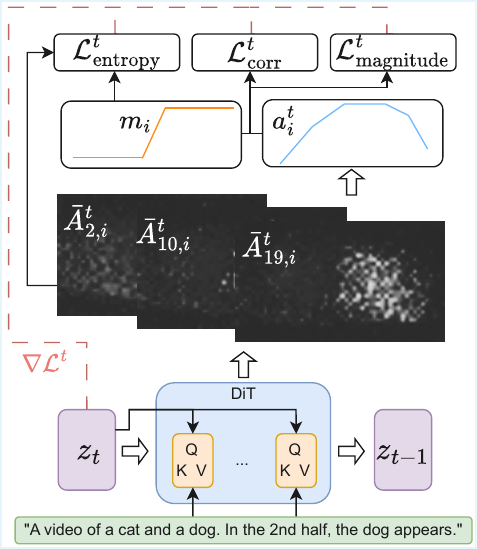}
    \caption{Illustration of \modelname. During a single denoising step $t$, we extract spatial attention maps $\bar{A}^t_{j,i}$ (for word $i$ at temporal index $j$), aggregate to a temporal attention signal $a_{i}^t$, and align it with the target mask vector $m_{i}$ via temporal and spatial losses. Gradients $\nabla \mathcal{L}$ are used to update the latent code $z_t$.}\vspace{-10pt}
    \label{fig:method}
\end{figure}

\section{Method}

Given a video $v$, our objective is to generate an edited version $v'$ such that specific tokens appear according to a temporal condition. To achieve fine-grained temporal control, we employ masks, which provide a more precise and less ambiguous mechanism than textual descriptions. We denote as input a binary mask $m_i = [m_{i,1}, \ldots, m_{i,T'}]$, where each element $m_{i,j} \in \{0, 1\}$ indicates whether token $p_i$ should appear at temporal position $j$ (see Figure~\ref{fig:atten}, bottom, orange line). In this example, the mask specifies that the token \textit{dog} should appear only in the second half of the video (i.e., frames 10–20). In general, $m_{i,j}$ may also take continuous values to represent the strength of appearance. In most experiments, we use binary masks, except in the audio-alignment setting, where the mask encodes sound intensity. The main steps of our approach are illustrated in Figure~\ref{fig:method} and discussed below. We also provide an algorithmic description in the appendix.

We steer the latent representation during the first $k$ denoising steps by applying $l$ gradient-based updates at each step, optimized with AdamW. Let $z_t$ denote the latent at step $t$; the update is:
\begin{equation}
z'_t = z_t - \alpha \nabla_{z_t} \mathcal{L}^t,
\end{equation}
where $\alpha$ is the learning rate.
The loss $\mathcal{L}^t$ consists of a single main component: a temporal correlation term that aligns the attention pattern with the target mask. While this term is most significant for aligning with the temporal condition, we also include two terms that further improve the results: (ii) a magnitude term that amplifies or suppresses token-level activation, and (iii) an attention entropy regularization term that preserves spatial consistency. We describe each component in detail below.

\paragraph{Temporal correlation term.}
The main loss term encourages the temporal attention vector $a_i^t$ to match the target mask $m_i$ using the Pearson correlation:
\begin{equation}
\mathcal{L}_{\text{corr}}^t = -\frac{\text{Cov}(m_i, \tilde{a}_i^t)}{\sigma_{m_i} \sigma_{\tilde{a}_i^t}},
\end{equation}
where $\tilde{a}_i^t$ is the min–max normalized version of $a_i^t$, defined as
$ \tilde{a}_i^t = \frac{a_i^t - \min(a_i^t)}{\max(a_i^t) - \min(a_i^t)} $.
Here, $\text{Cov}(\cdot,\cdot)$ denotes the empirical covariance, and $\sigma_{m_i}$ and $\sigma_{\tilde{a}_i^t}$ are the corresponding standard deviations. Note that if $m_i$ has zero variance, such as when encouraging an object to appear throughout the entire video, the correlation term becomes undefined; in such cases, it is omitted from the loss. 

The Pearson correlation measures the linear relationship between the mask $m_i$ and the attention term $a_i^t$, but it does not capture their scale. In our experiments, we observe two phenomena: (i) words that are initially invisible may remain so, and (ii) completely suppressing the appearance of a token is difficult. The following loss term is therefore designed to modulate the temporal strength of attention.

\paragraph{Attention magnitude term.}
The magnitude term adjusts the overall strength of attention for each word token $i$ according to the temporal mask $m_i$, complementing the correlation-based term by restoring scale sensitivity.  We employ the indicator function $\mathbb{1}_{\{\cdot\}}$, which equals one when the specified condition holds and zero otherwise. The positive magnitude component encourages attention in regions where the word is expected to be active:
\begin{equation}
\mathcal{L}_{\oplus}^t  = \frac{1}{T'} \sum_{j=1}^{T'} \mathbb{1}_{\{m_{i,j} > \tau\}} \cdot a_{i,j}^t.
\end{equation}
The negative magnitude component penalizes attention where the word should not be visible:
\begin{equation}
\mathcal{L}_{\ominus}^t  = \frac{1}{T'} \sum_{j=1}^{T'} \mathbb{1}_{\{m_{i,j} \leq \tau\}} \cdot a_{i,j}^t.
\end{equation}

The final magnitude loss is defined as the net difference:
\begin{equation}
\mathcal{L}_{\text{mag}}^t  =  \mathcal{L}_{\ominus}^t - \mathcal{L}_{\oplus}^t.
\end{equation}

\begin{figure}[t]
    \centering
    \includegraphics[width=1\linewidth]{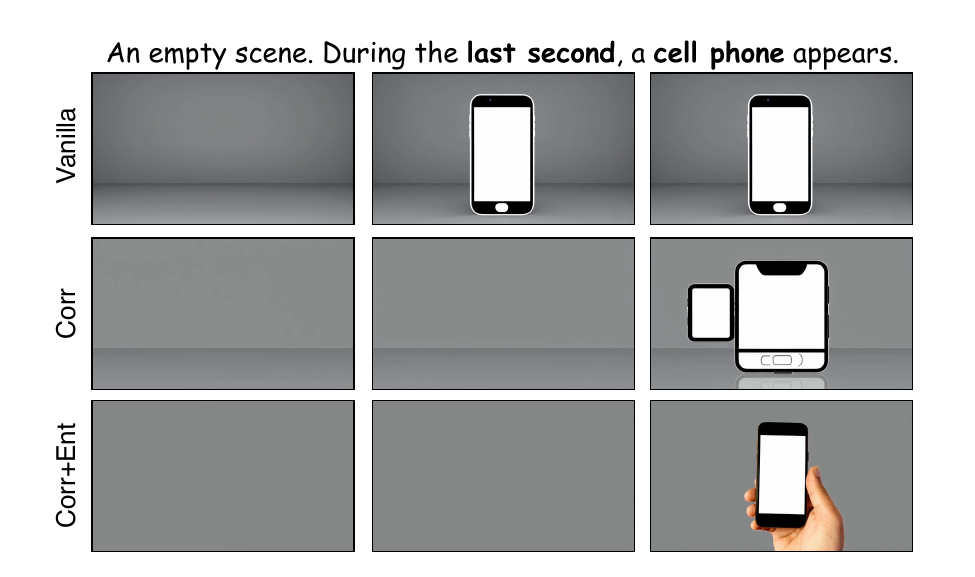}\vspace{-10pt}
    \caption{
Entropy regularization helps preserve semantic fidelity.
    }\vspace{-10pt}
    \label{fig:ablations}
\end{figure}

Here, $\tau \in [0,1]$ is a threshold that determines whether a given time index is considered active for token $p_i$. While $m_i$ is often binary (i.e., 0 or 1), it may also take continuous values (e.g., for soft audio-driven conditioning). In such cases, the indicator $\mathbb{1}_{\{m_{i,j} > \tau\}}$ selects time steps where the target activation is sufficiently strong.

Both the correlation and magnitude terms target the temporal objective but can harm spatial consistency. In our experiments, these terms sometimes corrupt object semantics, such as a smartphone turning into a distorted two-screen tablet (Figure~\ref{fig:ablations}; more examples in the appendix). To address this and preserve spatial coherence, we introduce an entropy-based regularization term.

\paragraph{Attention entropy regularization.}
Optimizing for the temporal objective may compromise spatial properties, often resulting in overly spread attention. We counter this by regularizing the entropy of the spatial attention maps at temporal indices where the word is active (i.e., $m_{i,j} > \tau$), resulting in:
\begin{equation}
\mathcal{L}_{\text{entropy}}^t = \frac{1}{T'} \sum_{j=1}^{T'} \mathbb{1}_{\{m_{i,j} > \tau\}} \cdot \mathcal{H}(\bar{A}^{t}_{j,i}).
\end{equation}
Here, $\mathcal{H}(\bar{A})$ denotes the Shannon entropy computed over the spatial dimensions.

In our experiments, we find that discouraging excessive spreading of attention not only improves temporal presence but also enhances overall image quality (see ablations), suggesting that this term may be beneficial for video generation more broadly.

\paragraph{Full objective.}
The combined loss at step $t$ is:
\begin{equation}
\mathcal{L}^t = \mathcal{L}_{\text{corr}}^t 
+ \lambda_1 \mathcal{L}_{\text{magnitude}}^t 
+ \lambda_2 \mathcal{L}_{\text{entropy}}^t,
\end{equation}
where $\lambda_1$ and $\lambda_2$ control the auxiliary terms. The correlation term is sufficient for most cases and provides strong temporal alignment. The magnitude term is helpful when the base model fails to accurately depict a token or when the token must be removed entirely. The entropy term maintains spatial focus and enhances video quality. The objective is optimized over latents during inference, while the model weights remain fixed.

\paragraph{Correlation-based early stopping.}
Some videos naturally align more closely with the temporal objective and therefore require fewer optimization steps. To adaptively allocate compute, we introduce a stopping rule based on the Pearson correlation in the temporal loss $\mathcal{L}_{\text{corr}}^t$. Specifically, for each denoising step $t$, we perform up to $l$ optimization iterations. If the correlation exceeds a threshold $\tau_{\text{corr}}$, we terminate optimization early for that step; otherwise, we continue refining until the limit is reached.

\section{Experiments}

\paragraph{Experimental setup.}
All experiments are conducted on a single NVIDIA B200 GPU. Due to limited resources, we focus on the smaller model. We optimize the latent representation during the first five denoising steps (see Appendix Sec.~\ref{sec:latency_ablation} for ablations), applying up to 10 gradient updates per step with a learning rate of $5\times10^{-4}$, increased to $1\times10^{-3}$ for the two-object setting. We use $\lambda_1{=}0.3$ and $\lambda_2{=}10$. See the appendix for ablation details.

\paragraph{Baseline.}

We evaluate our method on several text-to-video backbones, including Wan~2.1~(1.3B and 14B), Wan~2.2-TI2V~(5B), CogX1.5-L~(5B) and LTX-2~(19B). As a baseline, we use each model with explicit temporal cues in the prompt (e.g., ``\dots at the third second \dots''). The exact phrasing templates are provided in the appendix. Note that we also use the same timing templates in our approach for fair comparison. The lack of dedicated baselines stems from the novelty of our objective, which involves fine-grained temporal control and necessitates the definition of new evaluation metrics, discussed next.

\paragraph{Metrics.}

Current benchmarks such as VBench~\cite{huang2024vbench} do not measure temporal control. To address this, we introduce \emph{Temporal Accuracy}, which checks whether each token appears in the correct frames specified by the prompt, with separate scores for presence and absence. VBench relies on the GRiT detector~\cite{wu2024grit}, which we find unreliable for small or partially visible objects. Instead, we detect objects using the more recent YOLOv10~\cite{wang2024yolov10}.

For motion-centric prompts, Temporal Accuracy aligns the prompt-specified action time with the second exhibiting the strongest motion, as determined by per-second optical flow magnitude computed from videos sampled at 4 fps. 

We also report results on the VBench Multiple Object benchmark (detecting object appearance throughout the video),  and the Imaging Quality metric that uses a perceptual image-quality model.

\paragraph{Benchmark.}
In the single-object setup, we utilize 80 YOLOv10 classes, which span animals, everyday objects, vehicles, and more. For the two-object setting, we use 82 object pairs based on the VBench multiple-object benchmark. For movement, we curate 100 types of actions. Temporal masks are generated together with the prompt templates to ensure alignment. Full prompt patterns and categories are provided in the appendix.

\begin{table}[t]
\centering
\setlength{\tabcolsep}{4.5pt}
\renewcommand{\arraystretch}{0.97}
\caption{
Comparison on one-object, two-object, and movement temporal-control benchmarks.
Baselines rely on text-only temporal cues, while Ours applies \modelname\ on the corresponding backbone. Bold denotes the best result in each block.
}
\label{tab:all_qual}
\resizebox{\linewidth}{!}{
\begin{tabular}{lccccc}
\toprule
Method & Params & \multicolumn{3}{c}{Temporal} & Img $\uparrow$ \\
\cmidrule(lr){3-5}
& & Acc $\uparrow$ & Abs $\uparrow$ & Pres $\uparrow$ & \\
\midrule

\multicolumn{6}{c}{\textbf{One Object}} \\
\midrule
LTX-2          & 19B  & 74.23 & 85.23 & 62.12 & 42.79 \\
CogX-S           & 2B   & 58.15 & 74.77 & 39.88 & 45.16 \\
CogX1.5-L          & 5B   & 61.43 & 48.64 & 75.50 & 65.36 \\
Wan2.1-S       & 1.3B & 63.94 & 67.38 & 60.50 & 53.76 \\
Wan2.1-L       & 14B  & 64.23 & 55.34 & 74.00 & \textbf{66.89} \\
Wan2.2         & 5B   & 78.02 & \textbf{87.42} & 68.00 & 48.13 \\
\midrule
Ours (CogX-S)     & 2B & 63.75 & 78.41 & 47.62 & 46.67 \\
Ours (Wan2.1-S) & 1.3B & \textbf{83.56} & 87.38 & \textbf{79.75} & 56.92 \\

\midrule
\multicolumn{6}{c}{\textbf{Two Objects}} \\
\midrule
LTX-2          & 19B  & 39.31 & \textbf{64.88} & 16.08 & 58.62 \\
CogX1.5-L         & 5B   & 24.68 & 46.46 & 4.88  & 63.78 \\
Wan2.1-S       & 1.3B & 37.50 & 45.85 & 29.15 & 68.56 \\
Wan2.1-L       & 14B  & 40.07 & 54.15 & 27.27 & 69.01 \\
Wan2.2         & 5B   & 47.05 & 56.86 & 36.59 & 67.83 \\
\midrule
Ours~(Wan2.1-S)         & 1.3B & \textbf{53.17} & 57.32 & \textbf{49.02} & \textbf{70.82} \\

\midrule
\multicolumn{6}{c}{\textbf{Movement}} \\
\midrule
LTX-2          & 19B  & 28.00 & -- & -- & 51.61 \\
CogX1.5-L           & 5B   & 21.00 & -- & -- & 61.39 \\
Wan2.1-S       & 1.3B & 19.00 & -- & -- & 60.46 \\
Wan2.1-L       & 14B  & 18.00 & -- & -- & 61.31 \\
Wan2.2         & 5B   & 23.00 & -- & -- & 60.61 \\
\midrule
Ours~(Wan2.1-S)          & 1.3B & \textbf{54.00} & -- & -- & \textbf{63.24} \\

\bottomrule
\end{tabular}
}
\vspace{-15pt}
\end{table}

\subsection{Quantitative Results}

\paragraph{Single-object temporal control.}
Table~\ref{tab:all_qual} shows that \modelname\ significantly improves temporal control across backbones. On Wan2.1-S, we achieve a +19.62\% gain in Temporal Accuracy, with improvements of +20.0\% in absence and +19.25\% in presence accuracy, while also increasing Imaging Quality by $\sim$3\%. 

We find that Wan2.2 achieves high absence accuracy, but suffers from low Imaging Quality (48.13\%), largely due to explicit temporal phrases in the prompt that degrade visual fidelity and lead to missed detections (see Limitations). Removing these cues improves Imaging Quality to 57.78\% but reduces absence accuracy to 71.25\% (see Appendix).

Similar, though smaller, gains are observed on CogX-S reflecting a stronger bias toward its training distribution. In practice, this also makes more complex settings difficult to evaluate reliably: multi-object prompts often fail to generate all requested objects, and motion prompts frequently produce little or no discernible motion peak.

\paragraph{Two objects temporal control.} 

In Table~\ref{tab:all_qual} (Two Objects), the two-object setting is notably challenging. Wan2.1-S baseline achieves only 37.50\% temporal accuracy, while our method improves this to 53.17\% (+15.67). Absence and presence accuracies also increase by 11.47\% and 19.87\%, respectively, and Imaging Quality improves by 2.26\%.

Larger models also show similar degradation in temporal control in the two-object setting. Conversely, Imaging Quality is consistently higher across all methods, suggesting a possible bias in the perceptual metric toward multi-object scenes. Our method achieves the highest Imaging Quality.

\paragraph{Movement temporal control.}
Table~\ref{tab:all_qual} (Movement) evaluates temporal accuracy when conditioning is indicative of a peak of action. The temporal accuracy measures the timing of when the action peaks (see Metrics).  Wan2.1-S baseline achieves only 19\% temporal accuracy, indicating that the model does not follow action timing from text alone. \modelname improves this to 54\%, showing a substantial gain in aligning motion onset with the prompt.

Low performance in movement conditioning is consistent across all models. Our Imaging Quality remains the best in this setup as well.

\paragraph{Multiple objects benchmark.} We also report results on the VBench Multiple Objects benchmark in Table~\ref{tab:multipleobjects}, which tests the ability to generate two objects described in the input text throughout the entire video. We observe improved performance on the Multiple Object metric (76.37\% vs.\ 74.13\%). Detailed results are provided in the appendix.

\paragraph{Ablation.}
Table~\ref{tab:ablations} shows that each component contributes differently to temporal control and visual quality. Pearson correlation (C) drives strong temporal alignment, mainly through gains in absence accuracy, but degrades image quality and often leads to missed detections and inflated absence scores. Entropy (E) improves presence accuracy and stabilizes attention, while their combination (C+E) achieves a balanced trade-off. Notably, entropy alone yields the highest Imaging Quality (59.52\%), even surpassing the text baseline, highlighting the general benefit of attention smoothing. Additional ablations on loss weights, optimization steps and alternative objectives are provided in the appendix.

\paragraph{Human evaluation.}
We conducted a user study with 50 computer science graduate students, each comparing 16 video pairs (8 one-object and 8 two-object) generated by \modelname and Wan~2.1 with text-based temporal cues. Annotators judged temporal accuracy and visual quality, or indicated no difference, which we asked them to use sparingly (full protocol details are provided in the appendix). As shown in Table~\ref{tab:human}, our method is preferred for temporal accuracy (61.51\% vs.\ 16.94\%) and also improves visual quality. Notably, 21.55\% of temporal comparisons reported no preference, reflecting that humans struggle to perceive minor timing differences.

\begin{table}[t]
\centering
\setlength{\tabcolsep}{4.5pt}
\renewcommand{\arraystretch}{0.97}
\caption{
Comparison on the multiple-object benchmark.
Baselines rely on text-only temporal cues, while Ours applies \modelname.
Bold denotes the best result in each column.
}
\label{tab:multipleobjects}
\begin{tabular}{lcccc}
\toprule
Method & \multicolumn{2}{c}{Temporal} & Img $\uparrow$ \\
\cmidrule(lr){2-3}
& \makecell{GRiT $\uparrow$} & \makecell{YOLO $\uparrow$} & \\
\midrule
Text & 74.13\% & 61.54\% & \textbf{70.25\%} \\
Ours & \textbf{76.37\%} & \textbf{65.73\%} & 70.21\% \\
\bottomrule
\end{tabular}\vspace{-5pt}
\end{table}

\begin{table}[t]
\centering
\setlength{\tabcolsep}{4.5pt}
\renewcommand{\arraystretch}{0.97}
\caption{
Ablation study on the one-object benchmark.
C denotes Pearson correlation and E denotes entropy.
Bold denotes the best result in each column.
}
\label{tab:ablations}
\begin{tabular}{lccccc}
\toprule
Method & \multicolumn{3}{c}{Temporal} & Img $\uparrow$ \\
\cmidrule(lr){2-4}
& Acc $\uparrow$ & Abs $\uparrow$ & Pres $\uparrow$ & \\
\midrule
Text & 63.94\% & 67.38\% & 60.50\% & \textbf{53.76\%} \\
\midrule
Only C & 81.19\% & \textbf{91.50\%} & 70.88\% & 50.96\% \\
Only E & 72.94\% & 66.88\% & 79.00\% & \textbf{59.52\%} \\
C + E & 78.38\% & 77.38\% & 79.38\% & 57.60\% \\
Ours & \textbf{82.50\%} & 83.25\% & \textbf{81.75\%} & 56.51\% \\
\bottomrule
\end{tabular}\vspace{-5pt}
\end{table}

\begin{table}[t]
\small
\setlength{\tabcolsep}{4pt}
\centering
\caption{
User study comparing our method with Wan2.1 in terms of temporal accuracy and visual quality. 
}
\begin{tabular}{lcccc}
\toprule
\textbf{Question Type} & Ours & Baseline & Same &  \\
\midrule
Temporal Accuracy & \textbf{61.51\%} & 16.94\% & 21.55\%\\
Visual Quality     & \textbf{62.66\%} & 25.99\% & 11.35\% \\
\bottomrule
\end{tabular}\vspace{-10pt}
\label{tab:human}
\end{table}

\begin{figure*}[t]
    \centering
    \includegraphics[width=0.9\linewidth]{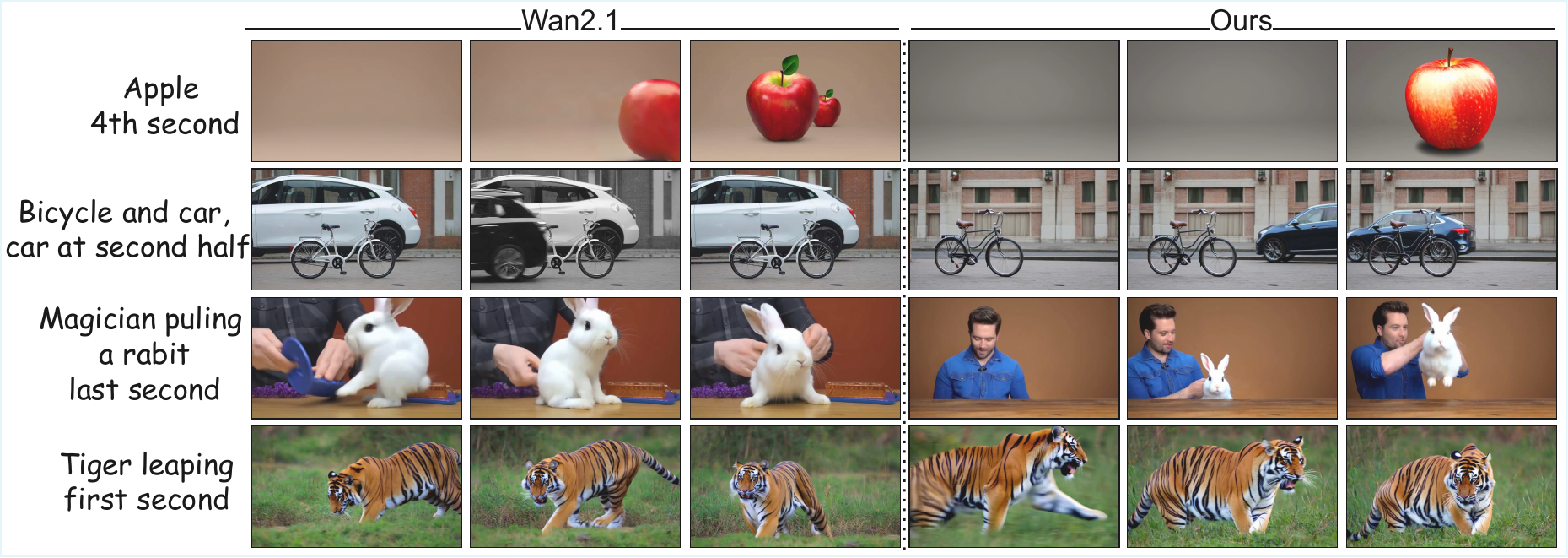}\vspace{-5pt}
\caption{Examples generated by Wan~2.1 (left) and our temporally conditioned method (right) for prompts with timing constraints. Prompts are simplified for brevity. Shown are frames from the first, middle, and final thirds of each video.}
    \label{fig:qual}
\end{figure*}

\begin{figure}[t]
    \centering
    \includegraphics[width=1\linewidth]{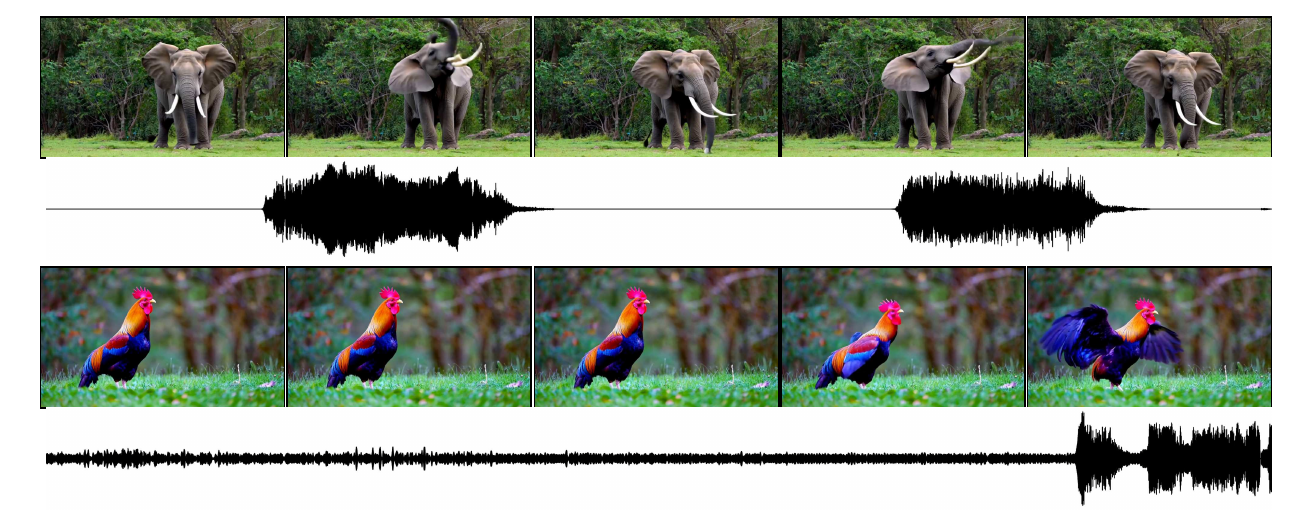}\vspace{-5pt}
    \caption{Examples of video alignment to an audio signal.}\vspace{-20pt}
    \label{fig:audio}
\end{figure}
\subsection{Qualitative Results}
In Figure~\ref{fig:qual}, each row illustrates a different temporal-control task.  
For a one-object prompt (apple appearing late), Wan~2.1 places the object near the middle, while our method correctly places it at the end.  
For a two-object prompt (car appearing in the second half), Wan~2.1 shows the car too early, whereas our method aligns it with the specified time.  
For movement-based prompts, Wan~2.1 fails to control timing (e.g., the rabbit appears throughout or the tiger leaps at incorrect times), while our method accurately localizes these events.  
Additional examples are provided in the supplementary material.

\subsection{Audio-Visual Alignment}
We also explore zero-shot alignment of video to external audio by using a preprocessed audio envelope as the temporal condition. Given a waveform $y$, we compute its onset-strength envelope $s_t$, pool it to the number of frames, and normalize it to $[0,1]$. Peaks are preserved through thresholding and otherwise smoothed with a Gaussian filter:
\begin{equation}
    \tilde{s}_t =
    \begin{cases}
    \hat{s}_t, & \hat{s}_t \ge \tau_{audio}, \\
    \text{GaussianSmooth}(\hat{s}_t; \sigma), & \text{otherwise}.
    \end{cases}
\end{equation}

Figure~\ref{fig:audio} shows examples where video motion aligns with peaks in the audio envelope, with additional results in the appendix. While tested on simple cases with clear peaks, these results suggest that audio-driven temporal conditioning is a promising approach for achieving audio-video alignment without requiring training on paired datasets.

\paragraph{Limitations.}
\modelname increases inference time, although it remains an inference-only method and avoids retraining. It can also introduce mild attribute shifts (e.g., color changes), as the current objectives do not explicitly enforce full semantic consistency.

Our evaluation also relies on proxy metrics based on object detection (YOLOv10) and optical flow. A limitation is that temporal absence can be sensitive to detection failures or degraded image quality, and may be artificially inflated. In contrast, temporal presence is more reliably captured. Notably, our improvements are primarily driven by gains in temporal presence rather than absence, making the evaluation more robust. Future work may incorporate semantic alignment metrics for a more comprehensive assessment.

\subsection{Conclusions}

Current text-to-video models generate high-quality videos but lack precise control over when objects and actions appear. We find that adding explicit timing cues does not improve temporal control and can substantially degrade visual quality (e.g., for Wan~2.1, Temporal Accuracy changes from 65.18\% without timing cues to 63.94\% with them, while Imaging Quality drops from 59.99\% to 53.76\%). To address this, we introduce a data-free temporal conditioning method that preserves visual quality while enabling fine-grained temporal control through attention steering.

These results suggest that temporal control is not a prompting problem. Even with explicit timing cues, models fail to produce correct temporal behavior, indicating a gap in how temporal information is represented. This motivates the use of explicit control mechanisms, such as ours, for reliable temporal alignment.

{
    \small
    \bibliographystyle{ieeenat_fullname}
    \bibliography{references}

\begin{thebibliography}{35}
\providecommand{\natexlab}[1]{#1}
\providecommand{\url}[1]{\texttt{#1}}
\expandafter\ifx\csname urlstyle\endcsname\relax
  \providecommand{\doi}[1]{doi: #1}\else
  \providecommand{\doi}{doi: \begingroup \urlstyle{rm}\Url}\fi

\bibitem[Avrahami et~al.(2025)Avrahami, Patashnik, Fried, Nemchinov, Aberman, Lischinski, and Cohen-Or]{avrahami2025stable}
Omri Avrahami, Or Patashnik, Ohad Fried, Egor Nemchinov, Kfir Aberman, Dani Lischinski, and Daniel Cohen-Or.
\newblock Stable flow: Vital layers for training-free image editing.
\newblock In \emph{Proceedings of the Computer Vision and Pattern Recognition Conference}, pages 7877--7888, 2025.

\bibitem[Bahmani et~al.(2025)Bahmani, Skorokhodov, Qian, Siarohin, Menapace, Tagliasacchi, Lindell, and Tulyakov]{bahmani2025ac3d}
Sherwin Bahmani, Ivan Skorokhodov, Guocheng Qian, Aliaksandr Siarohin, Willi Menapace, Andrea Tagliasacchi, David~B Lindell, and Sergey Tulyakov.
\newblock Ac3d: Analyzing and improving 3d camera control in video diffusion transformers.
\newblock In \emph{Proceedings of the Computer Vision and Pattern Recognition Conference}, pages 22875--22889, 2025.

\bibitem[Bar-Tal et~al.(2024)Bar-Tal, Chefer, Tov, Herrmann, Paiss, Zada, Ephrat, Hur, Liu, Raj, et~al.]{bar2024lumiere}
Omer Bar-Tal, Hila Chefer, Omer Tov, Charles Herrmann, Roni Paiss, Shiran Zada, Ariel Ephrat, Junhwa Hur, Guanghui Liu, Amit Raj, et~al.
\newblock Lumiere: A space-time diffusion model for video generation.
\newblock In \emph{SIGGRAPH Asia 2024 Conference Papers}, pages 1--11, 2024.

\bibitem[Battash et~al.(2024{\natexlab{a}})Battash, Rozner, Wolf, and Lindenbaum]{battash2024obtaining}
Barak Battash, Amit Rozner, Lior Wolf, and Ofir Lindenbaum.
\newblock Obtaining favorable layouts for multiple object generation.
\newblock \emph{arXiv preprint arXiv:2405.00791}, 2024{\natexlab{a}}.

\bibitem[Battash et~al.(2024{\natexlab{b}})Battash, Wolf, and Lindenbaum]{battash2024revisiting}
Barak Battash, Lior Wolf, and Ofir Lindenbaum.
\newblock Revisiting the noise model of stochastic gradient descent.
\newblock In \emph{International Conference on Artificial Intelligence and Statistics}, pages 4780--4788. PMLR, 2024{\natexlab{b}}.

\bibitem[Blattmann et~al.(2023)Blattmann, Rombach, Ling, Dockhorn, Kim, Fidler, and Kreis]{blattmann2023align}
Andreas Blattmann, Robin Rombach, Huan Ling, Tim Dockhorn, Seung~Wook Kim, Sanja Fidler, and Karsten Kreis.
\newblock Align your latents: High-resolution video synthesis with latent diffusion models.
\newblock In \emph{Proceedings of the IEEE/CVF Conference on Computer Vision and Pattern Recognition}, pages 22563--22575, 2023.

\bibitem[Chefer et~al.(2023)Chefer, Alaluf, Vinker, Wolf, and Cohen-Or]{chefer2023attend}
Hila Chefer, Yuval Alaluf, Yael Vinker, Lior Wolf, and Daniel Cohen-Or.
\newblock Attend-and-excite: Attention-based semantic guidance for text-to-image diffusion models.
\newblock \emph{ACM transactions on Graphics (TOG)}, 42\penalty0 (4):\penalty0 1--10, 2023.

\bibitem[Dahary et~al.(2024)Dahary, Patashnik, Aberman, and Cohen-Or]{dahary2024yourself}
Omer Dahary, Or Patashnik, Kfir Aberman, and Daniel Cohen-Or.
\newblock Be yourself: Bounded attention for multi-subject text-to-image generation.
\newblock In \emph{European Conference on Computer Vision}, pages 432--448. Springer, 2024.

\bibitem[Feng et~al.(2025)Feng, Liu, Liu, Wang, Vahdat, and Nie]{feng2025blobgen}
Weixi Feng, Chao Liu, Sifei Liu, William~Yang Wang, Arash Vahdat, and Weili Nie.
\newblock Blobgen-vid: Compositional text-to-video generation with blob video representations.
\newblock In \emph{Proceedings of the Computer Vision and Pattern Recognition Conference}, pages 12989--12998, 2025.

\bibitem[Gupta et~al.(2024)Gupta, Yu, Sohn, Gu, Hahn, Li, Essa, Jiang, and Lezama]{gupta2024photorealistic}
Agrim Gupta, Lijun Yu, Kihyuk Sohn, Xiuye Gu, Meera Hahn, Fei-Fei Li, Irfan Essa, Lu Jiang, and Jos{\'e} Lezama.
\newblock Photorealistic video generation with diffusion models.
\newblock In \emph{European Conference on Computer Vision}, pages 393--411. Springer, 2024.

\bibitem[He et~al.(2024)He, Xu, Guo, Wetzstein, Dai, Li, and Yang]{he2024cameractrl}
Hao He, Yinghao Xu, Yuwei Guo, Gordon Wetzstein, Bo Dai, Hongsheng Li, and Ceyuan Yang.
\newblock Cameractrl: Enabling camera control for text-to-video generation, 2024.

\bibitem[Hertz et~al.(2023)Hertz, Mokady, Tenenbaum, Aberman, Pritch, and Cohen-or]{hertz2022prompt}
Amir Hertz, Ron Mokady, Jay Tenenbaum, Kfir Aberman, Yael Pritch, and Daniel Cohen-or.
\newblock {Prompt-to-Prompt Image Editing with Cross-Attention Control}.
\newblock In \emph{The Eleventh International Conference on Learning Representations}, 2023.

\bibitem[Hou and Chen(2024)]{hou2024training}
Chen Hou and Zhibo Chen.
\newblock Training-free camera control for video generation.
\newblock \emph{arXiv preprint arXiv:2406.10126}, 2024.

\bibitem[Huang et~al.(2024)Huang, He, Yu, Zhang, Si, Jiang, Zhang, Wu, Jin, Chanpaisit, et~al.]{huang2024vbench}
Ziqi Huang, Yinan He, Jiashuo Yu, Fan Zhang, Chenyang Si, Yuming Jiang, Yuanhan Zhang, Tianxing Wu, Qingyang Jin, Nattapol Chanpaisit, et~al.
\newblock Vbench: Comprehensive benchmark suite for video generative models.
\newblock In \emph{Proceedings of the IEEE/CVF Conference on Computer Vision and Pattern Recognition}, pages 21807--21818, 2024.

\bibitem[Kim et~al.(2023)Kim, Lee, Kim, Ha, and Zhu]{densediffusion}
Yunji Kim, Jiyoung Lee, Jin-Hwa Kim, Jung-Woo Ha, and Jun-Yan Zhu.
\newblock Dense text-to-image generation with attention modulation.
\newblock In \emph{ICCV}, 2023.

\bibitem[Kuang et~al.(2024)Kuang, Cai, He, Xu, Li, Guibas, and Wetzstein]{kuang2024collaborative}
Zhengfei Kuang, Shengqu Cai, Hao He, Yinghao Xu, Hongsheng Li, Leonidas~J Guibas, and Gordon Wetzstein.
\newblock Collaborative video diffusion: Consistent multi-video generation with camera control.
\newblock \emph{Advances in Neural Information Processing Systems}, 37:\penalty0 16240--16271, 2024.

\bibitem[Li et~al.(2025)Li, Yu, Lin, Cho, Yoon, and Bansal]{li2025training}
Jialu Li, Shoubin Yu, Han Lin, Jaemin Cho, Jaehong Yoon, and Mohit Bansal.
\newblock Training-free guidance in text-to-video generation via multimodal planning and structured noise initialization.
\newblock \emph{arXiv preprint arXiv:2504.08641}, 2025.

\bibitem[Li et~al.(2023)Li, Keuper, Zhang, and Khoreva]{li2023divide}
Yumeng Li, Margret Keuper, Dan Zhang, and Anna Khoreva.
\newblock Divide \& bind your attention for improved generative semantic nursing.
\newblock \emph{arXiv preprint arXiv:2307.10864}, 2023.

\bibitem[Lian et~al.(2023)Lian, Shi, Yala, Darrell, and Li]{lian2023llm}
Long Lian, Baifeng Shi, Adam Yala, Trevor Darrell, and Boyi Li.
\newblock Llm-grounded video diffusion models.
\newblock \emph{arXiv preprint arXiv:2309.17444}, 2023.

\bibitem[Ling et~al.(2024)Ling, Bu, Zhang, Dong, Zang, Wu, Chen, Wang, and Jin]{ling2024motionclone}
Pengyang Ling, Jiazi Bu, Pan Zhang, Xiaoyi Dong, Yuhang Zang, Tong Wu, Huaian Chen, Jiaqi Wang, and Yi Jin.
\newblock Motionclone: Training-free motion cloning for controllable video generation.
\newblock \emph{arXiv preprint arXiv:2406.05338}, 2024.

\bibitem[Pondaven et~al.(2025)Pondaven, Siarohin, Tulyakov, Torr, and Pizzati]{pondaven2025video}
Alexander Pondaven, Aliaksandr Siarohin, Sergey Tulyakov, Philip Torr, and Fabio Pizzati.
\newblock Video motion transfer with diffusion transformers.
\newblock In \emph{Proceedings of the Computer Vision and Pattern Recognition Conference}, pages 22911--22921, 2025.

\bibitem[Shaulov et~al.(2025)Shaulov, Hazan, Wolf, and Chefer]{shaulov2025flowmo}
Ariel Shaulov, Itay Hazan, Lior Wolf, and Hila Chefer.
\newblock Flowmo: Variance-based flow guidance for coherent motion in video generation.
\newblock \emph{arXiv preprint arXiv:2506.01144}, 2025.

\bibitem[Singer et~al.(2022)Singer, Polyak, Hayes, Yin, An, Zhang, Hu, Yang, Ashual, Gafni, et~al.]{singer2022make}
Uriel Singer, Adam Polyak, Thomas Hayes, Xi Yin, Jie An, Songyang Zhang, Qiyuan Hu, Harry Yang, Oron Ashual, Oran Gafni, et~al.
\newblock Make-a-video: Text-to-video generation without text-video data.
\newblock \emph{arXiv preprint arXiv:2209.14792}, 2022.

\bibitem[Wan et~al.(2025)Wan, Wang, Ai, Wen, Mao, Xie, Chen, Yu, Zhao, Yang, et~al.]{wan2025wan}
Team Wan, Ang Wang, Baole Ai, Bin Wen, Chaojie Mao, Chen-Wei Xie, Di Chen, Feiwu Yu, Haiming Zhao, Jianxiao Yang, et~al.
\newblock Wan: Open and advanced large-scale video generative models.
\newblock \emph{arXiv preprint arXiv:2503.20314}, 2025.

\bibitem[Wang et~al.(2024{\natexlab{a}})Wang, Chen, Liu, Chen, Lin, Han, et~al.]{wang2024yolov10}
Ao Wang, Hui Chen, Lihao Liu, Kai Chen, Zijia Lin, Jungong Han, et~al.
\newblock Yolov10: Real-time end-to-end object detection.
\newblock \emph{Advances in Neural Information Processing Systems}, 37:\penalty0 107984--108011, 2024{\natexlab{a}}.

\bibitem[Wang et~al.(2024{\natexlab{b}})Wang, Zhang, Zou, Zeng, Wei, Yuan, and Li]{wang2024boximator}
Jiawei Wang, Yuchen Zhang, Jiaxin Zou, Yan Zeng, Guoqiang Wei, Liping Yuan, and Hang Li.
\newblock Boximator: Generating rich and controllable motions for video synthesis.
\newblock \emph{arXiv preprint arXiv:2402.01566}, 2024{\natexlab{b}}.

\bibitem[Wang et~al.(2023)Wang, Yuan, Zhang, Chen, Wang, Zhang, Shen, Zhao, and Zhou]{wang2023videocomposer}
Xiang Wang, Hangjie Yuan, Shiwei Zhang, Dayou Chen, Jiuniu Wang, Yingya Zhang, Yujun Shen, Deli Zhao, and Jingren Zhou.
\newblock Videocomposer: Compositional video synthesis with motion controllability.
\newblock \emph{Advances in Neural Information Processing Systems}, 36:\penalty0 7594--7611, 2023.

\bibitem[Wang et~al.(2024{\natexlab{c}})Wang, Yuan, Wang, Li, Chen, Xia, Luo, and Shan]{wang2024motionctrl}
Zhouxia Wang, Ziyang Yuan, Xintao Wang, Yaowei Li, Tianshui Chen, Menghan Xia, Ping Luo, and Ying Shan.
\newblock Motionctrl: A unified and flexible motion controller for video generation.
\newblock In \emph{ACM SIGGRAPH 2024 Conference Papers}, pages 1--11, 2024{\natexlab{c}}.

\bibitem[Wei et~al.(2025)Wei, Yu, Zhou, and Wang]{wei20253dv}
Min Wei, Chaohui Yu, Jingkai Zhou, and Fan Wang.
\newblock 3dv-ton: Textured 3d-guided consistent video try-on via diffusion models.
\newblock \emph{arXiv preprint arXiv:2504.17414}, 2025.

\bibitem[Wu et~al.(2024)Wu, Wang, Yang, Gan, Liu, Yuan, and Wang]{wu2024grit}
Jialian Wu, Jianfeng Wang, Zhengyuan Yang, Zhe Gan, Zicheng Liu, Junsong Yuan, and Lijuan Wang.
\newblock Grit: A generative region-to-text transformer for object understanding.
\newblock In \emph{European Conference on Computer Vision}, pages 207--224. Springer, 2024.

\bibitem[Wu et~al.(2025)Wu, Siarohin, Menapace, Skorokhodov, Fang, Chordia, Gilitschenski, and Tulyakov]{wu2025mind}
Ziyi Wu, Aliaksandr Siarohin, Willi Menapace, Ivan Skorokhodov, Yuwei Fang, Varnith Chordia, Igor Gilitschenski, and Sergey Tulyakov.
\newblock Mind the time: Temporally-controlled multi-event video generation.
\newblock In \emph{Proceedings of the Computer Vision and Pattern Recognition Conference}, pages 23989--24000, 2025.

\bibitem[Xiao et~al.(2024)Xiao, Zhou, Yang, and Pan]{xiao2024video}
Zeqi Xiao, Yifan Zhou, Shuai Yang, and Xingang Pan.
\newblock Video diffusion models are training-free motion interpreter and controller.
\newblock \emph{Advances in Neural Information Processing Systems}, 37:\penalty0 76115--76138, 2024.

\bibitem[Xu et~al.(2024)Xu, Nie, Liu, Liu, Kautz, Wang, and Vahdat]{xu2024camco}
Dejia Xu, Weili Nie, Chao Liu, Sifei Liu, Jan Kautz, Zhangyang Wang, and Arash Vahdat.
\newblock Camco: Camera-controllable 3d-consistent image-to-video generation.
\newblock \emph{arXiv preprint arXiv:2406.02509}, 2024.

\bibitem[Yang et~al.(2024{\natexlab{a}})Yang, Hou, Huang, Ma, Wan, Zhang, Chen, and Liao]{yang2024direct}
Shiyuan Yang, Liang Hou, Haibin Huang, Chongyang Ma, Pengfei Wan, Di Zhang, Xiaodong Chen, and Jing Liao.
\newblock Direct-a-video: Customized video generation with user-directed camera movement and object motion.
\newblock In \emph{ACM SIGGRAPH 2024 Conference Papers}, pages 1--12, 2024{\natexlab{a}}.

\bibitem[Yang et~al.(2024{\natexlab{b}})Yang, Teng, Zheng, Ding, Huang, Xu, Yang, Hong, Zhang, Feng, et~al.]{yang2024cogvideox}
Zhuoyi Yang, Jiayan Teng, Wendi Zheng, Ming Ding, Shiyu Huang, Jiazheng Xu, Yuanming Yang, Wenyi Hong, Xiaohan Zhang, Guanyu Feng, et~al.
\newblock Cogvideox: Text-to-video diffusion models with an expert transformer.
\newblock \emph{arXiv preprint arXiv:2408.06072}, 2024{\natexlab{b}}.

\end{thebibliography}
}

\clearpage

\onecolumn

\appendix
\section*{Supplementary Material for \modelname{}: Temporal Attention Guidance for Text-to-Video Models}

This supplementary material includes algorithmic pseudocode, ablation studies, experimental design details, application-specific configurations, extended quantitative and qualitative results, and the human-evaluation setup. Videos and code are also provided in the supplementary package. 

The sections of the supplementary material are as follows:

\begin{enumerate}
    \item Algorithm pseudocode (Section~\ref{sec:appendix-algorithm}).
    \item Experimental setup and hyperparameters (Section~\ref{sec:appendix-expsetup}).
    \item Configuration details for each application setting (Section~\ref{sec:appendix-applications}).
    \item Human evaluation protocol and interface (Section~\ref{sec:appendix-human-eval}).
    \item Latency analysis and trade-off (Section~\ref{sec:latency_ablation}).
    \item Comparison to Dense Diffusion (Section~\ref{sec:appendix_dense}).
    \item Optimization steps and update frequency (Section~\ref{sec:appendix_steps}).
    \item Extended multi-object evaluation (Section~\ref{sec:appendix-multitoken}).
    \item Ablation of loss components (Section~\ref{sec:appendix-ablations}).
    \item Comparison to cosine similarity objective (Section~\ref{sec:appendix_cosine}).
    \item Limitations of explicit temporal cues in text-to-video models (Section~\ref{sec:appendix-time-prompts}).
    \item Additional qualitative results (Section~\ref{sec:appendix-qualitative}).
\end{enumerate}

\clearpage

\subsection{Algorithm Pseudo-code}
\label{sec:appendix-algorithm}

Algorithm~\ref{alg:attend-align} outlines our inference-time optimization procedure. Starting from noise $z_T$, we iteratively steer the latent code during the first $k$ denoising steps of the diffusion process. At each step $t$, we compute cross-attention maps from the video-text model $\operatorname{DiT}$ and derive the scalar temporal attention $a_i^t$ for each token $p_i$.

The optimization loss $\mathcal{L}$ combines three components: correlation alignment $\mathcal{L}_{\text{c}}^t$, magnitude modulation $\mathcal{L}_{\text{enr}}^t$, entropy regularization and $\mathcal{L}_{\text{ent}}^t$. The latent code is updated via gradient descent for up to $l$ steps per denoising timestep, or until the Pearson correlation exceeds a threshold $\tau_{\text{corr}}$. After optimization, the standard denoising process resumes to produce the final latent $z_0$.

\begin{algorithm}[h]
\caption{\modelname: Temporal Attention-Guided Inference-Time Steering}
\label{alg:attend-align}
\textbf{Input:} Prompt $\mathcal{P}$, tokens $p_i$ with temporal masks $m_i$, total diffusion steps $T$, optimized steps $k$, max updates per step $l$, correlation threshold $\tau_{\text{corr}}$, model $\operatorname{DiT}$\\
\textbf{Output:} Final latent $z_0$
\begin{algorithmic}[1]
\State Initialize $z_T \sim \mathcal{N}(0, I)$
\For{$t = T, T{-}1, \dots, T{-}k{+}1$}
    \For{$u = 1$ to $l$}
    \State Extract cross-attentions $A^t$ from $\operatorname{DiT}(z_t, \mathcal{P}, t)$
        \State Average heads and layers: $\bar{A}^t$
        \State $\mathcal{L} \gets 0$
        \For{each $p_i$ with $m_i$}
\State $a_i^t \gets [\langle \bar{A}^t_{j,i} \rangle_{x,y}]_{j=1}^{T'}$
            \State $\mathcal{L}_{\text{c}}^t \gets -\text{Corr}(m_i, \text{minmax}(a_i^t))$
            \State $\mathcal{L}_{\oplus}^t \gets \frac{1}{T'} \sum_{j=1}^{T'} \mathbb{1}[m_{i,j} > \tau] \cdot a_{i,j}^t$
            \State $\mathcal{L}_{\ominus}^t \gets \frac{1}{T'} \sum_{j=1}^{T'} \mathbb{1}[m_{i,j} \leq \tau] \cdot a_{i,j}^t$
            \State $\mathcal{L}_{\text{mag}}^t \gets  \mathcal{L}_{\ominus}^t - \mathcal{L}_{\oplus}^t$ 
\State $\mathcal{L}_{\text{ent}}^t \gets \frac{1}{T'} \sum_{j=1}^{T'} \mathbb{1}[m_{i,j} > \tau] \cdot \mathcal{H}(\bar{A}^t_{j,i})$
\State $\mathcal{L} \gets \mathcal{L} + \mathcal{L}_{\text{c}}^t + \lambda_1 \mathcal{L}_{\text{enr}}^t + \lambda_2 \mathcal{L}_{\text{ent}}^t$
        \EndFor
        \State $z_t' \gets z_t - \alpha_t \nabla_{z_t} (\mathcal{L} / N)$
        \If{All $\mathcal{L}_{\text{corr}}^t \leq \tau_{\text{corr}}$} \label{alg:pearson-check}
            \State \textbf{break}
        \EndIf
        \State $z_t \gets z_t'$
    \EndFor
    \State $z_{t-1} \gets \operatorname{DiT}(z_t', \mathcal{P}, t)$
\EndFor
\State Continue denoising using the standard denoising diffusion process for $t = T{-}k, \dots, 1$
\State \Return  $z_0$
\end{algorithmic}
\end{algorithm}

\clearpage

\subsection{Experimental Setup}
\label{sec:appendix-expsetup}

All experiments use \texttt{Wan 2.1} as the backbone model. We apply a classifier-free guidance scale of 6 and a sample shift of 3 during generation.

For single-object scenes, we use $\lambda_1{=}0.3$, $\lambda_2{=}10$, a learning rate of $5{\times}10^{-4}$, a Pearson stopping threshold of $\tau_{\text{corr}}{=}0.9$, and perform optimization over the first $l{=}5$ denoising steps with up to $k{=}10$ gradient updates per step.

For two-object scenes, which demand stronger guidance, we increase the learning rate to $10^{-3}$ to improve convergence. For motion-centric prompts, we relax the Pearson threshold to $\tau_{\text{corr}}{=}0.85$ to enable more flexible optimization.

\subsection{Applications Setup}
\label{sec:appendix-applications}

\paragraph{Single-Object Temporal Control.}
To evaluate temporal grounding for individual objects, we construct a dataset based on the 80 object categories detectable by the YOLO detector. These categories cover common entities such as animals, vehicles, household items, and everyday objects. The complete class list is provided below:

\begin{quote}
\small
\begin{tabular}{p{0.95\linewidth}}
person, bicycle, car, motorcycle, airplane, bus, train, truck, boat, traffic light, fire hydrant, stop sign, parking meter, bench, bird, cat, dog, horse, sheep, cow, elephant, bear, zebra, giraffe, backpack, umbrella, handbag, tie, suitcase, frisbee, skis, snowboard, sports ball, kite, baseball bat, baseball glove, skateboard, surfboard, tennis racket, bottle, wine glass, cup, fork, knife, spoon, bowl, banana, apple, sandwich, orange, broccoli, carrot, hot dog, pizza, donut, cake, chair, couch, potted plant, bed, dining table, toilet, tv, laptop, mouse, remote, keyboard, cell phone, microwave, oven, toaster, sink, refrigerator, book, clock, vase, scissors, teddy bear, hair drier, toothbrush.
\end{tabular}
\end{quote}

Videos are generated from prompts of the form \textit{`An empty scene. Suddenly, $\langle o_1 \rangle$ appears out of nowhere, drawing all attention.'} Object classes are uniformly sampled across different temporal intervals.

\paragraph{Two-Object Temporal Control.} 
We evaluate \modelname{} on 82 prompts derived from the VBench protocol, each consisting of a pair of objects. The complete numbered list of pairs is provided below:

\begin{quote}
\small
\begin{enumerate}
\item \textbf{a bird} and \textbf{a cat}
\item \textbf{a cat} and \textbf{a dog}
\item \textbf{a dog} and \textbf{a horse}
\item \textbf{a horse} and \textbf{a sheep}
\item \textbf{a sheep} and \textbf{a cow}
\item \textbf{a cow} and \textbf{an elephant}
\item \textbf{an elephant} and \textbf{a bear}
\item \textbf{a bear} and \textbf{a zebra}
\item \textbf{a zebra} and \textbf{a giraffe}
\item \textbf{a giraffe} and \textbf{a bird}
\item \textbf{a chair} and \textbf{a couch}
\item \textbf{a couch} and \textbf{a potted plant}
\item \textbf{a potted plant} and \textbf{a tv}
\item \textbf{a tv} and \textbf{a laptop}
\item \textbf{a laptop} and \textbf{a remote}
\item \textbf{a remote} and \textbf{a keyboard}
\item \textbf{a keyboard} and \textbf{a cell phone}
\item \textbf{a cell phone} and \textbf{a book}
\item \textbf{a book} and \textbf{a clock}
\item \textbf{a clock} and \textbf{a backpack}
\item \textbf{a backpack} and \textbf{an umbrella}
\item \textbf{an umbrella} and \textbf{a handbag}
\item \textbf{a handbag} and \textbf{a tie}
\item \textbf{a tie} and \textbf{a suitcase}
\item \textbf{a suitcase} and \textbf{a vase}
\item \textbf{a vase} and \textbf{scissors}
\item \textbf{scissors} and \textbf{a teddy bear}
\item \textbf{a teddy bear} and \textbf{a frisbee}
\item \textbf{a frisbee} and \textbf{skis}
\item \textbf{skis} and \textbf{a snowboard}
\item \textbf{a snowboard} and \textbf{a sports ball}
\item \textbf{a sports ball} and \textbf{a kite}
\item \textbf{a kite} and \textbf{a baseball bat}
\item \textbf{a baseball bat} and \textbf{a baseball glove}
\item \textbf{a baseball glove} and \textbf{a skateboard}
\item \textbf{a skateboard} and \textbf{a surfboard}
\item \textbf{a surfboard} and \textbf{a tennis racket}
\item \textbf{a tennis racket} and \textbf{a bottle}
\item \textbf{a bottle} and \textbf{a chair}
\item \textbf{an airplane} and \textbf{a train}
\item \textbf{a train} and \textbf{a boat}
\item \textbf{a boat} and \textbf{an airplane}
\item \textbf{a bicycle} and \textbf{a car}
\item \textbf{a car} and \textbf{a motorcycle}
\item \textbf{a motorcycle} and \textbf{a bus}
\item \textbf{a bus} and \textbf{a traffic light}
\item \textbf{a traffic light} and \textbf{a fire hydrant}
\item \textbf{a fire hydrant} and \textbf{a stop sign}
\item \textbf{a stop sign} and \textbf{a parking meter}
\item \textbf{a parking meter} and \textbf{a truck}
\item \textbf{a truck} and \textbf{a bicycle}
\item \textbf{a toilet} and \textbf{a hair drier}
\item \textbf{a hair drier} and \textbf{a toothbrush}
\item \textbf{a toothbrush} and \textbf{a sink}
\item \textbf{a sink} and \textbf{a toilet}
\item \textbf{a wine glass} and \textbf{a chair}
\item \textbf{a cup} and \textbf{a couch}
\item \textbf{a fork} and \textbf{a potted plant}
\item \textbf{a knife} and \textbf{a tv}
\item \textbf{a spoon} and \textbf{a laptop}
\item \textbf{a bowl} and \textbf{a remote}
\item \textbf{a banana} and \textbf{a keyboard}
\item \textbf{an apple} and \textbf{a cell phone}
\item \textbf{a sandwich} and \textbf{a book}
\item \textbf{an orange} and \textbf{a clock}
\item \textbf{broccoli} and \textbf{a backpack}
\item \textbf{a carrot} and \textbf{an umbrella}
\item \textbf{a hot dog} and \textbf{a handbag}
\item \textbf{a pizza} and \textbf{a tie}
\item \textbf{a donut} and \textbf{a suitcase}
\item \textbf{a cake} and \textbf{a vase}
\item \textbf{an oven} and \textbf{scissors}
\item \textbf{a toaster} and \textbf{a teddy bear}
\item \textbf{a microwave} and \textbf{a frisbee}
\item \textbf{a refrigerator} and \textbf{skis}
\item \textbf{a bicycle} and \textbf{an airplane}
\item \textbf{a car} and \textbf{a train}
\item \textbf{a motorcycle} and \textbf{a boat}
\item \textbf{a person} and \textbf{a toilet}
\item \textbf{a person} and \textbf{a hair drier}
\item \textbf{a person} and \textbf{a toothbrush}
\item \textbf{a person} and \textbf{a sink}
\end{enumerate}
\end{quote}

Prompts follow the structure:
\textit{`The video begins with a serene view centered on $\langle o_1 \rangle$, with no sign of $\langle o_2 \rangle$. In the second half, $\langle o_2 \rangle$ unexpectedly appears, altering the dynamic of the scene.'}

A static control signal (all ones) is used for $\langle o_1 \rangle$. We sample 20 evenly spaced frames and require $\langle o_1 \rangle$ to be present throughout. A frame is counted as successful if both objects are detected when $\langle o_2 \rangle$ is active, and only $\langle o_1 \rangle$ is detected when $\langle o_2 \rangle$ is inactive.

\paragraph{Motion Temporal Control.}
To evaluate fine-grained temporal modulation of actions, we generate 100 videos using prompts of the form:
\textit{`A video of $\langle s \rangle$ $\langle v \rangle$ $\langle a \rangle$ with a strong movement at the $\langle t \rangle$ second,'}
where $\langle s \rangle$, $\langle v \rangle$, $\langle a \rangle$, and $\langle t \rangle$ denote the subject, verb, adverb, and time reference, respectively.

The full list of subject–verb–adverb triplets is given below:

\begin{quote}
\small
\begin{enumerate}
\item \textbf{a cat} pouncing \textbf{silently}
\item \textbf{a dancer} twirling \textbf{elegantly}
\item \textbf{a bird} flapping \textbf{frantically}
\item \textbf{a lion} roaring \textbf{loudly}
\item \textbf{a wave} crashing \textbf{violently}
\item \textbf{a ballerina} leaping \textbf{lightly}
\item \textbf{a helicopter} spinning \textbf{noisily}
\item \textbf{a snake} slithering \textbf{smoothly}
\item \textbf{a bear} growling \textbf{deeply}
\item \textbf{a man} running \textbf{fast}
\item \textbf{a woman} laughing \textbf{brightly}
\item \textbf{a child} crawling \textbf{slowly}
\item \textbf{a monkey} swinging \textbf{playfully}
\item \textbf{a firework} exploding \textbf{suddenly}
\item \textbf{a wolf} howling \textbf{hauntingly}
\item \textbf{a dolphin} diving \textbf{swiftly}
\item \textbf{a plane} ascending \textbf{steadily}
\item \textbf{a kangaroo} hopping \textbf{energetically}
\item \textbf{a fox} darting \textbf{quickly}
\item \textbf{a shadow} moving \textbf{mysteriously}
\item \textbf{a squirrel} scampering \textbf{nervously}
\item \textbf{a truck} crashing \textbf{loudly}
\item \textbf{a child} screaming \textbf{unexpectedly}
\item \textbf{a penguin} waddling \textbf{adorably}
\item \textbf{a frog} croaking \textbf{rhythmically}
\item \textbf{a leaf} drifting \textbf{slowly}
\item \textbf{a shark} circling \textbf{menacingly}
\item \textbf{a comet} streaking \textbf{brightly}
\item \textbf{a dancer} collapsing \textbf{dramatically}
\item \textbf{a boxer} punching \textbf{fiercely}
\item \textbf{a tree} swaying \textbf{gently}
\item \textbf{a man} lifting weights \textbf{powerfully}
\item \textbf{a windmill} rotating \textbf{steadily}
\item \textbf{a girl} skipping \textbf{cheerfully}
\item \textbf{a car} drifting \textbf{dangerously}
\item \textbf{a train} accelerating \textbf{fast}
\item \textbf{a snake} striking \textbf{quickly}
\item \textbf{a fire} burning \textbf{intensely}
\item \textbf{a glacier} cracking \textbf{slowly}
\item \textbf{a bee} buzzing \textbf{constantly}
\item \textbf{a deer} sprinting \textbf{fearfully}
\item \textbf{a volcano} erupting \textbf{violently}
\item \textbf{a runner} collapsing \textbf{from exhaustion}
\item \textbf{a gymnast} flipping \textbf{smoothly}
\item \textbf{a rocket} launching \textbf{thunderously}
\item \textbf{a drummer} hitting \textbf{rapidly}
\item \textbf{a magician} vanishing \textbf{mysteriously}
\item \textbf{a spider} crawling \textbf{delicately}
\item \textbf{a tiger} pacing \textbf{restlessly}
\item \textbf{a surfer} balancing \textbf{skillfully}
\item \textbf{a leopard} growling \textbf{softly}
\item \textbf{a swimmer} diving \textbf{gracefully}
\item \textbf{a baby} clapping \textbf{happily}
\item \textbf{a mime} gesturing \textbf{expressively}
\item \textbf{a crow} cawing \textbf{sharply}
\item \textbf{a goat} headbutting \textbf{suddenly}
\item \textbf{a girl} blowing bubbles \textbf{gently}
\item \textbf{a chef} chopping \textbf{rapidly}
\item \textbf{a horse} shaking its mane \textbf{proudly}
\item \textbf{a robot} malfunctioning \textbf{erratically}
\item \textbf{a meteor} falling \textbf{fast}
\item \textbf{a snake} coiling \textbf{tightly}
\item \textbf{a rabbit} thumping \textbf{nervously}
\item \textbf{a man} collapsing \textbf{dramatically}
\item \textbf{a car} braking \textbf{suddenly}
\item \textbf{a hawk} diving \textbf{precisely}
\item \textbf{a wolf} snarling \textbf{viciously}
\item \textbf{a plane} landing \textbf{smoothly}
\item \textbf{a firetruck} speeding \textbf{urgently}
\item \textbf{a chimpanzee} clapping \textbf{playfully}
\item \textbf{a dancer} stomping \textbf{rhythmically}
\item \textbf{a snail} inching \textbf{slowly}
\item \textbf{a lioness} crouching \textbf{quietly}
\item \textbf{a storm cloud} rolling \textbf{ominously}
\item \textbf{a bee} landing \textbf{precisely}
\item \textbf{a magician} pulling a rabbit \textbf{suddenly}
\item \textbf{a glacier} melting \textbf{steadily}
\item \textbf{a jellyfish} floating \textbf{gracefully}
\item \textbf{a tree} falling \textbf{loudly}
\item \textbf{a drummer} drumming \textbf{fiercely}
\item \textbf{a torch} blazing \textbf{brightly}
\item \textbf{a woman} spinning \textbf{dramatically}
\item \textbf{a child} running \textbf{barefoot}
\item \textbf{a fox} sniffing \textbf{cautiously}
\item \textbf{a man} throwing a ball \textbf{forcefully}
\item \textbf{a bear} shaking off water \textbf{heavily}
\item \textbf{a lion} walking \textbf{regally}
\item \textbf{a horse} rearing \textbf{suddenly}
\item \textbf{a falcon} gliding \textbf{silently}
\item \textbf{an elephant} raising its trunk \textbf{highly}
\item \textbf{a dog} barking \textbf{forcefully}
\item \textbf{a man} sneezing \textbf{powerfully}
\item \textbf{a child} jumping \textbf{joyfully}
\item \textbf{a tiger} leaping \textbf{fiercely}
\item \textbf{a woman} spinning \textbf{gracefully}
\item \textbf{a robot} marching \textbf{stiffly}
\item \textbf{a horse} galloping \textbf{wildly}
\item \textbf{an eagle} soaring \textbf{majestically}
\item \textbf{a flame} flickering \textbf{rapidly}
\item \textbf{a skateboarder} flipping \textbf{expertly}
\end{enumerate}
\end{quote}

The verb is modulated using a binary temporal mask (set to 1 only at time $\langle t \rangle$), while the subject remains active throughout the video.

\clearpage

\subsection{Human Evaluation Protocol}
\label{sec:appendix-human-eval}
To assess temporal alignment and visual quality, we conducted a human preference study using side-by-side video comparisons. All annotators were graduate students conducting research in computer vision or closely related areas.

For each prompt, two videos were shown: one generated by our method and one by the text-only baseline, with their order (Video~A or Video~B) randomized to avoid bias.

Annotators were instructed to read the prompt and answer two questions:  
(1) Which video is more temporally accurate (i.e., does the object appear at the correct time and remain visible)?  
(2) Which video is more visually appealing?  
A ``Same (use sparingly)'' option was available when no meaningful difference was perceived.

After the experiments, several annotators noted that judging temporal control could be challenging when the difference in timing was only 1--2 seconds. They reported that knowing the exact expected trigger time from the control mask would have helped them assess temporal accuracy more reliably. While we chose not to repeat the study, we include this observation as guidance for future temporal evaluation protocols.

\begin{figure}[h]
\centering
\includegraphics[width=0.5\linewidth]{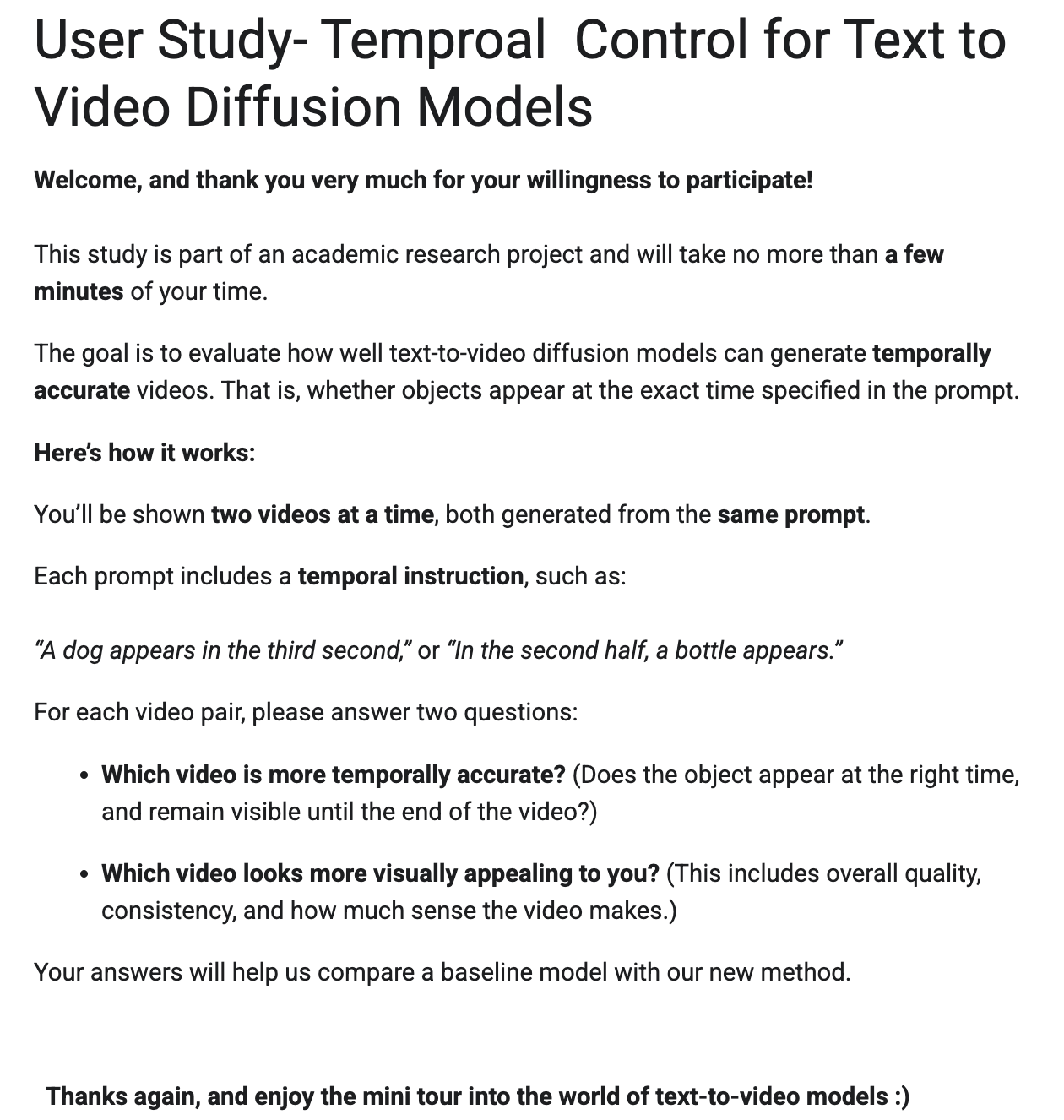}
\caption{Evaluation interface showing the prompt and two videos (A and B) for comparison.}
\label{fig:ui-eval1}
\end{figure}

\begin{figure}[h]
\centering
\includegraphics[width=0.5\linewidth]{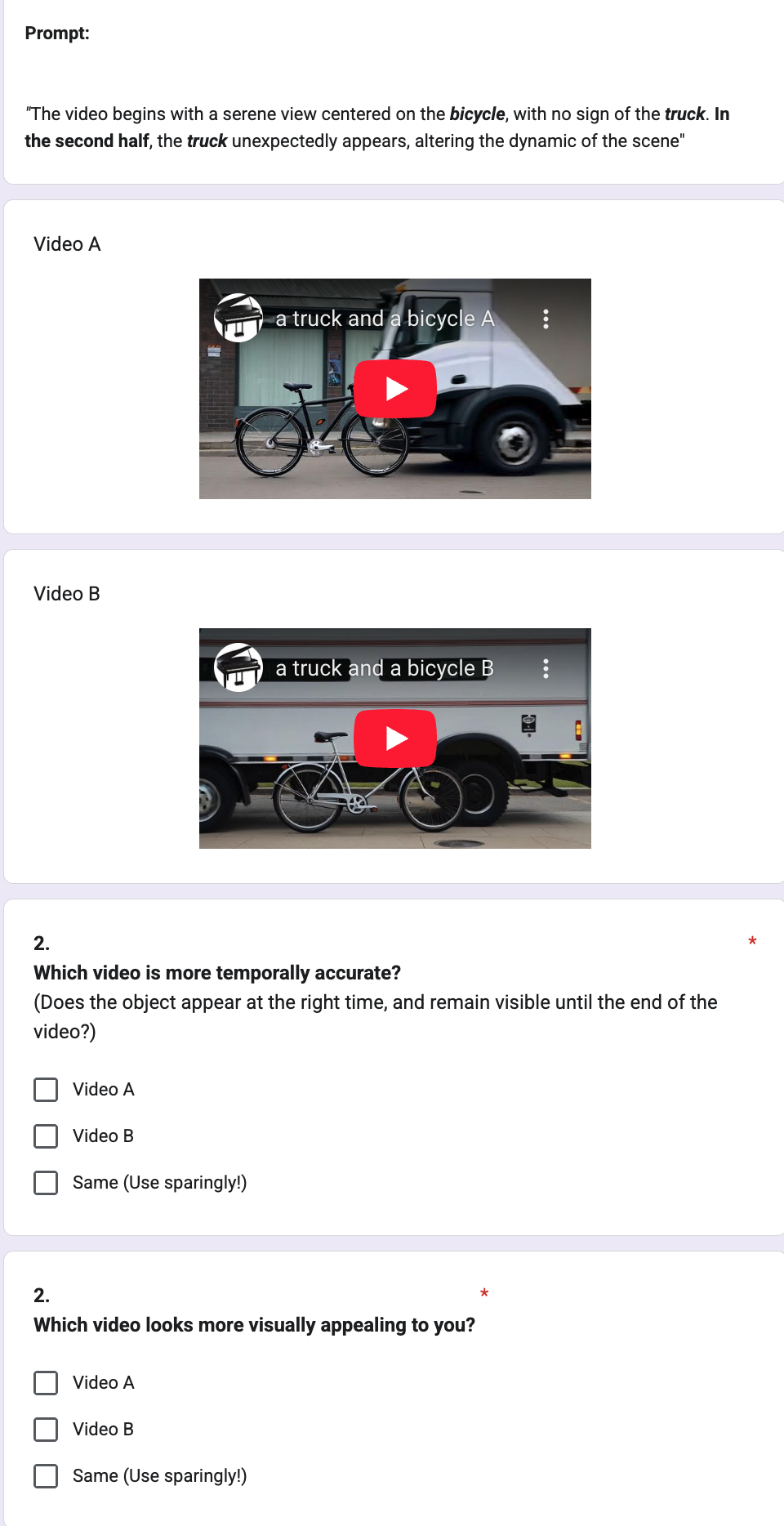}
\caption{Second example of the human evaluation interface.}
\label{fig:ui-eval2}
\end{figure}

Example screenshots of the evaluation interface are shown in Figures~\ref{fig:ui-eval1} and~\ref{fig:ui-eval2}.

\clearpage

\subsection{Latency Discussion}
\label{sec:latency_ablation}

\begin{figure}[h]
    \centering
    \includegraphics[width=\linewidth]{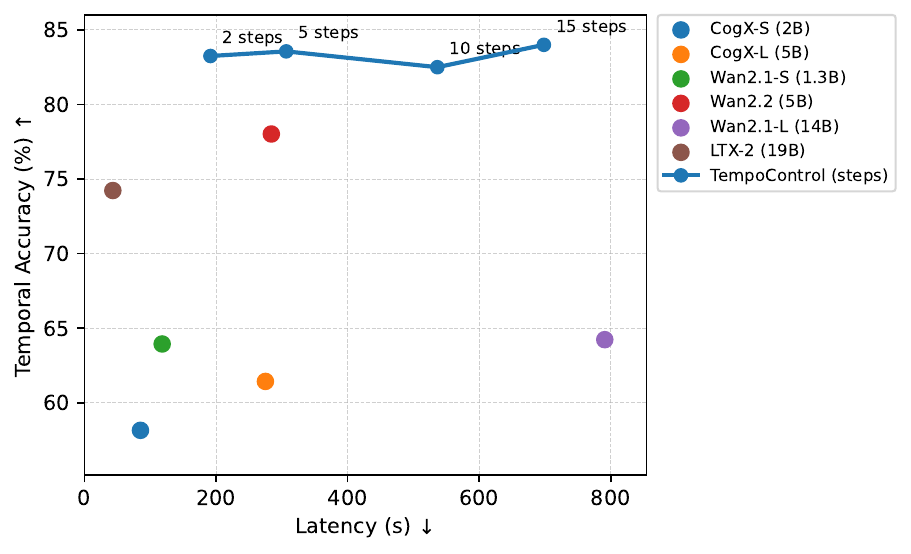}
    \vspace{-8pt}
    \caption{
    Latency vs.\ temporal accuracy. Each baseline is shown as a single point, while \modelname forms a tradeoff curve as the number of inference steps increases. Labels denote the number of inference steps.
    }
    \label{fig:latency_tradeoff}
    \vspace{-10pt}
\end{figure}

We analyze the tradeoff between inference latency and temporal control by varying the number of inference steps in \modelname.
Figure~\ref{fig:latency_tradeoff} visualizes the latency--accuracy frontier, while Table~\ref{tab:latency_ablation} reports the corresponding temporal, absence, and presence accuracies.

As shown, \modelname consistently outperforms all baselines in comparable latency regimes.
Even with only 2 inference steps, \modelname achieves 83.25\% temporal accuracy, surpassing all baselines while operating within a similar latency range to mid-size models.
Increasing the number of steps further improves performance, with 5 steps providing the best overall balance between latency and accuracy (83.56\% temporal accuracy).
Beyond this point, the gains are marginal, with the performance largely saturating after 5--10 steps.

Importantly, \modelname achieves superior temporal control while using a significantly smaller backbone (1.3B parameters), compared to baselines that scale up to 19B parameters.
This demonstrates that strong temporal reasoning can be achieved efficiently without requiring large-scale models.

\begin{table}[h]
\centering
\setlength{\tabcolsep}{4pt}
\renewcommand{\arraystretch}{0.97}
\caption{
Latency--performance tradeoff. We compare \modelname (Wan 1.3B) under different inference steps against representative baselines.
}
\label{tab:latency_ablation}
\begin{tabular}{lccccc}
\toprule
Method & Latency (s) $\downarrow$ & Temp $\uparrow$ & Abs $\uparrow$ & Pres $\uparrow$ \\
\midrule
CogX (2B)        & $\sim85$  & 58.15 & 74.77 & 39.88 \\
Wan2.1-S (1.3B)  & $\sim118$ & 63.94 & 67.38 & 60.50 \\
Wan2.2 (5B)      & $\sim284$ & 78.02 & \textbf{87.42} & 68.00 \\
Wan2.1-L (14B)   & $\sim791$ & 64.23 & 55.34 & 74.00 \\
LTX-2 (19B)      & $\sim43$  & 74.23 & 85.23 & 62.12 \\
\midrule
Ours (2 steps)   & $\sim191$ & 83.25 & 86.12 & 80.38 \\
Ours (5 steps)   & $\sim307$ & \textbf{83.56} & 87.38 & 79.75 \\
Ours (10 steps)  & $\sim536$ & 82.50 & 83.25 & \textbf{81.75} \\
Ours (15 steps)  & $\sim698$ & 84.00 & 86.38 & 81.62 \\
\bottomrule
\end{tabular}
\vspace{-8pt}
\end{table}

\clearpage
\subsection{Dense Diffusion vs.\ TempoControl}
\label{sec:appendix_dense}

We compare TempoControl to Dense Diffusion~\cite{densediffusion}, which applies direct spatiotemporal masking $R_i(t,x,y)$ to cross-attention maps based on a temporal control signal, without explicitly optimizing temporal alignment.

While this enables temporal gating, it does not align the attention distribution $A_i(t)$ with the desired pattern $m_{i,t}$. Moreover, the lack of regularization leads to spatially incoherent attention, producing visual artifacts. These artifacts degrade object appearance and hinder detection, resulting in artificially inflated absence accuracy rather than correct temporal control.

In contrast, TempoControl explicitly optimizes temporal alignment via correlation, controls activation strength via a magnitude term, and enforces spatial coherence through entropy regularization. This yields temporally aligned and visually consistent generations, improving presence accuracy while maintaining image quality.

\begin{table}[h]
\centering
\begin{tabular}{lcccc}
\toprule
Method & Temp. Acc. & Abs. Acc. & Pres. Acc. & Img. Qual. \\
\midrule
Dense diffusion & 77.94 & \textbf{86.25} & 69.62 & 51.02 \\
TempoControl & \textbf{82.50} & 83.25 & \textbf{81.75} & \textbf{56.51} \\
\bottomrule
\end{tabular}
\caption{Comparison between Dense Diffusion and TempoControl with Wan2.1 for the one object setting..}
\end{table}

\clearpage
\subsection{Optimization Steps and Update Frequency}
\label{sec:appendix_steps}

We analyze the effect of optimization depth for \modelname. We first ablate the number of gradient updates per step in the one-object setting (Table~\ref{tab:updates}). Based on this, we fix the number of updates to 5 and vary the number of optimized diffusion steps across tasks and backbones (Table~\ref{tab:steps}).

Increasing the number of gradient updates improves temporal alignment, but fewer than five updates per step leads to under-optimized attention, while additional updates yield diminishing returns.

The impact of optimizing different diffusion steps depends on when temporal structure is established in the underlying model. In Wan-based one- and two-object settings, performance saturates after 5 steps, as object appearance is determined early in the denoising process. In contrast, movement control and CogVideoX benefit from additional steps, since temporal structure emerges later. Overall, optimizing 5 steps provides a strong trade-off between performance and latency.

\begin{table}[h]
\centering
\setlength{\tabcolsep}{4.5pt}
\renewcommand{\arraystretch}{0.97}
\caption{
Effect of gradient updates per step in the one-object setting.
Bold denotes the best result.
}
\label{tab:updates}
\begin{tabular}{lcccc}
\toprule
Setting & Acc $\uparrow$ & Abs $\uparrow$ & Pres $\uparrow$ & Img $\uparrow$ \\
\midrule
2 steps / 2 updates & 72.00\% & 74.50\% & 69.50\% & 57.87\% \\
2 steps / 5 updates & 77.75\% & 79.00\% & 76.50\% & 57.41\% \\
5 steps / 2 updates & 74.12\% & 73.62\% & 74.62\% &\textbf{ 58.03\%} \\
5 steps / 5 updates & 78.81\% & 81.50\% & 76.12\% & 57.43\%  \\
5 steps / 10 updates (Ours) & \textbf{83.56\%} & \textbf{87.38\%} & \textbf{79.75\%} & 56.92\% \\
\bottomrule
\end{tabular}
\end{table}

\begin{table}[h]
\centering
\setlength{\tabcolsep}{4.5pt}
\renewcommand{\arraystretch}{0.97}
\caption{
Effect of optimized diffusion steps across tasks and backbones (5 updates).
Bold denotes the best result within each block.
}
\label{tab:steps}
\begin{tabular}{lcccc}
\toprule
Setting & Acc $\uparrow$ & Abs $\uparrow$ & Pres $\uparrow$ & Img $\uparrow$ \\
\midrule

\multicolumn{5}{c}{\textbf{One Object (Wan)}} \\
\midrule
2 steps & 83.25\% & 86.12\% & 80.38\% & 56.72\% \\
5 steps (Ours) & \textbf{83.56\%} & \textbf{87.38\%} & 79.75\% & \textbf{56.92\%} \\
10 steps & 82.50\% & 83.25\% & \textbf{81.75\%} & 56.51\% \\

\midrule
\multicolumn{5}{c}{\textbf{One Object (CogVideoX)}} \\
\midrule
2 steps & 60.12\% & 78.07\% & 40.38\% & 46.35\% \\
5 steps (Ours) & 63.75\% & \textbf{78.41\%} & 47.62\% & 46.67\% \\
10 steps & \textbf{65.48\%} & 77.27\% & \textbf{52.50\%} & \textbf{47.88\%} \\

\midrule
\multicolumn{5}{c}{\textbf{Two Objects (Wan)}} \\
\midrule
2 steps & 50.98\% & \textbf{59.88\%} & 42.07\% & 70.61\% \\
5 steps (Ours) & \textbf{53.17\%} & 57.32\% & \textbf{49.02\%} & 70.82\% \\
10 steps & 52.93\% & 58.05\% & 47.80\% & \textbf{71.09\%} \\

\midrule
\multicolumn{5}{c}{\textbf{Movement (Wan)}} \\
\midrule
2 steps & 38\% & -- & -- & 62.85\% \\
5 steps (Ours) & 54\% & -- & -- & 63.24\% \\
10 steps & \textbf{57\%} & -- & -- & \textbf{63.38\%} \\

\bottomrule
\end{tabular}
\end{table}

\clearpage
\subsection{Multi-Object Evaluation}
\label{sec:appendix-multitoken}
We report extended results on multi-object text-to-video generation using the VBench benchmark. In this setting, prompts reference multiple entities (e.g., `A dog and a cat'), and the evaluation measures whether all specified objects appear consistently throughout the video. Each video is sampled at 16 evenly spaced frames, and a frame is counted as successful if both target objects are detected; the final score is the average per-frame success rate. In the context of our method, this setup corresponds to applying a constant mask of one for both objects across all frames.

For completeness, we briefly summarize the VBench metrics reported in Table~\ref{tab:full-multipleobjects}:
\begin{itemize}
    \item \textbf{Multiple-Objects Accuracy (GriT / YOLO).} Measures whether all referenced objects appear correctly in each sampled frame, using open-vocabulary detectors (GriT) or standard detectors (YOLO). Higher values indicate more reliable object visibility and grounding.
    \item \textbf{Subject Consistency.} Quantifies how consistently the appearance of the main subject is preserved across frames. It is computed using feature similarity both between consecutive frames and relative to the first frame of the video; higher values indicate a more stable and coherent subject identity.

    \item \textbf{Background Consistency.} Measures the stability of the background across the video. High scores indicate coherent scene structure without abrupt changes.
\item \textbf{Dynamic Degree.} Captures the amount of motion or temporal change in the video. The metric is binary at the video level: a video is marked as dynamic if its average optical flow surpasses a predefined threshold. Higher values indicate more motion, while lower values reflect greater temporal stability.

\end{itemize}

Table~\ref{tab:full-multipleobjects} summarizes the results. Our method achieves substantial gains in object-centric metrics: multi-object accuracy improves from 74.1\% to 76.4\% (GriT) and from 61.5\% to 65.7\% (YOLO), demonstrating better grounding and more reliable visibility of both entities. We also observe improvements across consistency metrics, including subject consistency (from 97.2\% to 97.8\%) and background consistency (from 97.6\% to 98.1\%).

The Dynamic Degree decreases from 30.5\% to 18.8\%, reflecting a trade-off: our method promotes temporal stability, often leading to more consistent object appearance across the video, at the cost of reduced motion. This highlights a limitation of this evaluation protocol: VBench counts an object as correct only when it is present in \emph{every} sampled frame; hence, the benchmark implicitly favors unnatural videos in which objects persist throughout the sequence.

\begin{table*}[h]
\caption{Full results for Multiple object benchmark results. Best results per column are in bold.}
\label{tab:full-multipleobjects}
\centering
\begin{adjustbox}{width=0.9\textwidth}
\begin{tabular}{lcccccccc}
\toprule
Method &
\makecell{Multiple\\Object (GriT)} & 
\makecell{Multiple\\Object (YOLO)} &
\makecell{Subject\\Consistency} & 
\makecell{Background\\Consistency} & 
\makecell{Motion\\Smoothness} & 
\makecell{Dynamic\\Degree} & 
\makecell{Aesthetic\\Quality} & 
\makecell{Imaging\\Quality} \\
\midrule
Text  & 74.13\% & 61.54\% & 97.22\% & 97.56\% & 99.18\% & \textbf{30.49\%} & \textbf{62.84}\% & \textbf{70.25}\%\\
Ours  & \textbf{76.37\%} & \textbf{65.73\%} & \textbf{97.81\%} & \textbf{98.10\%} & \textbf{99.40\%} & 18.78\% & 62.52\% & 70.21\%\\
\bottomrule
\end{tabular}
\end{adjustbox}
\end{table*}

\clearpage

\subsection{Loss Component Ablation}
\label{sec:appendix-ablations}

We conduct an ablation study to analyze the influence of the magnitude weight~\(\lambda_1\) and the entropy weight~\(\lambda_2\) on temporal control and visual quality. The results reveal clear and complementary effects of the two loss components. Increasing the magnitude term~\(\lambda_1\) reduces Imaging Quality while improving T. Absence Accuracy. This suggests that stronger magnitude guidance more strictly enforces the temporal mask, but may also reduce overall visual fidelity.

In contrast, increasing the entropy term~\(\lambda_2\) improves Imaging Quality and raises T. Presence Accuracy, but it decreases T. Absence Accuracy, indicating that stronger entropy regularization encourages more persistent object appearance.

Overall, our chosen hyperparameters (\(\lambda_1 = 0.3\), \(\lambda_2 = 10\)) provide a balanced tradeoff between these competing effects, achieving strong Temporal Accuracy while maintaining high Imaging Quality.

\begin{table*}[h]
\caption{Ablation results with Wan2.1-S for the one object setting. All variants use the same temporal Pearson term; we ablate the magnitude weight \(\lambda_1\) and the entropy weight \(\lambda_2\). Best scores per column are in \textbf{bold}.}
\label{tab:full-ablations}
\centering

\begin{subtable}{\textwidth}
\centering
\caption{Effect of magnitude weight \(\lambda_1\) with fixed entropy weight \(\lambda_2 = 10\).}
\label{tab:ablations-magnitude}

\begin{tabular}{lcccc}
\toprule
Method &
\makecell{Temporal\\Accuracy} & 
\makecell{T. Absence\\Accuracy} &
\makecell{T. Presence\\Accuracy} &
\makecell{Imaging\\Quality} \\
\midrule
Text baseline & 63.94\% & 67.38\% & 60.50\% & 53.76\% \\ 

Ours (\(\lambda_1{=}0.3, \lambda_2{=}10\))  
& \textbf{82.50\%} & 83.25\% & \textbf{81.75\%} & \textbf{56.51\%} \\

\midrule
\(\lambda_1{=}0.1, \lambda_2{=}10\)  
& 79.50\% & 80.62\% & 78.38\% & 56.37\% \\

\(\lambda_1{=}0.65, \lambda_2{=}10\) 
& 81.06\% & 86.50\% & 75.62\% & 53.94\% \\

\(\lambda_1{=}1.0, \lambda_2{=}10\) 
& 83.12\% & \textbf{89.00\%} & 77.25\% & 53.74\% \\

\bottomrule
\end{tabular}

\end{subtable}

\vspace{0.75em}

\begin{subtable}{\textwidth}
\centering
\caption{Effect of entropy weight \(\lambda_2\) with fixed magnitude weight \(\lambda_1 = 0.3\).}
\label{tab:ablations-entropy}

\begin{tabular}{lcccc}
\toprule
Method &
\makecell{Temporal\\Accuracy} & 
\makecell{T. Absence\\Accuracy} &
\makecell{T. Presence\\Accuracy} &
\makecell{Imaging\\Quality} \\
\midrule
Text baseline & 63.94\% & 67.38\% & 60.50\% & 53.76\% \\ 

Ours (\(\lambda_1{=}0.3, \lambda_2{=}10\))  
& \textbf{82.50\%} & 83.25\% & \textbf{81.75\%} & \textbf{56.51\%} \\

\midrule
\(\lambda_1{=}0.3, \lambda_2{=}1\)  
& 80.38\% & \textbf{91.25\%} & 69.50\% & 52.01\% \\

\(\lambda_1{=}0.3, \lambda_2{=}5\)  
& 81.88\% & 88.50\% & 75.25\% & 54.04\% \\

\(\lambda_1{=}0.3, \lambda_2{=}15\) 
& 81.56\% & 81.62\% & 81.50\% & 56.31\% \\

\bottomrule
\end{tabular}

\end{subtable}

\end{table*}

\clearpage

\subsection{Cosine Similarity vs.\ TempoControl}
\label{sec:appendix_cosine}

We evaluate a variant in which the pearson correlation term is replaced with cosine similarity, a standard measure of alignment between attention patterns.

Cosine similarity jointly captures alignment and magnitude, effectively entangling temporal synchronization with activation strength. This coupling allows improvements in the objective without necessarily correcting temporal misalignment, since increased activation can artificially boost similarity. In contrast, TempoControl explicitly separates correlation (for alignment) and magnitude (for visibility), enabling more precise and targeted optimization. This decoupling improves temporal accuracy and presence. 

\begin{table}[h]
\centering
\begin{tabular}{lcccc}
\toprule
Method & Temp. Acc. & Abs. Acc. & Pres. Acc. & Img. Qual. \\
\midrule
Cosine similarity & 79.00 & 82.38 & 75.62 & 55.97 \\
TempoControl & \textbf{82.50} & \textbf{83.25} & \textbf{81.75} & \textbf{56.51} \\
\bottomrule
\end{tabular}
\caption{Comparison between cosine similarity objective and TempoControl with Wan2.1-S for the one object setting.}
\end{table}

\clearpage

\begin{figure}[h]
    \centering
    \includegraphics[width=1\linewidth]{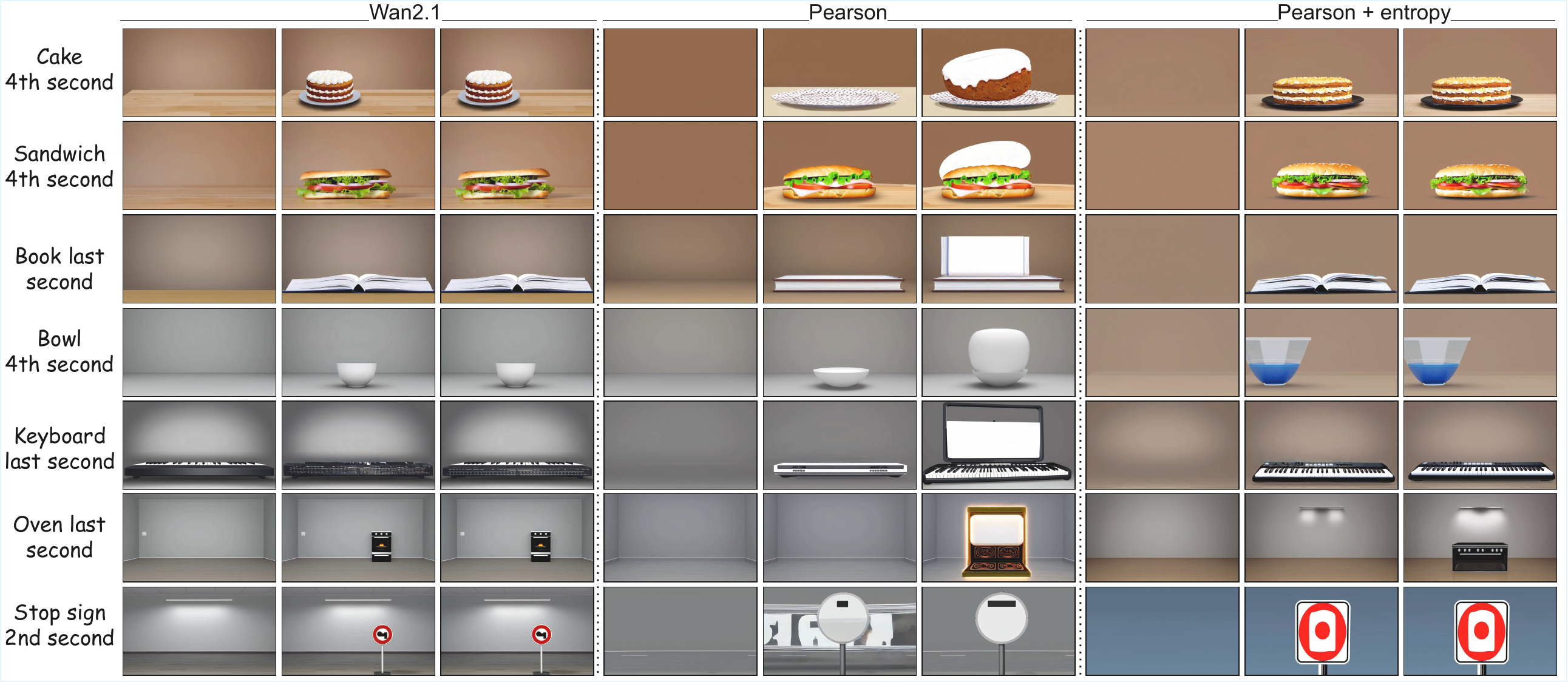}
    \caption{Comparison between Wan2.1 and Pearson-only optimization for single-object temporal control. Pearson correlation alone often satisfies the temporal constraint but alters object semantics, producing distorted or inconsistent appearances (e.g., the cake becoming misshaped, the sandwich warping, the book changing structure, or the keyboard turning into a laptop). Adding entropy regularization prevents these failures by stabilizing the object's identity across time.}
    \label{fig:entropy-qual}
\end{figure}

Figure~\ref{fig:entropy-qual} illustrates ablation results for the entropy term, revealing the limitation of using the Pearson correlation term alone without entropy regularization. While Pearson-based optimization successfully forces the object to appear at the desired time, it often alters the object's semantics, leading to noticeable inconsistencies. For example, in the ``Cake'' row, the Pearson-only variant transforms the cake into a noticeably different shape and texture; in the ``Sandwich'' row, the sandwich becomes distorted across frames; and in the ``Book'' row, the open book becomes a closed stack with altered proportions. Similar semantic drift is observed in the ``Keyboard,'' ``Oven,'' and ``Stop sign'' examples.

These results highlight that Pearson correlation alone can enforce temporal timing but does not sufficiently constrain the appearance space, resulting in unwanted semantic changes. The entropy regularization helps mitigate those artifacts.

\clearpage
\subsection{Limitations of Explicit Temporal Cues in State-of-the-Art Text-to-Video Models}
\label{sec:appendix-time-prompts}
We analyze the effect of using explicit temporal phrasing (e.g., `in the third second', `in the second part of the video') in the prompt for the one-object setup. Table~\ref{tab:comparetime} compares Wan~2.1, Wan~2.2, and our method, each evaluated with and without such phrasing.

For Wan~2.1, explicit temporal cues do not improve temporal grounding. In fact, removing them increases Temporal Accuracy (63.94\% to 65.18\%) and leads to a clear gain in Imaging Quality (53.76\% to 59.99\%), indicating that the model struggles to interpret precise timing expressions and performs better without them.

Wan~2.2 shows a more nuanced pattern: it handles temporal phrasing better than Wan~2.1, likely due to improved data or training techniques. However, the drop in quality when explicit timing is introduced is still \emph{significant}. Imaging Quality decreases sharply from 57.78\% (without time) to 48.13\% (with time), showing that temporal phrasing introduces confusion even for stronger models. Presence Accuracy also falls (79.50\% to 68\%), further highlighting that explicit timing signals continue to degrade performance.

Our method remains the most robust and accurate across settings. Across all conditions, our method provides the most reliable and precise temporal control, regardless of prompt style. Interestingly, although our approach builds on Wan~2.1 as the base model, explicit temporal phrasing does increase the temporal accuracy. This suggests that our optimization procedure may curate the base model's timing capabilities.

\begin{table*}[h]
\caption{Effect of temporal phrasing in prompts for the one-object setup. Removing explicit timing improves video quality.}
\label{tab:comparetime}
\centering
\begin{tabular}{lcccc}
\toprule
\textbf{Method} &
\textbf{Temp. Acc.} & 
\textbf{T. Abs. Acc.} &
\textbf{T. Pres. Acc.} &
\textbf{Imaging Quality} \\
\midrule
Wan2.1 with time    & 63.94\% & \textbf{67.38\%} & 60.50\% & 53.76\% \\ 
Wan2.1 without time  & \textbf{65.18\%} & 65.12\% & \textbf{65.25\%} & \textbf{59.99\%} \\
\midrule
Wan2.2 with time & \textbf{78.02\%} & \textbf{87.42\%} & 68.00\% & 48.13\% \\
Wan2.2 without time & 75.24\% & 71.25\% & \textbf{79.50\%} & \textbf{57.78\%} \\
\midrule
Ours with time & \textbf{82.50\%} & \textbf{83.25\%} & \textbf{81.75\%} & 56.51\%  \\
Ours without time & 78.75\% & 82.25\% & 75.25\% & \textbf{57.14\%} \\
\bottomrule
\end{tabular}
\end{table*}

\clearpage

\subsection{Additional Qualitative Examples}
\label{sec:appendix-qualitative}

We provide qualitative results across the one-object, two-object, and motion-based setups. For each example, we show sampled frames from the generated videos, illustrating the temporal grounding of object appearances. Prompts are abbreviated for brevity, indicating the key object and its expected time of appearance. Note that once introduced, the object is expected to persist until the end of the video. \textbf{Videos can be viewed on the project page.}

\begin{figure*}[t]
\centering
\begin{minipage}{\textwidth}
\begin{minipage}[c]{2cm}\vspace*{0pt}\vfill\raggedright\textit{\small a baseball glove and a skateboard}\vfill\end{minipage}
\includegraphics[width=0.1\linewidth]{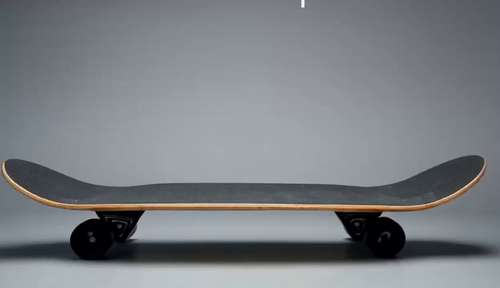}
\includegraphics[width=0.1\linewidth]{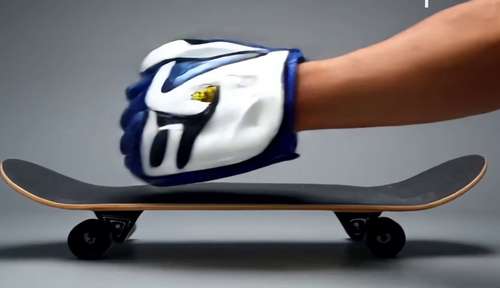}
\includegraphics[width=0.1\linewidth]{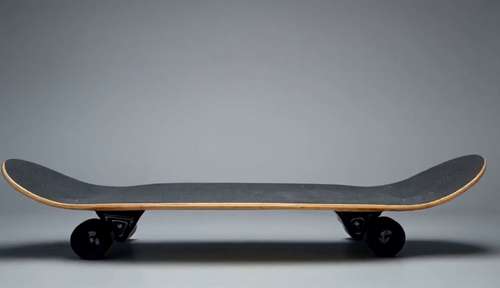}
\includegraphics[width=0.1\linewidth]{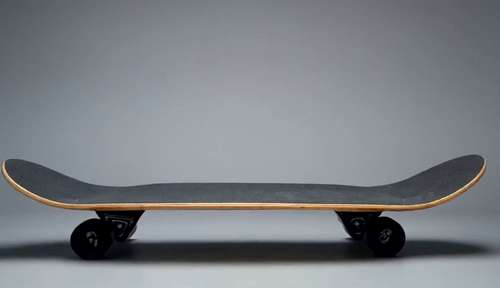}
\hspace{2mm}
\includegraphics[width=0.1\linewidth]{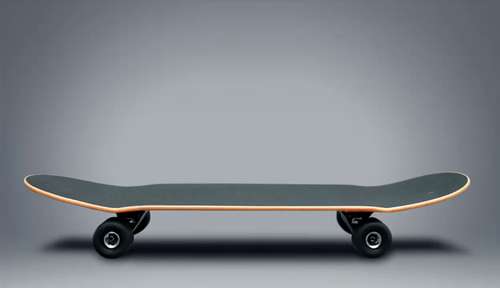}
\includegraphics[width=0.1\linewidth]{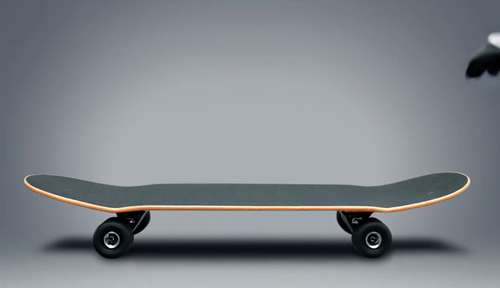}
\includegraphics[width=0.1\linewidth]{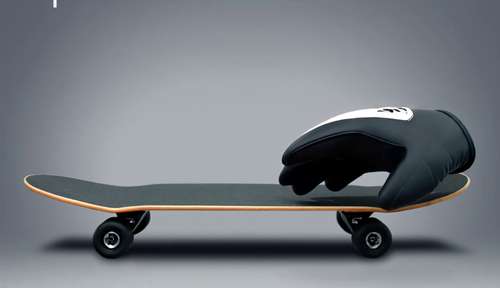}
\includegraphics[width=0.1\linewidth]{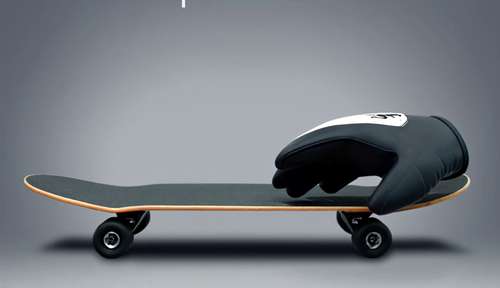}
\\[1.5ex]
\begin{minipage}[c]{2cm}\vspace*{0pt}\vfill\raggedright\textit{\small a bicycle and a car}\vfill\end{minipage}
\includegraphics[width=0.1\linewidth]{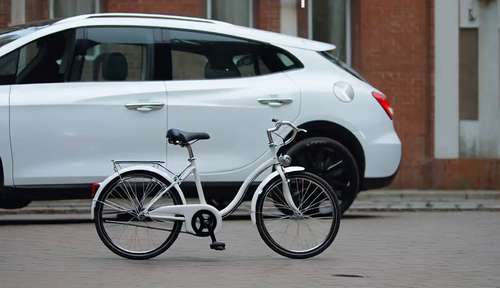}
\includegraphics[width=0.1\linewidth]{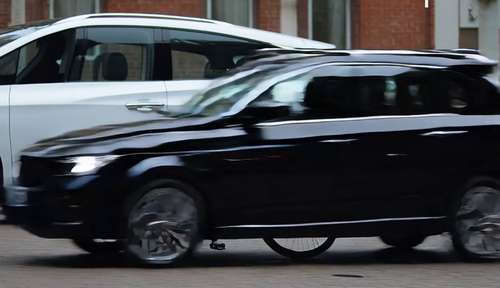}
\includegraphics[width=0.1\linewidth]{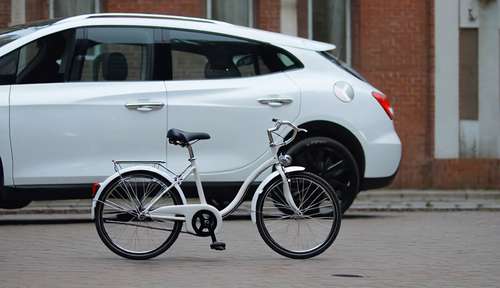}
\includegraphics[width=0.1\linewidth]{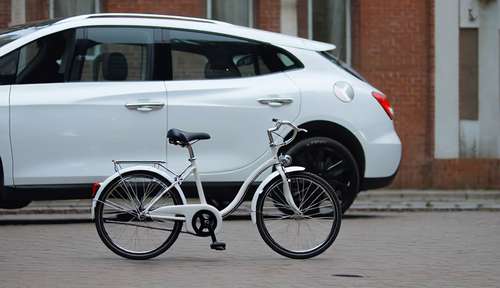}
\hspace{2mm}
\includegraphics[width=0.1\linewidth]{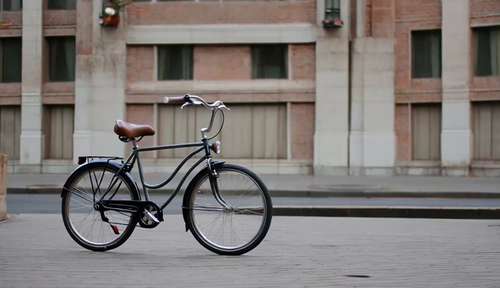}
\includegraphics[width=0.1\linewidth]{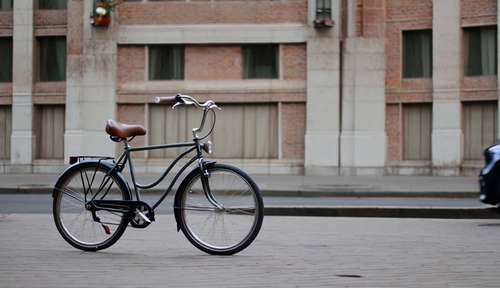}
\includegraphics[width=0.1\linewidth]{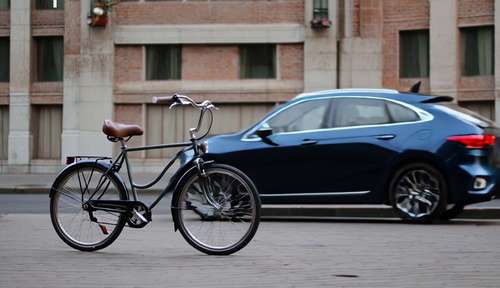}
\includegraphics[width=0.1\linewidth]{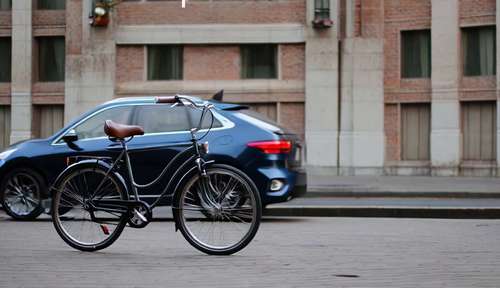}
\\[1.5ex]
\begin{minipage}[c]{2cm}\vspace*{0pt}\vfill\raggedright\textit{\small a bicycle and an airplane}\vfill\end{minipage}
\includegraphics[width=0.1\linewidth]{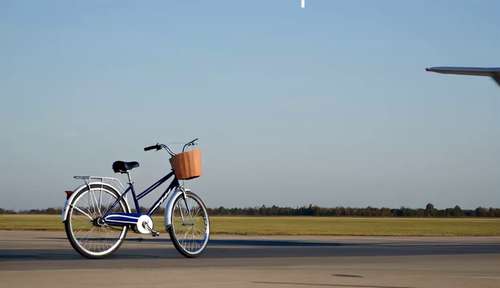}
\includegraphics[width=0.1\linewidth]{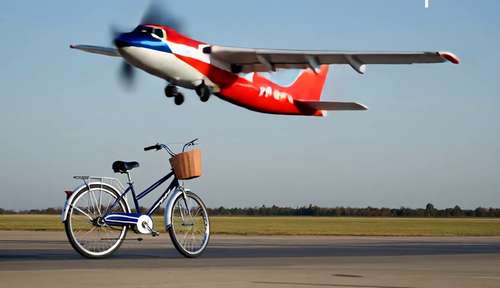}
\includegraphics[width=0.1\linewidth]{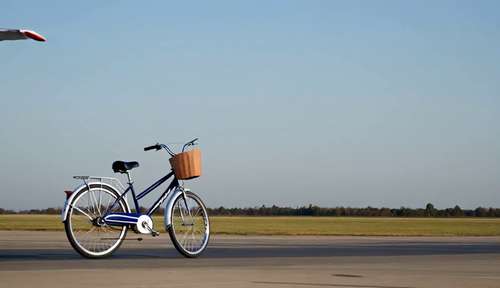}
\includegraphics[width=0.1\linewidth]{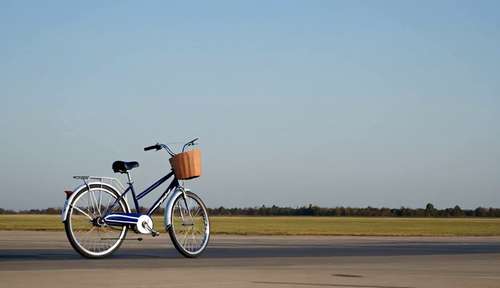}
\hspace{2mm}
\includegraphics[width=0.1\linewidth]{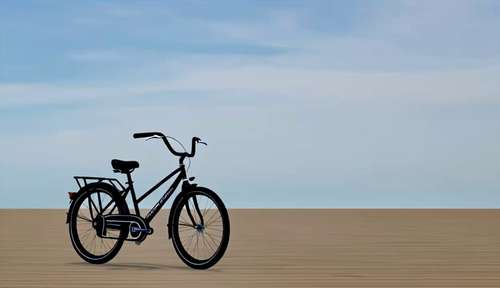}
\includegraphics[width=0.1\linewidth]{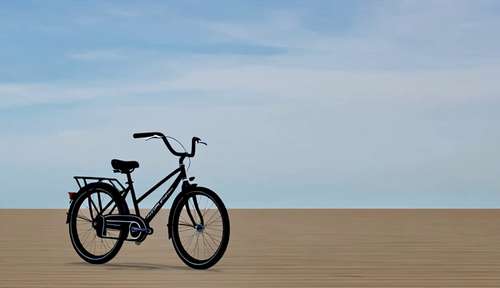}
\includegraphics[width=0.1\linewidth]{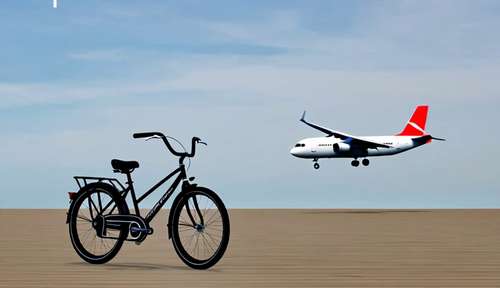}
\includegraphics[width=0.1\linewidth]{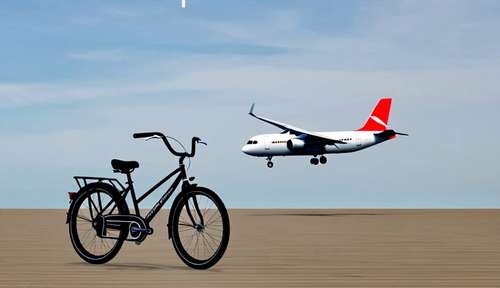}
\\[1.5ex]
\begin{minipage}[c]{2cm}\vspace*{0pt}\vfill\raggedright\textit{\small a bird and a cat}\vfill\end{minipage}
\includegraphics[width=0.1\linewidth]{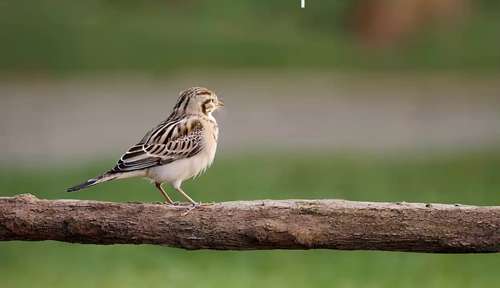}
\includegraphics[width=0.1\linewidth]{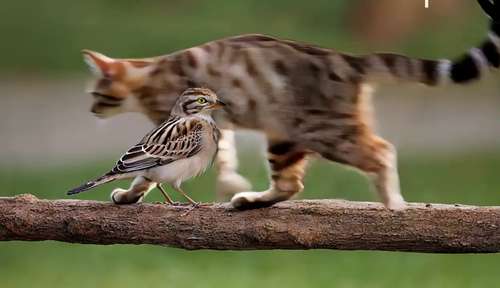}
\includegraphics[width=0.1\linewidth]{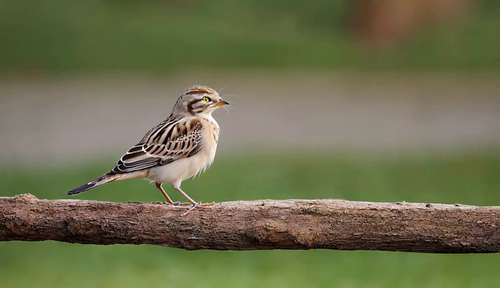}
\includegraphics[width=0.1\linewidth]{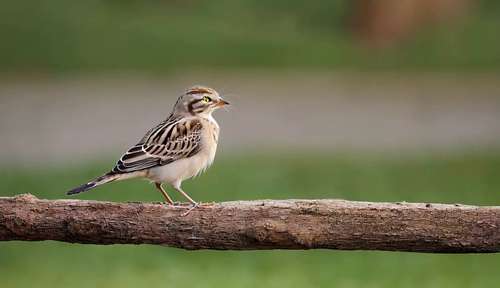}
\hspace{2mm}
\includegraphics[width=0.1\linewidth]{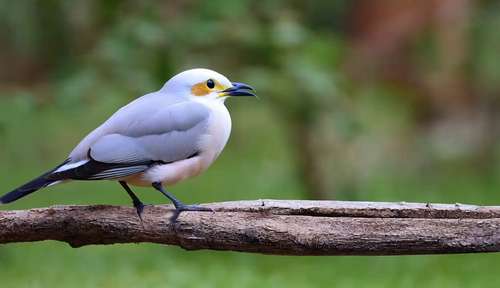}
\includegraphics[width=0.1\linewidth]{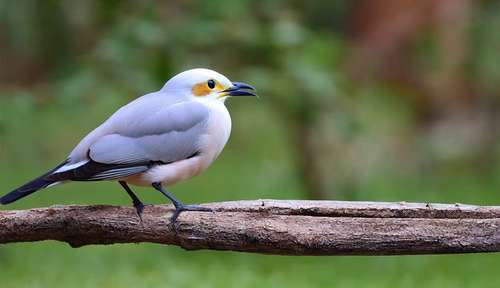}
\includegraphics[width=0.1\linewidth]{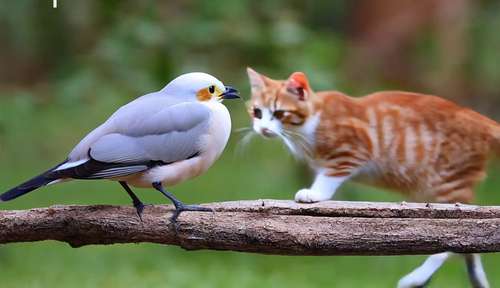}
\includegraphics[width=0.1\linewidth]{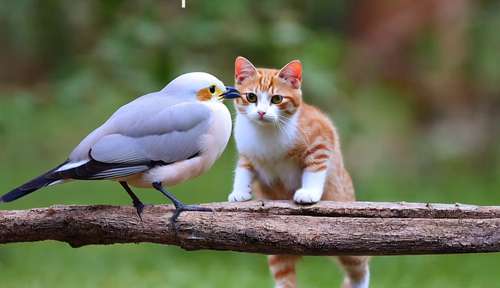}
\\[1.5ex]
\begin{minipage}[c]{2cm}\vspace*{0pt}\vfill\raggedright\textit{\small a boat and an airplane}\vfill\end{minipage}
\includegraphics[width=0.1\linewidth]{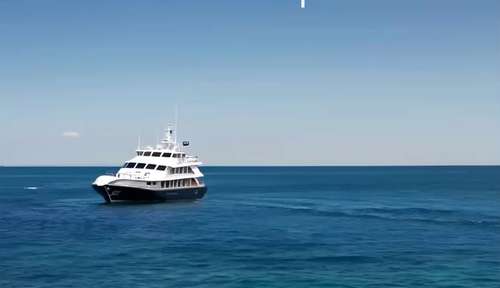}
\includegraphics[width=0.1\linewidth]{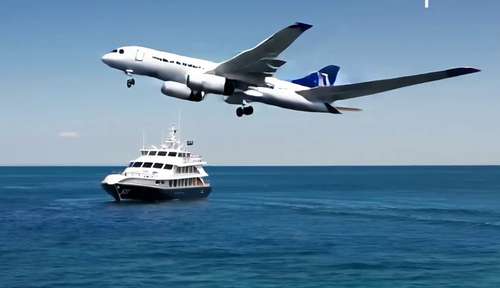}
\includegraphics[width=0.1\linewidth]{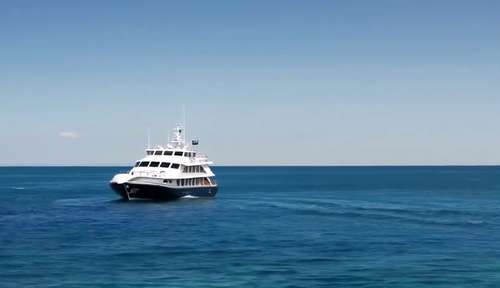}
\includegraphics[width=0.1\linewidth]{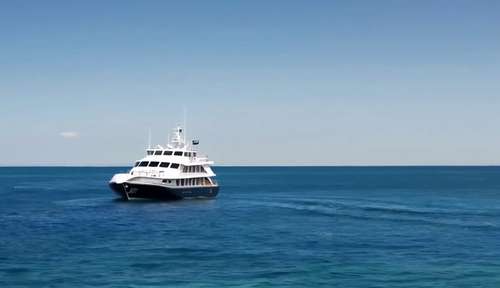}
\hspace{2mm}
\includegraphics[width=0.1\linewidth]{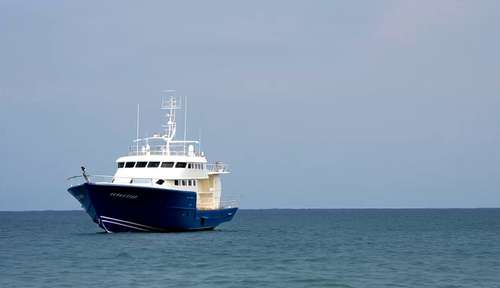}
\includegraphics[width=0.1\linewidth]{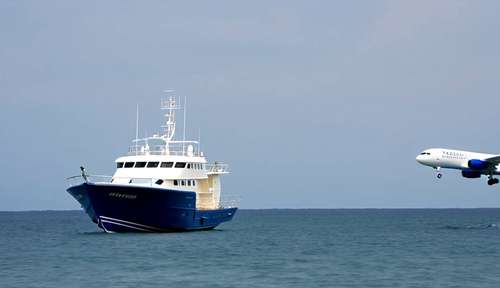}
\includegraphics[width=0.1\linewidth]{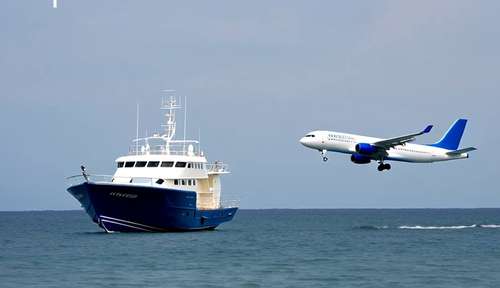}
\includegraphics[width=0.1\linewidth]{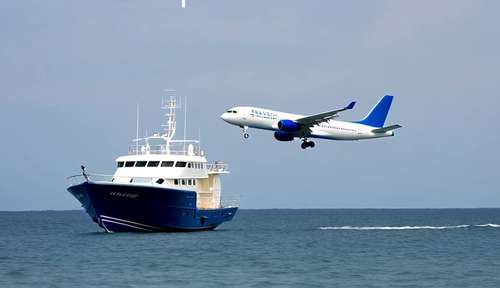}
\\[1.5ex]
\begin{minipage}[c]{2cm}\vspace*{0pt}\vfill\raggedright\textit{\small a book and a clock}\vfill\end{minipage}
\includegraphics[width=0.1\linewidth]{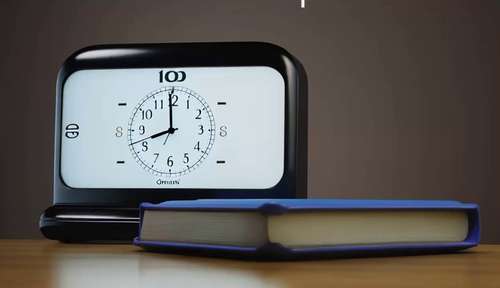}
\includegraphics[width=0.1\linewidth]{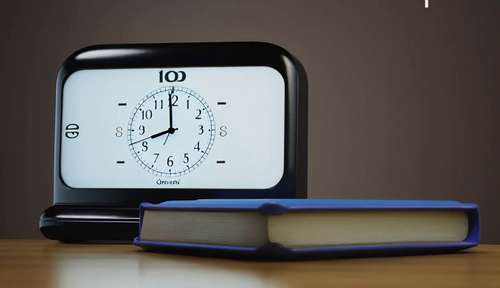}
\includegraphics[width=0.1\linewidth]{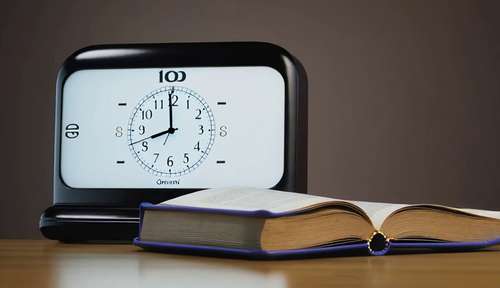}
\includegraphics[width=0.1\linewidth]{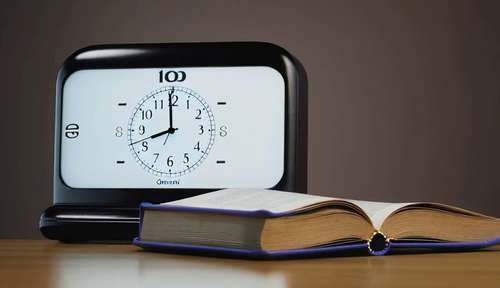}
\hspace{2mm}
\includegraphics[width=0.1\linewidth]{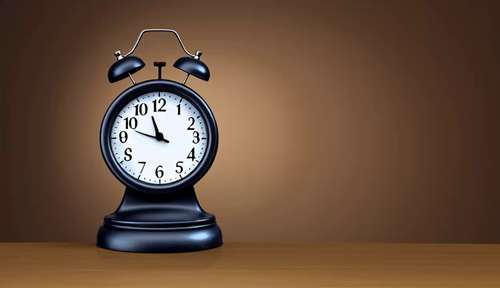}
\includegraphics[width=0.1\linewidth]{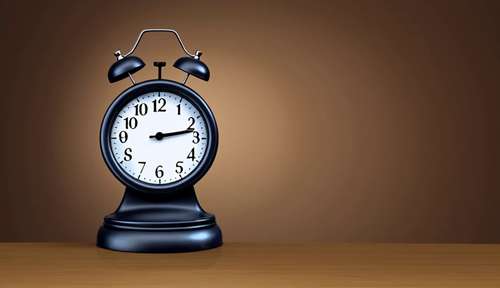}
\includegraphics[width=0.1\linewidth]{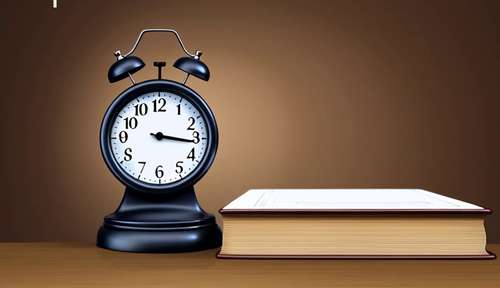}
\includegraphics[width=0.1\linewidth]{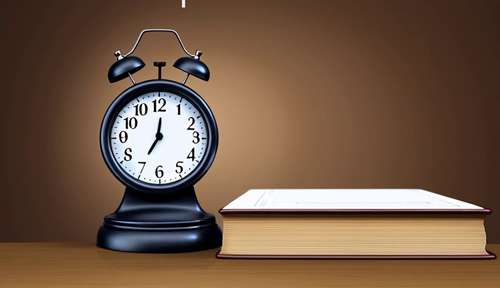}
\\[1.5ex]
\begin{minipage}[c]{2cm}\vspace*{0pt}\vfill\raggedright\textit{\small a bowl and a remote}\vfill\end{minipage}
\includegraphics[width=0.1\linewidth]{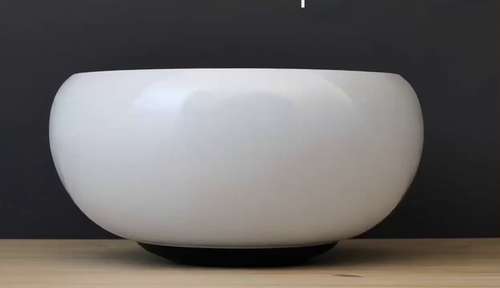}
\includegraphics[width=0.1\linewidth]{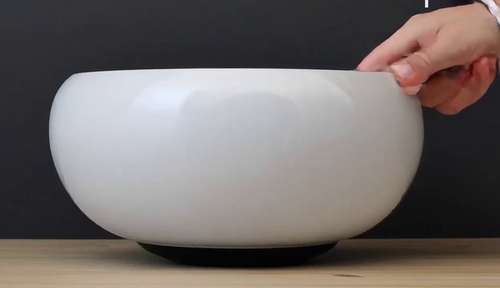}
\includegraphics[width=0.1\linewidth]{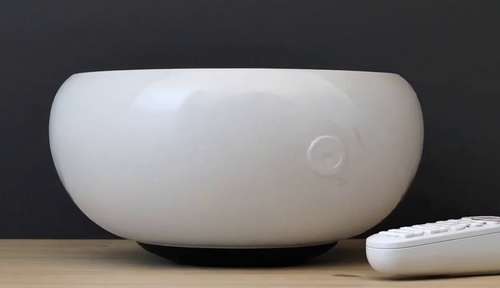}
\includegraphics[width=0.1\linewidth]{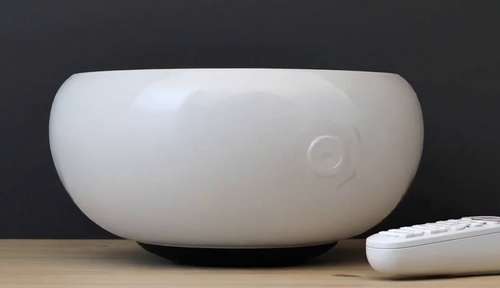}
\hspace{2mm}
\includegraphics[width=0.1\linewidth]{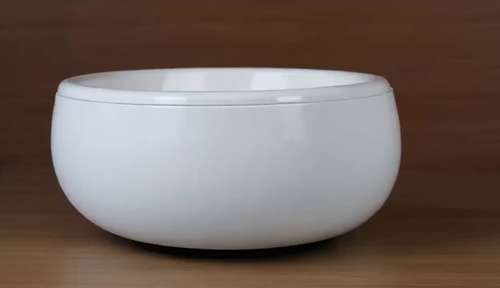}
\includegraphics[width=0.1\linewidth]{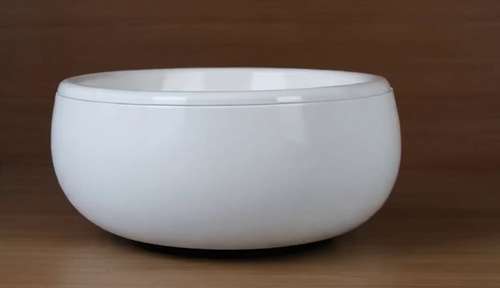}
\includegraphics[width=0.1\linewidth]{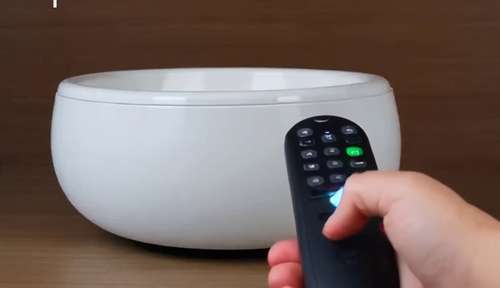}
\includegraphics[width=0.1\linewidth]{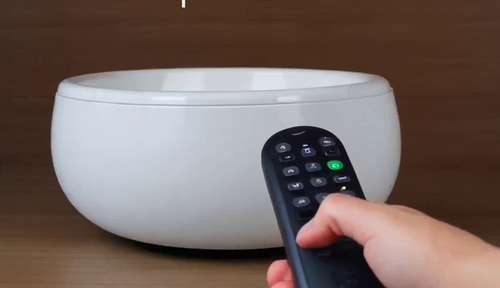}
\\[1.5ex]
\begin{minipage}[c]{2cm}\vspace*{0pt}\vfill\raggedright\textit{\small a bus and a traffic light}\vfill\end{minipage}
\includegraphics[width=0.1\linewidth]{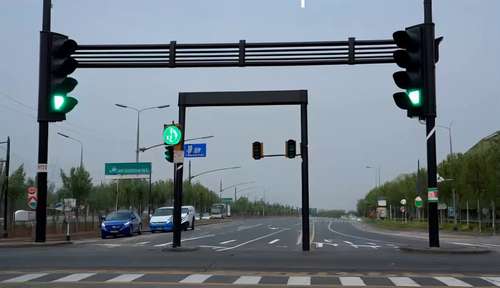}
\includegraphics[width=0.1\linewidth]{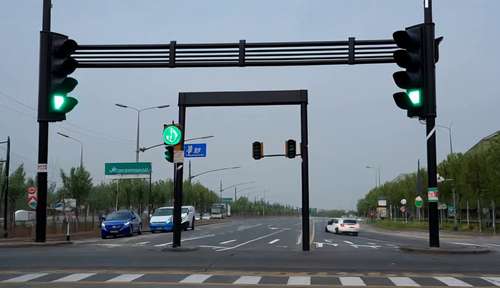}
\includegraphics[width=0.1\linewidth]{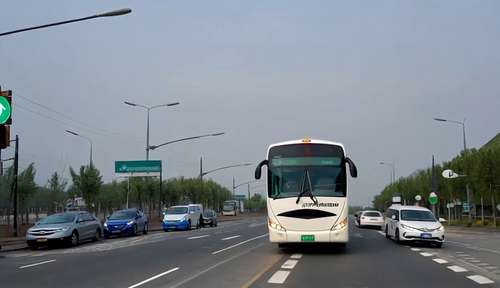}
\includegraphics[width=0.1\linewidth]{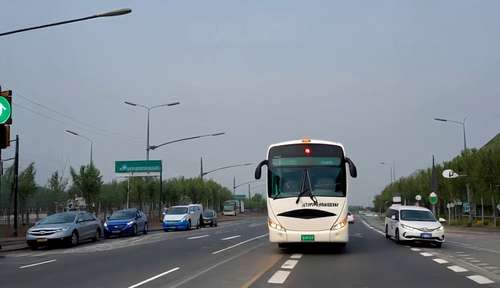}
\hspace{2mm}
\includegraphics[width=0.1\linewidth]{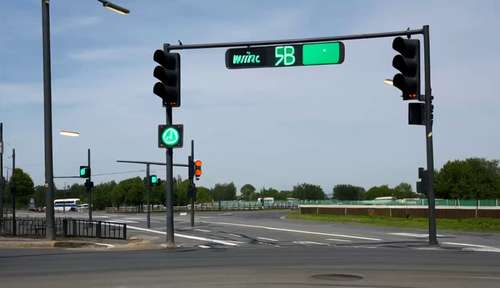}
\includegraphics[width=0.1\linewidth]{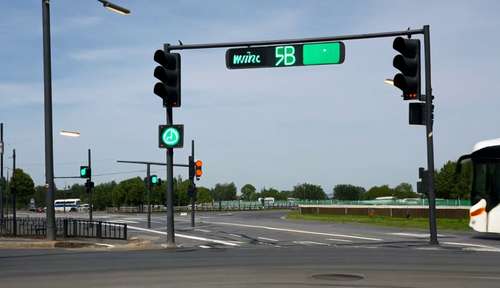}
\includegraphics[width=0.1\linewidth]{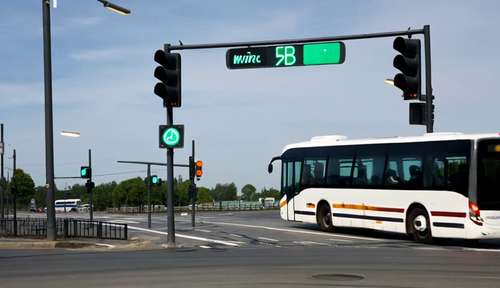}
\includegraphics[width=0.1\linewidth]{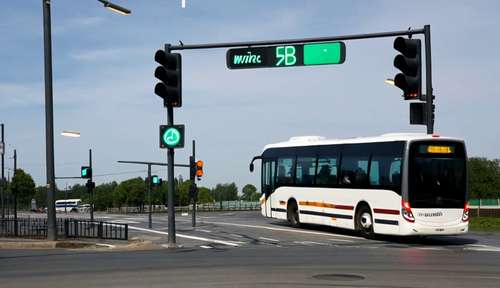}
\\[1.5ex]
\begin{minipage}[c]{2cm}\vspace*{0pt}\vfill\raggedright\textit{\small a cake and a vase}\vfill\end{minipage}
\includegraphics[width=0.1\linewidth]{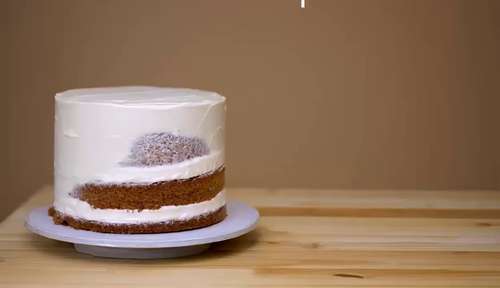}
\includegraphics[width=0.1\linewidth]{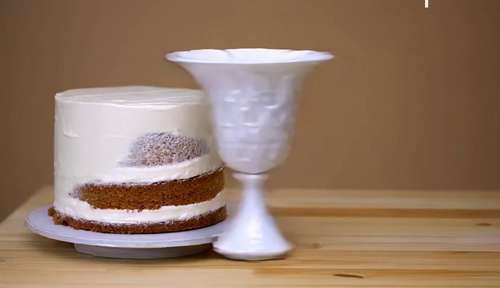}
\includegraphics[width=0.1\linewidth]{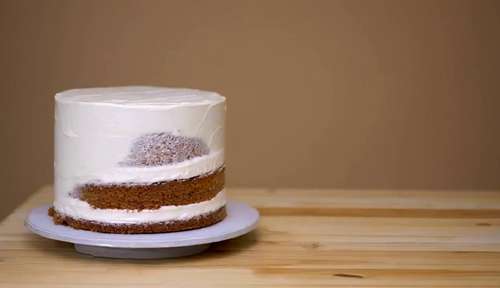}
\includegraphics[width=0.1\linewidth]{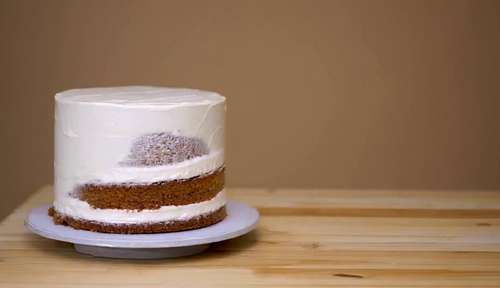}
\hspace{2mm}
\includegraphics[width=0.1\linewidth]{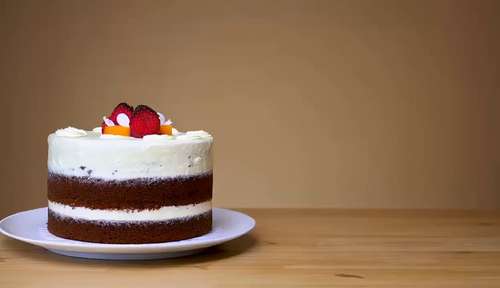}
\includegraphics[width=0.1\linewidth]{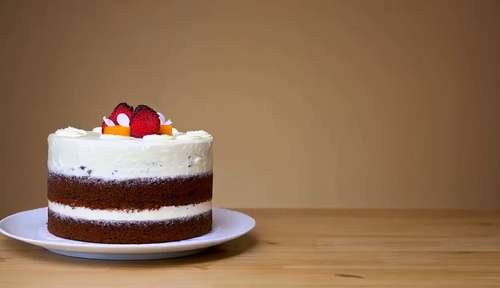}
\includegraphics[width=0.1\linewidth]{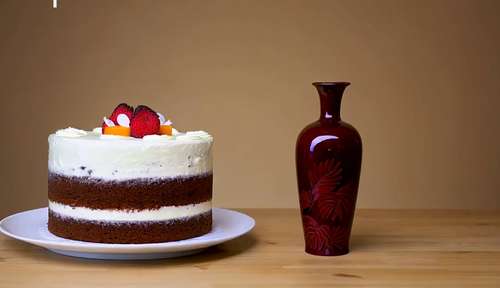}
\includegraphics[width=0.1\linewidth]{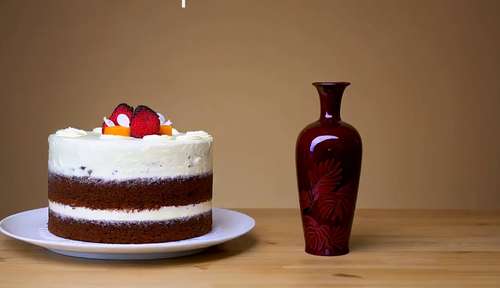}
\\[1.5ex]
\begin{minipage}[c]{2cm}\vspace*{0pt}\vfill\raggedright\textit{\small a car and a motorcycle}\vfill\end{minipage}
\includegraphics[width=0.1\linewidth]{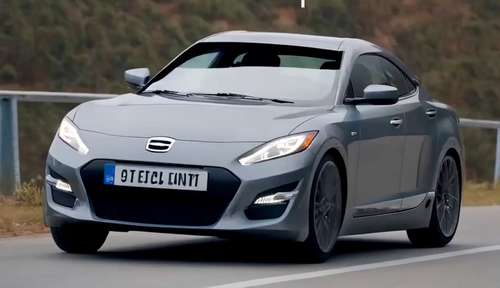}
\includegraphics[width=0.1\linewidth]{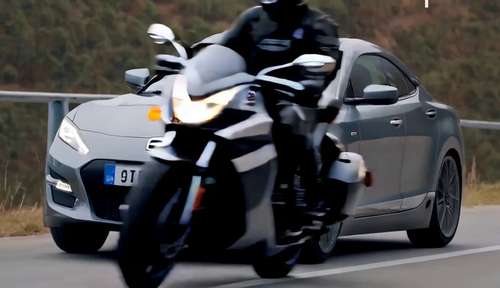}
\includegraphics[width=0.1\linewidth]{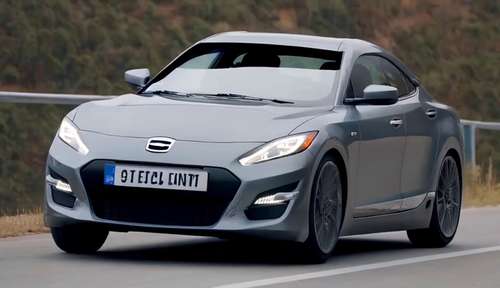}
\includegraphics[width=0.1\linewidth]{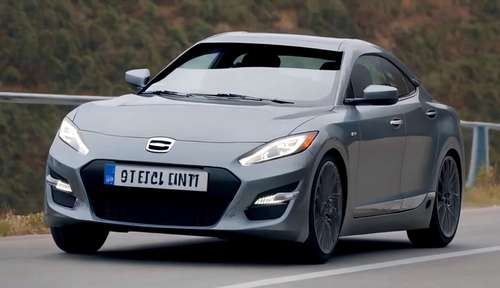}
\hspace{2mm}
\includegraphics[width=0.1\linewidth]{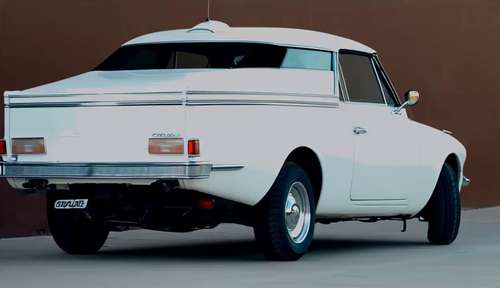}
\includegraphics[width=0.1\linewidth]{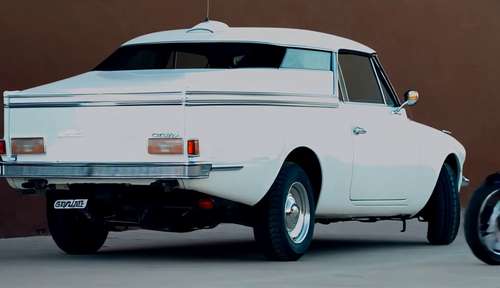}
\includegraphics[width=0.1\linewidth]{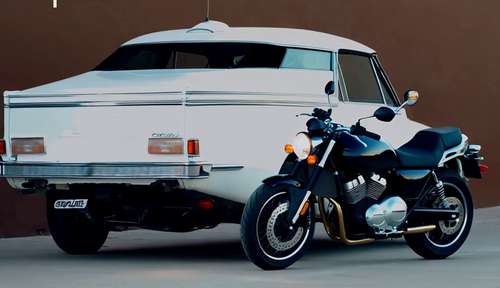}
\includegraphics[width=0.1\linewidth]{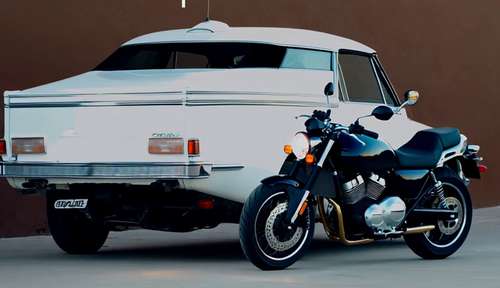}
\\[1.5ex]
\begin{minipage}[c]{2cm}\vspace*{0pt}\vfill\raggedright\textit{\small a cat and a dog}\vfill\end{minipage}
\includegraphics[width=0.1\linewidth]{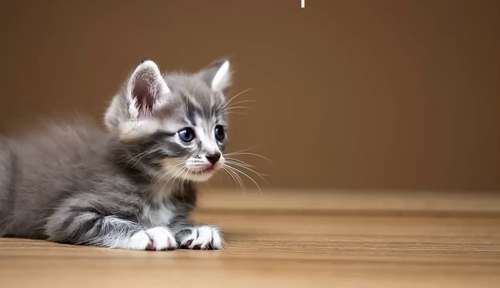}
\includegraphics[width=0.1\linewidth]{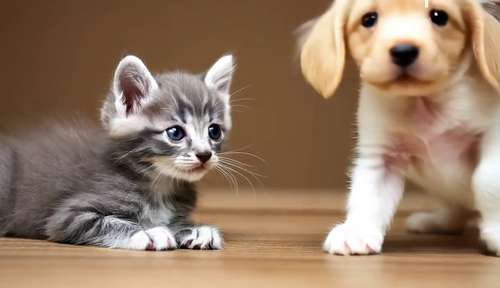}
\includegraphics[width=0.1\linewidth]{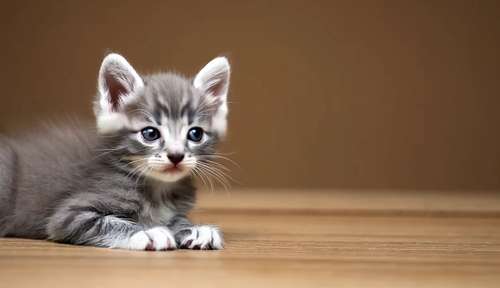}
\includegraphics[width=0.1\linewidth]{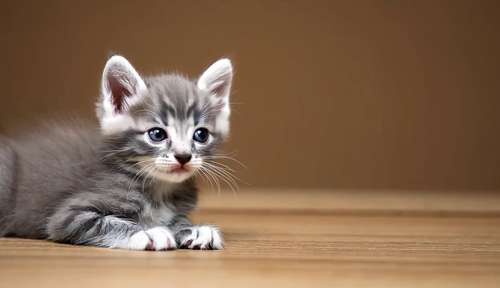}
\hspace{2mm}
\includegraphics[width=0.1\linewidth]{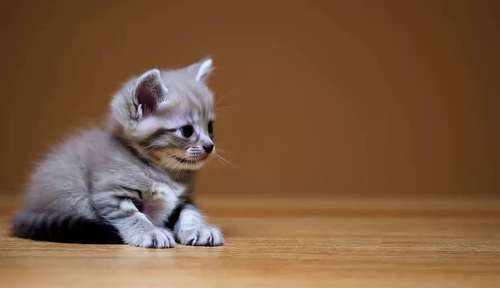}
\includegraphics[width=0.1\linewidth]{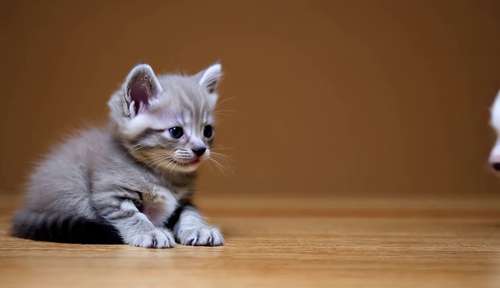}
\includegraphics[width=0.1\linewidth]{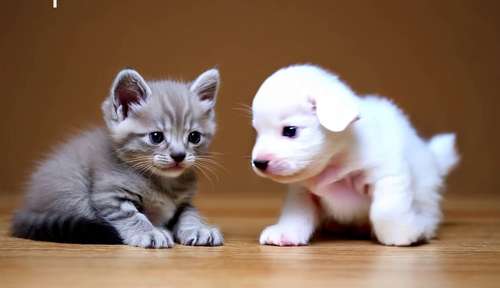}
\includegraphics[width=0.1\linewidth]{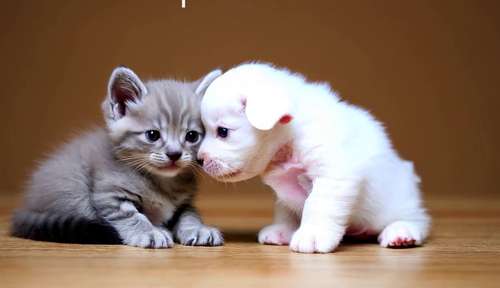}
\\[1.5ex]
\begin{minipage}[c]{2cm}\vspace*{0pt}\vfill\raggedright\textit{\small a cell phone and a book}\vfill\end{minipage}
\includegraphics[width=0.1\linewidth]{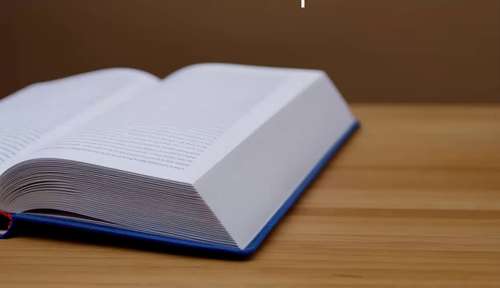}
\includegraphics[width=0.1\linewidth]{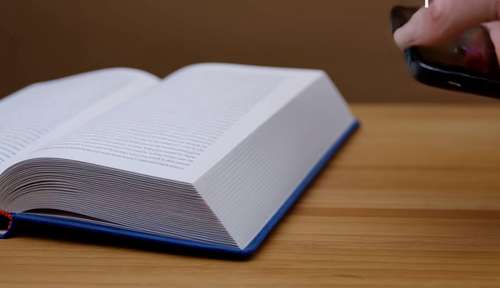}
\includegraphics[width=0.1\linewidth]{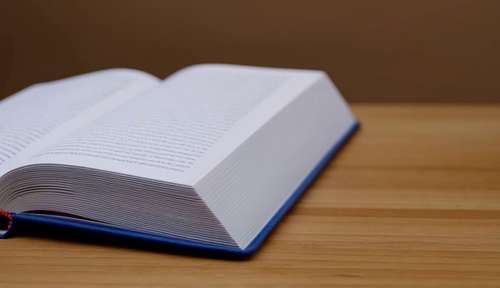}
\includegraphics[width=0.1\linewidth]{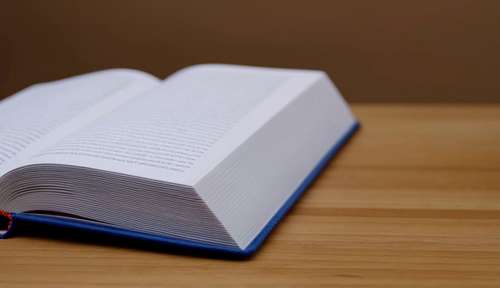}
\hspace{2mm}
\includegraphics[width=0.1\linewidth]{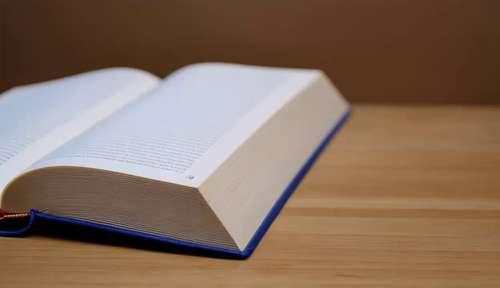}
\includegraphics[width=0.1\linewidth]{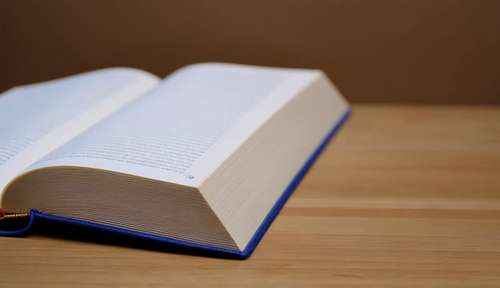}
\includegraphics[width=0.1\linewidth]{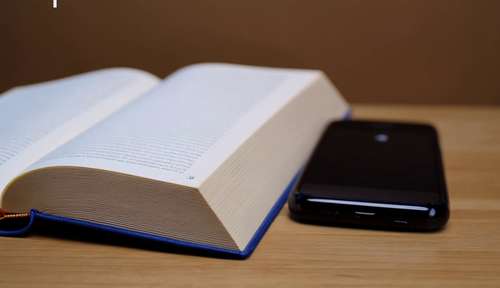}
\includegraphics[width=0.1\linewidth]{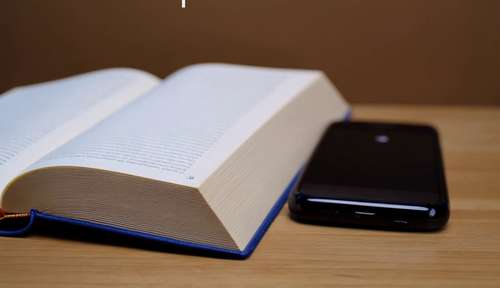}
\\[1.5ex]
\begin{minipage}[c]{2cm}\vspace*{0pt}\vfill\raggedright\textit{\small a donut and a suitcase}\vfill\end{minipage}
\includegraphics[width=0.1\linewidth]{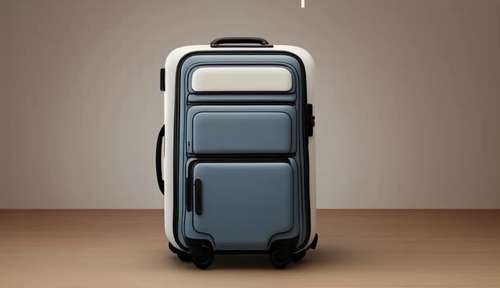}
\includegraphics[width=0.1\linewidth]{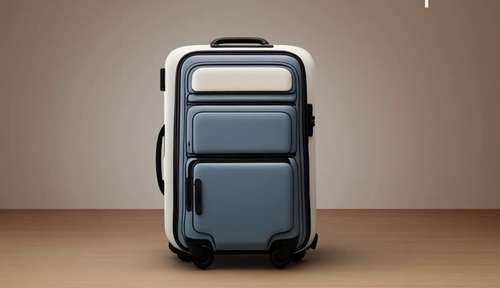}
\includegraphics[width=0.1\linewidth]{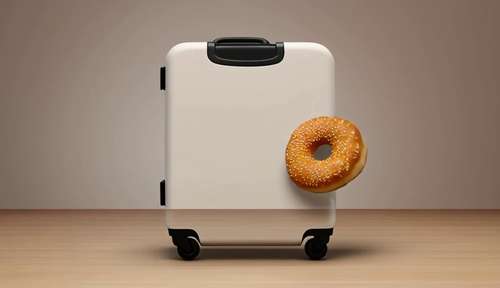}
\includegraphics[width=0.1\linewidth]{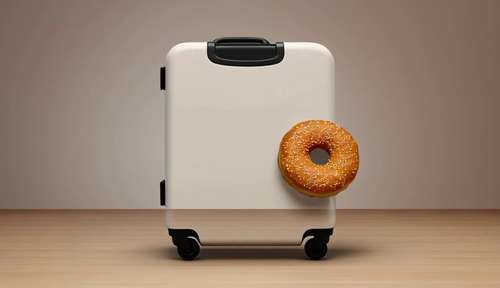}
\hspace{2mm}
\includegraphics[width=0.1\linewidth]{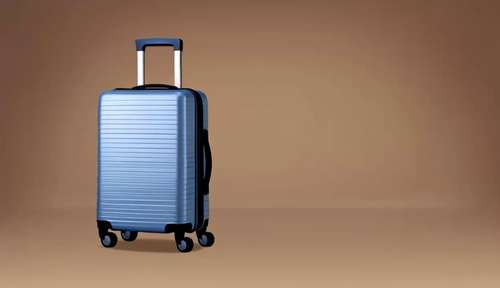}
\includegraphics[width=0.1\linewidth]{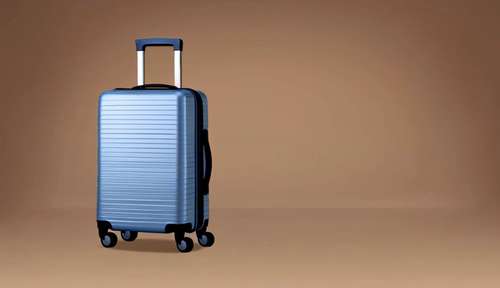}
\includegraphics[width=0.1\linewidth]{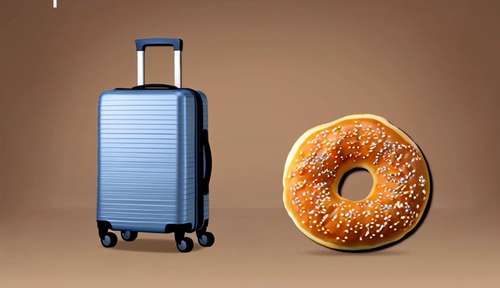}
\includegraphics[width=0.1\linewidth]{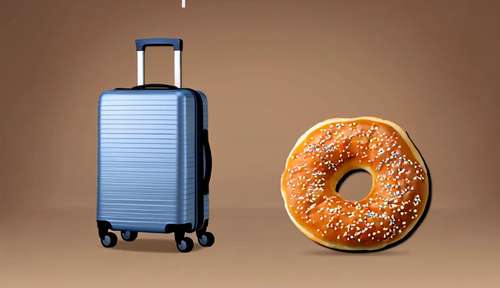}
\\[1.5ex]
\end{minipage}
\caption{Two Objects Text vs Ours.}
\label{fig:two_objects_text_vs_ours}
\end{figure*}

\begin{figure*}[t]
\centering
\begin{minipage}{\textwidth}
\makebox[2cm][l]{}\makebox[0.40\linewidth]{\small Text}\hspace{2mm}\makebox[0.40\linewidth]{\small Ours}\\[1.5ex]
\begin{minipage}[c]{2cm}\vspace*{0pt}\vfill\raggedright\textit{\small a giraffe and a bird}\vfill\end{minipage}
\includegraphics[width=0.1\linewidth]{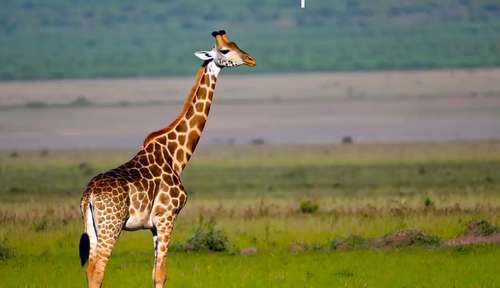}
\includegraphics[width=0.1\linewidth]{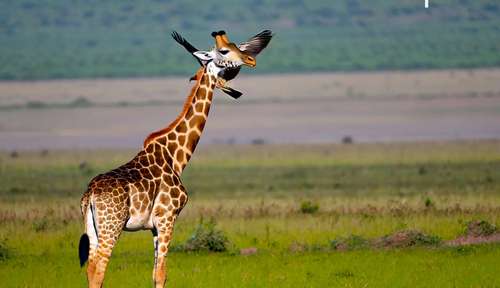}
\includegraphics[width=0.1\linewidth]{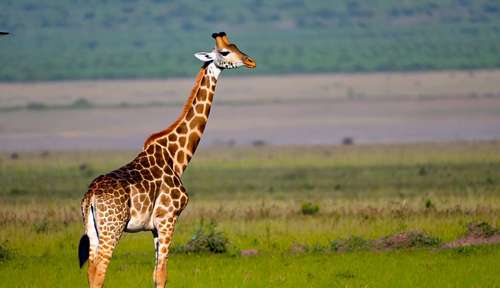}
\includegraphics[width=0.1\linewidth]{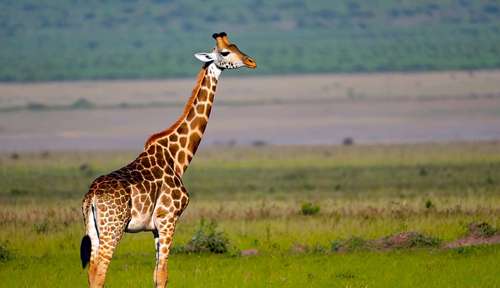}
\hspace{2mm}
\includegraphics[width=0.1\linewidth]{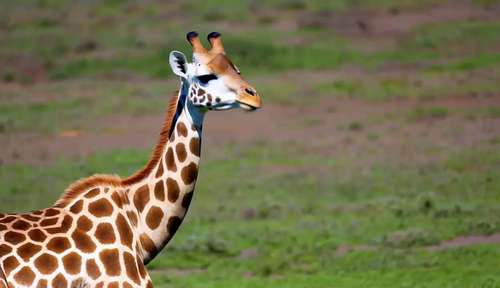}
\includegraphics[width=0.1\linewidth]{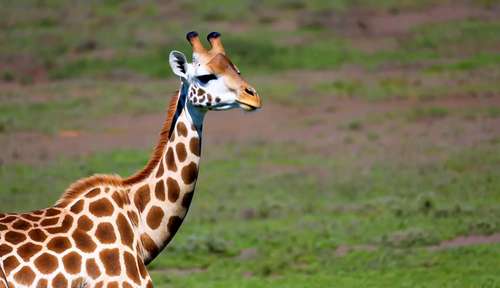}
\includegraphics[width=0.1\linewidth]{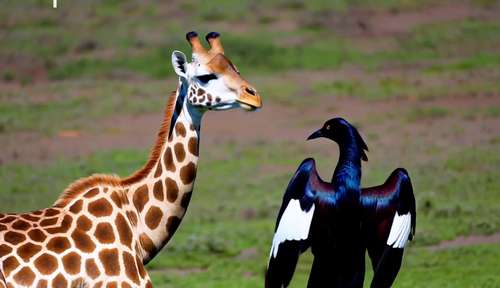}
\includegraphics[width=0.1\linewidth]{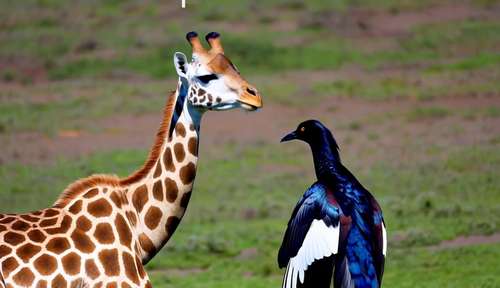}
\\[1.5ex]
\begin{minipage}[c]{2cm}\vspace*{0pt}\vfill\raggedright\textit{\small a handbag and a tie}\vfill\end{minipage}
\includegraphics[width=0.1\linewidth]{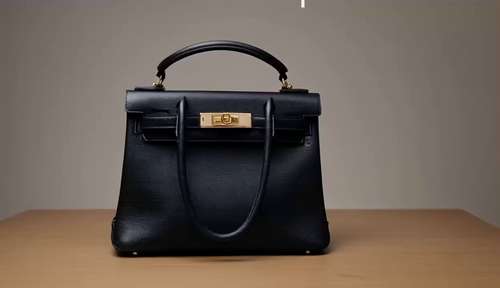}
\includegraphics[width=0.1\linewidth]{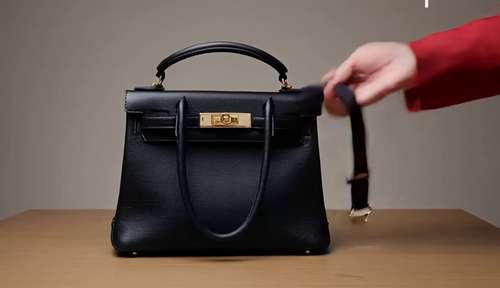}
\includegraphics[width=0.1\linewidth]{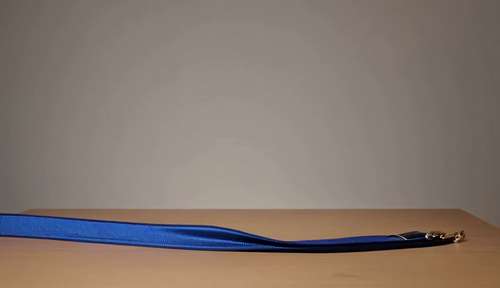}
\includegraphics[width=0.1\linewidth]{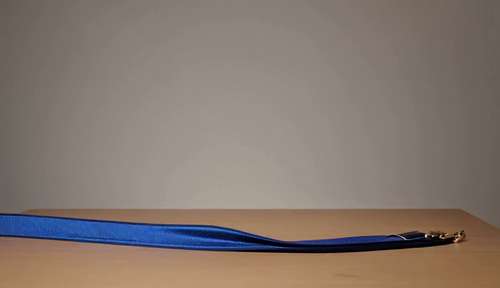}
\hspace{2mm}
\includegraphics[width=0.1\linewidth]{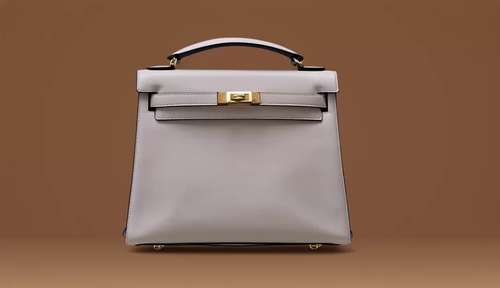}
\includegraphics[width=0.1\linewidth]{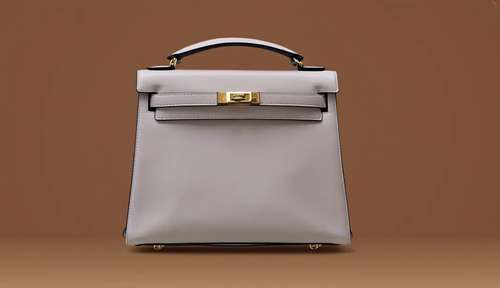}
\includegraphics[width=0.1\linewidth]{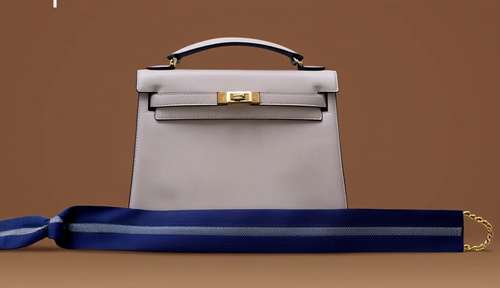}
\includegraphics[width=0.1\linewidth]{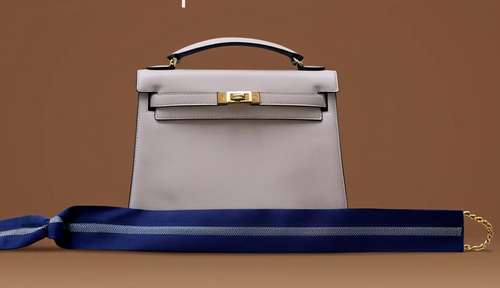}
\\[1.5ex]
\begin{minipage}[c]{2cm}\vspace*{0pt}\vfill\raggedright\textit{\small a horse and a sheep}\vfill\end{minipage}
\includegraphics[width=0.1\linewidth]{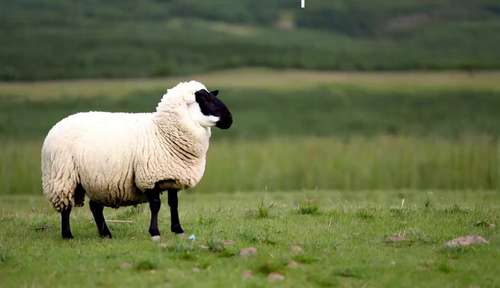}
\includegraphics[width=0.1\linewidth]{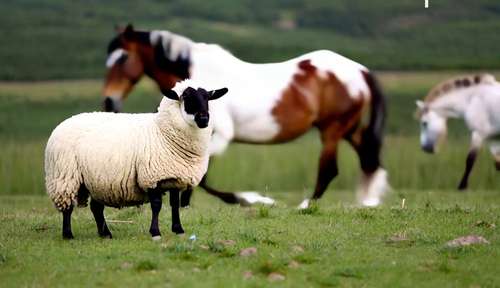}
\includegraphics[width=0.1\linewidth]{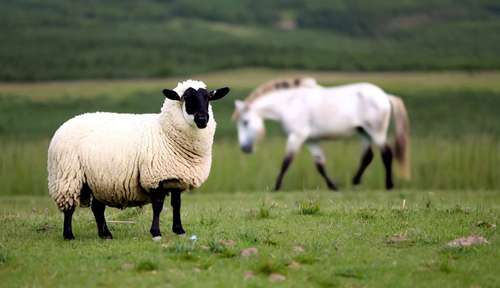}
\includegraphics[width=0.1\linewidth]{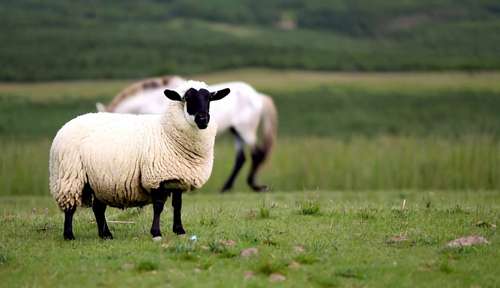}
\hspace{2mm}
\includegraphics[width=0.1\linewidth]{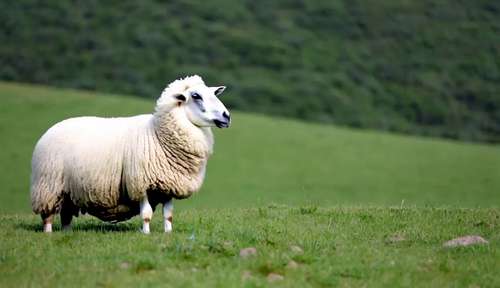}
\includegraphics[width=0.1\linewidth]{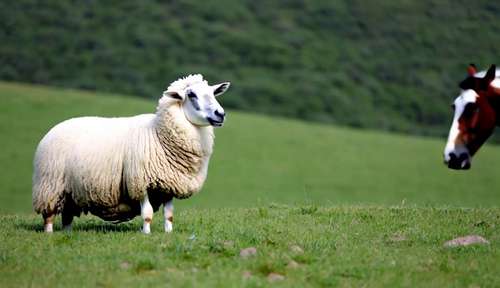}
\includegraphics[width=0.1\linewidth]{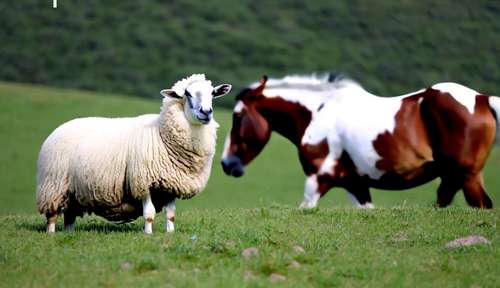}
\includegraphics[width=0.1\linewidth]{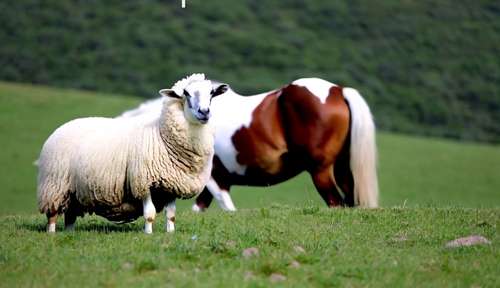}
\\[1.5ex]
\begin{minipage}[c]{2cm}\vspace*{0pt}\vfill\raggedright\textit{\small a laptop and a remote}\vfill\end{minipage}
\includegraphics[width=0.1\linewidth]{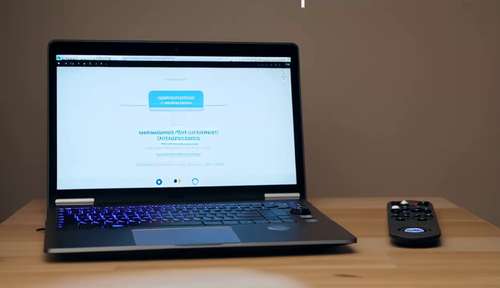}
\includegraphics[width=0.1\linewidth]{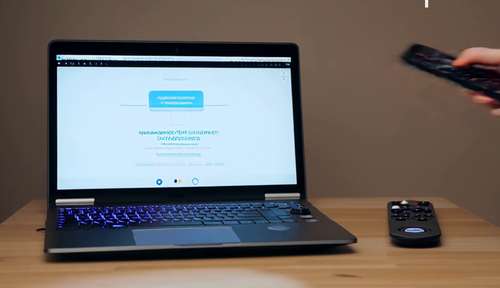}
\includegraphics[width=0.1\linewidth]{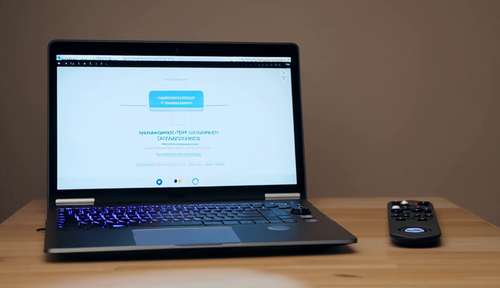}
\includegraphics[width=0.1\linewidth]{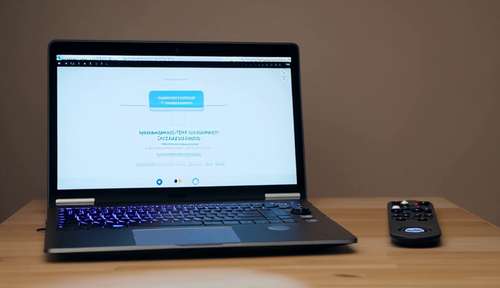}
\hspace{2mm}
\includegraphics[width=0.1\linewidth]{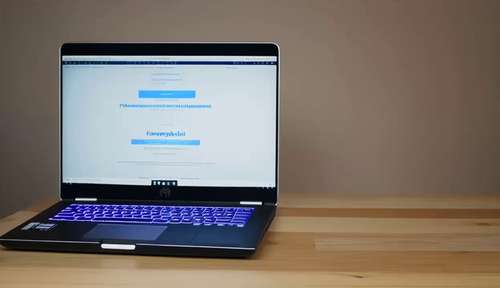}
\includegraphics[width=0.1\linewidth]{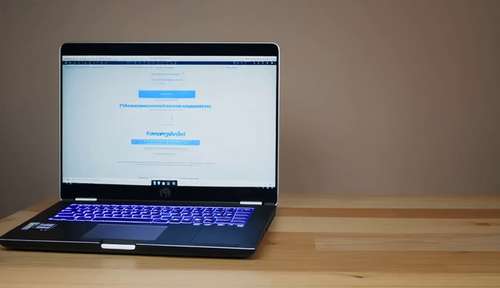}
\includegraphics[width=0.1\linewidth]{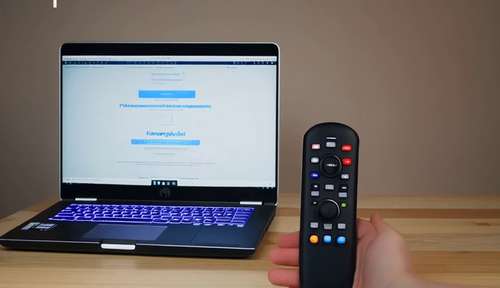}
\includegraphics[width=0.1\linewidth]{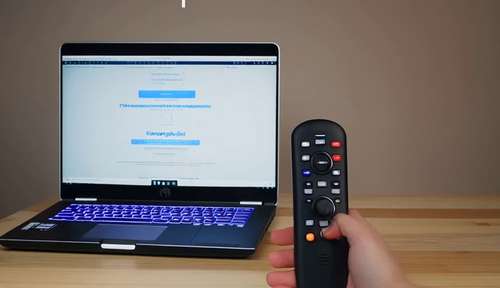}
\\[1.5ex]
\begin{minipage}[c]{2cm}\vspace*{0pt}\vfill\raggedright\textit{\small a motorcycle and a boat}\vfill\end{minipage}
\includegraphics[width=0.1\linewidth]{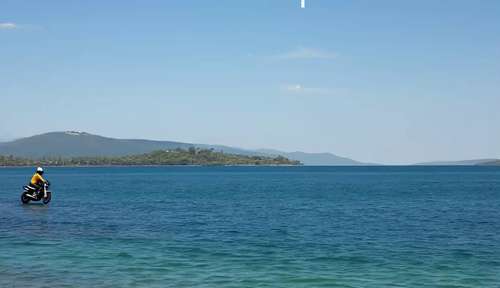}
\includegraphics[width=0.1\linewidth]{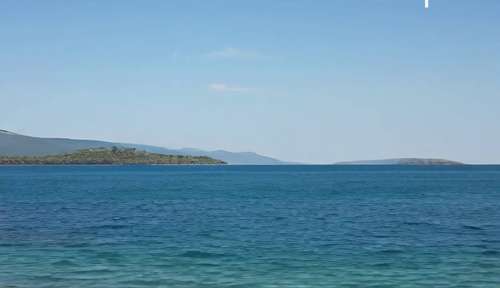}
\includegraphics[width=0.1\linewidth]{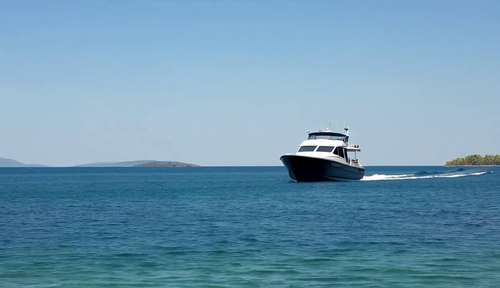}
\includegraphics[width=0.1\linewidth]{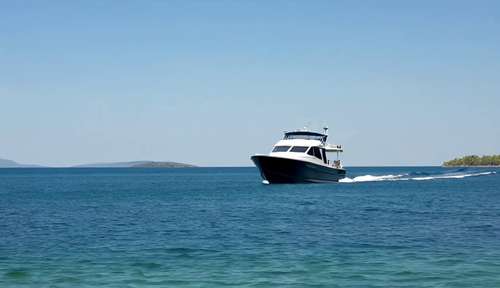}
\hspace{2mm}
\includegraphics[width=0.1\linewidth]{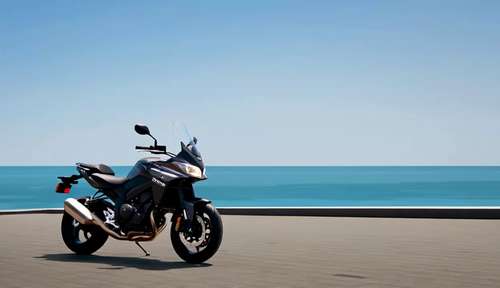}
\includegraphics[width=0.1\linewidth]{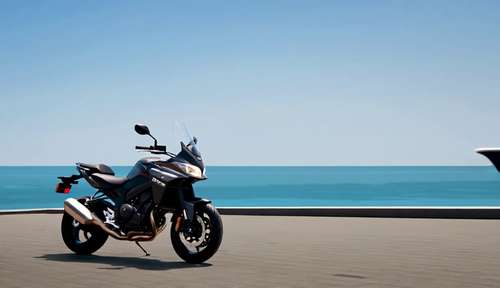}
\includegraphics[width=0.1\linewidth]{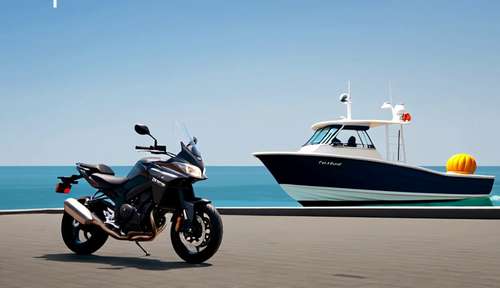}
\includegraphics[width=0.1\linewidth]{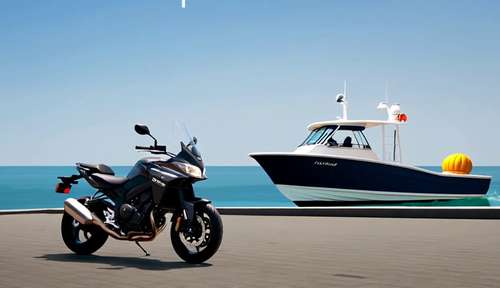}
\\[1.5ex]
\begin{minipage}[c]{2cm}\vspace*{0pt}\vfill\raggedright\textit{\small a motorcycle and a bus}\vfill\end{minipage}
\includegraphics[width=0.1\linewidth]{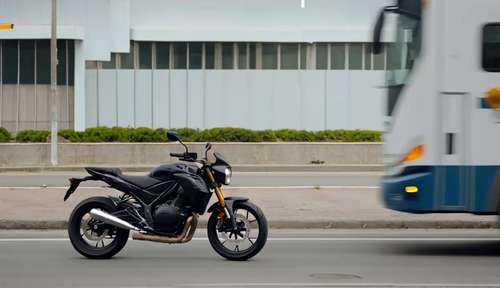}
\includegraphics[width=0.1\linewidth]{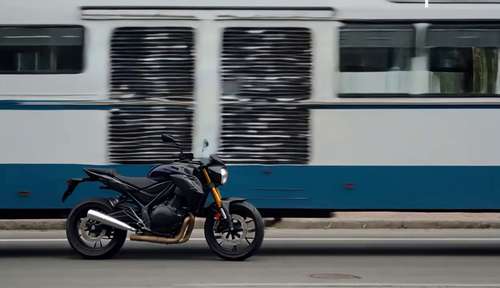}
\includegraphics[width=0.1\linewidth]{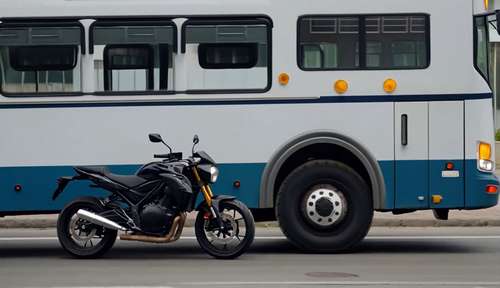}
\includegraphics[width=0.1\linewidth]{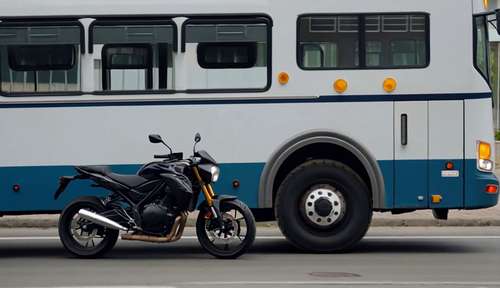}
\hspace{2mm}
\includegraphics[width=0.1\linewidth]{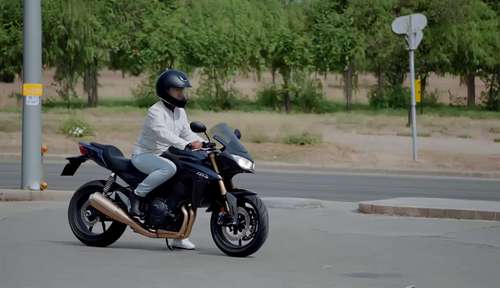}
\includegraphics[width=0.1\linewidth]{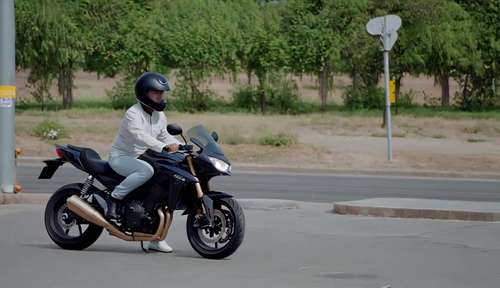}
\includegraphics[width=0.1\linewidth]{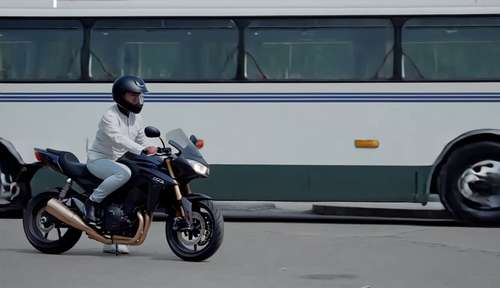}
\includegraphics[width=0.1\linewidth]{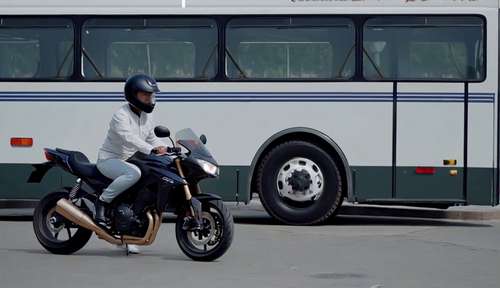}
\\[1.5ex]
\begin{minipage}[c]{2cm}\vspace*{0pt}\vfill\raggedright\textit{\small a person and a toothbrush}\vfill\end{minipage}
\includegraphics[width=0.1\linewidth]{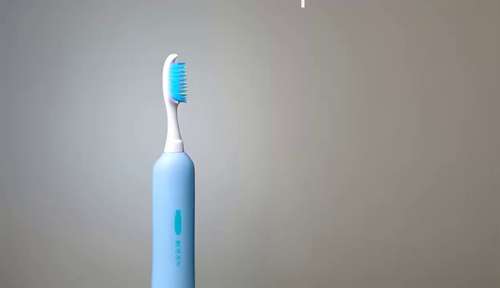}
\includegraphics[width=0.1\linewidth]{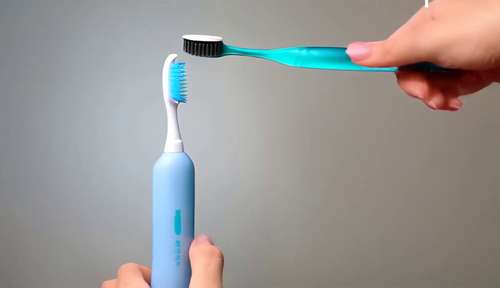}
\includegraphics[width=0.1\linewidth]{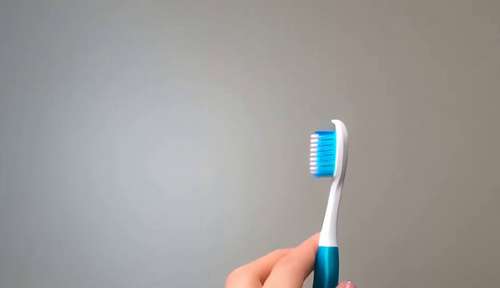}
\includegraphics[width=0.1\linewidth]{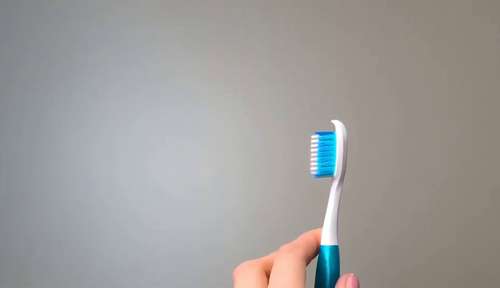}
\hspace{2mm}
\includegraphics[width=0.1\linewidth]{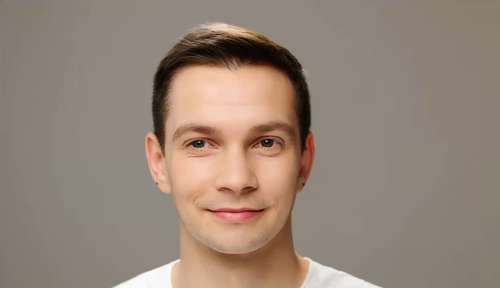}
\includegraphics[width=0.1\linewidth]{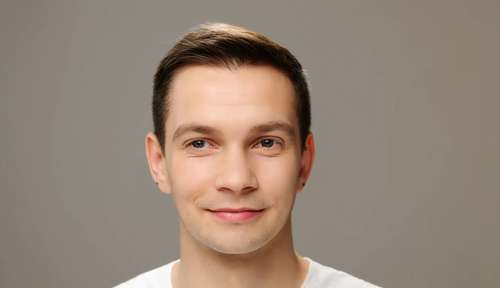}
\includegraphics[width=0.1\linewidth]{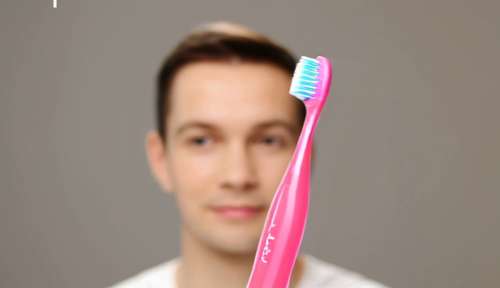}
\includegraphics[width=0.1\linewidth]{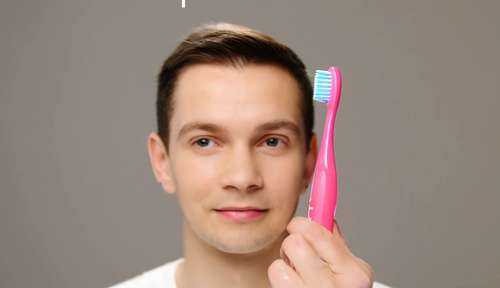}
\\[1.5ex]
\begin{minipage}[c]{2cm}\vspace*{0pt}\vfill\raggedright\textit{\small a pizza and a tie}\vfill\end{minipage}
\includegraphics[width=0.1\linewidth]{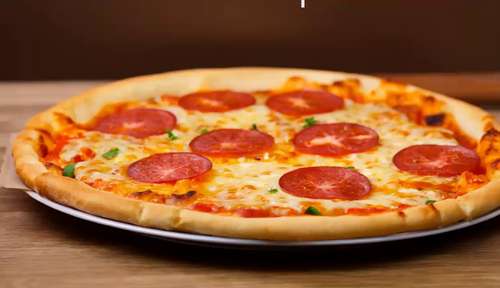}
\includegraphics[width=0.1\linewidth]{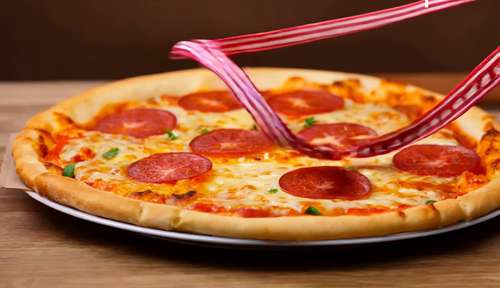}
\includegraphics[width=0.1\linewidth]{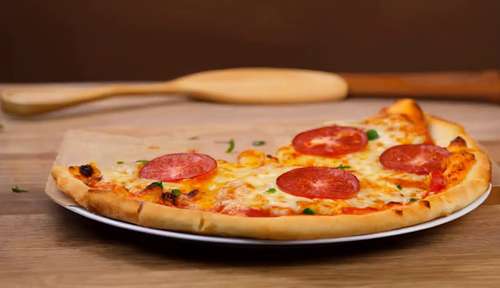}
\includegraphics[width=0.1\linewidth]{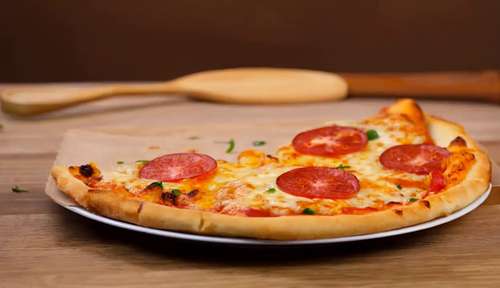}
\hspace{2mm}
\includegraphics[width=0.1\linewidth]{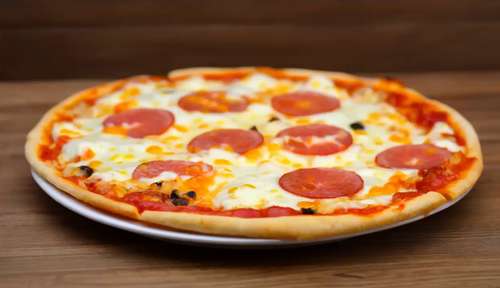}
\includegraphics[width=0.1\linewidth]{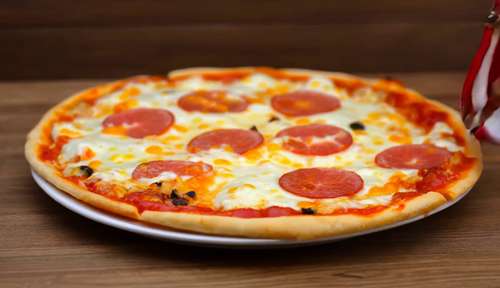}
\includegraphics[width=0.1\linewidth]{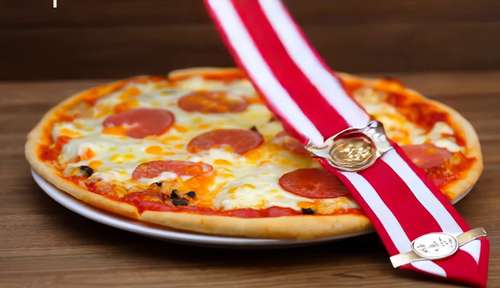}
\includegraphics[width=0.1\linewidth]{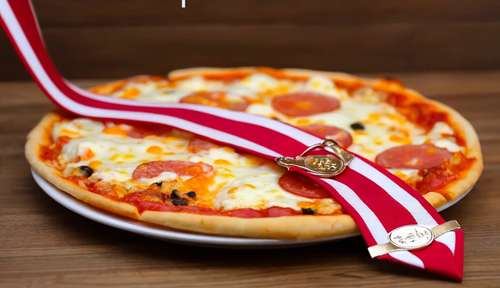}
\\[1.5ex]
\begin{minipage}[c]{2cm}\vspace*{0pt}\vfill\raggedright\textit{\small a potted plant and a tv}\vfill\end{minipage}
\includegraphics[width=0.1\linewidth]{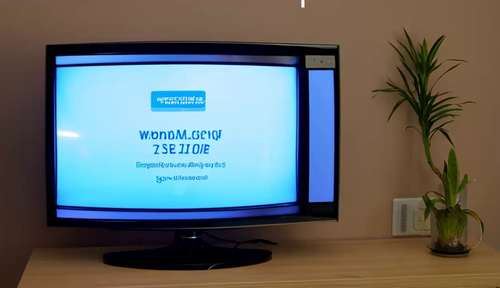}
\includegraphics[width=0.1\linewidth]{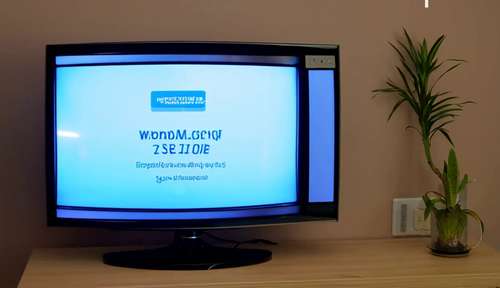}
\includegraphics[width=0.1\linewidth]{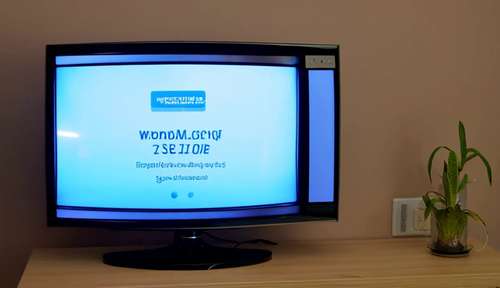}
\includegraphics[width=0.1\linewidth]{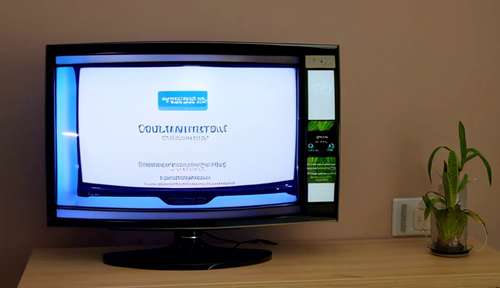}
\hspace{2mm}
\includegraphics[width=0.1\linewidth]{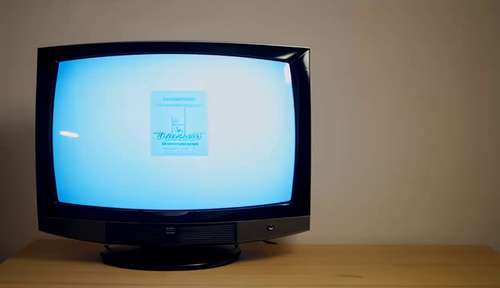}
\includegraphics[width=0.1\linewidth]{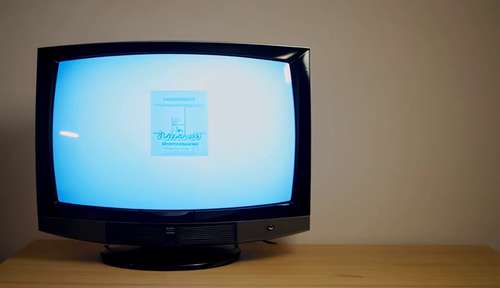}
\includegraphics[width=0.1\linewidth]{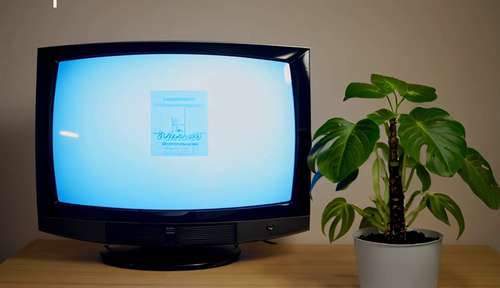}
\includegraphics[width=0.1\linewidth]{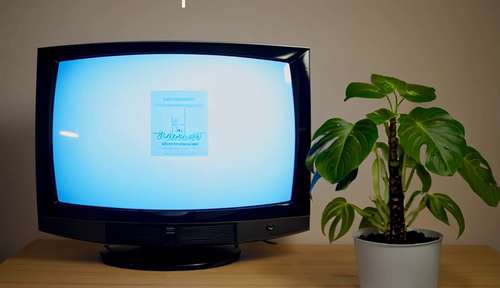}
\\[1.5ex]
\begin{minipage}[c]{2cm}\vspace*{0pt}\vfill\raggedright\textit{\small a spoon and a laptop}\vfill\end{minipage}
\includegraphics[width=0.1\linewidth]{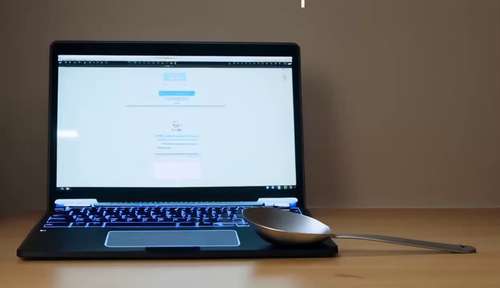}
\includegraphics[width=0.1\linewidth]{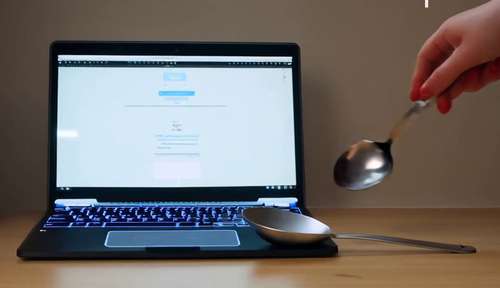}
\includegraphics[width=0.1\linewidth]{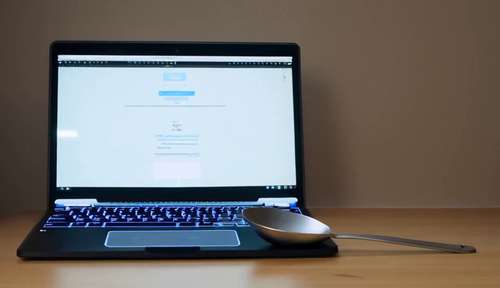}
\includegraphics[width=0.1\linewidth]{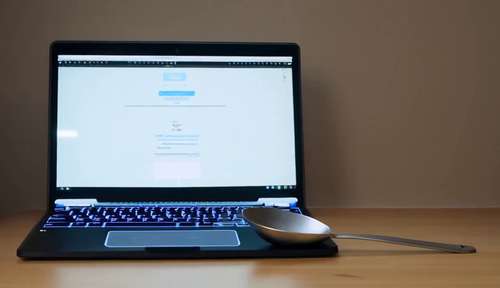}
\hspace{2mm}
\includegraphics[width=0.1\linewidth]{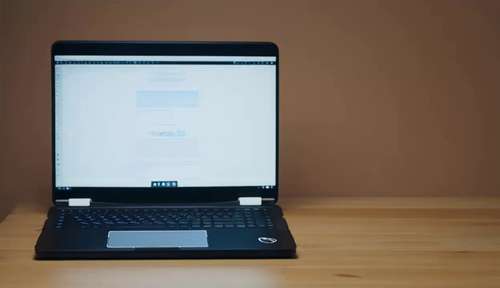}
\includegraphics[width=0.1\linewidth]{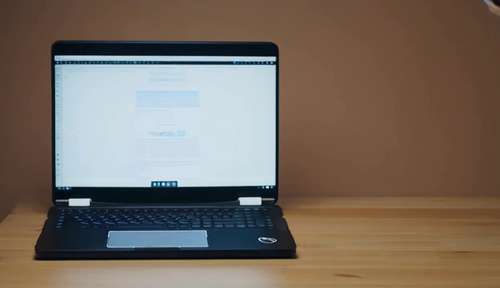}
\includegraphics[width=0.1\linewidth]{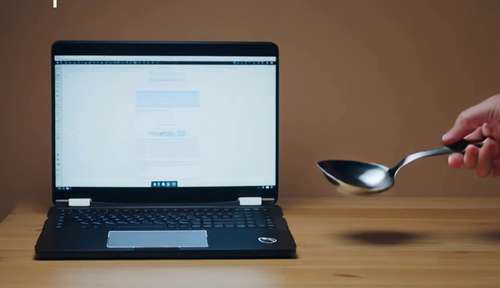}
\includegraphics[width=0.1\linewidth]{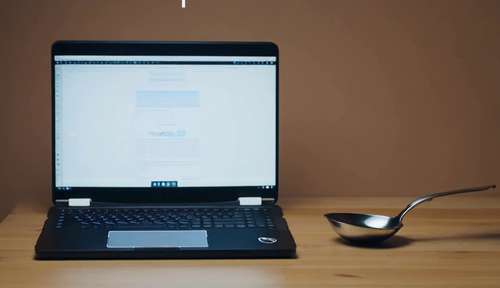}
\\[1.5ex]
\begin{minipage}[c]{2cm}\vspace*{0pt}\vfill\raggedright\textit{\small a tennis racket and a bottle}\vfill\end{minipage}
\includegraphics[width=0.1\linewidth]{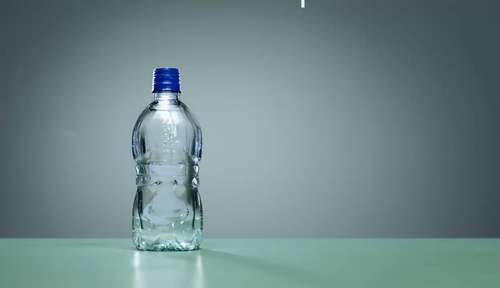}
\includegraphics[width=0.1\linewidth]{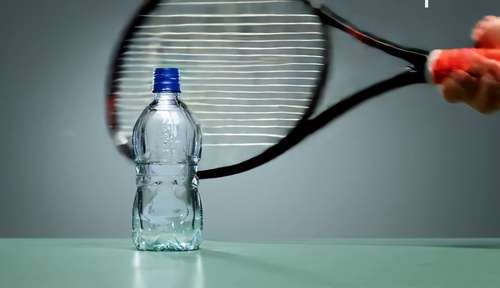}
\includegraphics[width=0.1\linewidth]{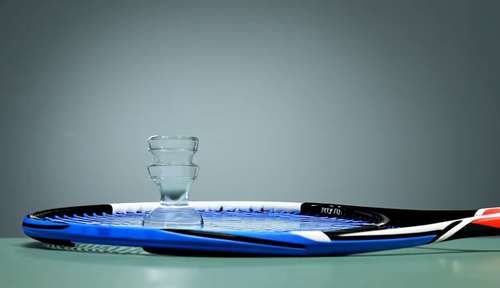}
\includegraphics[width=0.1\linewidth]{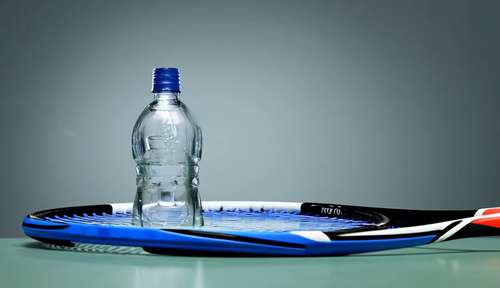}
\hspace{2mm}
\includegraphics[width=0.1\linewidth]{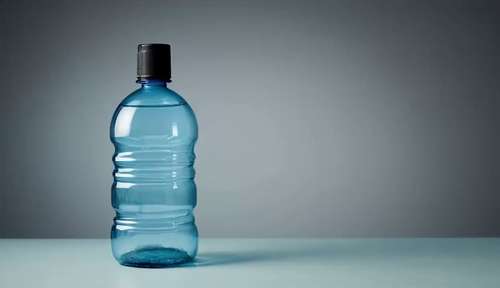}
\includegraphics[width=0.1\linewidth]{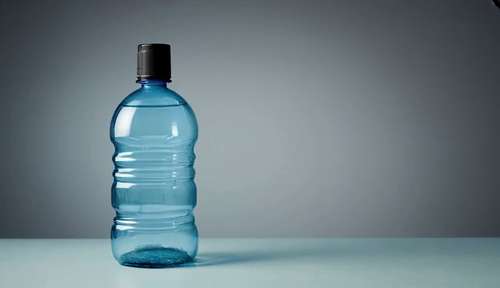}
\includegraphics[width=0.1\linewidth]{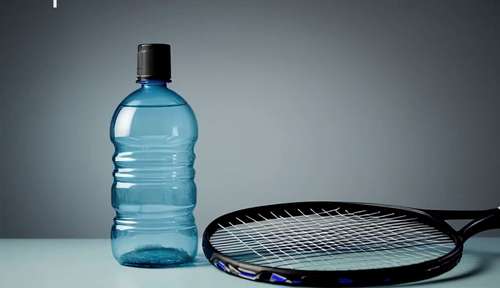}
\includegraphics[width=0.1\linewidth]{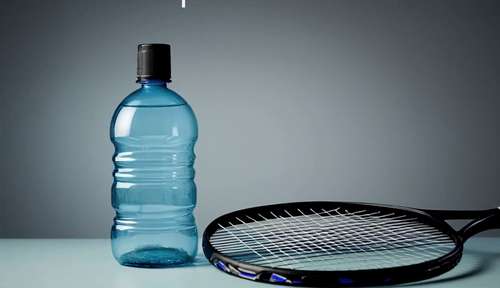}
\\[1.5ex]
\begin{minipage}[c]{2cm}\vspace*{0pt}\vfill\raggedright\textit{\small a toaster and a teddy bear}\vfill\end{minipage}
\includegraphics[width=0.1\linewidth]{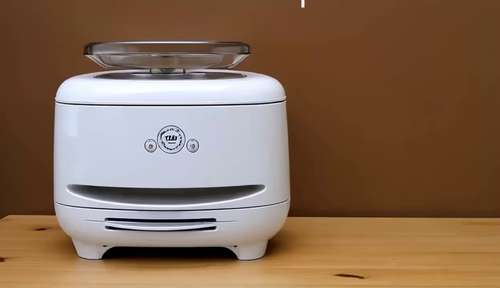}
\includegraphics[width=0.1\linewidth]{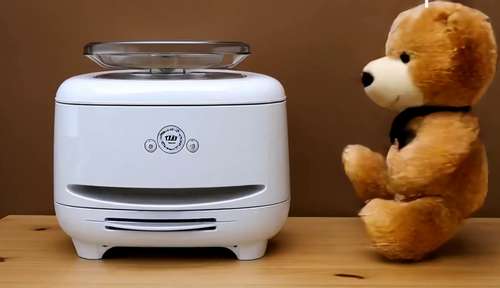}
\includegraphics[width=0.1\linewidth]{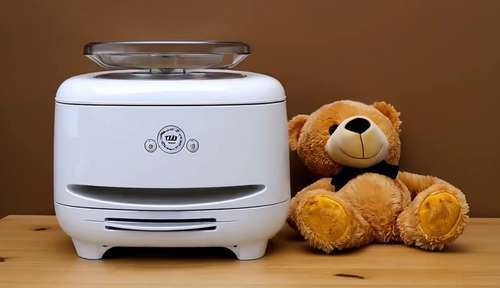}
\includegraphics[width=0.1\linewidth]{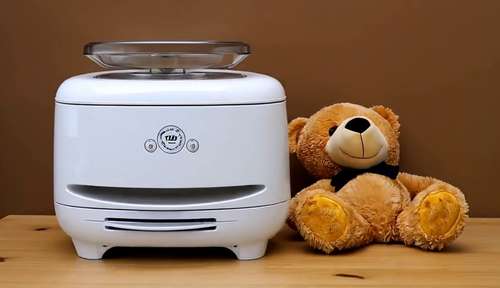}
\hspace{2mm}
\includegraphics[width=0.1\linewidth]{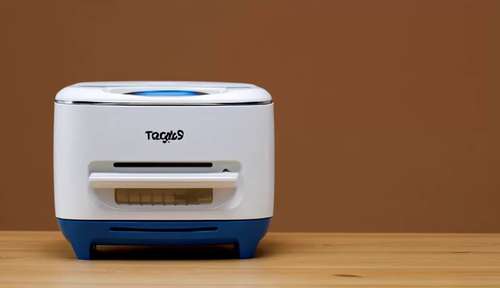}
\includegraphics[width=0.1\linewidth]{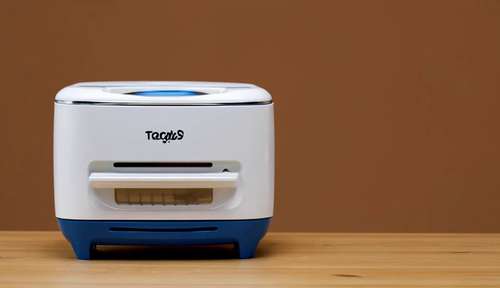}
\includegraphics[width=0.1\linewidth]{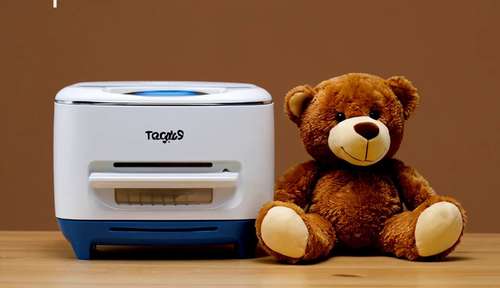}
\includegraphics[width=0.1\linewidth]{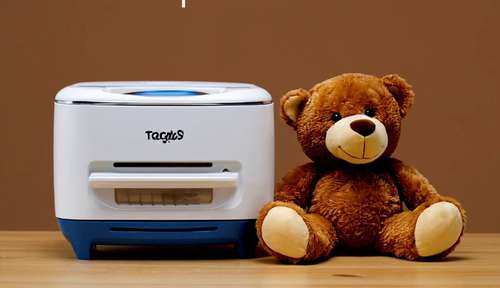}
\\[1.5ex]
\begin{minipage}[c]{2cm}\vspace*{0pt}\vfill\raggedright\textit{\small a truck and a bicycle}\vfill\end{minipage}
\includegraphics[width=0.1\linewidth]{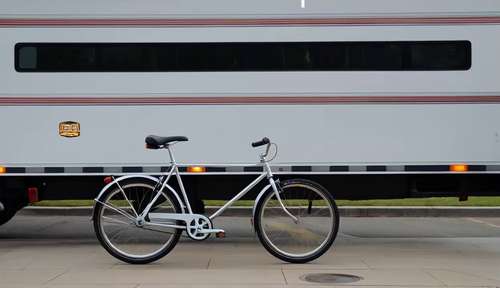}
\includegraphics[width=0.1\linewidth]{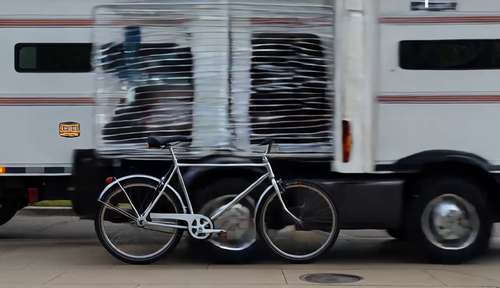}
\includegraphics[width=0.1\linewidth]{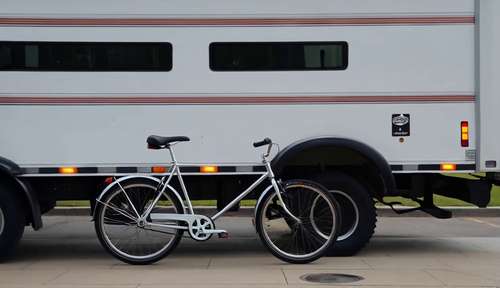}
\includegraphics[width=0.1\linewidth]{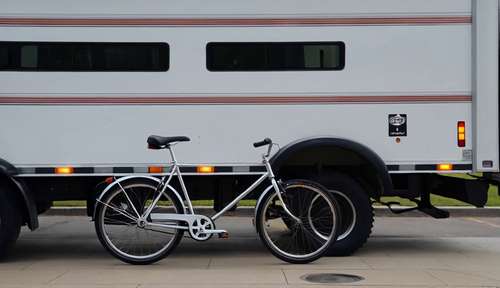}
\hspace{2mm}
\includegraphics[width=0.1\linewidth]{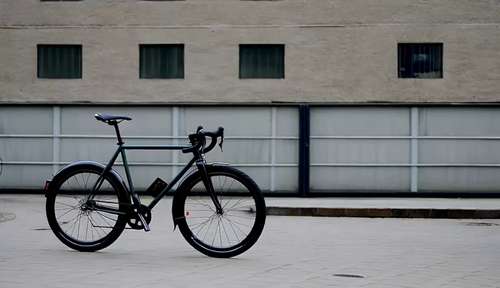}
\includegraphics[width=0.1\linewidth]{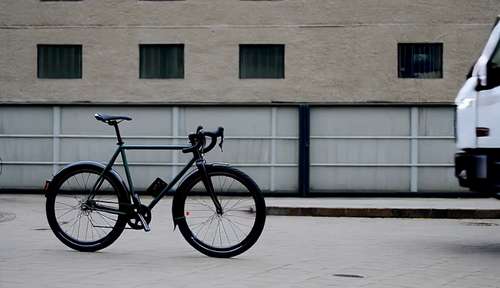}
\includegraphics[width=0.1\linewidth]{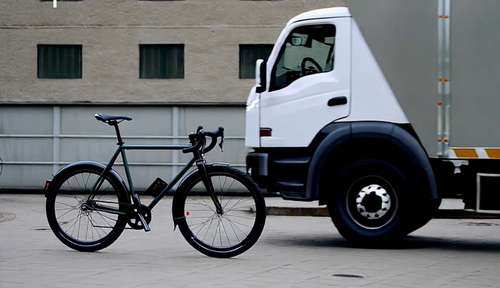}
\includegraphics[width=0.1\linewidth]{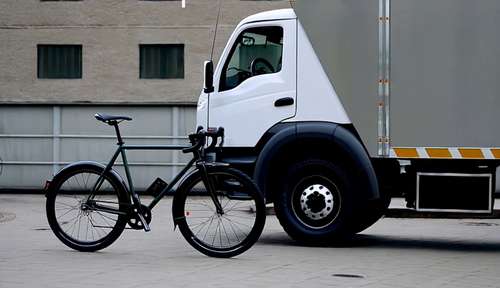}
\\[1.5ex]
\begin{minipage}[c]{2cm}\vspace*{0pt}\vfill\raggedright\textit{\small a zebra and a giraffe}\vfill\end{minipage}
\includegraphics[width=0.1\linewidth]{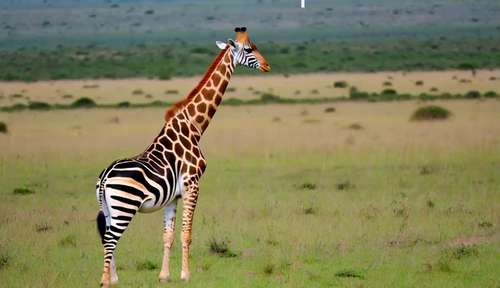}
\includegraphics[width=0.1\linewidth]{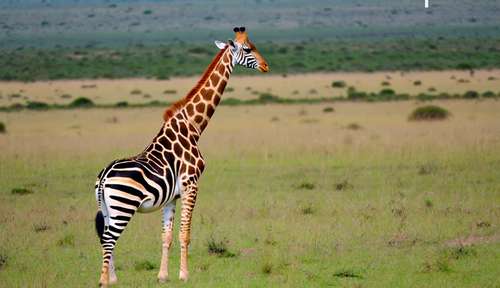}
\includegraphics[width=0.1\linewidth]{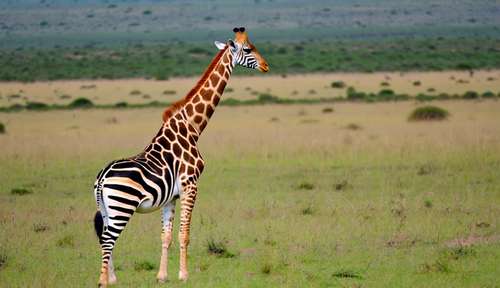}
\includegraphics[width=0.1\linewidth]{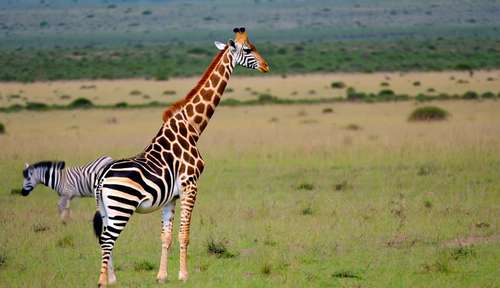}
\hspace{2mm}
\includegraphics[width=0.1\linewidth]{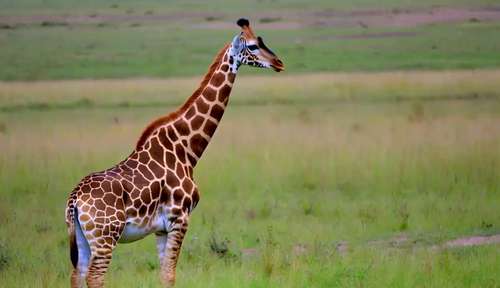}
\includegraphics[width=0.1\linewidth]{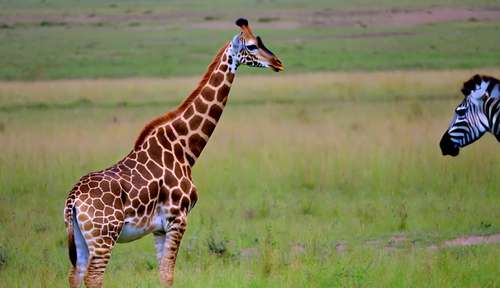}
\includegraphics[width=0.1\linewidth]{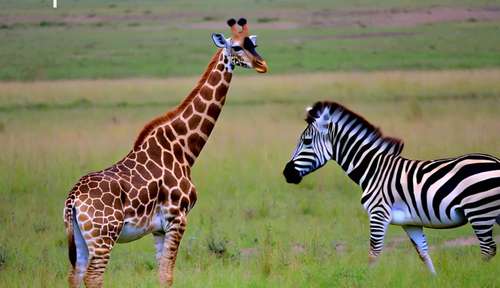}
\includegraphics[width=0.1\linewidth]{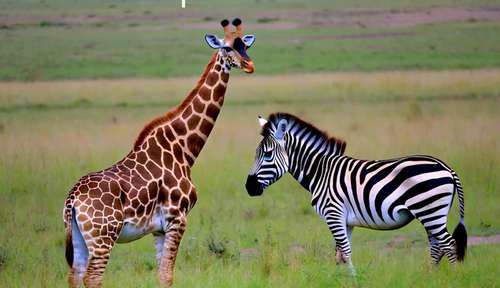}
\\[1.5ex]
\end{minipage}
\caption{Two Objects Text vs Ours.}
\label{fig:two_objects2_text_vs_ours}
\end{figure*}

\begin{figure*}[t]
\centering
\begin{minipage}{\textwidth}
\makebox[2cm][l]{}\makebox[0.40\linewidth]{\small Text}\hspace{2mm}\makebox[0.40\linewidth]{\small Ours}\\[1.5ex]
\begin{minipage}[c]{2cm}\vspace*{0pt}\vfill\raggedright\textit{\small airplane 2nd second}\vfill\end{minipage}
\includegraphics[width=0.1\linewidth]{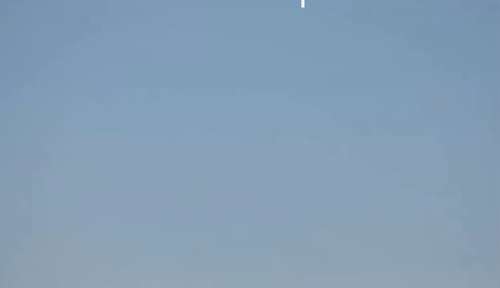}
\includegraphics[width=0.1\linewidth]{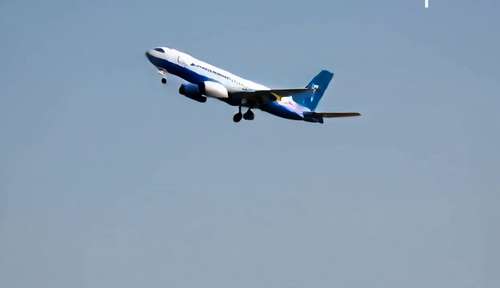}
\includegraphics[width=0.1\linewidth]{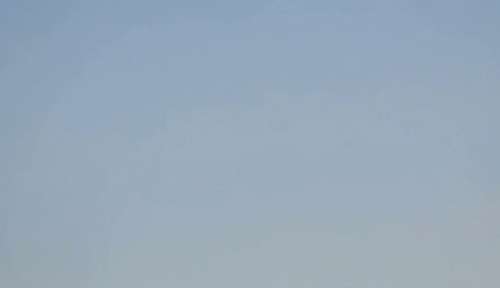}
\includegraphics[width=0.1\linewidth]{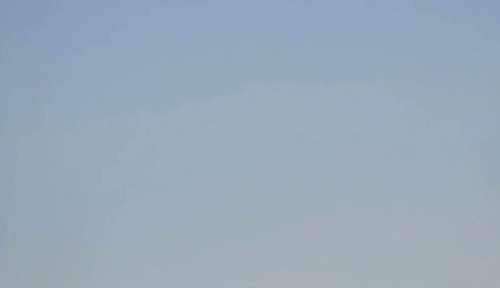}
\hspace{2mm}
\includegraphics[width=0.1\linewidth]{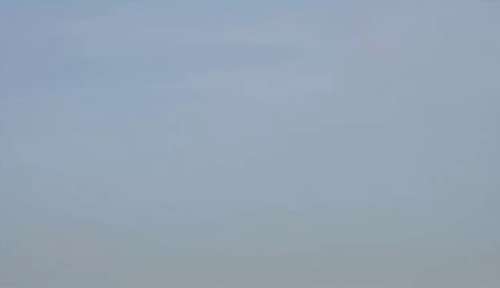}
\includegraphics[width=0.1\linewidth]{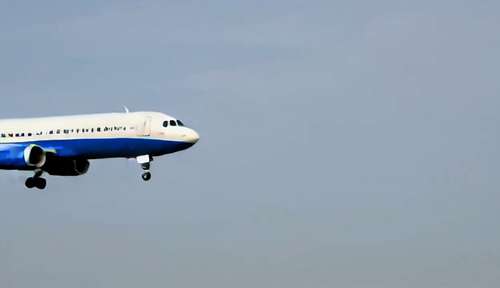}
\includegraphics[width=0.1\linewidth]{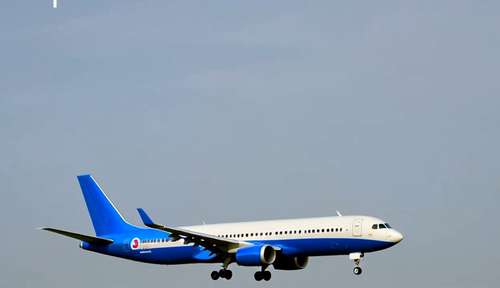}
\includegraphics[width=0.1\linewidth]{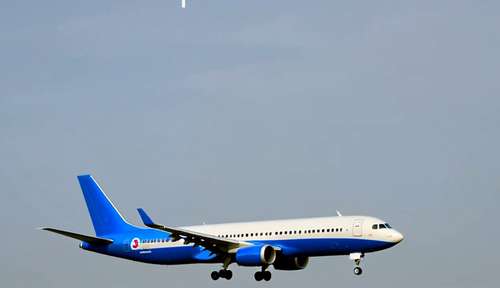}
\\[1.5ex]
\begin{minipage}[c]{2cm}\vspace*{0pt}\vfill\raggedright\textit{\small apple 4th second}\vfill\end{minipage}
\includegraphics[width=0.1\linewidth]{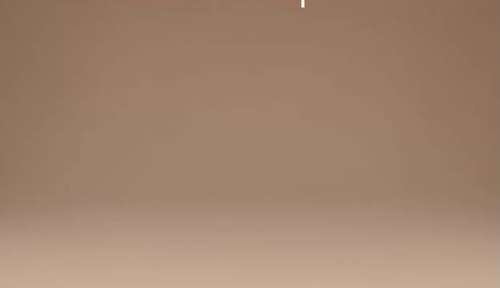}
\includegraphics[width=0.1\linewidth]{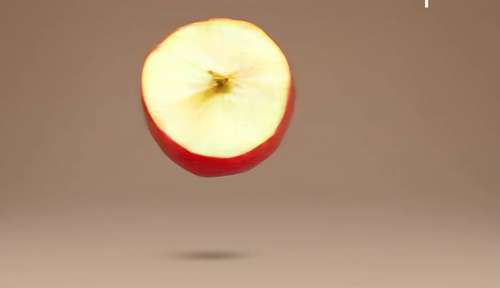}
\includegraphics[width=0.1\linewidth]{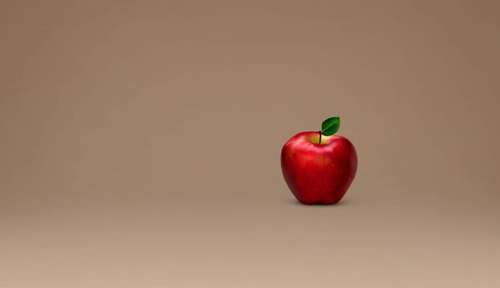}
\includegraphics[width=0.1\linewidth]{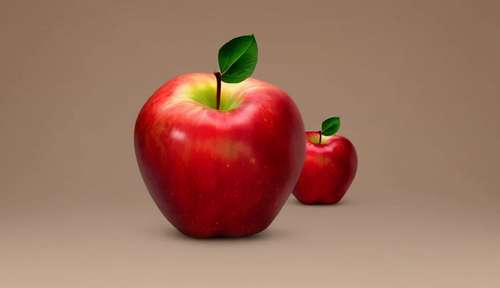}
\hspace{2mm}
\includegraphics[width=0.1\linewidth]{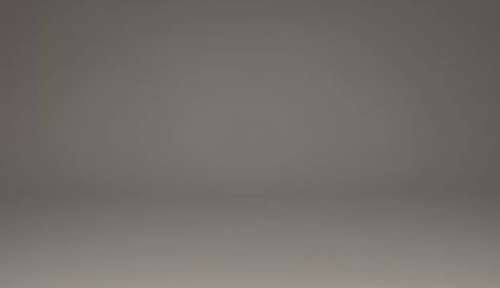}
\includegraphics[width=0.1\linewidth]{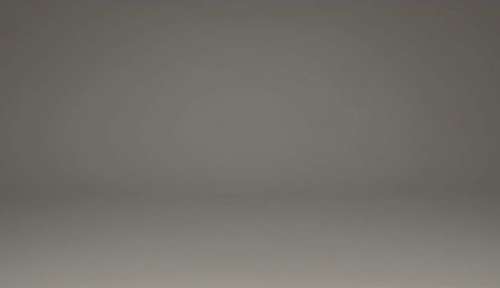}
\includegraphics[width=0.1\linewidth]{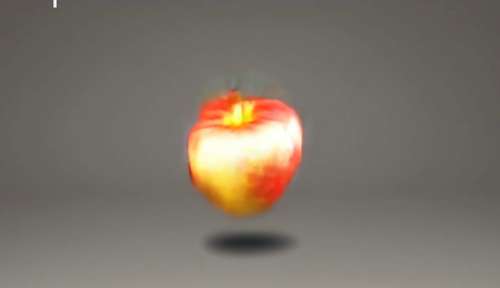}
\includegraphics[width=0.1\linewidth]{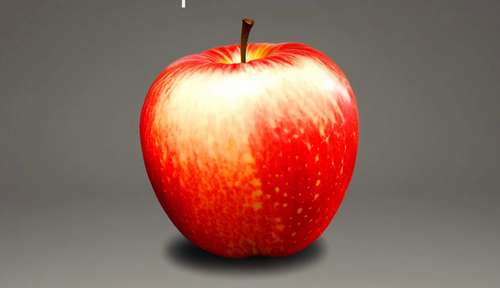}
\\[1.5ex]
\begin{minipage}[c]{2cm}\vspace*{0pt}\vfill\raggedright\textit{\small backpack 3rd second}\vfill\end{minipage}
\includegraphics[width=0.1\linewidth]{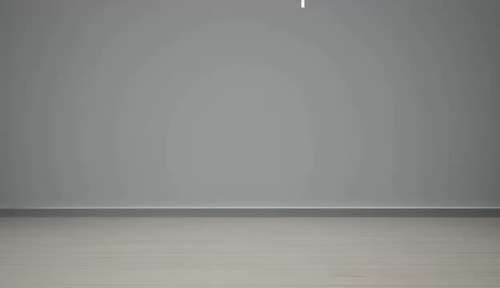}
\includegraphics[width=0.1\linewidth]{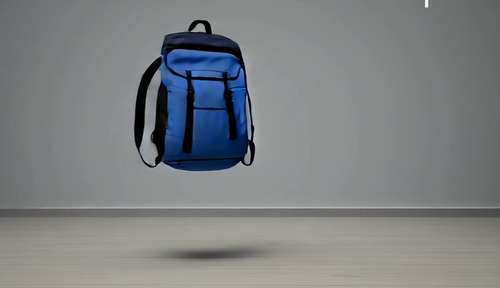}
\includegraphics[width=0.1\linewidth]{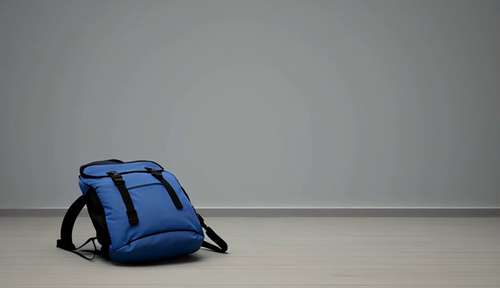}
\includegraphics[width=0.1\linewidth]{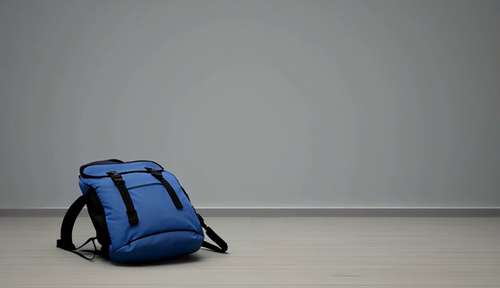}
\hspace{2mm}
\includegraphics[width=0.1\linewidth]{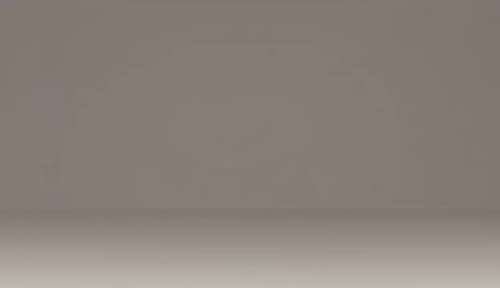}
\includegraphics[width=0.1\linewidth]{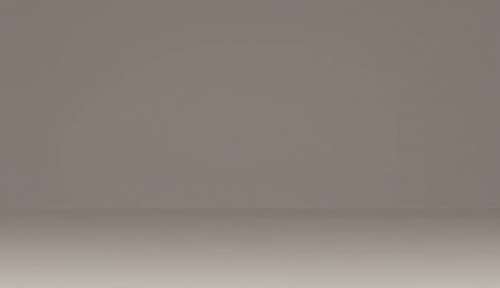}
\includegraphics[width=0.1\linewidth]{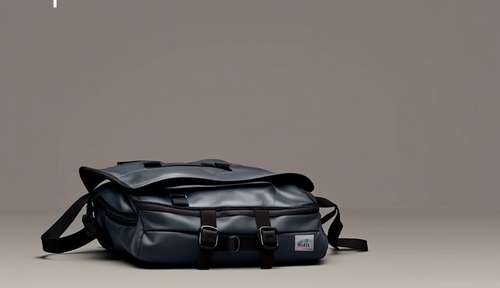}
\includegraphics[width=0.1\linewidth]{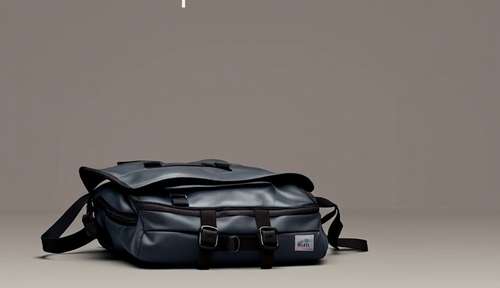}
\\[1.5ex]
\begin{minipage}[c]{2cm}\vspace*{0pt}\vfill\raggedright\textit{\small bench 2nd second}\vfill\end{minipage}
\includegraphics[width=0.1\linewidth]{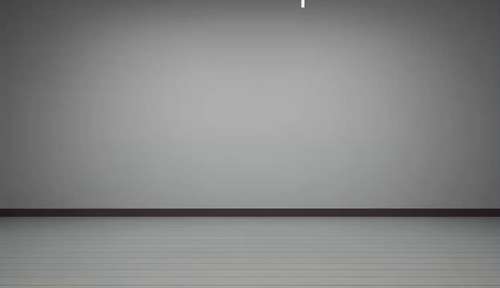}
\includegraphics[width=0.1\linewidth]{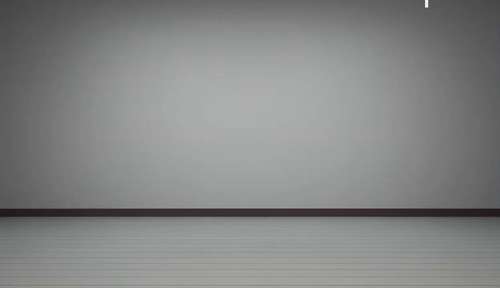}
\includegraphics[width=0.1\linewidth]{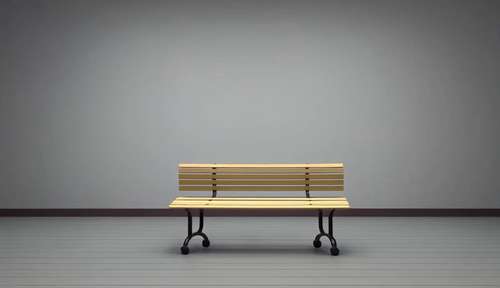}
\includegraphics[width=0.1\linewidth]{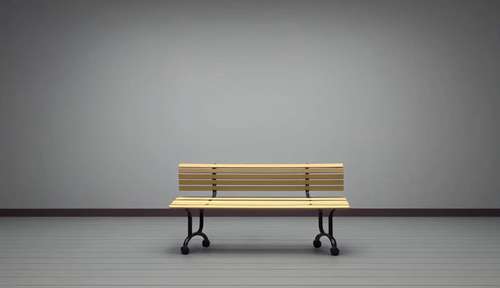}
\hspace{2mm}
\includegraphics[width=0.1\linewidth]{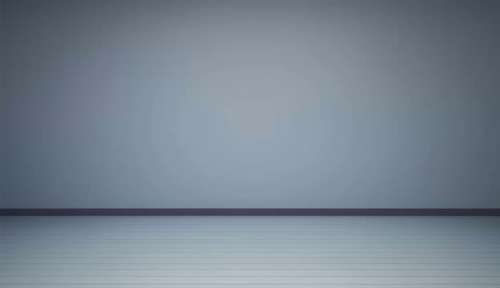}
\includegraphics[width=0.1\linewidth]{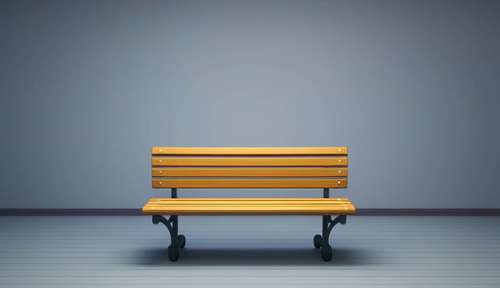}
\includegraphics[width=0.1\linewidth]{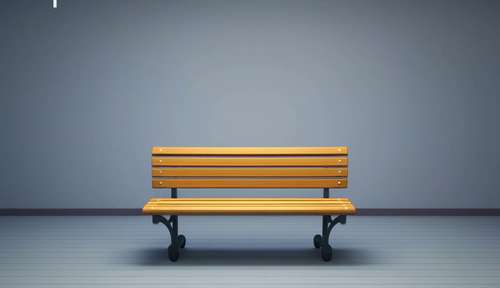}
\includegraphics[width=0.1\linewidth]{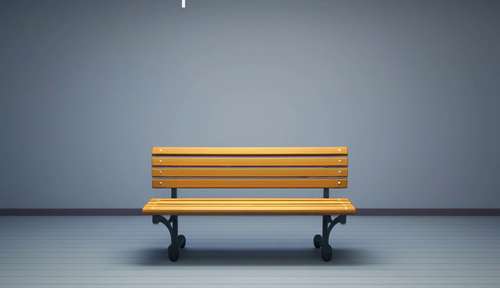}
\\[1.5ex]
\begin{minipage}[c]{2cm}\vspace*{0pt}\vfill\raggedright\textit{\small bicycle 2nd second}\vfill\end{minipage}
\includegraphics[width=0.1\linewidth]{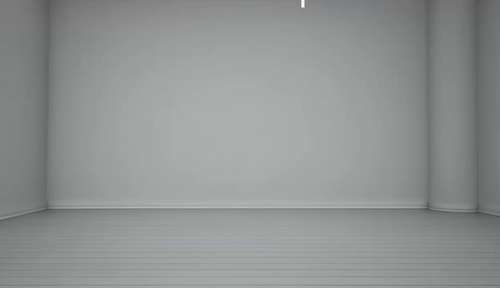}
\includegraphics[width=0.1\linewidth]{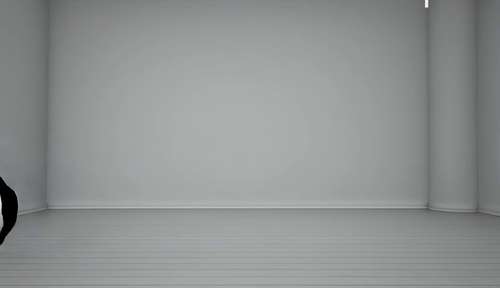}
\includegraphics[width=0.1\linewidth]{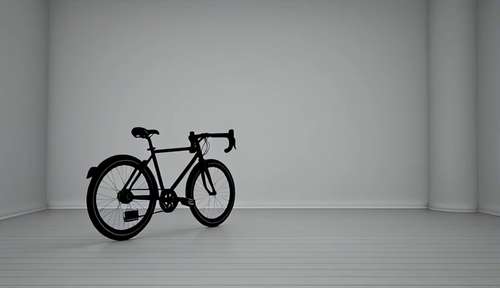}
\includegraphics[width=0.1\linewidth]{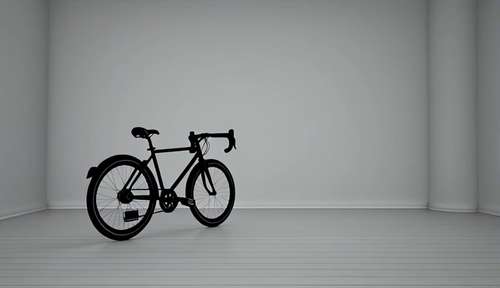}
\hspace{2mm}
\includegraphics[width=0.1\linewidth]{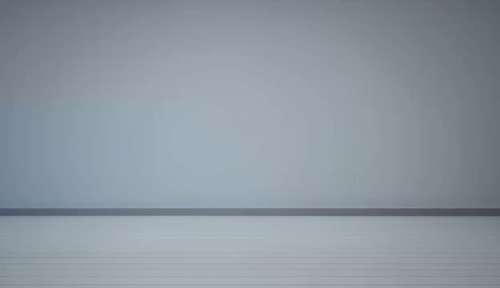}
\includegraphics[width=0.1\linewidth]{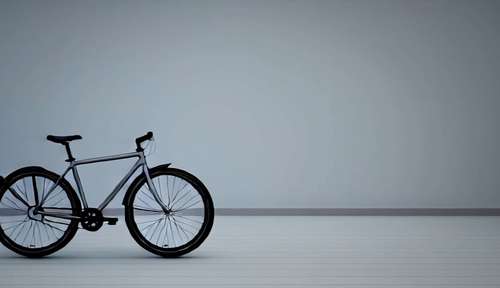}
\includegraphics[width=0.1\linewidth]{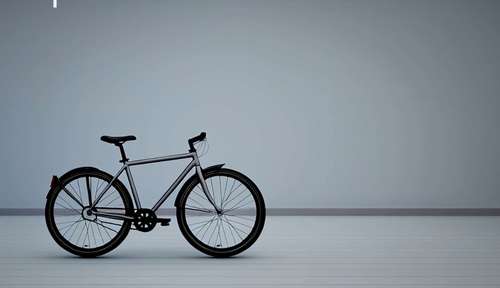}
\includegraphics[width=0.1\linewidth]{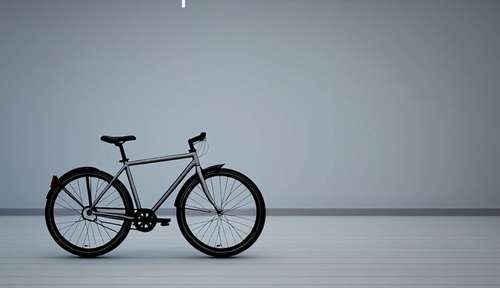}
\\[1.5ex]
\begin{minipage}[c]{2cm}\vspace*{0pt}\vfill\raggedright\textit{\small bowl 4th second}\vfill\end{minipage}
\includegraphics[width=0.1\linewidth]{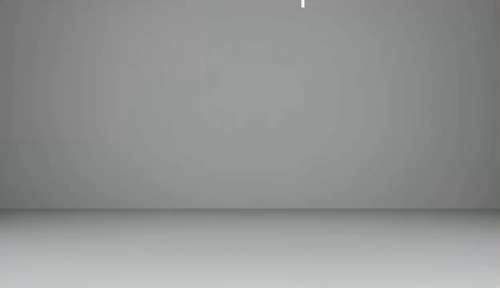}
\includegraphics[width=0.1\linewidth]{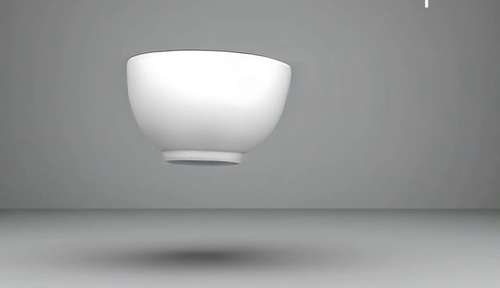}
\includegraphics[width=0.1\linewidth]{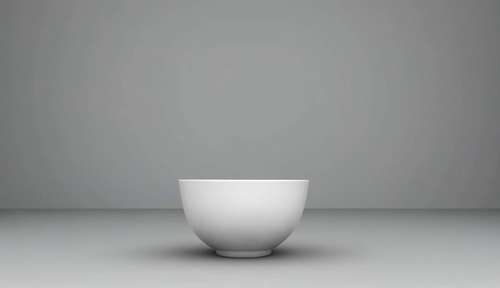}
\includegraphics[width=0.1\linewidth]{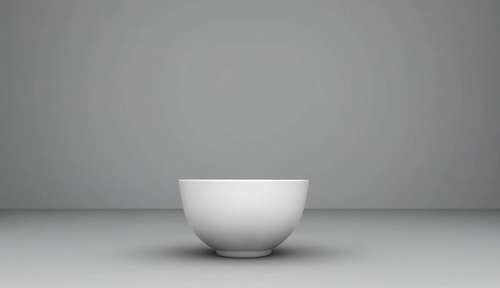}
\hspace{2mm}
\includegraphics[width=0.1\linewidth]{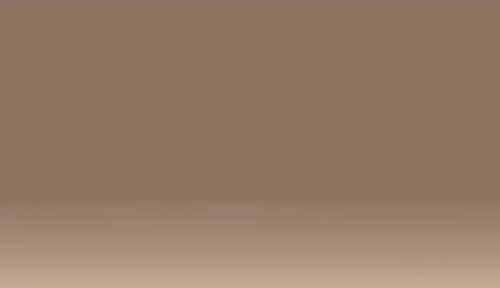}
\includegraphics[width=0.1\linewidth]{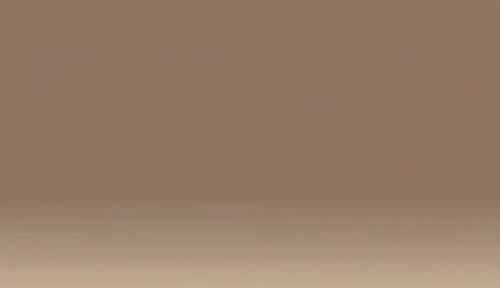}
\includegraphics[width=0.1\linewidth]{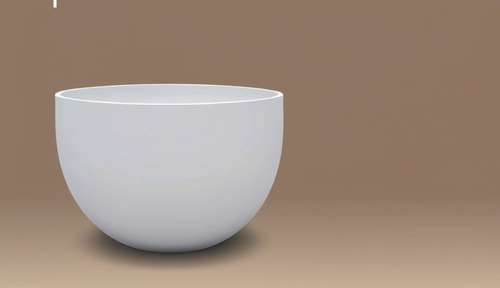}
\includegraphics[width=0.1\linewidth]{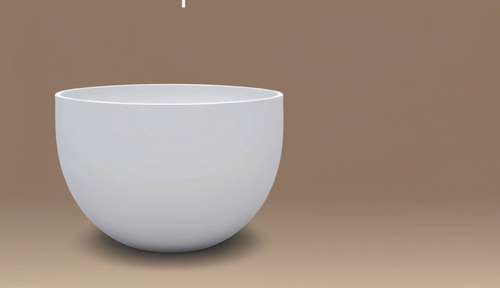}
\\[1.5ex]
\begin{minipage}[c]{2cm}\vspace*{0pt}\vfill\raggedright\textit{\small bus 2nd second}\vfill\end{minipage}
\includegraphics[width=0.1\linewidth]{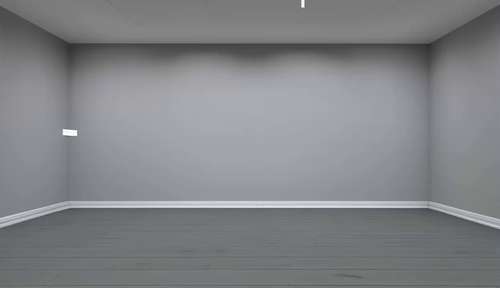}
\includegraphics[width=0.1\linewidth]{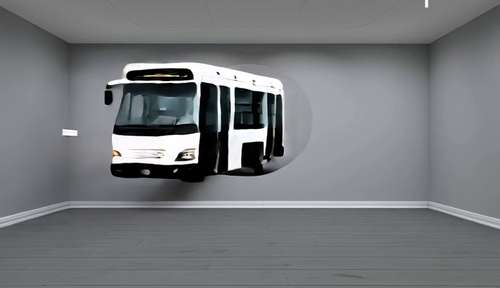}
\includegraphics[width=0.1\linewidth]{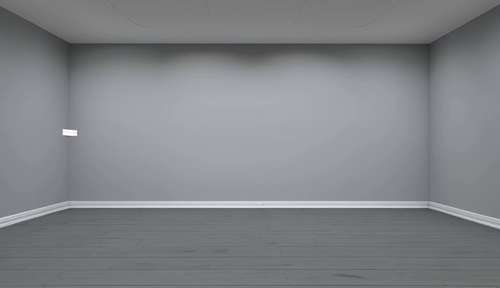}
\includegraphics[width=0.1\linewidth]{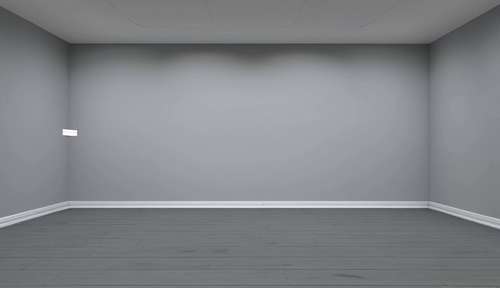}
\hspace{2mm}
\includegraphics[width=0.1\linewidth]{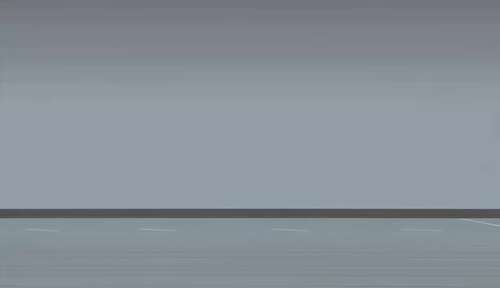}
\includegraphics[width=0.1\linewidth]{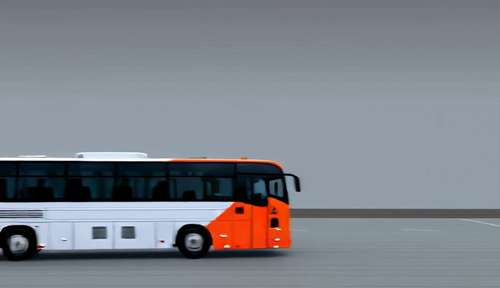}
\includegraphics[width=0.1\linewidth]{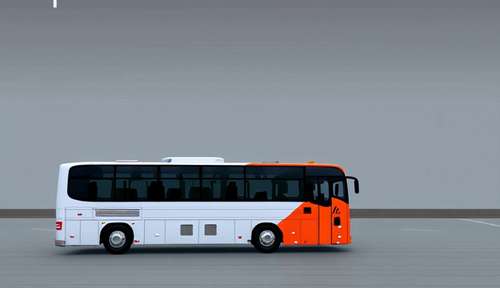}
\includegraphics[width=0.1\linewidth]{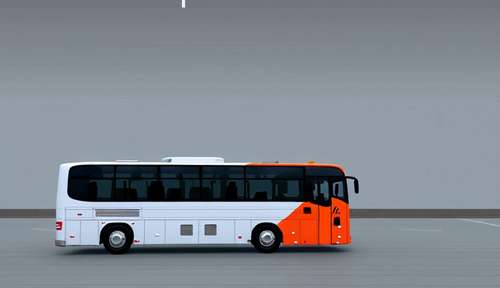}
\\[1.5ex]
\begin{minipage}[c]{2cm}\vspace*{0pt}\vfill\raggedright\textit{\small cell phone last second}\vfill\end{minipage}
\includegraphics[width=0.1\linewidth]{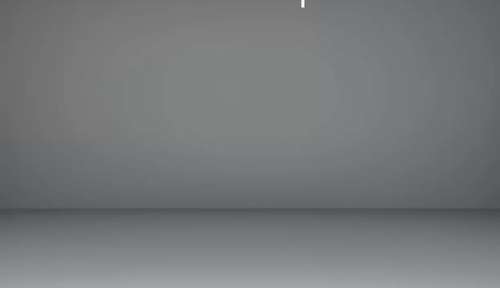}
\includegraphics[width=0.1\linewidth]{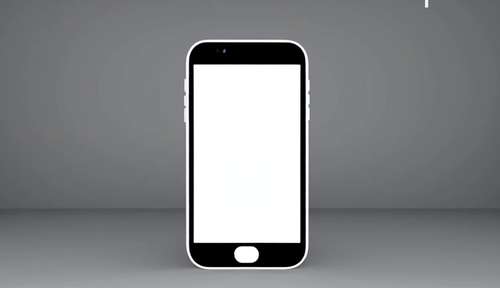}
\includegraphics[width=0.1\linewidth]{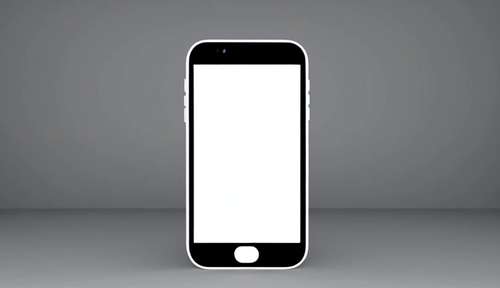}
\includegraphics[width=0.1\linewidth]{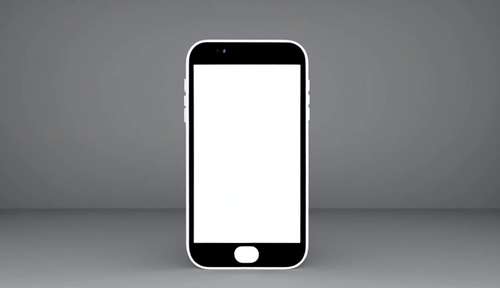}
\hspace{2mm}
\includegraphics[width=0.1\linewidth]{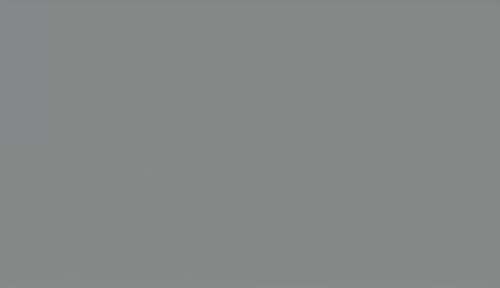}
\includegraphics[width=0.1\linewidth]{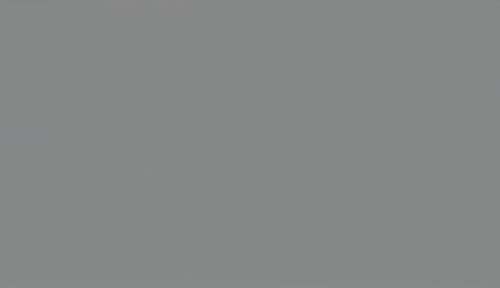}
\includegraphics[width=0.1\linewidth]{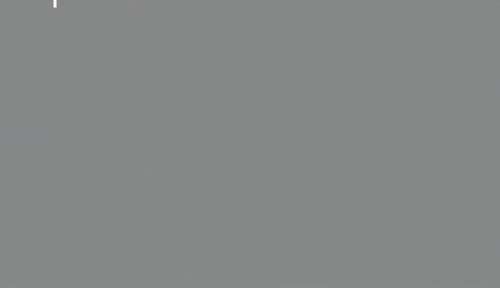}
\includegraphics[width=0.1\linewidth]{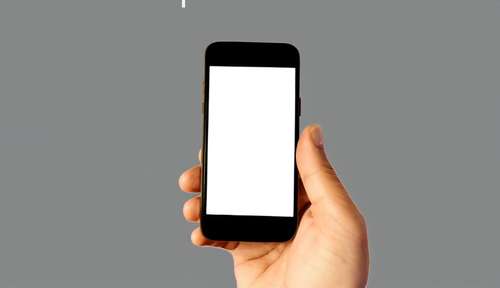}
\\[1.5ex]
\begin{minipage}[c]{2cm}\vspace*{0pt}\vfill\raggedright\textit{\small clock last second}\vfill\end{minipage}
\includegraphics[width=0.1\linewidth]{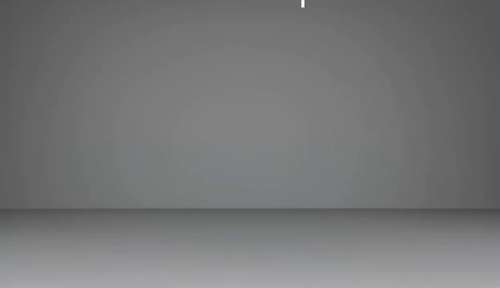}
\includegraphics[width=0.1\linewidth]{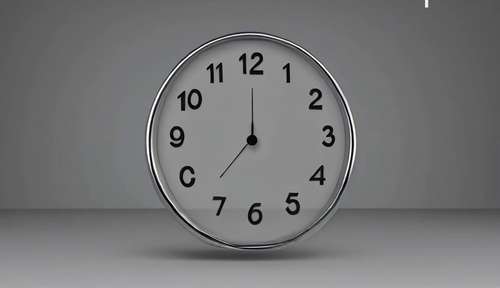}
\includegraphics[width=0.1\linewidth]{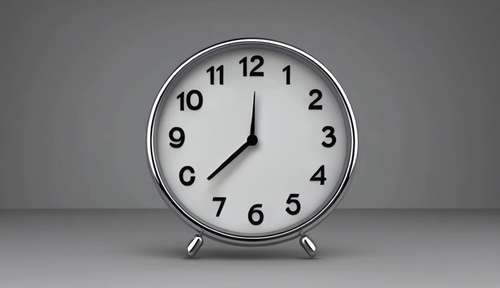}
\includegraphics[width=0.1\linewidth]{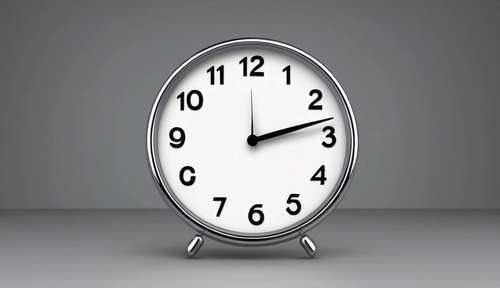}
\hspace{2mm}
\includegraphics[width=0.1\linewidth]{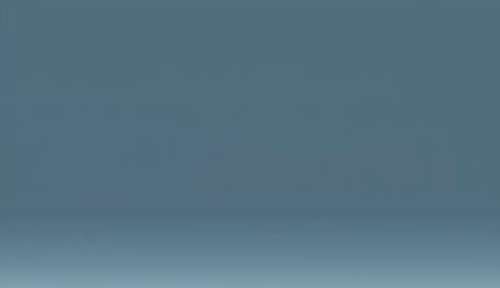}
\includegraphics[width=0.1\linewidth]{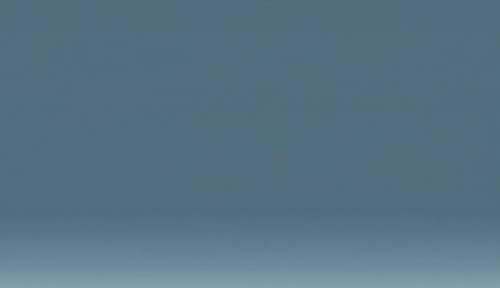}
\includegraphics[width=0.1\linewidth]{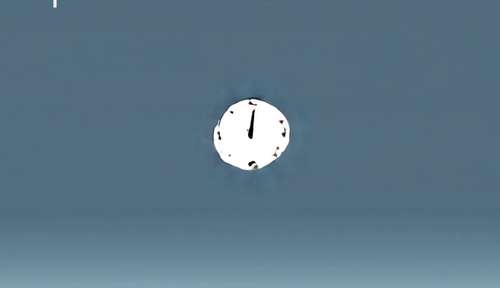}
\includegraphics[width=0.1\linewidth]{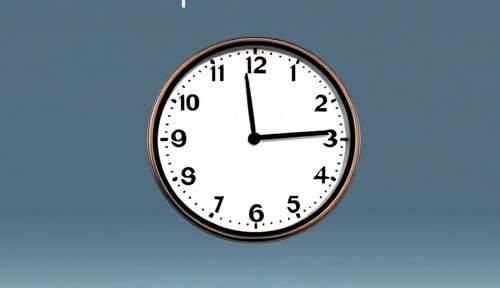}
\\[1.5ex]
\begin{minipage}[c]{2cm}\vspace*{0pt}\vfill\raggedright\textit{\small cow 2nd second}\vfill\end{minipage}
\includegraphics[width=0.1\linewidth]{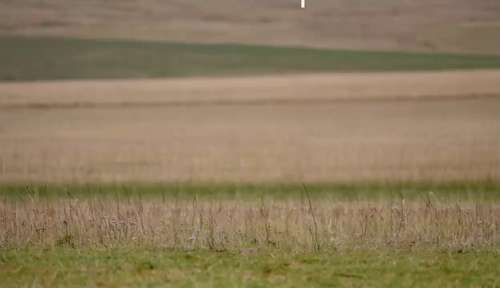}
\includegraphics[width=0.1\linewidth]{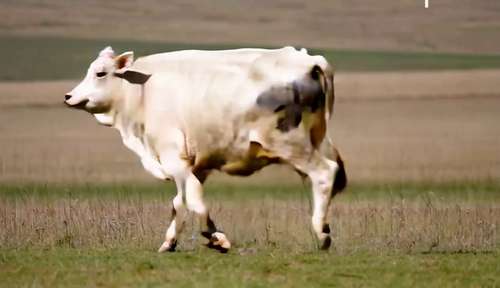}
\includegraphics[width=0.1\linewidth]{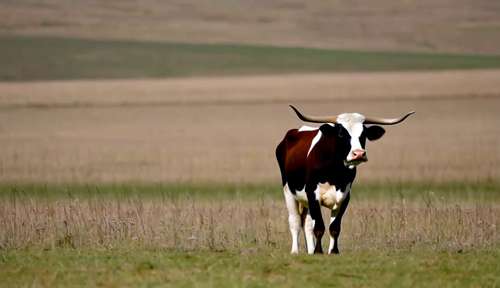}
\includegraphics[width=0.1\linewidth]{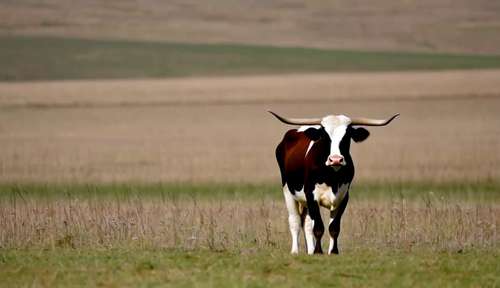}
\hspace{2mm}
\includegraphics[width=0.1\linewidth]{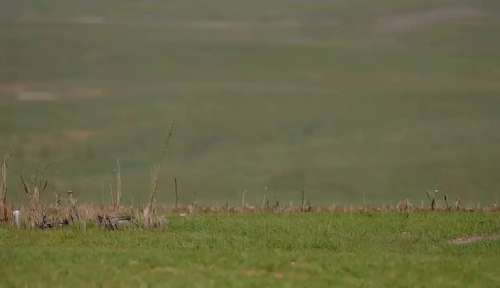}
\includegraphics[width=0.1\linewidth]{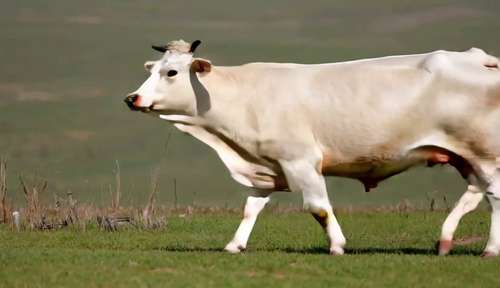}
\includegraphics[width=0.1\linewidth]{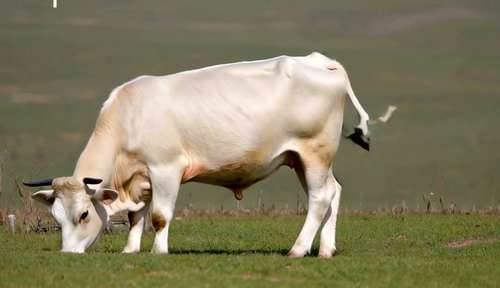}
\includegraphics[width=0.1\linewidth]{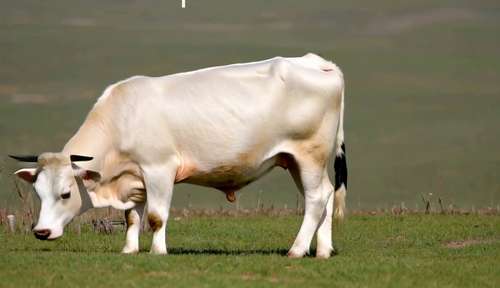}
\\[1.5ex]
\begin{minipage}[c]{2cm}\vspace*{0pt}\vfill\raggedright\textit{\small dining table last second}\vfill\end{minipage}
\includegraphics[width=0.1\linewidth]{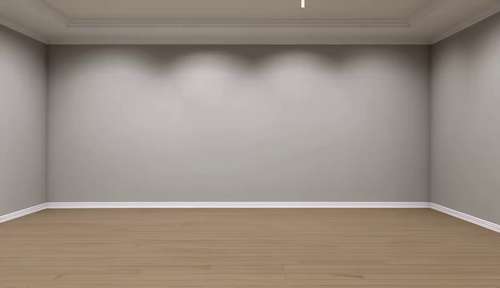}
\includegraphics[width=0.1\linewidth]{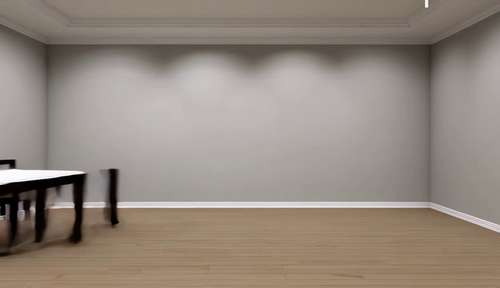}
\includegraphics[width=0.1\linewidth]{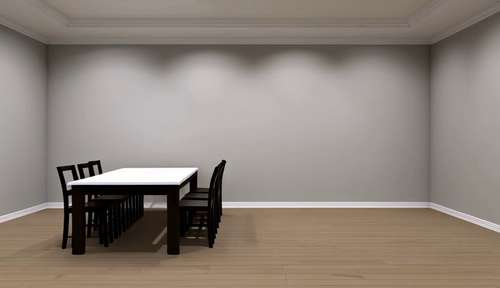}
\includegraphics[width=0.1\linewidth]{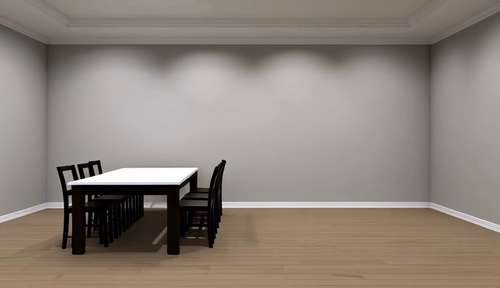}
\hspace{2mm}
\includegraphics[width=0.1\linewidth]{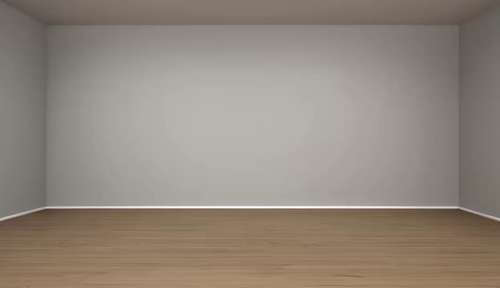}
\includegraphics[width=0.1\linewidth]{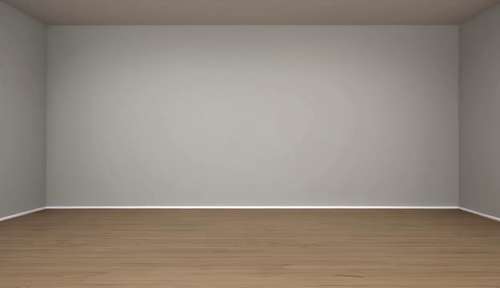}
\includegraphics[width=0.1\linewidth]{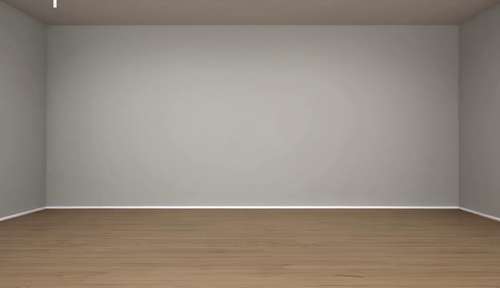}
\includegraphics[width=0.1\linewidth]{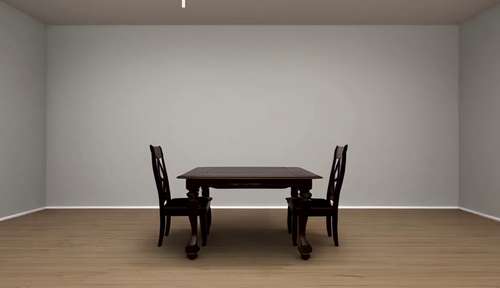}
\\[1.5ex]
\begin{minipage}[c]{2cm}\vspace*{0pt}\vfill\raggedright\textit{\small dog last second}\vfill\end{minipage}
\includegraphics[width=0.1\linewidth]{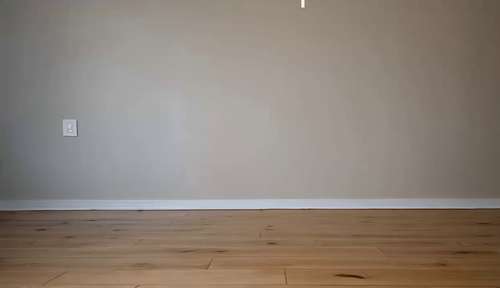}
\includegraphics[width=0.1\linewidth]{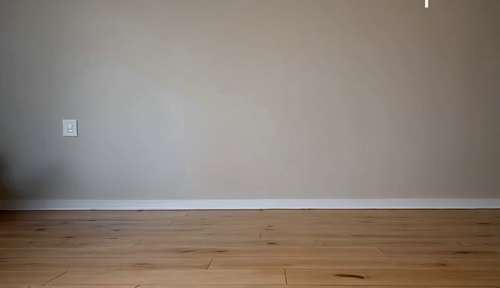}
\includegraphics[width=0.1\linewidth]{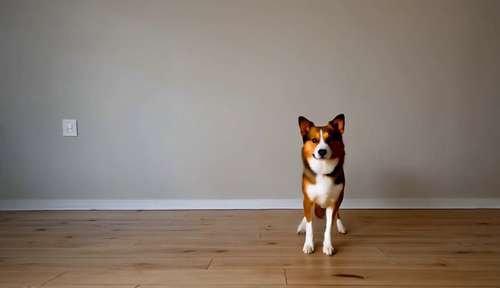}
\includegraphics[width=0.1\linewidth]{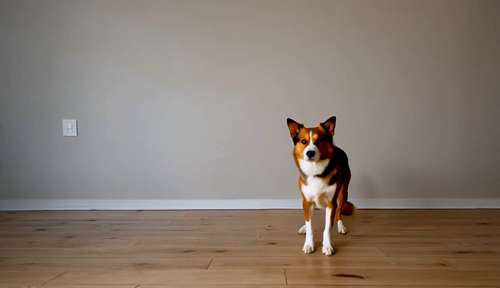}
\hspace{2mm}
\includegraphics[width=0.1\linewidth]{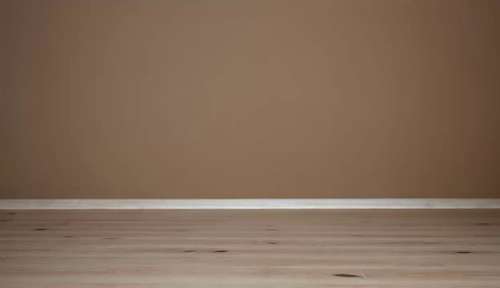}
\includegraphics[width=0.1\linewidth]{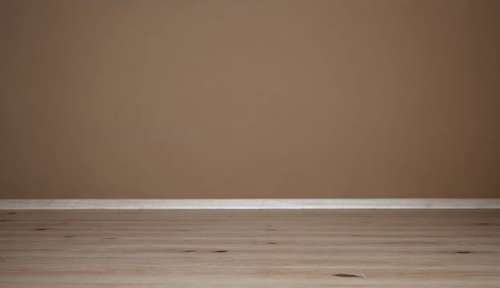}
\includegraphics[width=0.1\linewidth]{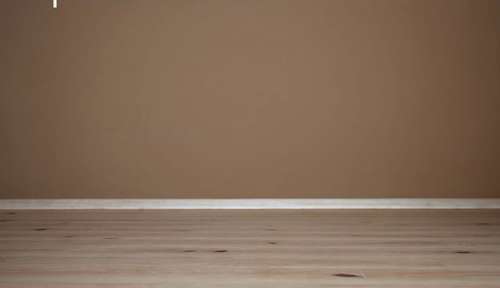}
\includegraphics[width=0.1\linewidth]{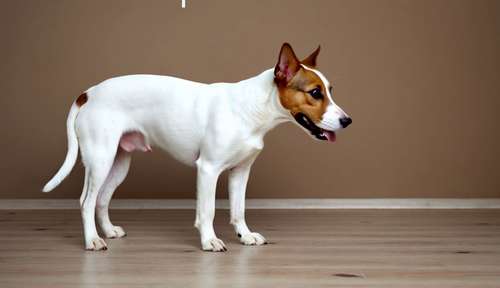}
\\[1.5ex]
\begin{minipage}[c]{2cm}\vspace*{0pt}\vfill\raggedright\textit{\small donut 4th second}\vfill\end{minipage}
\includegraphics[width=0.1\linewidth]{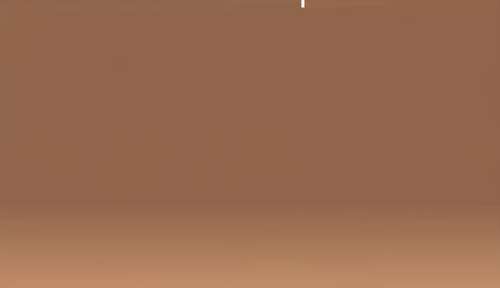}
\includegraphics[width=0.1\linewidth]{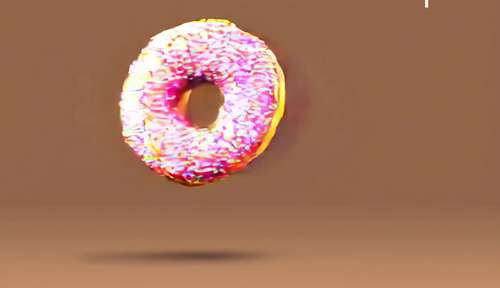}
\includegraphics[width=0.1\linewidth]{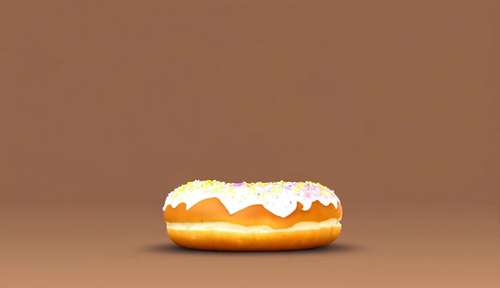}
\includegraphics[width=0.1\linewidth]{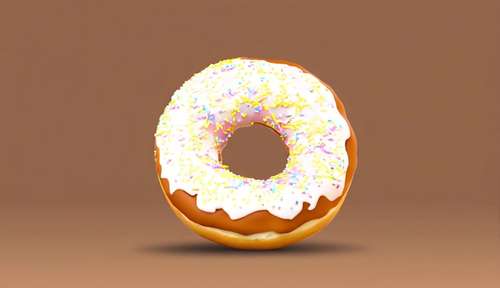}
\hspace{2mm}
\includegraphics[width=0.1\linewidth]{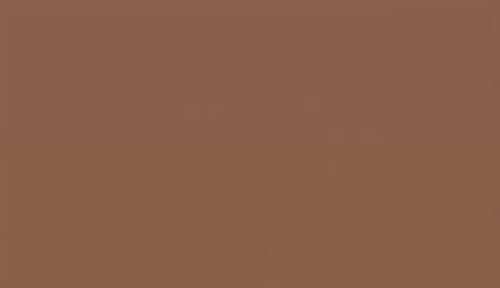}
\includegraphics[width=0.1\linewidth]{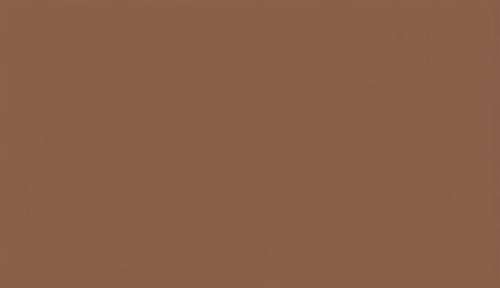}
\includegraphics[width=0.1\linewidth]{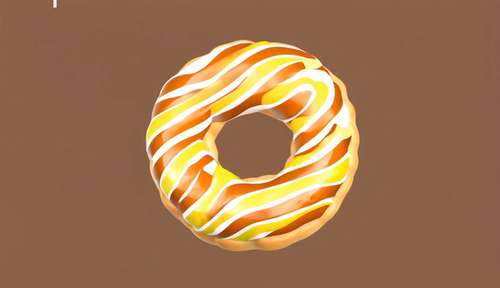}
\includegraphics[width=0.1\linewidth]{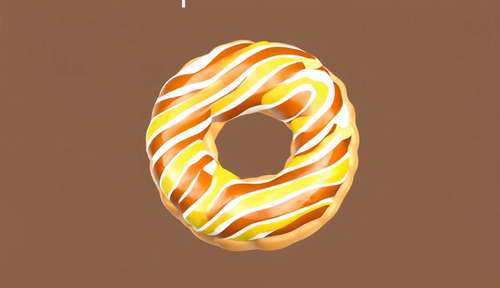}
\\[1.5ex]
\begin{minipage}[c]{2cm}\vspace*{0pt}\vfill\raggedright\textit{\small fork 4th second}\vfill\end{minipage}
\includegraphics[width=0.1\linewidth]{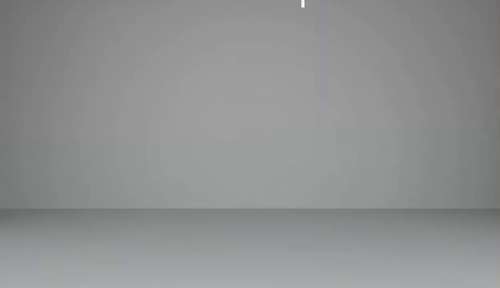}
\includegraphics[width=0.1\linewidth]{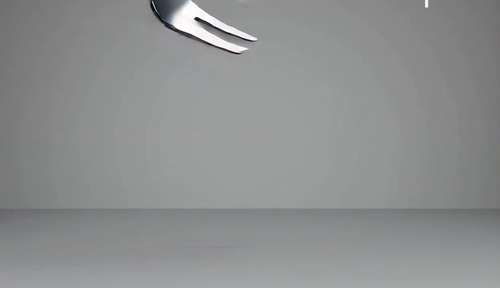}
\includegraphics[width=0.1\linewidth]{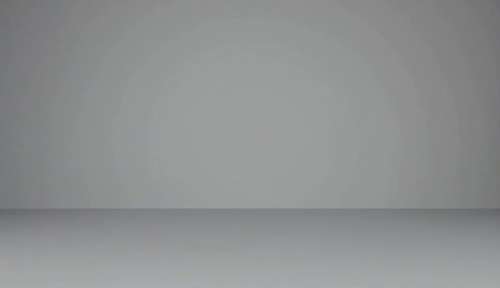}
\includegraphics[width=0.1\linewidth]{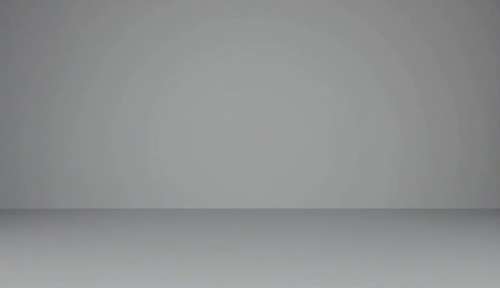}
\hspace{2mm}
\includegraphics[width=0.1\linewidth]{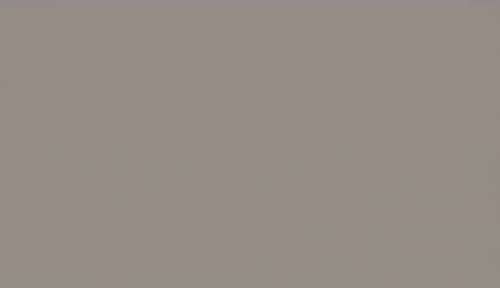}
\includegraphics[width=0.1\linewidth]{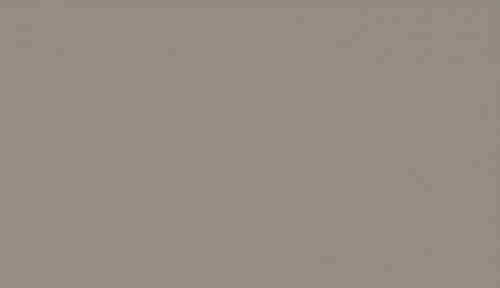}
\includegraphics[width=0.1\linewidth]{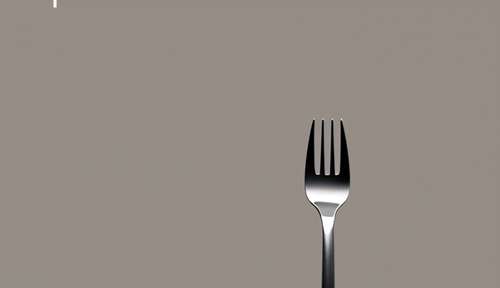}
\includegraphics[width=0.1\linewidth]{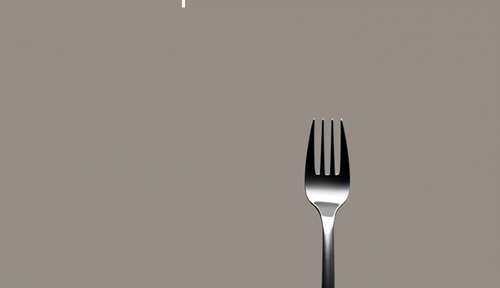}
\\[1.5ex]
\end{minipage}
\caption{One Object Text vs Ours.}
\label{fig:one_object_text_vs_ours}
\end{figure*}

\begin{figure*}[t]
\centering
\begin{minipage}{\textwidth}
\makebox[2cm][l]{}\makebox[0.40\linewidth]{\small Text}\hspace{2mm}\makebox[0.40\linewidth]{\small Ours}\\[1.5ex]
\begin{minipage}[c]{2cm}\vspace*{0pt}\vfill\raggedright\textit{\small hair drier last second}\vfill\end{minipage}
\includegraphics[width=0.1\linewidth]{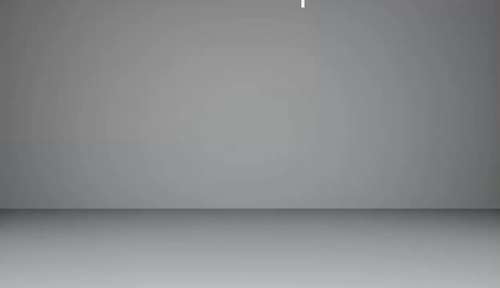}
\includegraphics[width=0.1\linewidth]{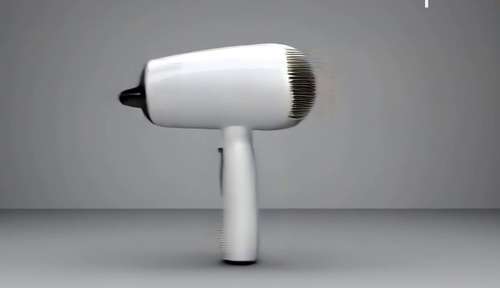}
\includegraphics[width=0.1\linewidth]{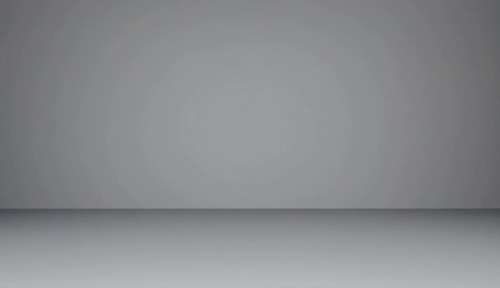}
\includegraphics[width=0.1\linewidth]{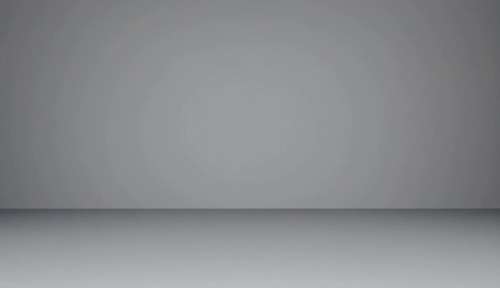}
\hspace{2mm}
\includegraphics[width=0.1\linewidth]{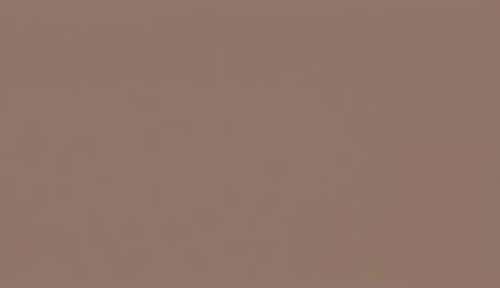}
\includegraphics[width=0.1\linewidth]{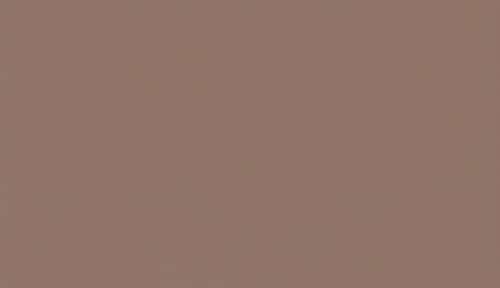}
\includegraphics[width=0.1\linewidth]{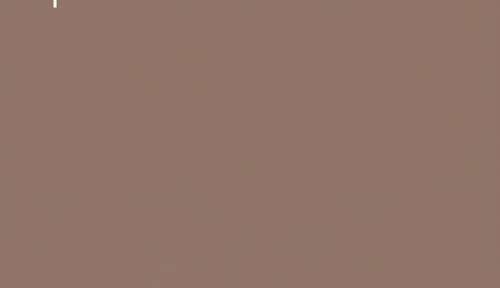}
\includegraphics[width=0.1\linewidth]{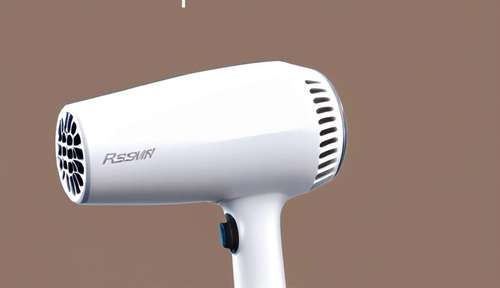}
\\[1.5ex]
\begin{minipage}[c]{2cm}\vspace*{0pt}\vfill\raggedright\textit{\small handbag 3rd second}\vfill\end{minipage}
\includegraphics[width=0.1\linewidth]{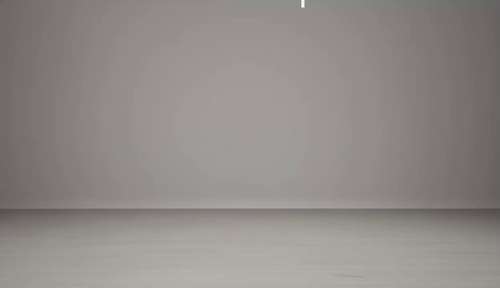}
\includegraphics[width=0.1\linewidth]{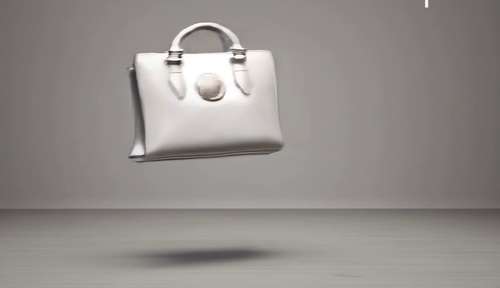}
\includegraphics[width=0.1\linewidth]{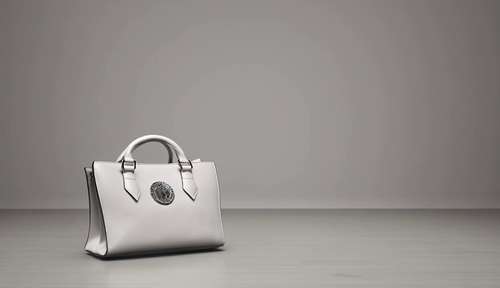}
\includegraphics[width=0.1\linewidth]{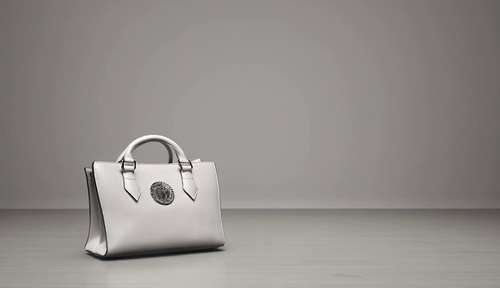}
\hspace{2mm}
\includegraphics[width=0.1\linewidth]{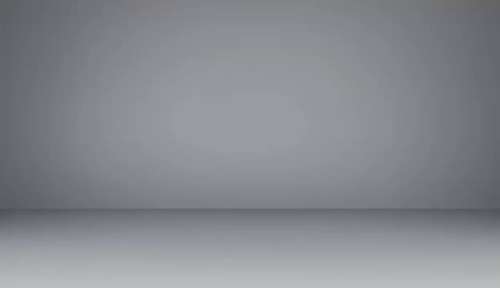}
\includegraphics[width=0.1\linewidth]{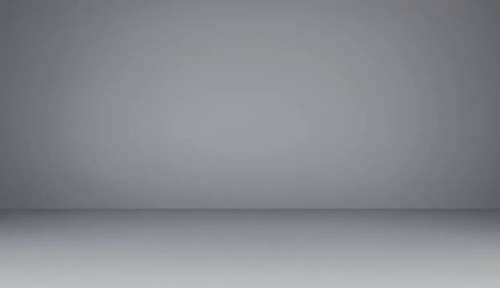}
\includegraphics[width=0.1\linewidth]{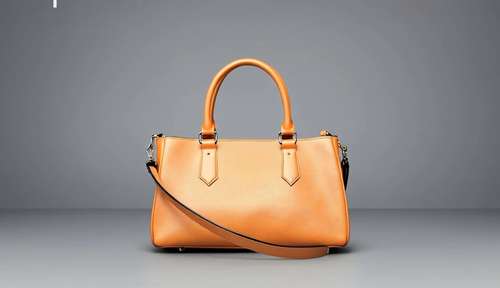}
\includegraphics[width=0.1\linewidth]{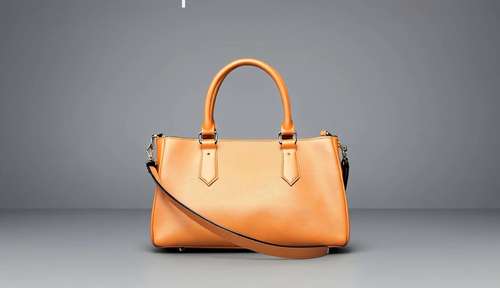}
\\[1.5ex]
\begin{minipage}[c]{2cm}\vspace*{0pt}\vfill\raggedright\textit{\small kite 3rd second}\vfill\end{minipage}
\includegraphics[width=0.1\linewidth]{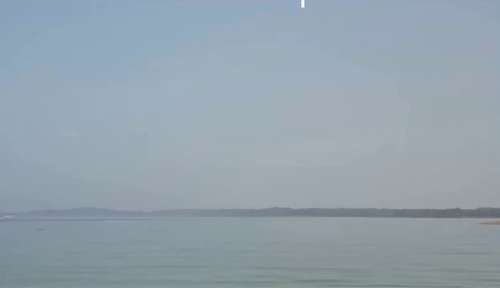}
\includegraphics[width=0.1\linewidth]{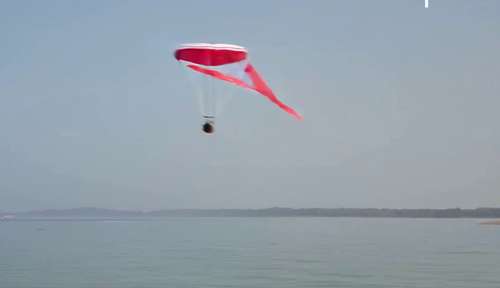}
\includegraphics[width=0.1\linewidth]{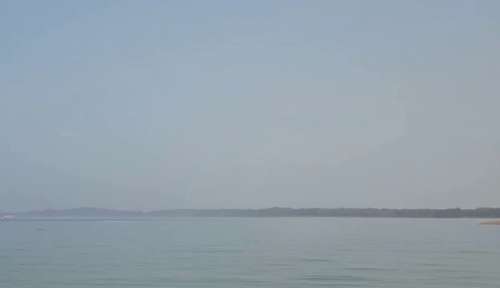}
\includegraphics[width=0.1\linewidth]{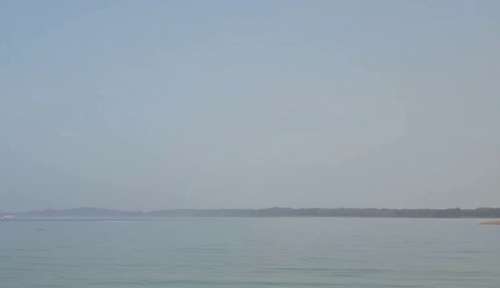}
\hspace{2mm}
\includegraphics[width=0.1\linewidth]{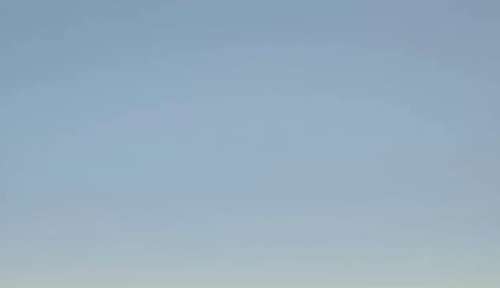}
\includegraphics[width=0.1\linewidth]{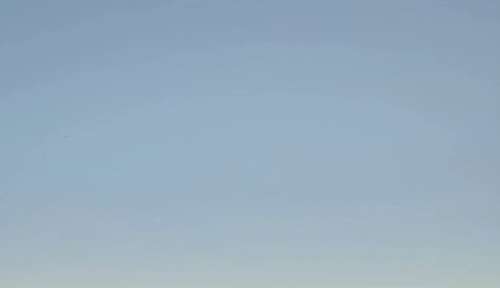}
\includegraphics[width=0.1\linewidth]{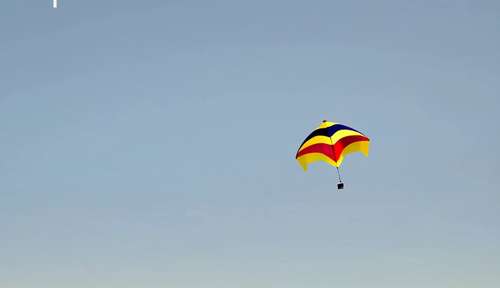}
\includegraphics[width=0.1\linewidth]{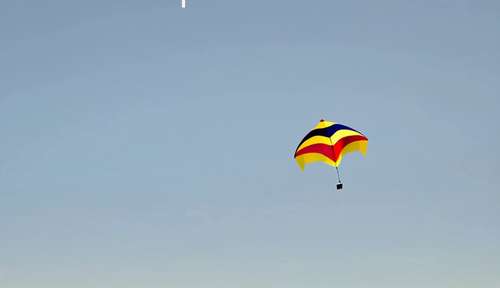}
\\[1.5ex]
\begin{minipage}[c]{2cm}\vspace*{0pt}\vfill\raggedright\textit{\small knife 4th second}\vfill\end{minipage}
\includegraphics[width=0.1\linewidth]{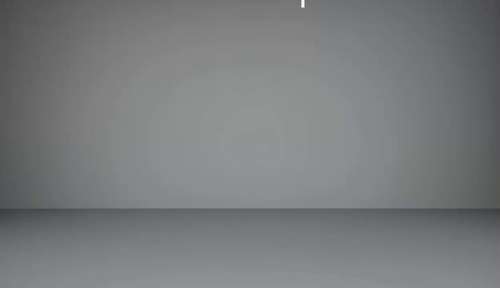}
\includegraphics[width=0.1\linewidth]{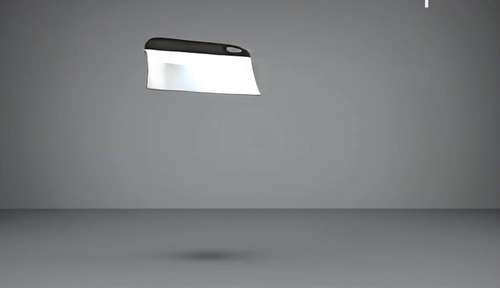}
\includegraphics[width=0.1\linewidth]{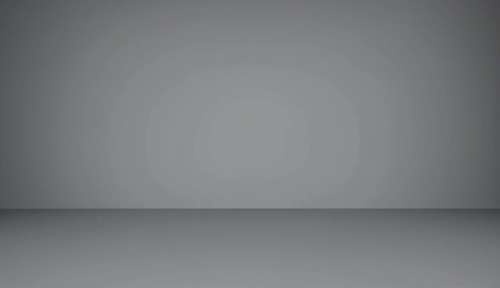}
\includegraphics[width=0.1\linewidth]{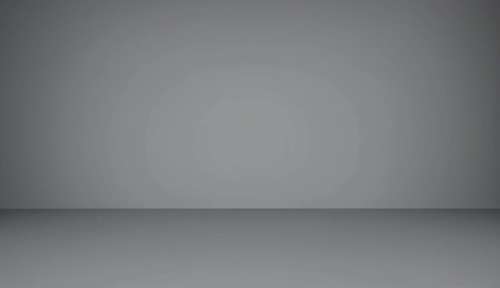}
\hspace{2mm}
\includegraphics[width=0.1\linewidth]{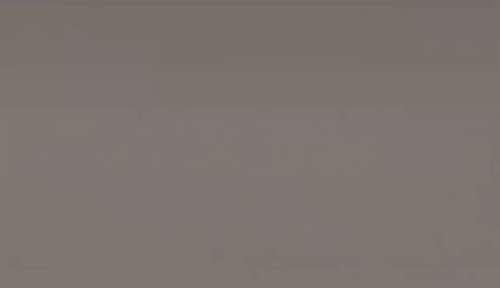}
\includegraphics[width=0.1\linewidth]{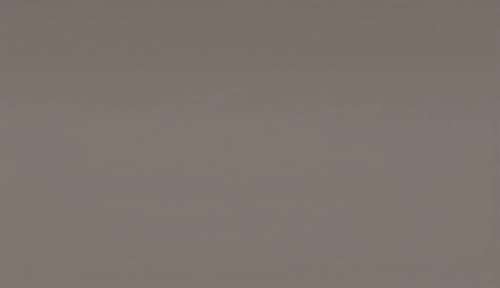}
\includegraphics[width=0.1\linewidth]{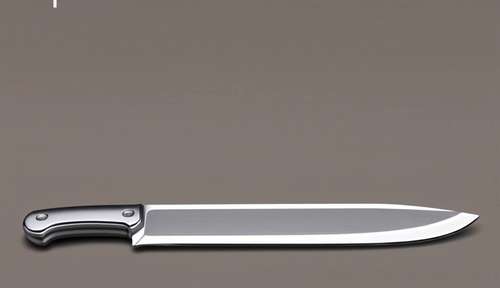}
\includegraphics[width=0.1\linewidth]{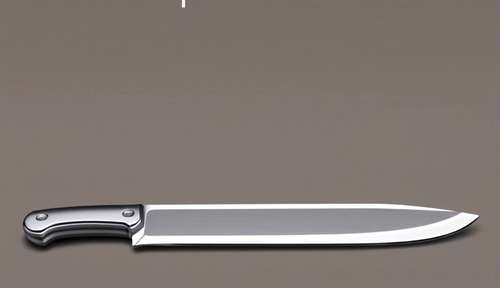}
\\[1.5ex]
\begin{minipage}[c]{2cm}\vspace*{0pt}\vfill\raggedright\textit{\small scissors last second}\vfill\end{minipage}
\includegraphics[width=0.1\linewidth]{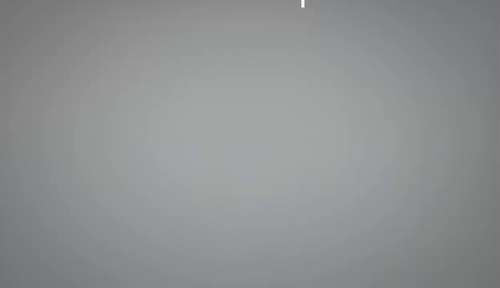}
\includegraphics[width=0.1\linewidth]{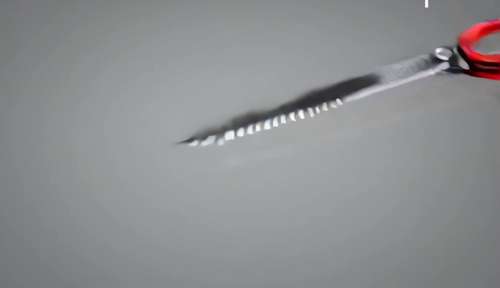}
\includegraphics[width=0.1\linewidth]{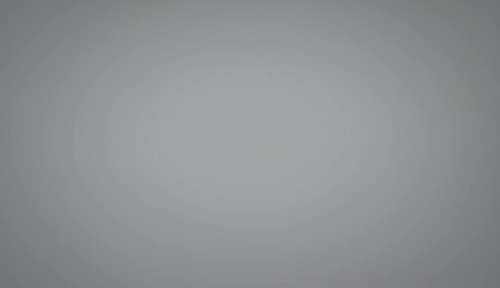}
\includegraphics[width=0.1\linewidth]{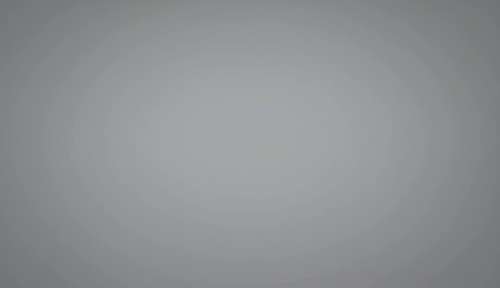}
\hspace{2mm}
\includegraphics[width=0.1\linewidth]{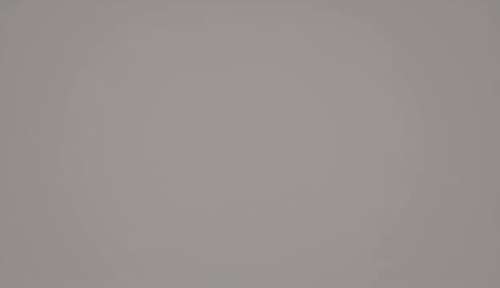}
\includegraphics[width=0.1\linewidth]{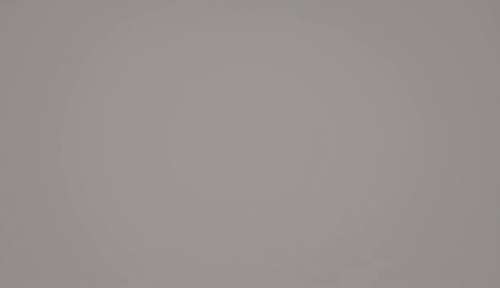}
\includegraphics[width=0.1\linewidth]{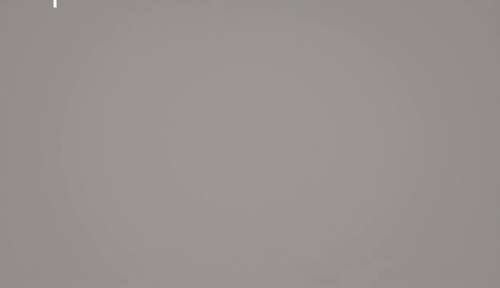}
\includegraphics[width=0.1\linewidth]{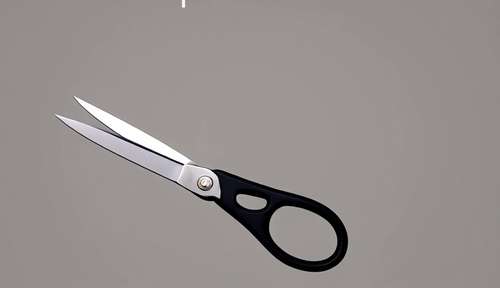}
\\[1.5ex]
\begin{minipage}[c]{2cm}\vspace*{0pt}\vfill\raggedright\textit{\small sheep 2nd second}\vfill\end{minipage}
\includegraphics[width=0.1\linewidth]{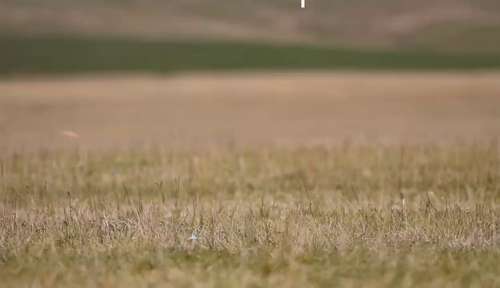}
\includegraphics[width=0.1\linewidth]{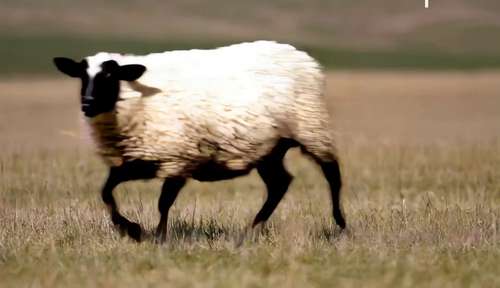}
\includegraphics[width=0.1\linewidth]{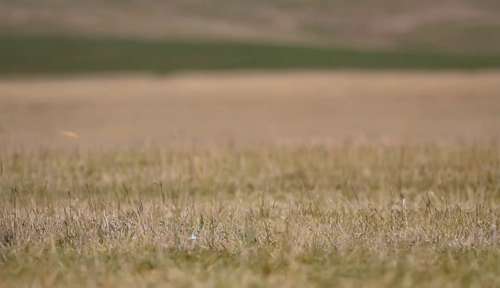}
\includegraphics[width=0.1\linewidth]{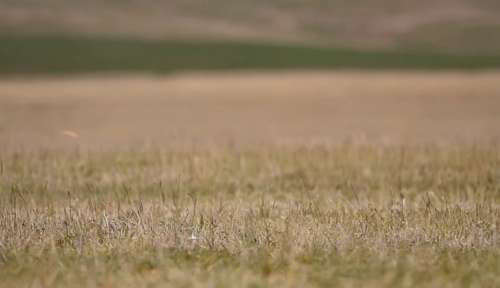}
\hspace{2mm}
\includegraphics[width=0.1\linewidth]{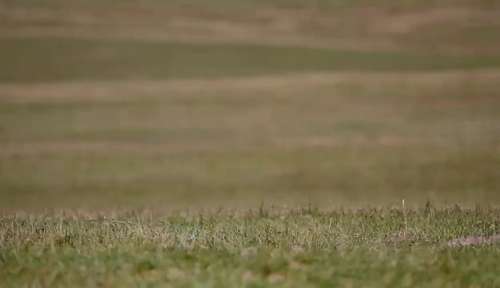}
\includegraphics[width=0.1\linewidth]{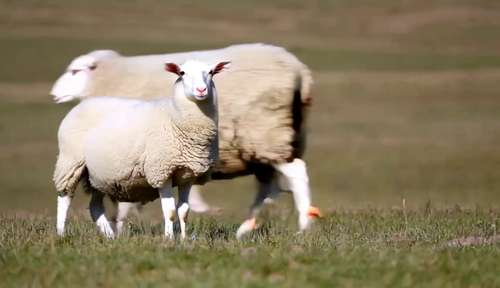}
\includegraphics[width=0.1\linewidth]{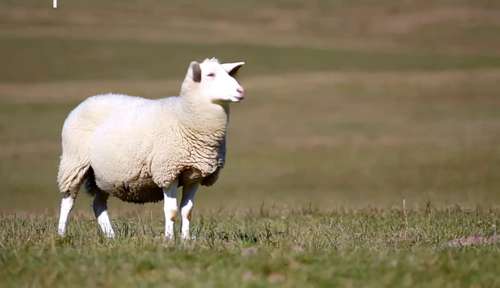}
\includegraphics[width=0.1\linewidth]{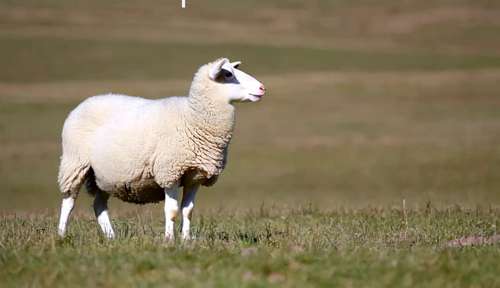}
\\[1.5ex]
\begin{minipage}[c]{2cm}\vspace*{0pt}\vfill\raggedright\textit{\small skateboard 3rd second}\vfill\end{minipage}
\includegraphics[width=0.1\linewidth]{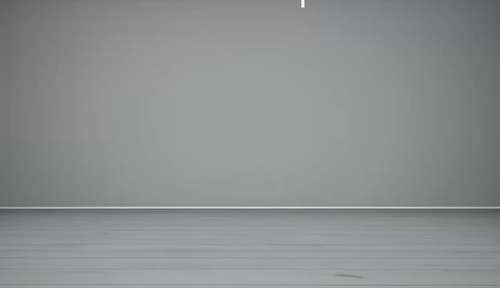}
\includegraphics[width=0.1\linewidth]{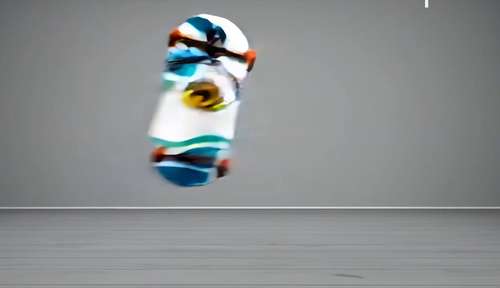}
\includegraphics[width=0.1\linewidth]{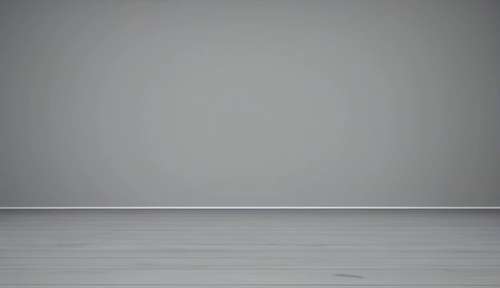}
\includegraphics[width=0.1\linewidth]{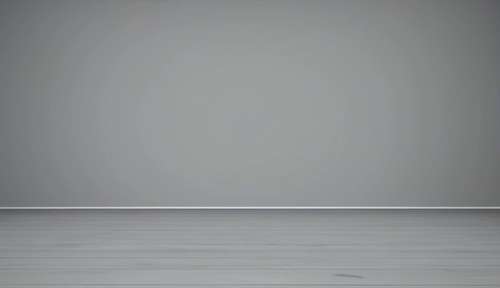}
\hspace{2mm}
\includegraphics[width=0.1\linewidth]{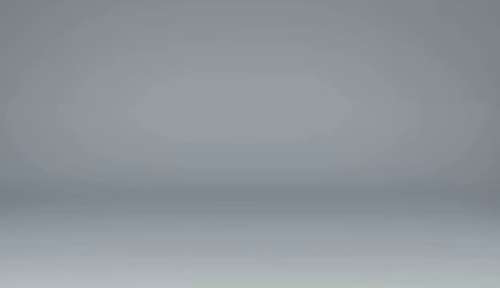}
\includegraphics[width=0.1\linewidth]{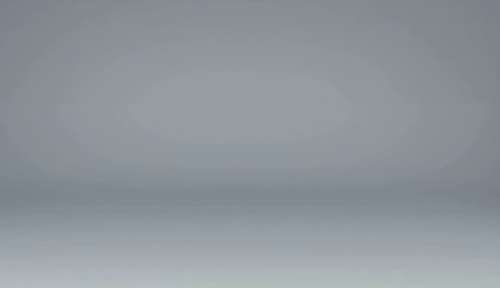}
\includegraphics[width=0.1\linewidth]{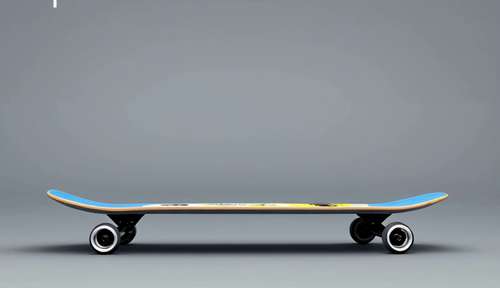}
\includegraphics[width=0.1\linewidth]{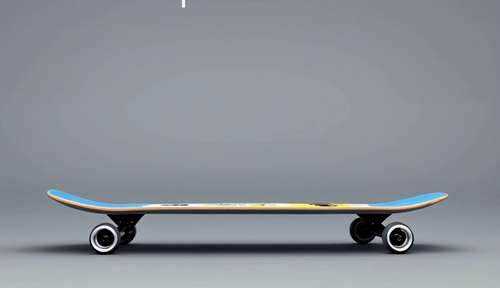}
\\[1.5ex]
\begin{minipage}[c]{2cm}\vspace*{0pt}\vfill\raggedright\textit{\small snowboard 3rd second}\vfill\end{minipage}
\includegraphics[width=0.1\linewidth]{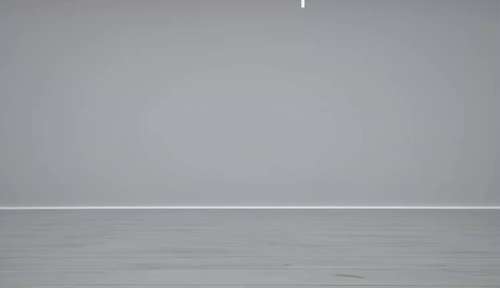}
\includegraphics[width=0.1\linewidth]{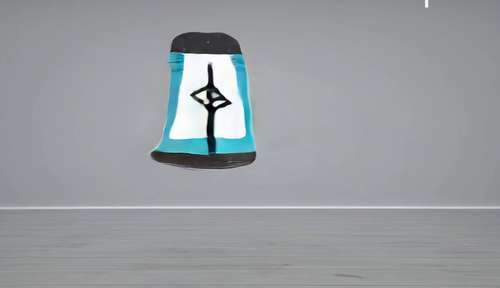}
\includegraphics[width=0.1\linewidth]{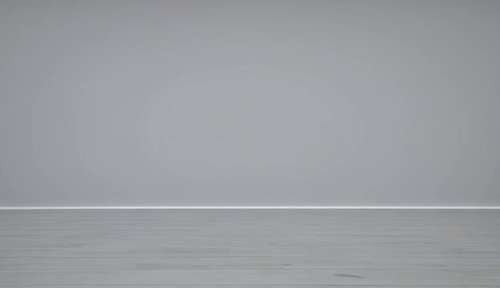}
\includegraphics[width=0.1\linewidth]{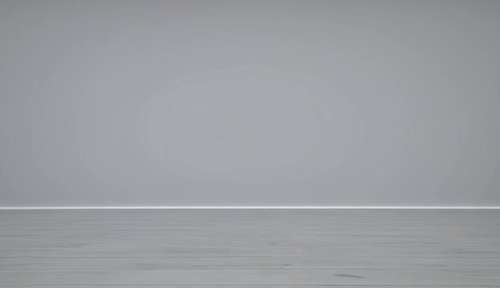}
\hspace{2mm}
\includegraphics[width=0.1\linewidth]{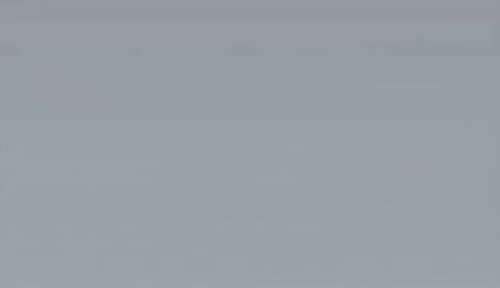}
\includegraphics[width=0.1\linewidth]{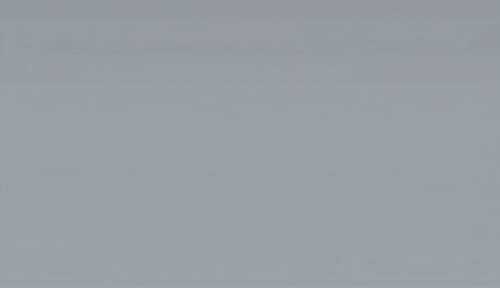}
\includegraphics[width=0.1\linewidth]{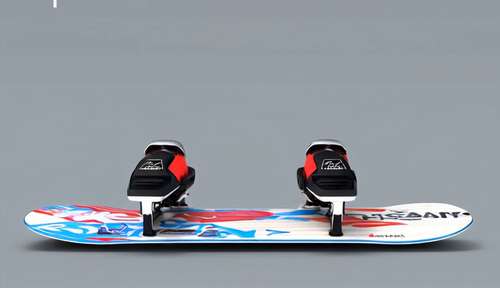}
\includegraphics[width=0.1\linewidth]{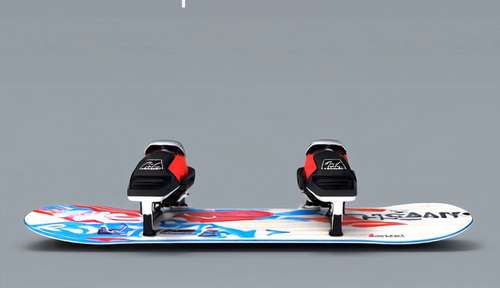}
\\[1.5ex]
\begin{minipage}[c]{2cm}\vspace*{0pt}\vfill\raggedright\textit{\small sports ball 3rd second}\vfill\end{minipage}
\includegraphics[width=0.1\linewidth]{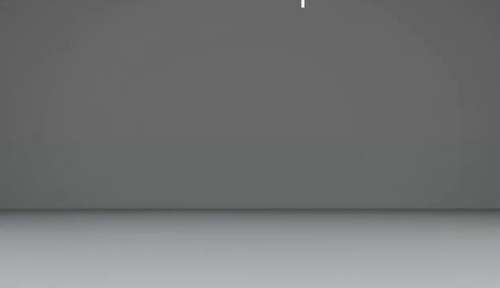}
\includegraphics[width=0.1\linewidth]{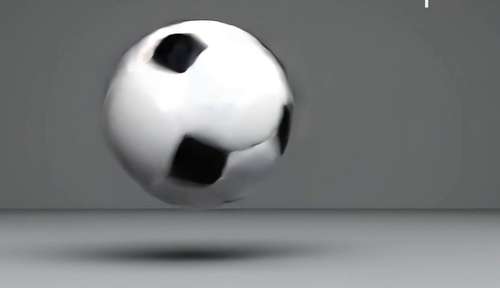}
\includegraphics[width=0.1\linewidth]{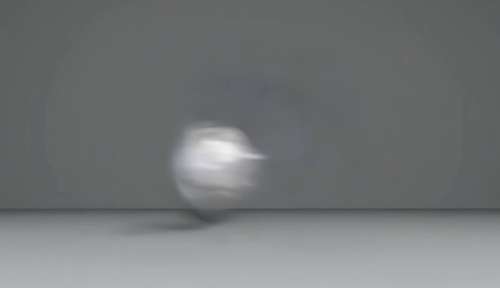}
\includegraphics[width=0.1\linewidth]{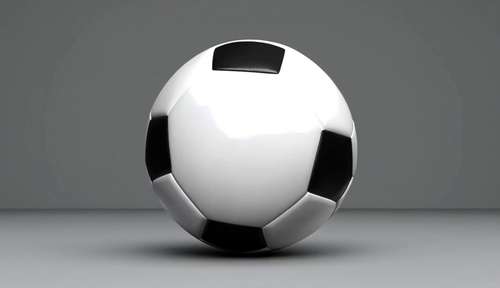}
\hspace{2mm}
\includegraphics[width=0.1\linewidth]{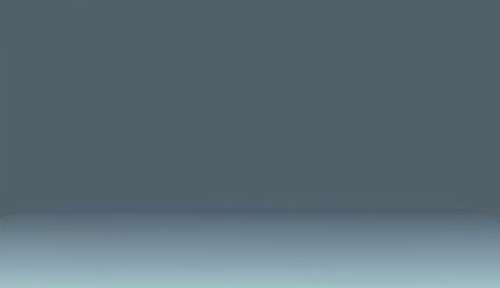}
\includegraphics[width=0.1\linewidth]{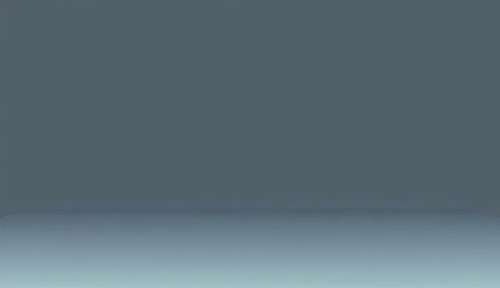}
\includegraphics[width=0.1\linewidth]{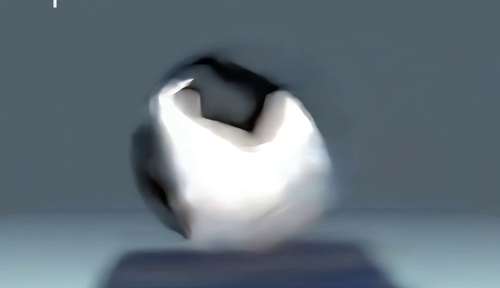}
\includegraphics[width=0.1\linewidth]{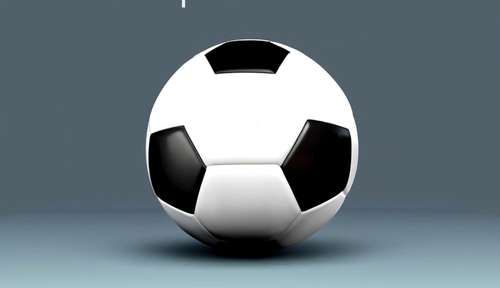}
\\[1.5ex]
\begin{minipage}[c]{2cm}\vspace*{0pt}\vfill\raggedright\textit{\small suitcase 3rd second}\vfill\end{minipage}
\includegraphics[width=0.1\linewidth]{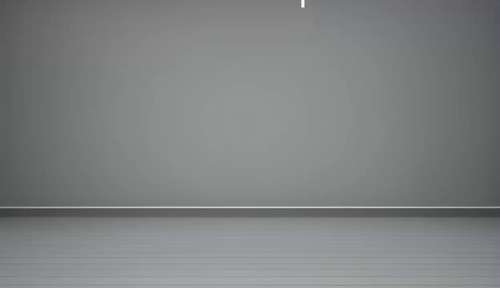}
\includegraphics[width=0.1\linewidth]{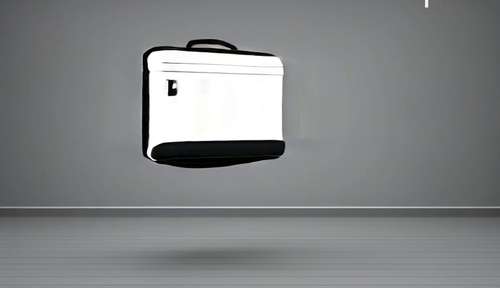}
\includegraphics[width=0.1\linewidth]{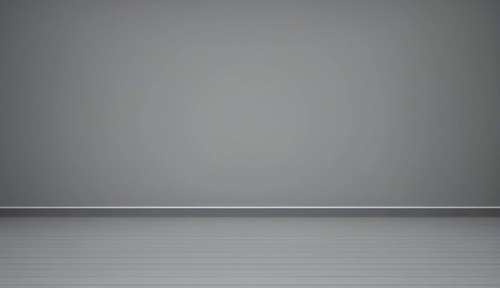}
\includegraphics[width=0.1\linewidth]{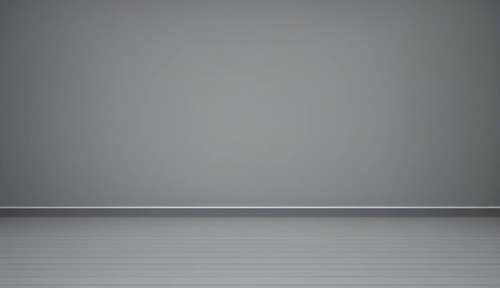}
\hspace{2mm}
\includegraphics[width=0.1\linewidth]{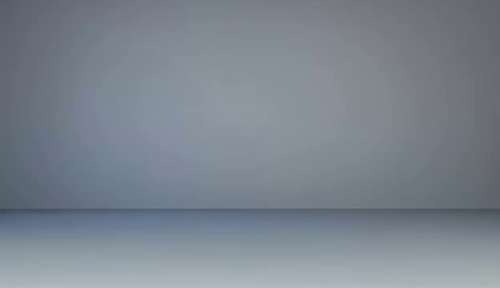}
\includegraphics[width=0.1\linewidth]{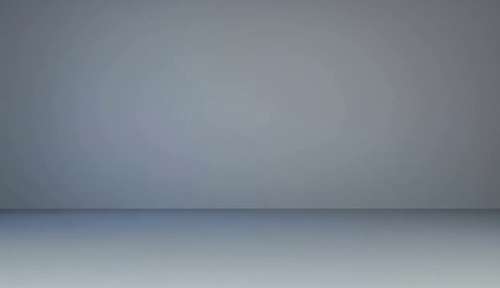}
\includegraphics[width=0.1\linewidth]{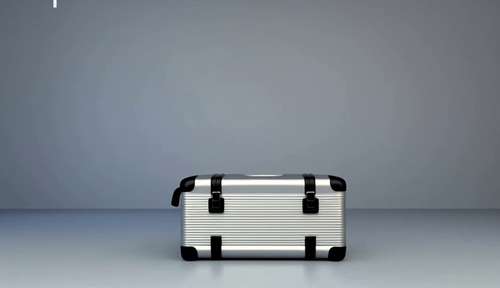}
\includegraphics[width=0.1\linewidth]{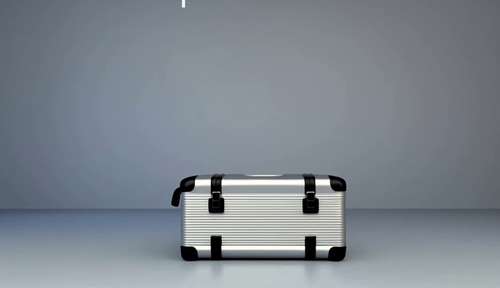}
\\[1.5ex]
\begin{minipage}[c]{2cm}\vspace*{0pt}\vfill\raggedright\textit{\small toilet last second}\vfill\end{minipage}
\includegraphics[width=0.1\linewidth]{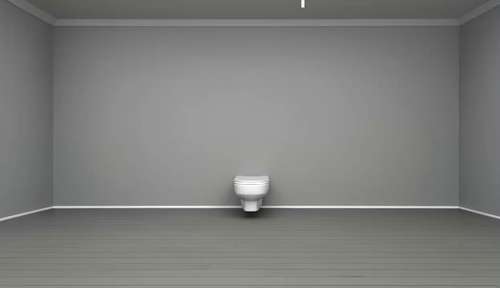}
\includegraphics[width=0.1\linewidth]{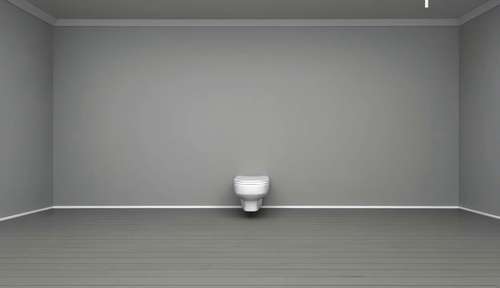}
\includegraphics[width=0.1\linewidth]{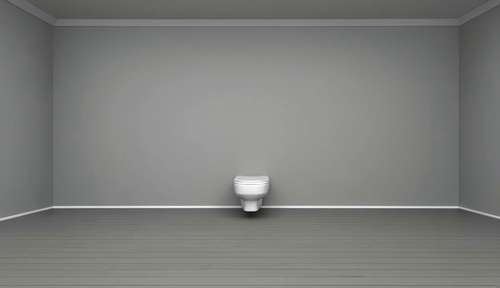}
\includegraphics[width=0.1\linewidth]{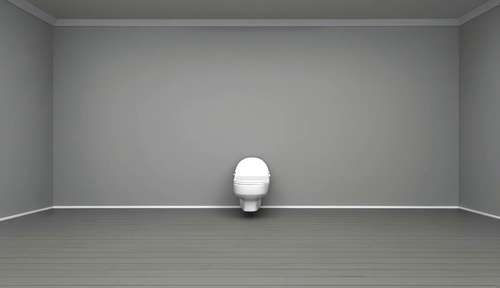}
\hspace{2mm}
\includegraphics[width=0.1\linewidth]{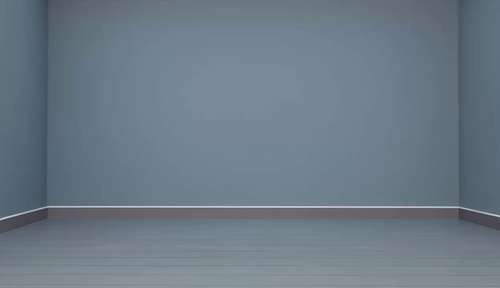}
\includegraphics[width=0.1\linewidth]{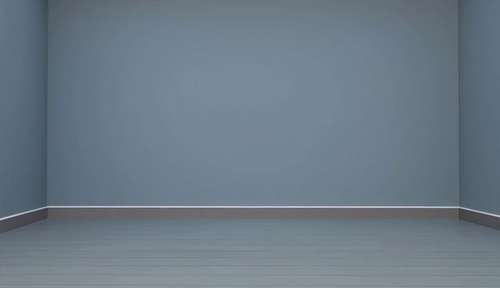}
\includegraphics[width=0.1\linewidth]{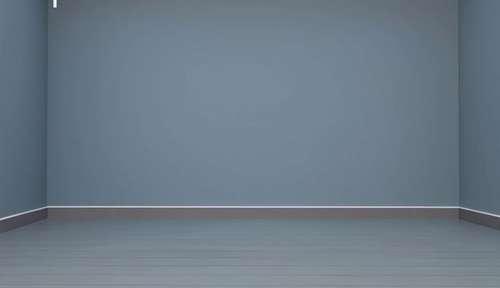}
\includegraphics[width=0.1\linewidth]{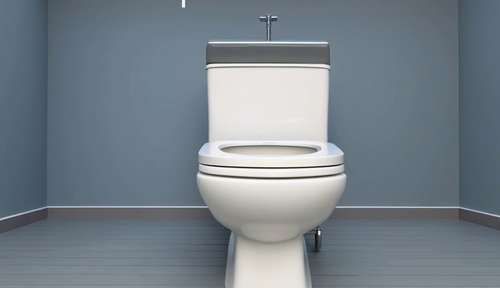}
\\[1.5ex]
\begin{minipage}[c]{2cm}\vspace*{0pt}\vfill\raggedright\textit{\small truck 2nd second}\vfill\end{minipage}
\includegraphics[width=0.1\linewidth]{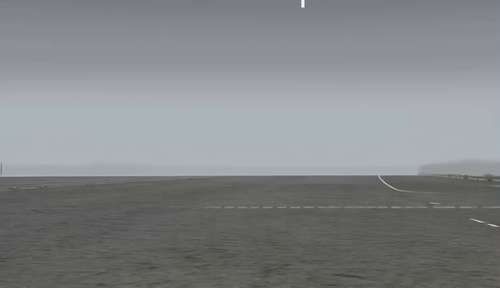}
\includegraphics[width=0.1\linewidth]{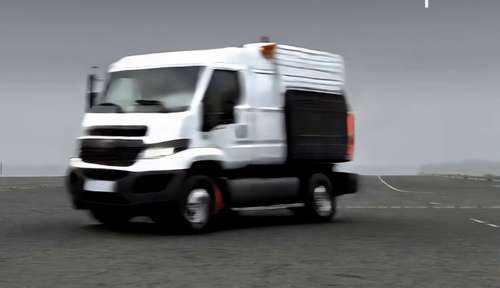}
\includegraphics[width=0.1\linewidth]{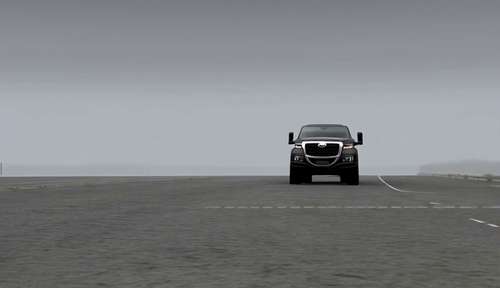}
\includegraphics[width=0.1\linewidth]{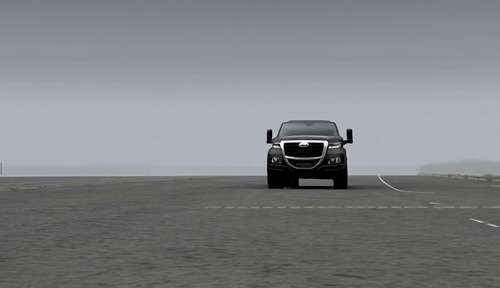}
\hspace{2mm}
\includegraphics[width=0.1\linewidth]{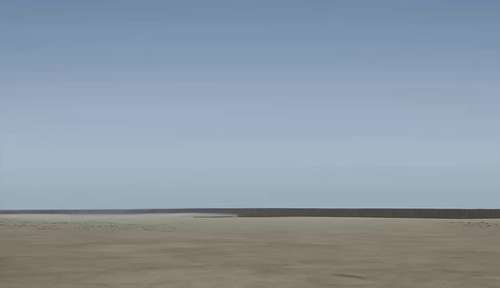}
\includegraphics[width=0.1\linewidth]{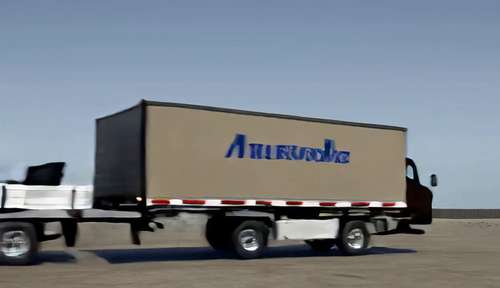}
\includegraphics[width=0.1\linewidth]{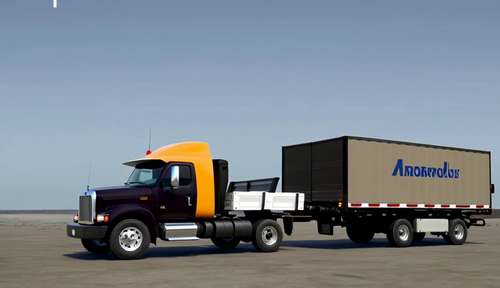}
\includegraphics[width=0.1\linewidth]{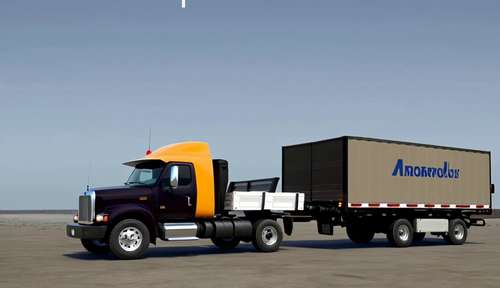}
\\[1.5ex]
\begin{minipage}[c]{2cm}\vspace*{0pt}\vfill\raggedright\textit{\small umbrella 3rd second}\vfill\end{minipage}
\includegraphics[width=0.1\linewidth]{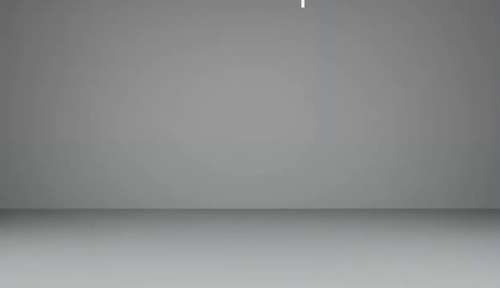}
\includegraphics[width=0.1\linewidth]{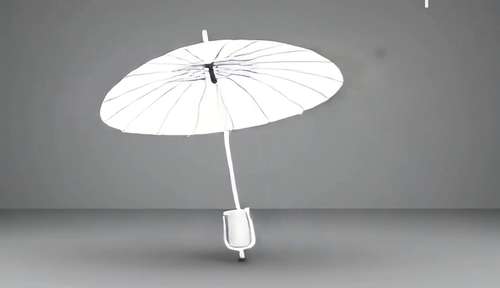}
\includegraphics[width=0.1\linewidth]{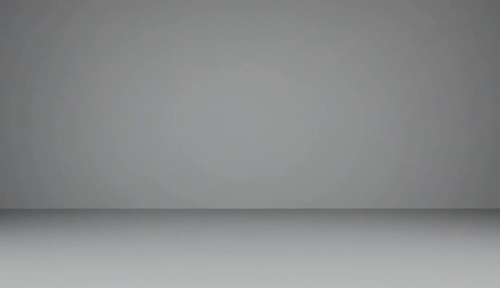}
\includegraphics[width=0.1\linewidth]{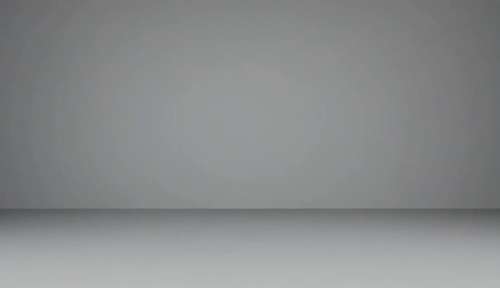}
\hspace{2mm}
\includegraphics[width=0.1\linewidth]{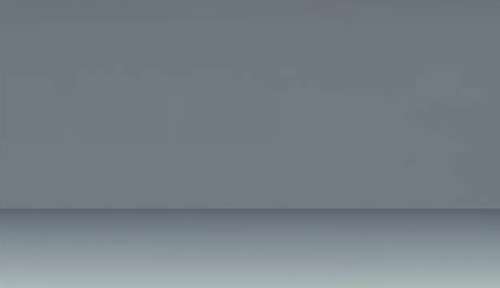}
\includegraphics[width=0.1\linewidth]{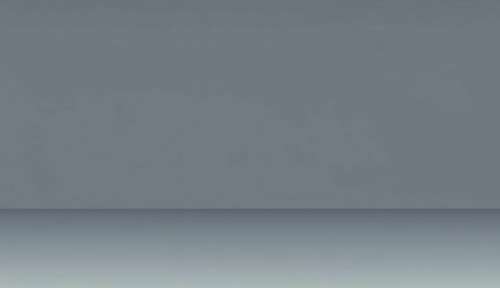}
\includegraphics[width=0.1\linewidth]{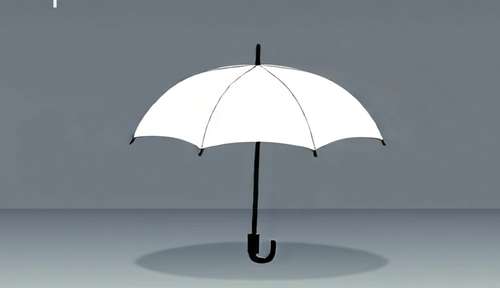}
\includegraphics[width=0.1\linewidth]{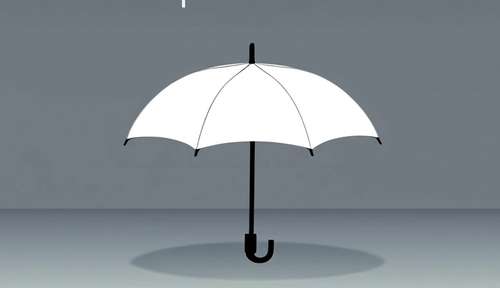}
\\[1.5ex]
\end{minipage}
\caption{One Object Text vs Ours.}
\label{fig:one_object2_text_vs_ours}
\end{figure*}

\begin{figure*}[t]
\centering
\begin{minipage}{\textwidth}
\makebox[2cm][l]{}\makebox[0.40\linewidth]{\small Text}\hspace{2mm}\makebox[0.40\linewidth]{\small Ours}\\[1.5ex]
\begin{minipage}[c]{2cm}\vspace*{0pt}\vfill\raggedright\textit{\small bee buzzing 3rd second}\vfill\end{minipage}
\includegraphics[width=0.1\linewidth]{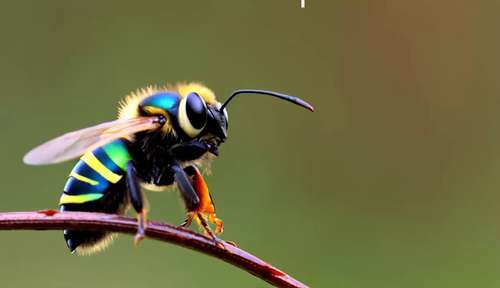}
\includegraphics[width=0.1\linewidth]{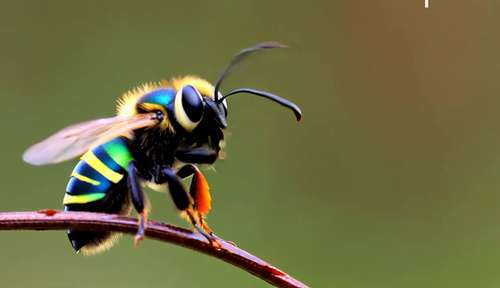}
\includegraphics[width=0.1\linewidth]{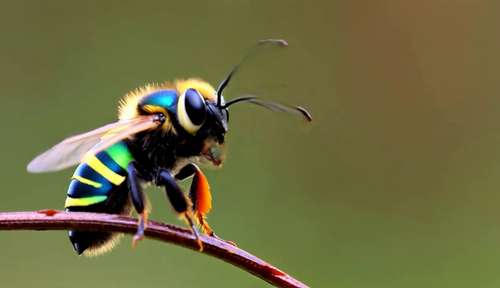}
\includegraphics[width=0.1\linewidth]{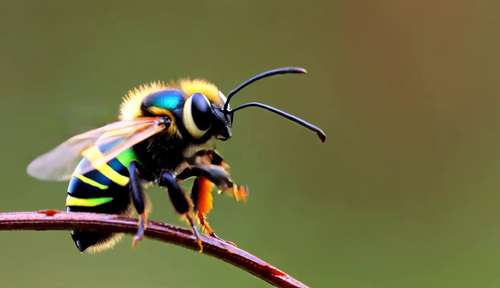}
\hspace{2mm}
\includegraphics[width=0.1\linewidth]{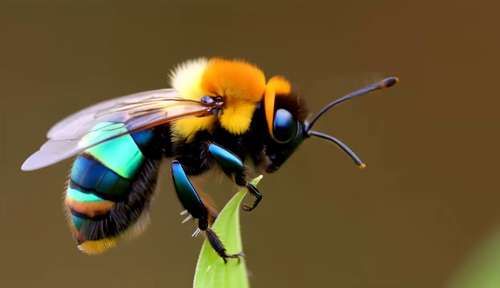}
\includegraphics[width=0.1\linewidth]{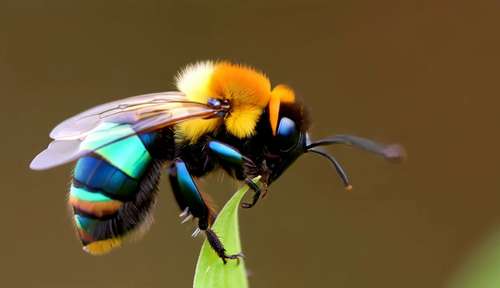}
\includegraphics[width=0.1\linewidth]{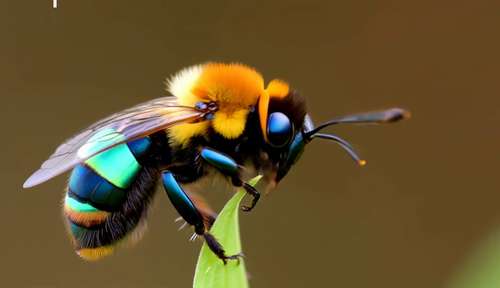}
\includegraphics[width=0.1\linewidth]{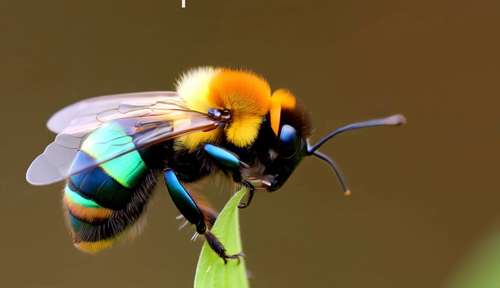}
\\[1.5ex]
\begin{minipage}[c]{2cm}\vspace*{0pt}\vfill\raggedright\textit{\small bird flapping 2nd second}\vfill\end{minipage}
\includegraphics[width=0.1\linewidth]{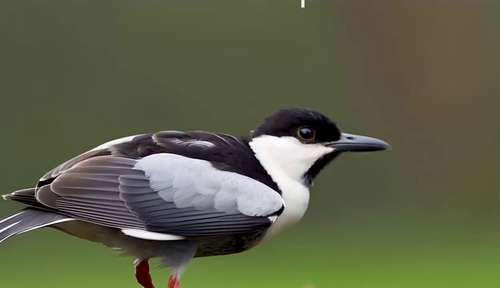}
\includegraphics[width=0.1\linewidth]{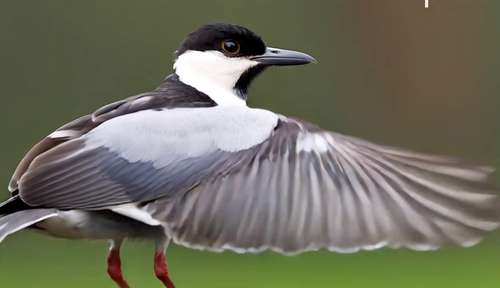}
\includegraphics[width=0.1\linewidth]{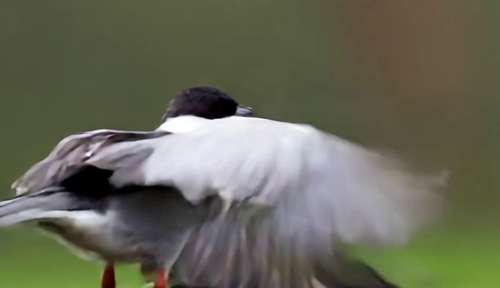}
\includegraphics[width=0.1\linewidth]{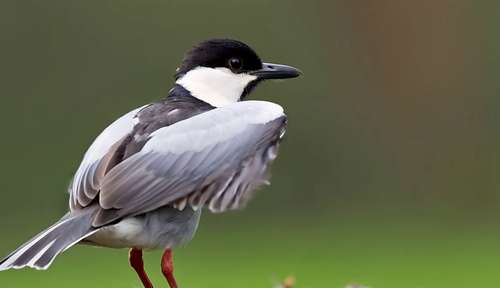}
\hspace{2mm}
\includegraphics[width=0.1\linewidth]{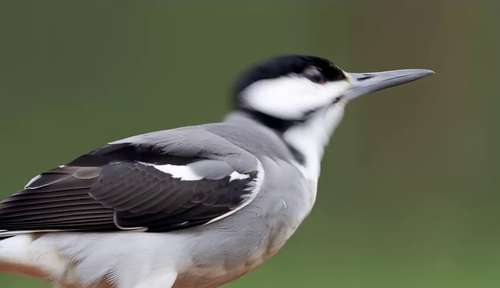}
\includegraphics[width=0.1\linewidth]{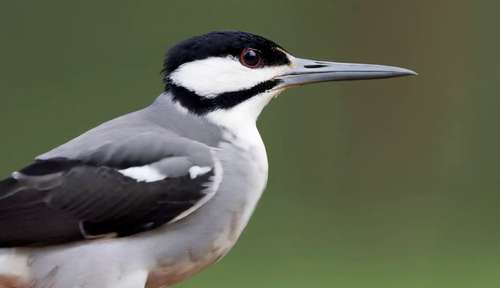}
\includegraphics[width=0.1\linewidth]{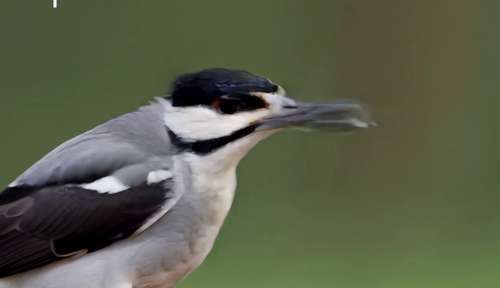}
\includegraphics[width=0.1\linewidth]{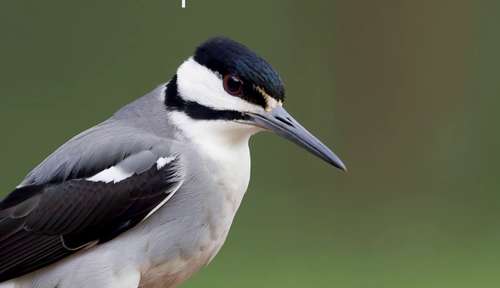}
\\[1.5ex]
\begin{minipage}[c]{2cm}\vspace*{0pt}\vfill\raggedright\textit{\small child crawling 2nd second}\vfill\end{minipage}
\includegraphics[width=0.1\linewidth]{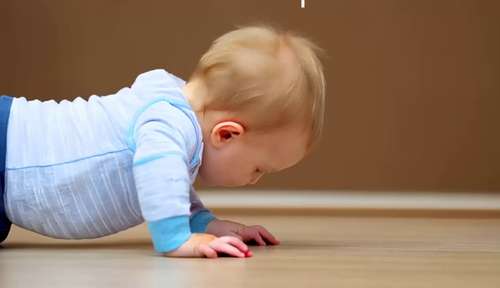}
\includegraphics[width=0.1\linewidth]{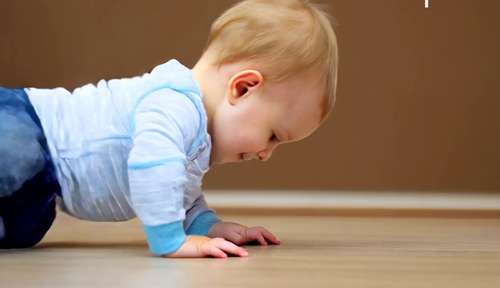}
\includegraphics[width=0.1\linewidth]{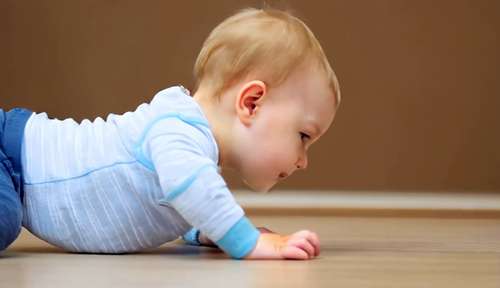}
\includegraphics[width=0.1\linewidth]{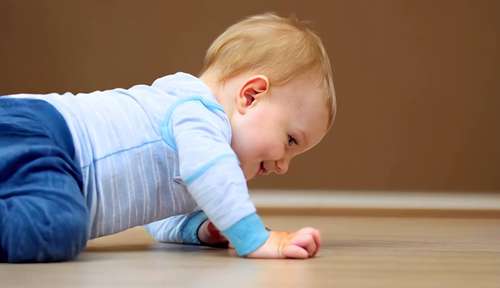}
\hspace{2mm}
\includegraphics[width=0.1\linewidth]{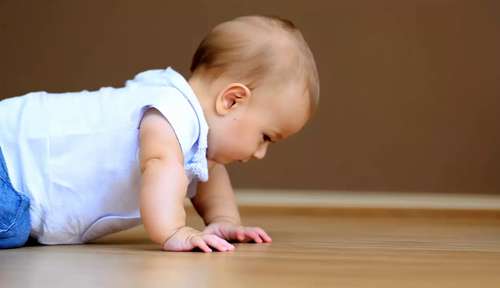}
\includegraphics[width=0.1\linewidth]{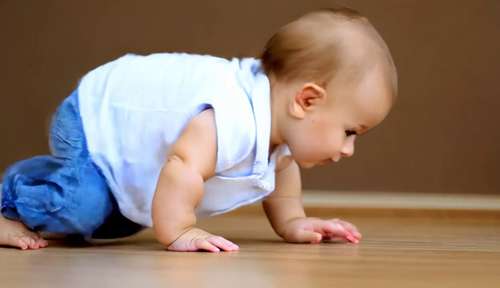}
\includegraphics[width=0.1\linewidth]{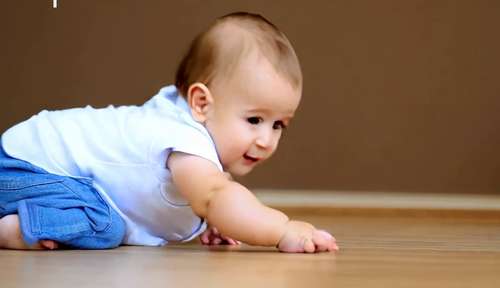}
\includegraphics[width=0.1\linewidth]{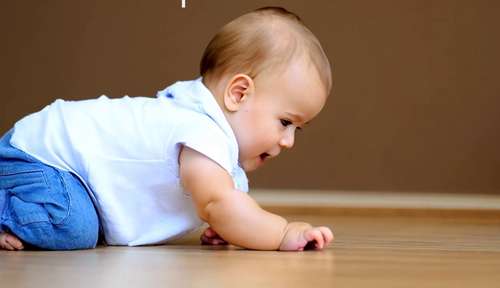}
\\[1.5ex]
\begin{minipage}[c]{2cm}\vspace*{0pt}\vfill\raggedright\textit{\small child jumping 1st second}\vfill\end{minipage}
\includegraphics[width=0.1\linewidth]{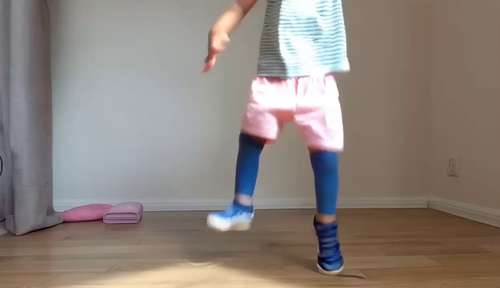}
\includegraphics[width=0.1\linewidth]{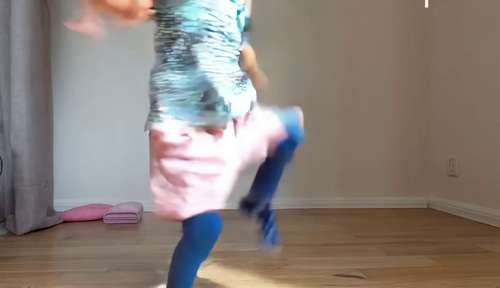}
\includegraphics[width=0.1\linewidth]{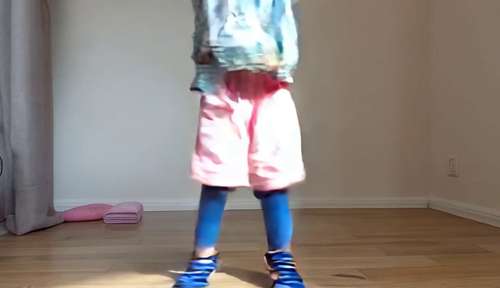}
\includegraphics[width=0.1\linewidth]{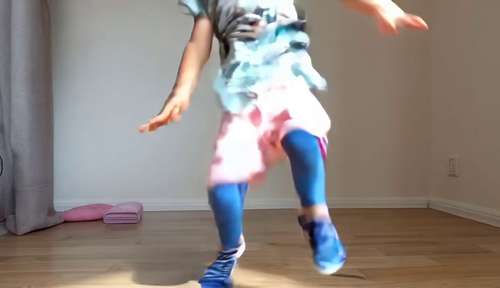}
\hspace{2mm}
\includegraphics[width=0.1\linewidth]{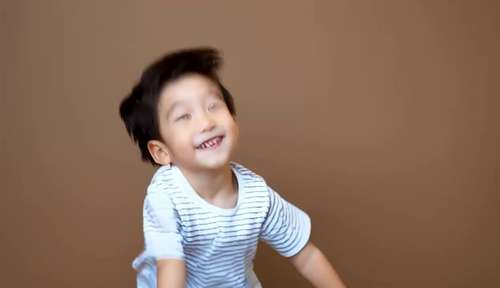}
\includegraphics[width=0.1\linewidth]{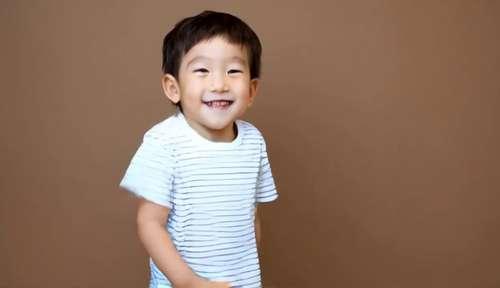}
\includegraphics[width=0.1\linewidth]{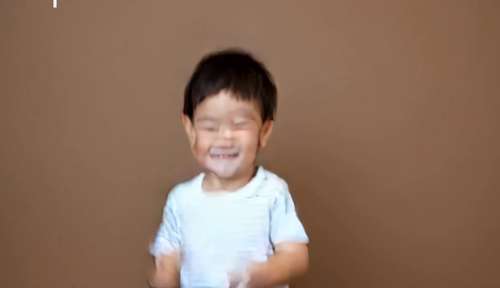}
\includegraphics[width=0.1\linewidth]{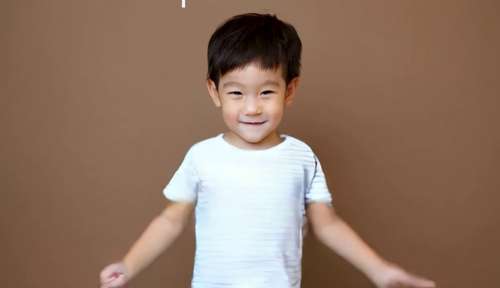}
\\[1.5ex]
\begin{minipage}[c]{2cm}\vspace*{0pt}\vfill\raggedright\textit{\small child screaming 2nd second}\vfill\end{minipage}
\includegraphics[width=0.1\linewidth]{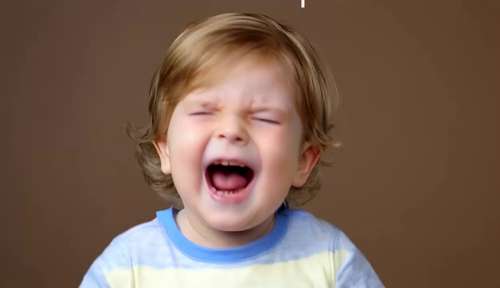}
\includegraphics[width=0.1\linewidth]{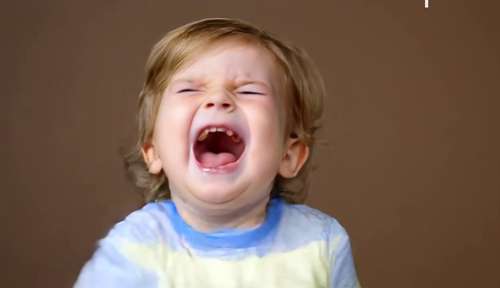}
\includegraphics[width=0.1\linewidth]{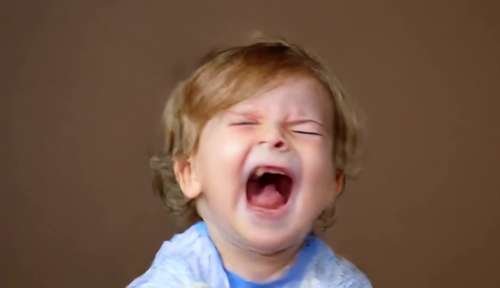}
\includegraphics[width=0.1\linewidth]{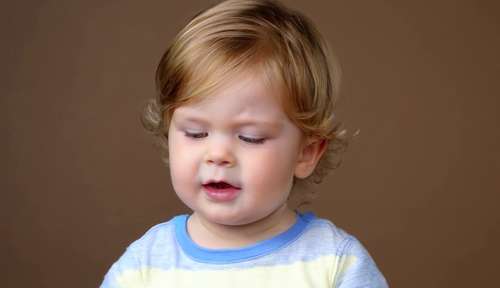}
\hspace{2mm}
\includegraphics[width=0.1\linewidth]{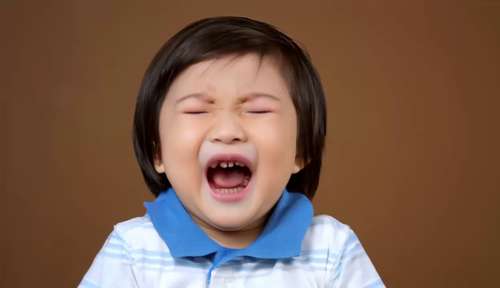}
\includegraphics[width=0.1\linewidth]{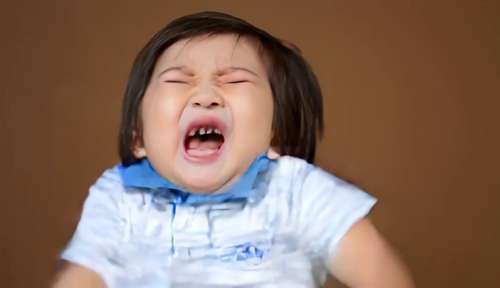}
\includegraphics[width=0.1\linewidth]{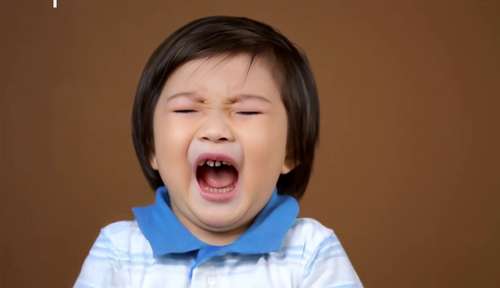}
\includegraphics[width=0.1\linewidth]{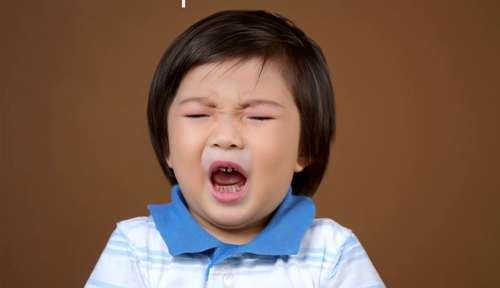}
\\[1.5ex]
\begin{minipage}[c]{2cm}\vspace*{0pt}\vfill\raggedright\textit{\small chimpanzee clapping 1st second}\vfill\end{minipage}
\includegraphics[width=0.1\linewidth]{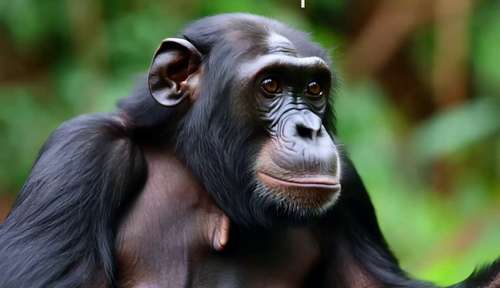}
\includegraphics[width=0.1\linewidth]{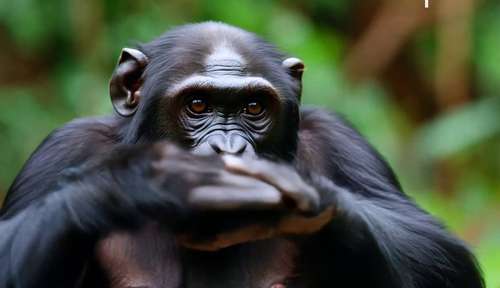}
\includegraphics[width=0.1\linewidth]{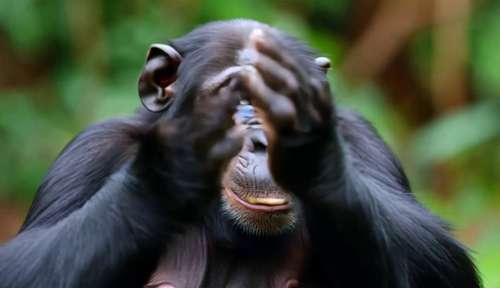}
\includegraphics[width=0.1\linewidth]{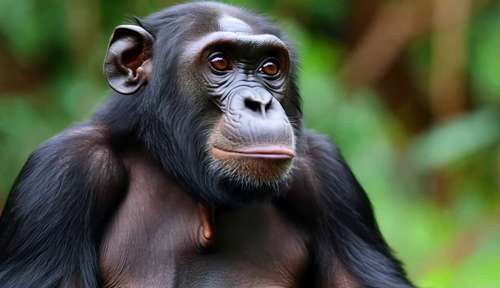}
\hspace{2mm}
\includegraphics[width=0.1\linewidth]{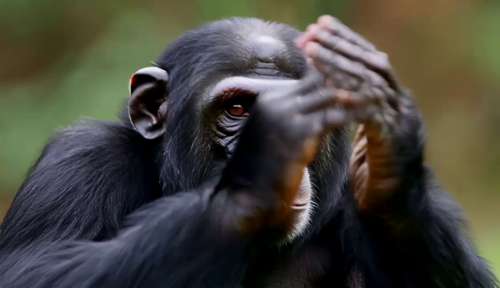}
\includegraphics[width=0.1\linewidth]{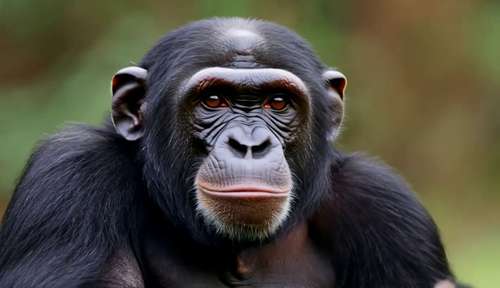}
\includegraphics[width=0.1\linewidth]{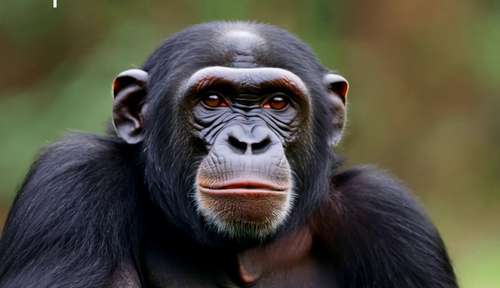}
\includegraphics[width=0.1\linewidth]{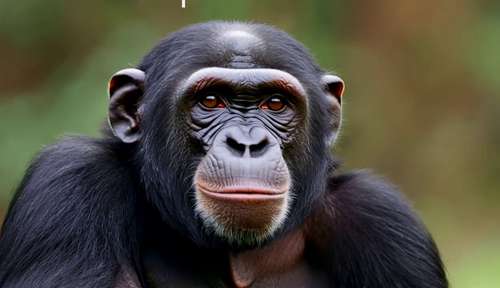}
\\[1.5ex]
\begin{minipage}[c]{2cm}\vspace*{0pt}\vfill\raggedright\textit{\small crow cawing 4th second}\vfill\end{minipage}
\includegraphics[width=0.1\linewidth]{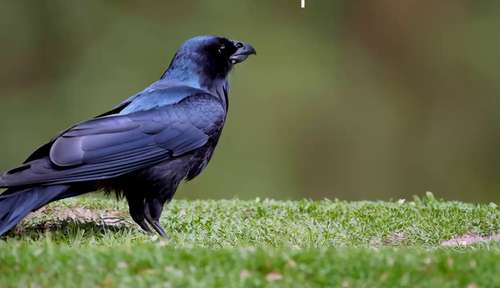}
\includegraphics[width=0.1\linewidth]{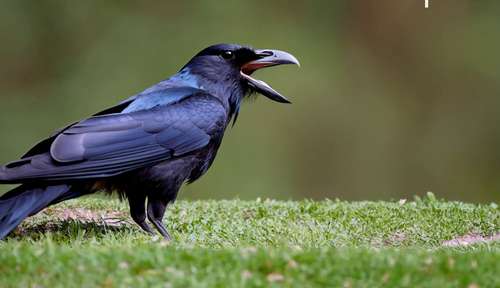}
\includegraphics[width=0.1\linewidth]{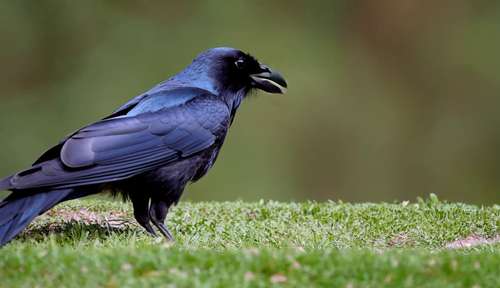}
\includegraphics[width=0.1\linewidth]{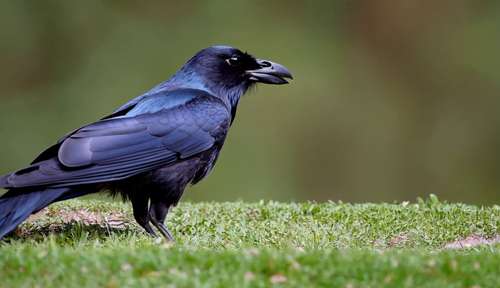}
\hspace{2mm}
\includegraphics[width=0.1\linewidth]{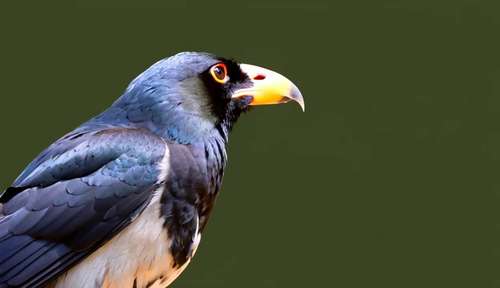}
\includegraphics[width=0.1\linewidth]{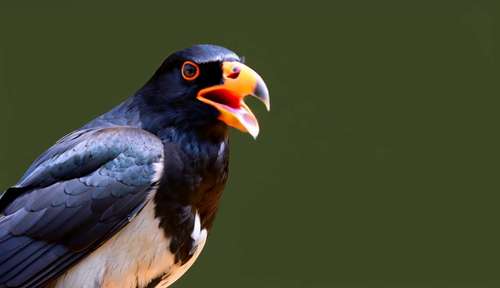}
\includegraphics[width=0.1\linewidth]{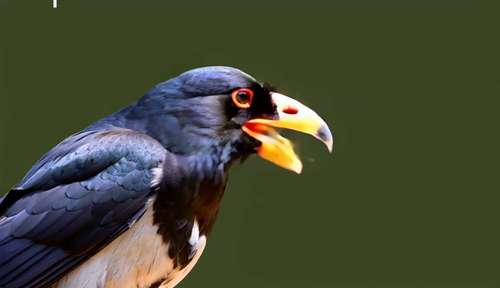}
\includegraphics[width=0.1\linewidth]{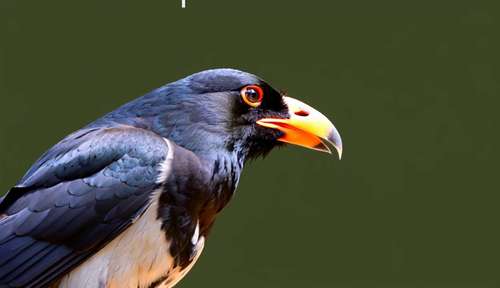}
\\[1.5ex]
\begin{minipage}[c]{2cm}\vspace*{0pt}\vfill\raggedright\textit{\small falcon gliding last second}\vfill\end{minipage}
\includegraphics[width=0.1\linewidth]{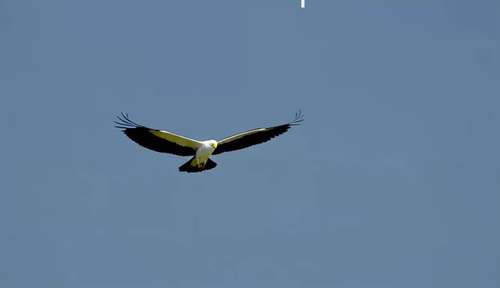}
\includegraphics[width=0.1\linewidth]{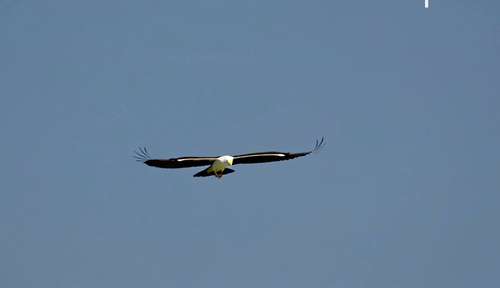}
\includegraphics[width=0.1\linewidth]{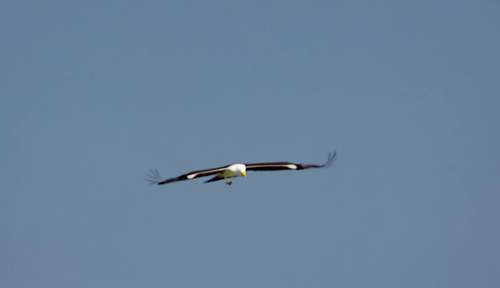}
\includegraphics[width=0.1\linewidth]{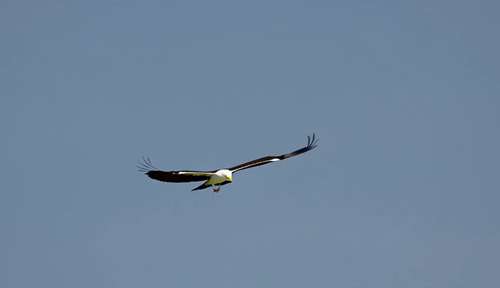}
\hspace{2mm}
\includegraphics[width=0.1\linewidth]{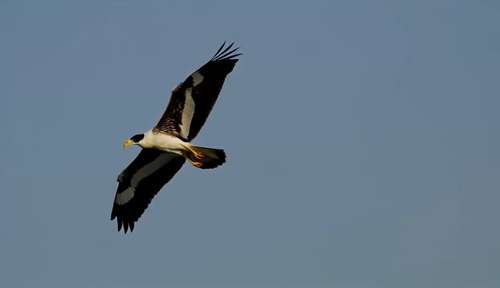}
\includegraphics[width=0.1\linewidth]{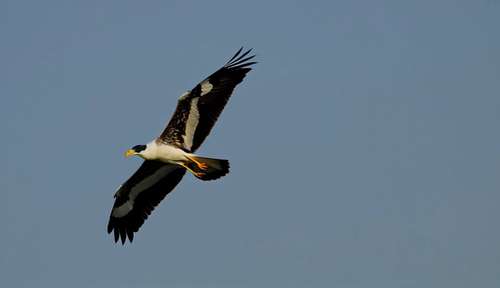}
\includegraphics[width=0.1\linewidth]{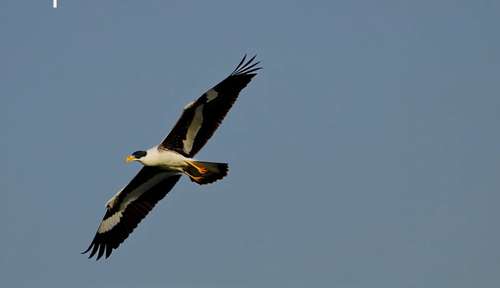}
\includegraphics[width=0.1\linewidth]{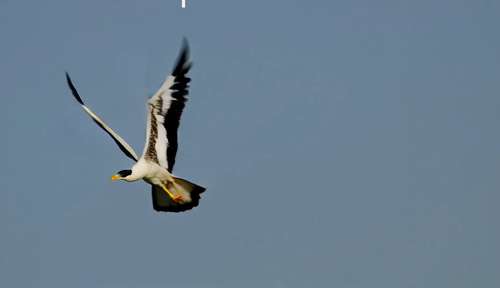}
\\[1.5ex]
\begin{minipage}[c]{2cm}\vspace*{0pt}\vfill\raggedright\textit{\small fox darting 2nd second}\vfill\end{minipage}
\includegraphics[width=0.1\linewidth]{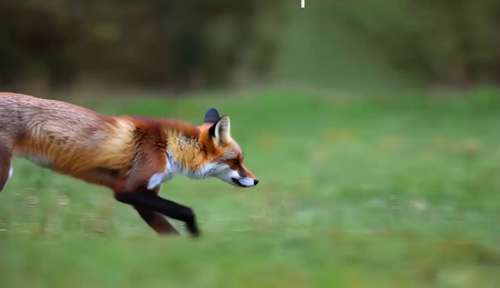}
\includegraphics[width=0.1\linewidth]{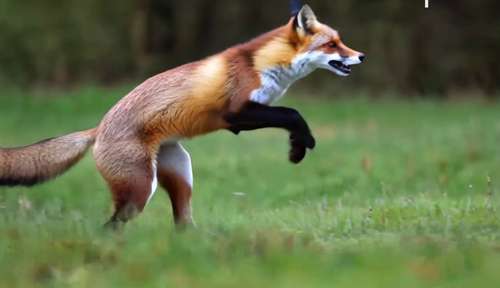}
\includegraphics[width=0.1\linewidth]{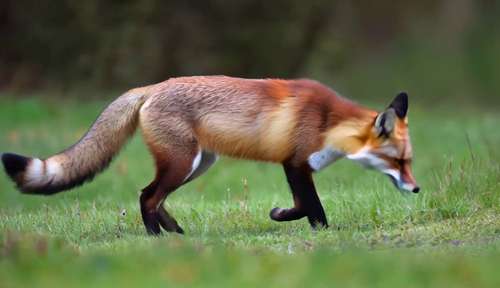}
\includegraphics[width=0.1\linewidth]{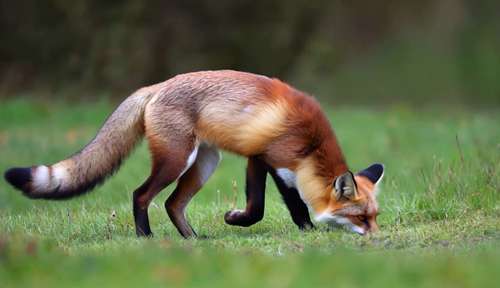}
\hspace{2mm}
\includegraphics[width=0.1\linewidth]{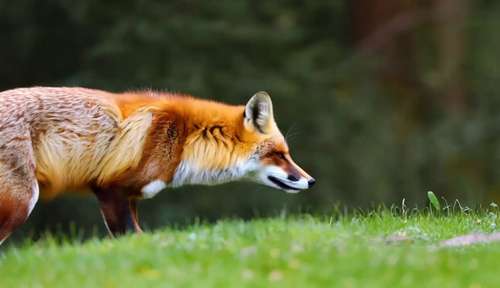}
\includegraphics[width=0.1\linewidth]{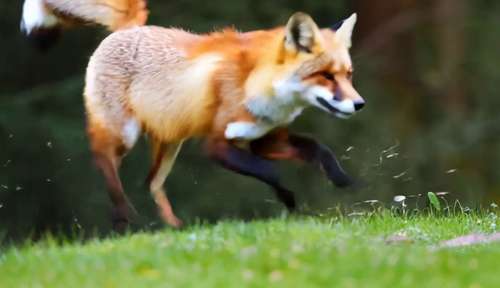}
\includegraphics[width=0.1\linewidth]{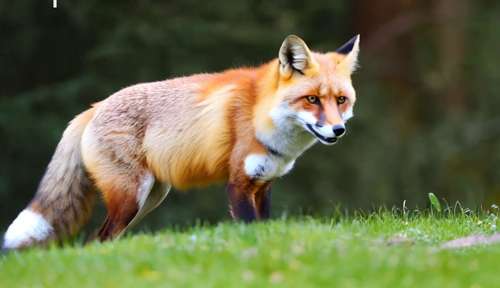}
\includegraphics[width=0.1\linewidth]{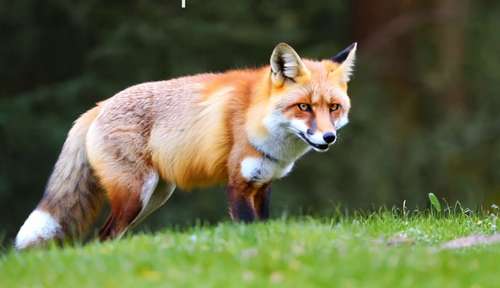}
\\[1.5ex]
\begin{minipage}[c]{2cm}\vspace*{0pt}\vfill\raggedright\textit{\small horse shaking its mane 4th second}\vfill\end{minipage}
\includegraphics[width=0.1\linewidth]{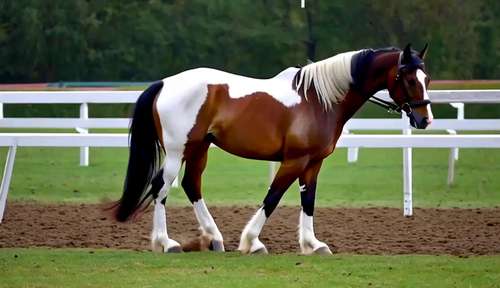}
\includegraphics[width=0.1\linewidth]{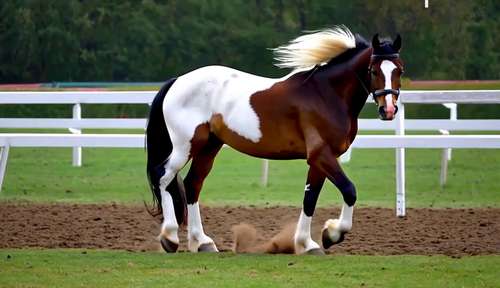}
\includegraphics[width=0.1\linewidth]{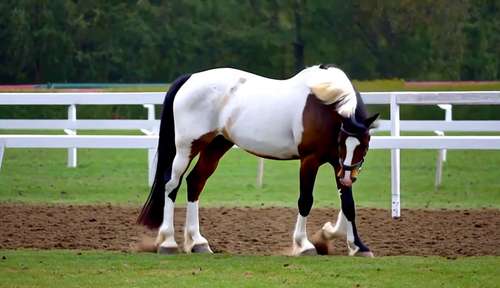}
\includegraphics[width=0.1\linewidth]{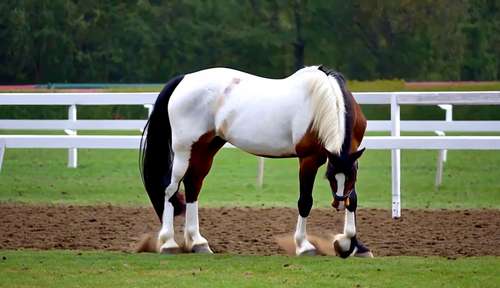}
\hspace{2mm}
\includegraphics[width=0.1\linewidth]{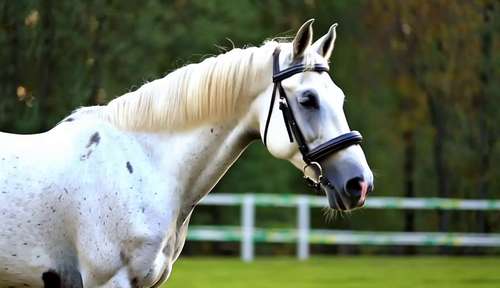}
\includegraphics[width=0.1\linewidth]{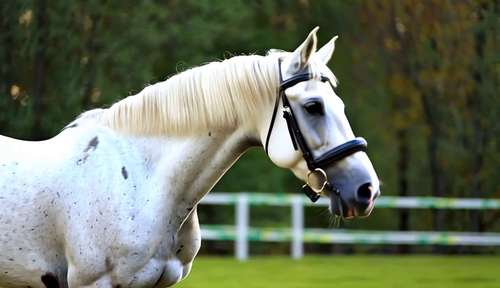}
\includegraphics[width=0.1\linewidth]{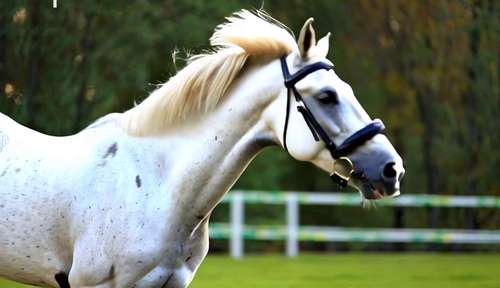}
\includegraphics[width=0.1\linewidth]{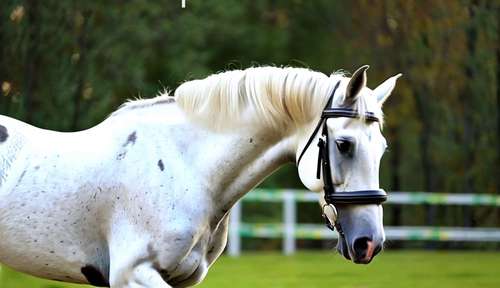}
\\[1.5ex]
\begin{minipage}[c]{2cm}\vspace*{0pt}\vfill\raggedright\textit{\small jellyfish floating last second}\vfill\end{minipage}
\includegraphics[width=0.1\linewidth]{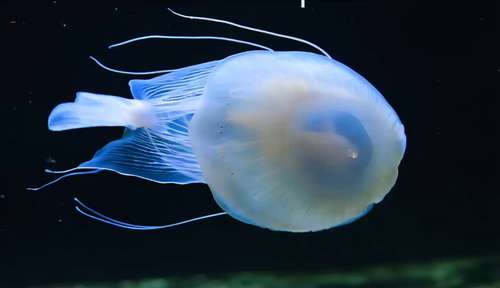}
\includegraphics[width=0.1\linewidth]{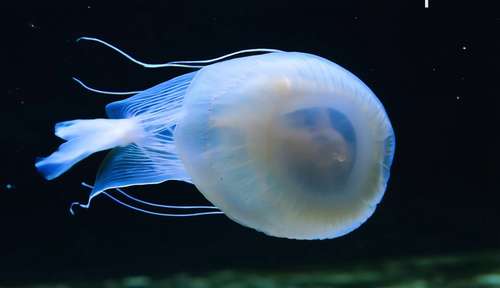}
\includegraphics[width=0.1\linewidth]{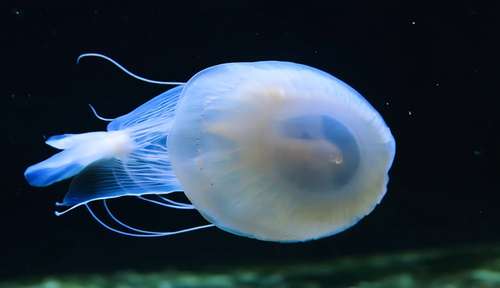}
\includegraphics[width=0.1\linewidth]{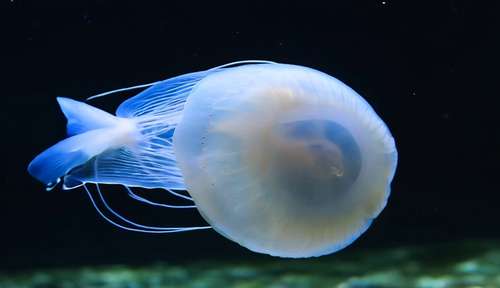}
\hspace{2mm}
\includegraphics[width=0.1\linewidth]{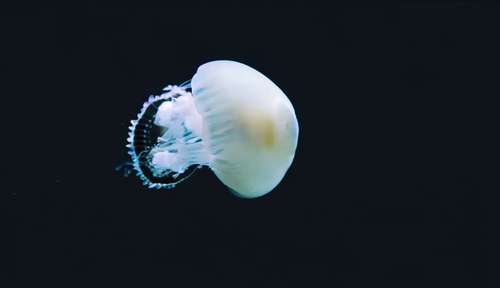}
\includegraphics[width=0.1\linewidth]{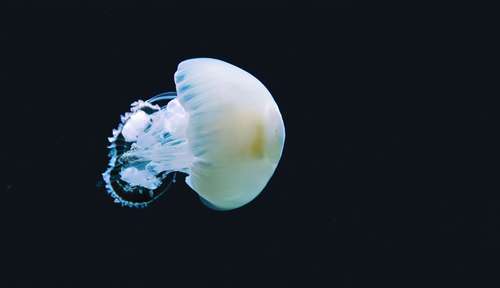}
\includegraphics[width=0.1\linewidth]{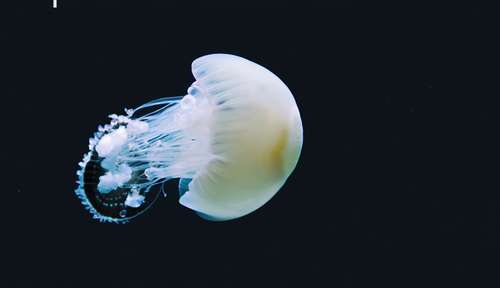}
\includegraphics[width=0.1\linewidth]{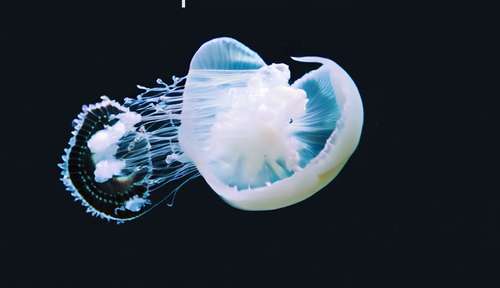}
\\[1.5ex]
\begin{minipage}[c]{2cm}\vspace*{0pt}\vfill\raggedright\textit{\small kangaroo hopping 2nd second}\vfill\end{minipage}
\includegraphics[width=0.1\linewidth]{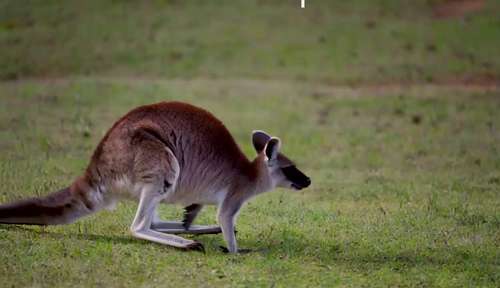}
\includegraphics[width=0.1\linewidth]{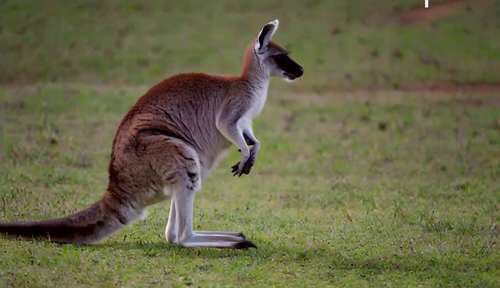}
\includegraphics[width=0.1\linewidth]{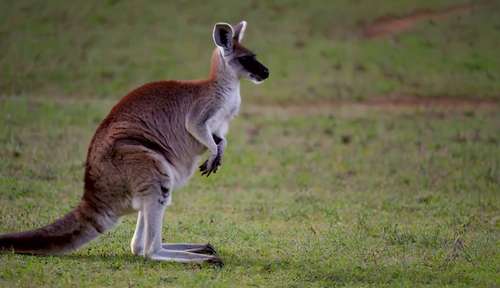}
\includegraphics[width=0.1\linewidth]{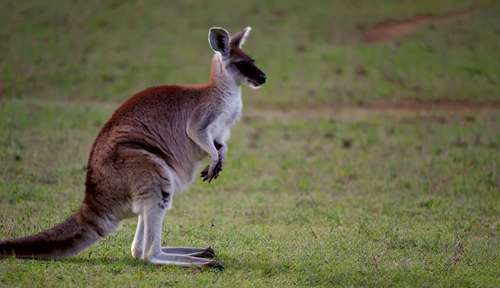}
\hspace{2mm}
\includegraphics[width=0.1\linewidth]{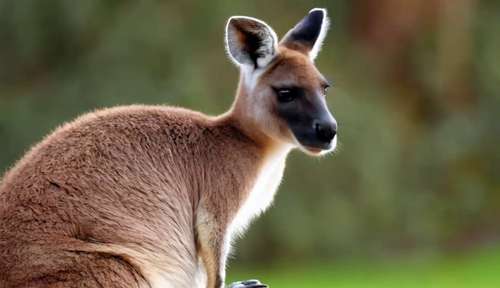}
\includegraphics[width=0.1\linewidth]{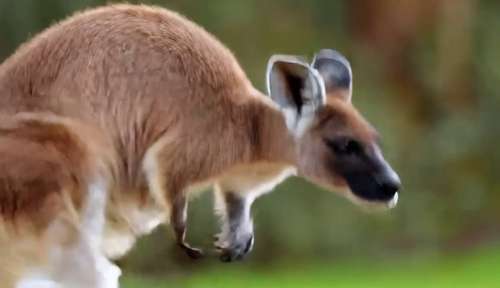}
\includegraphics[width=0.1\linewidth]{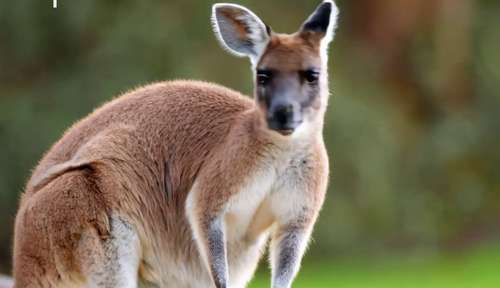}
\includegraphics[width=0.1\linewidth]{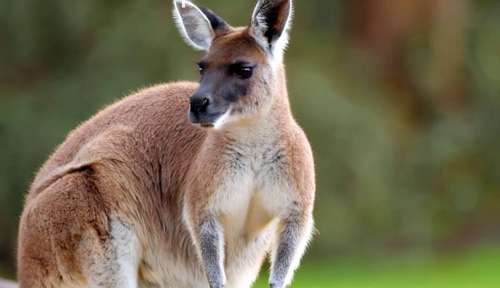}
\\[1.5ex]
\begin{minipage}[c]{2cm}\vspace*{0pt}\vfill\raggedright\textit{\small leopard growling 4th second}\vfill\end{minipage}
\includegraphics[width=0.1\linewidth]{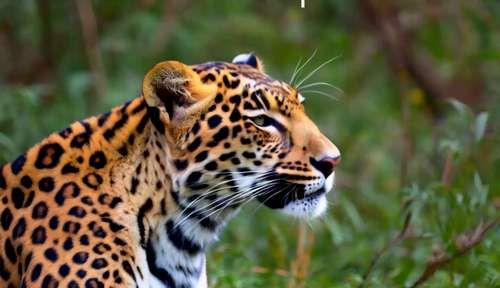}
\includegraphics[width=0.1\linewidth]{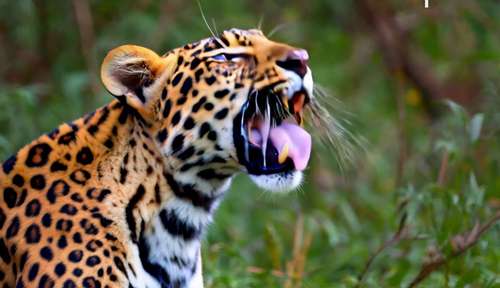}
\includegraphics[width=0.1\linewidth]{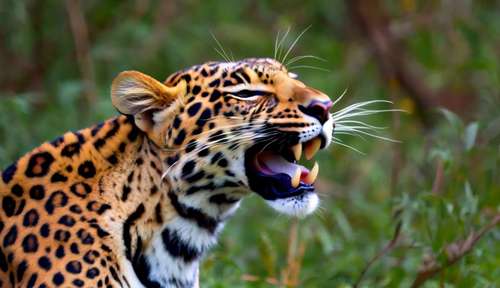}
\includegraphics[width=0.1\linewidth]{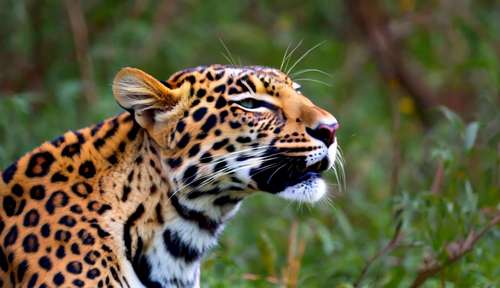}
\hspace{2mm}
\includegraphics[width=0.1\linewidth]{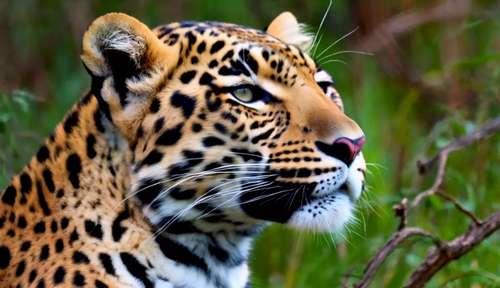}
\includegraphics[width=0.1\linewidth]{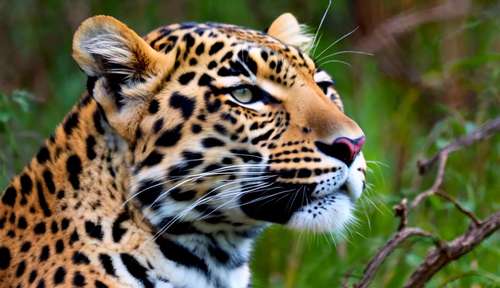}
\includegraphics[width=0.1\linewidth]{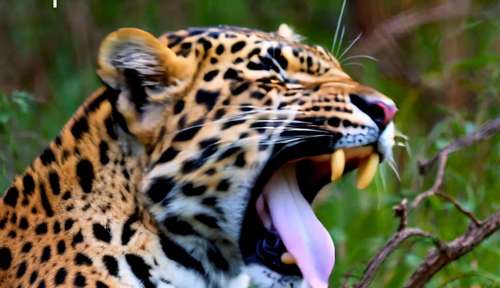}
\includegraphics[width=0.1\linewidth]{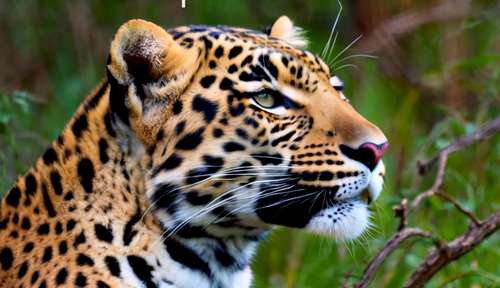}
\\[1.5ex]
\end{minipage}
\caption{Movement Text vs Ours.}
\label{fig:movement_text_vs_ours}
\end{figure*}

\begin{figure*}[t]
\centering
\begin{minipage}{\textwidth}
\makebox[2cm][l]{}\makebox[0.40\linewidth]{\small Text}\hspace{2mm}\makebox[0.40\linewidth]{\small Ours}\\[1.5ex]
\begin{minipage}[c]{2cm}\vspace*{0pt}\vfill\raggedright\textit{\small lion roaring 1st second}\vfill\end{minipage}
\includegraphics[width=0.1\linewidth]{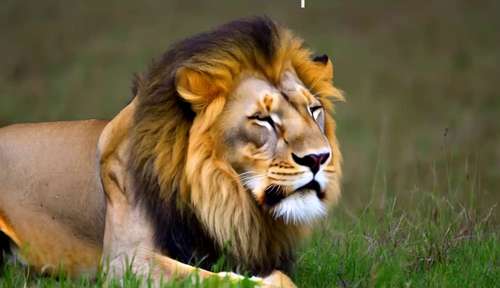}
\includegraphics[width=0.1\linewidth]{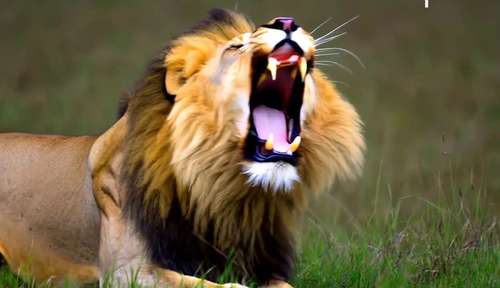}
\includegraphics[width=0.1\linewidth]{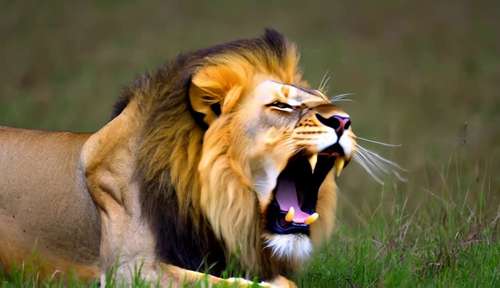}
\includegraphics[width=0.1\linewidth]{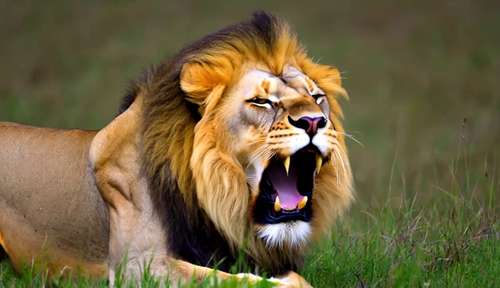}
\hspace{2mm}
\includegraphics[width=0.1\linewidth]{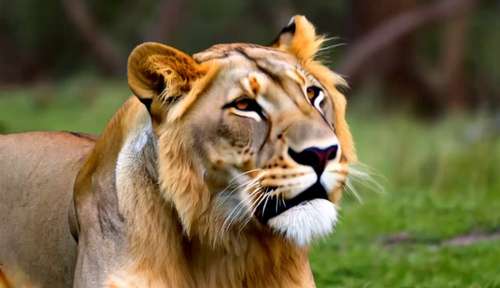}
\includegraphics[width=0.1\linewidth]{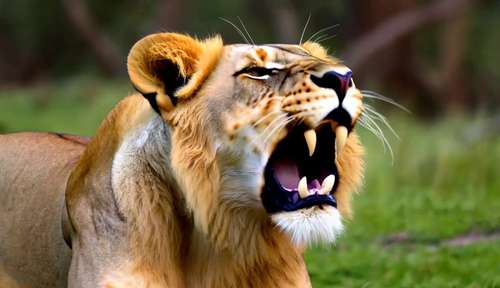}
\includegraphics[width=0.1\linewidth]{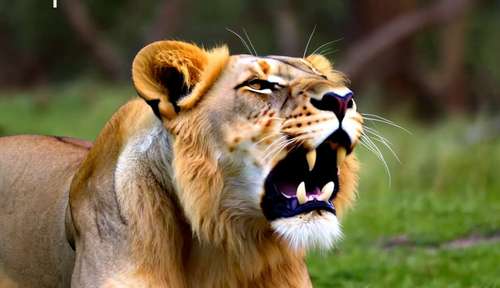}
\includegraphics[width=0.1\linewidth]{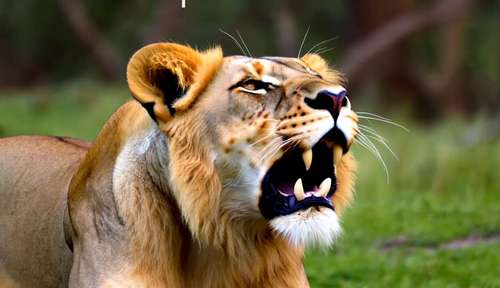}
\\[1.5ex]
\begin{minipage}[c]{2cm}\vspace*{0pt}\vfill\raggedright\textit{\small lion walking last second}\vfill\end{minipage}
\includegraphics[width=0.1\linewidth]{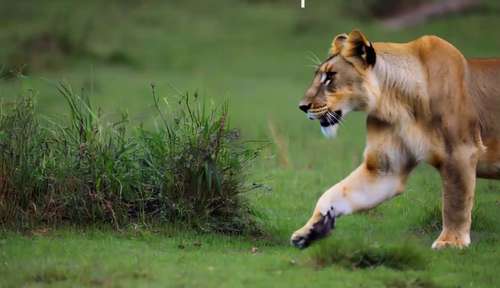}
\includegraphics[width=0.1\linewidth]{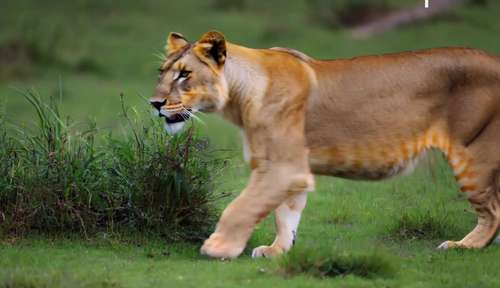}
\includegraphics[width=0.1\linewidth]{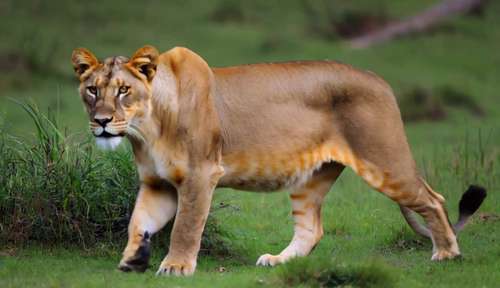}
\includegraphics[width=0.1\linewidth]{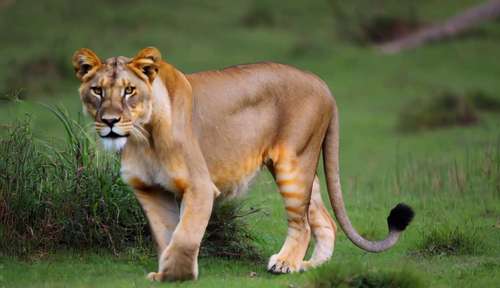}
\hspace{2mm}
\includegraphics[width=0.1\linewidth]{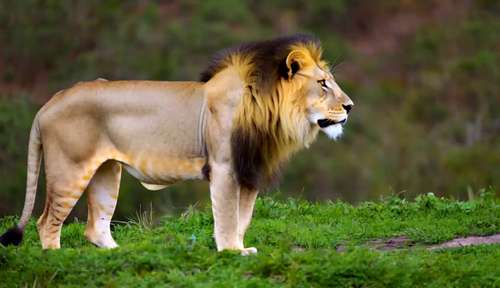}
\includegraphics[width=0.1\linewidth]{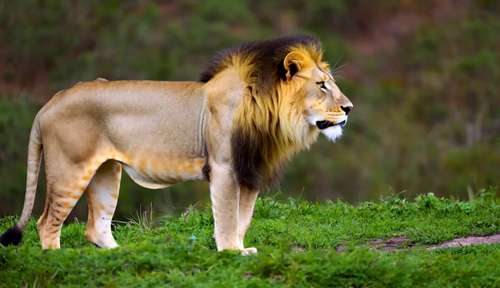}
\includegraphics[width=0.1\linewidth]{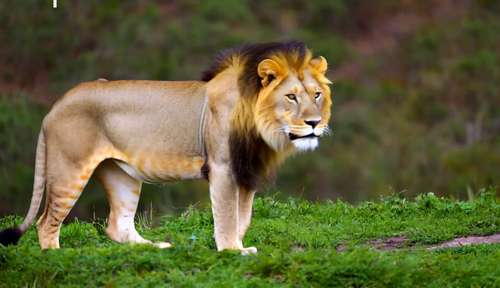}
\includegraphics[width=0.1\linewidth]{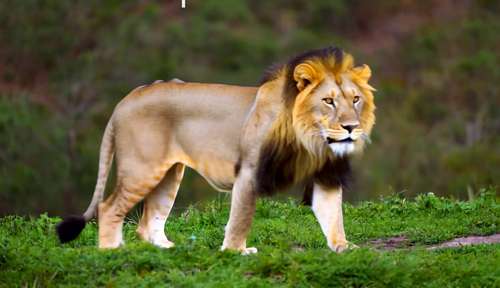}
\\[1.5ex]
\begin{minipage}[c]{2cm}\vspace*{0pt}\vfill\raggedright\textit{\small magician pulling rabbit last second}\vfill\end{minipage}
\includegraphics[width=0.1\linewidth]{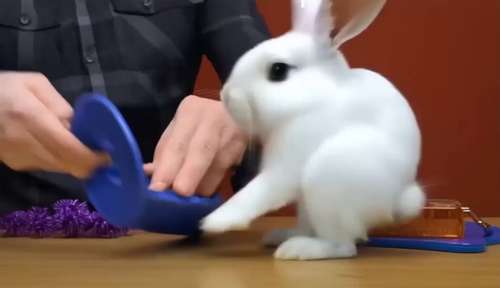}
\includegraphics[width=0.1\linewidth]{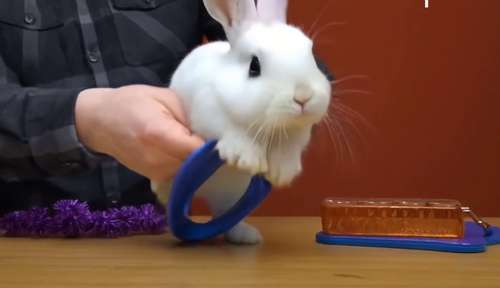}
\includegraphics[width=0.1\linewidth]{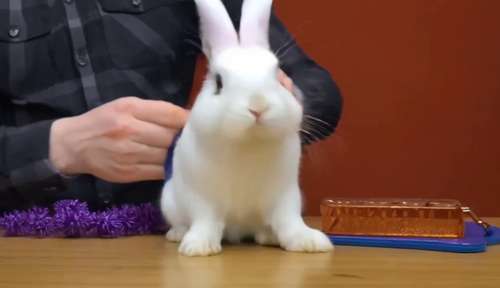}
\includegraphics[width=0.1\linewidth]{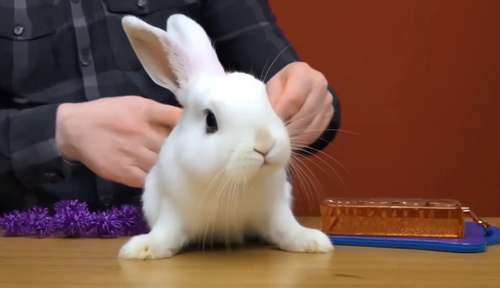}
\hspace{2mm}
\includegraphics[width=0.1\linewidth]{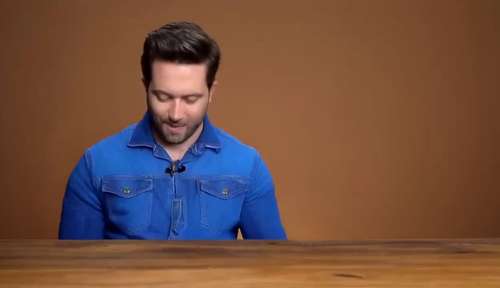}
\includegraphics[width=0.1\linewidth]{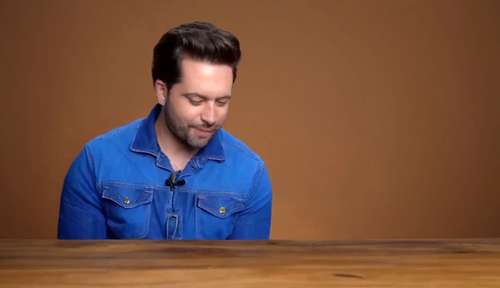}
\includegraphics[width=0.1\linewidth]{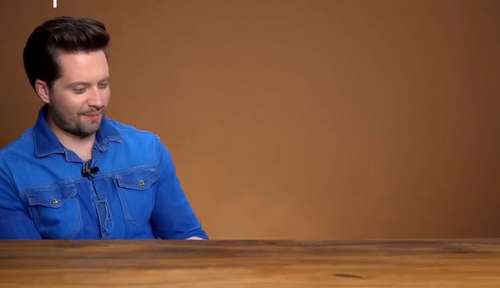}
\includegraphics[width=0.1\linewidth]{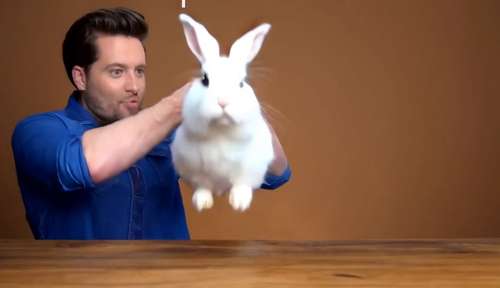}
\\[1.5ex]
\begin{minipage}[c]{2cm}\vspace*{0pt}\vfill\raggedright\textit{\small penguin waddling 1st second}\vfill\end{minipage}
\includegraphics[width=0.1\linewidth]{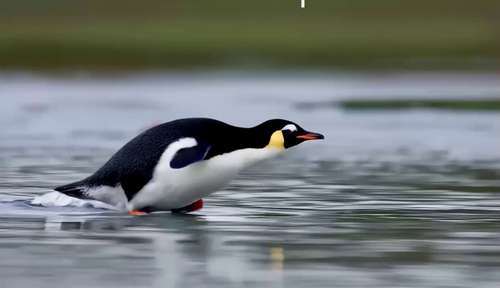}
\includegraphics[width=0.1\linewidth]{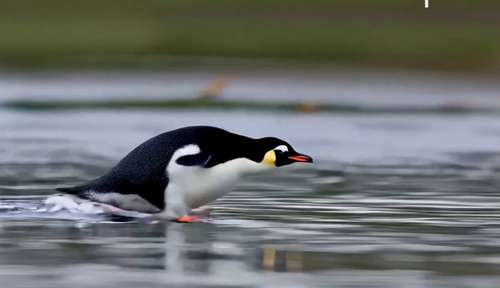}
\includegraphics[width=0.1\linewidth]{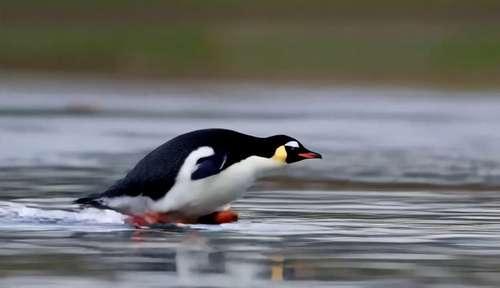}
\includegraphics[width=0.1\linewidth]{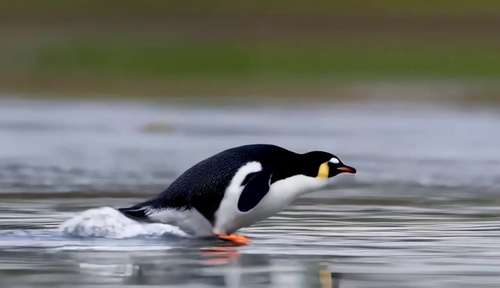}
\hspace{2mm}
\includegraphics[width=0.1\linewidth]{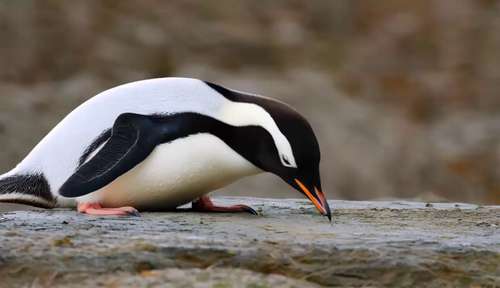}
\includegraphics[width=0.1\linewidth]{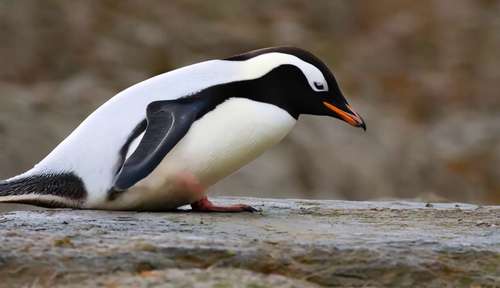}
\includegraphics[width=0.1\linewidth]{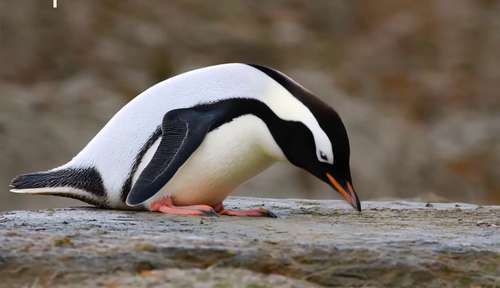}
\includegraphics[width=0.1\linewidth]{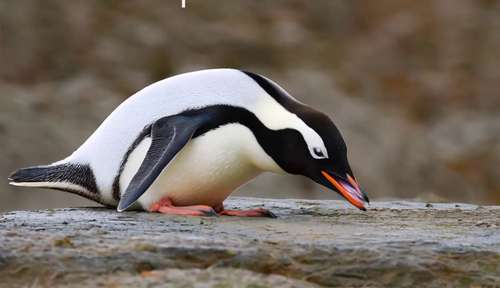}
\\[1.5ex]
\begin{minipage}[c]{2cm}\vspace*{0pt}\vfill\raggedright\textit{\small rabbit thumping 4th second}\vfill\end{minipage}
\includegraphics[width=0.1\linewidth]{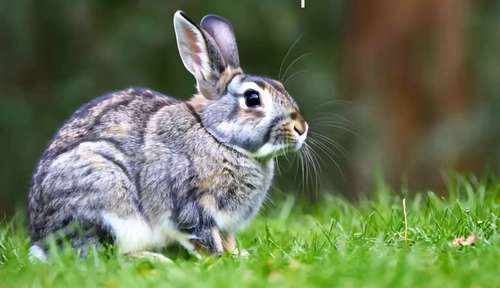}
\includegraphics[width=0.1\linewidth]{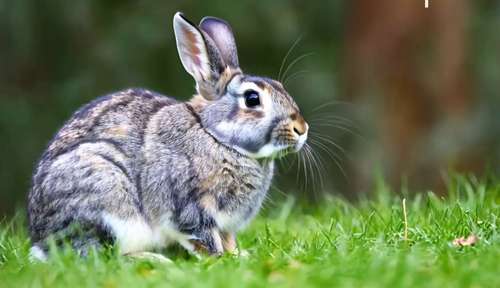}
\includegraphics[width=0.1\linewidth]{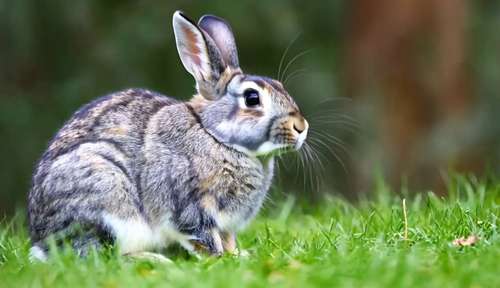}
\includegraphics[width=0.1\linewidth]{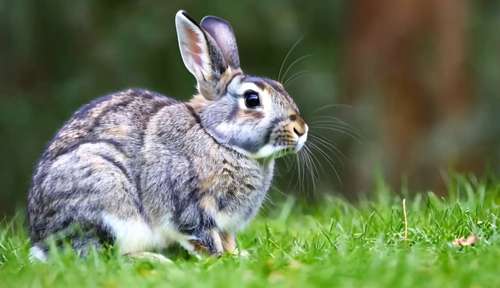}
\hspace{2mm}
\includegraphics[width=0.1\linewidth]{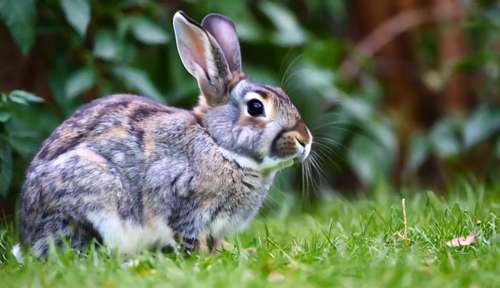}
\includegraphics[width=0.1\linewidth]{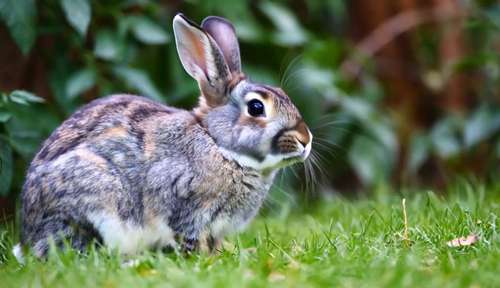}
\includegraphics[width=0.1\linewidth]{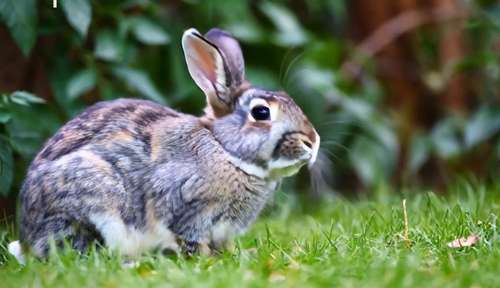}
\includegraphics[width=0.1\linewidth]{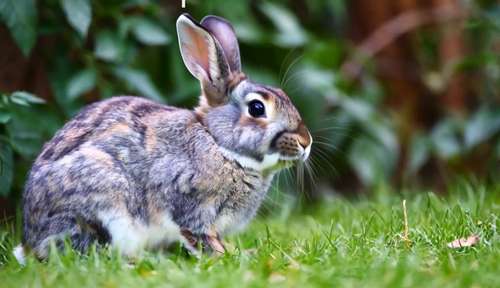}
\\[1.5ex]
\begin{minipage}[c]{2cm}\vspace*{0pt}\vfill\raggedright\textit{\small robot marching 1st second}\vfill\end{minipage}
\includegraphics[width=0.1\linewidth]{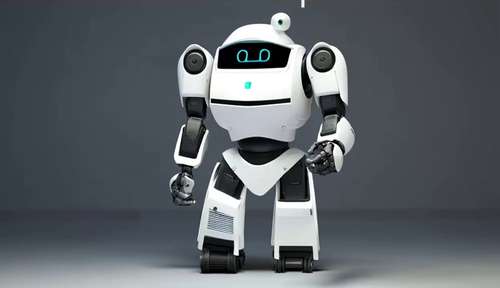}
\includegraphics[width=0.1\linewidth]{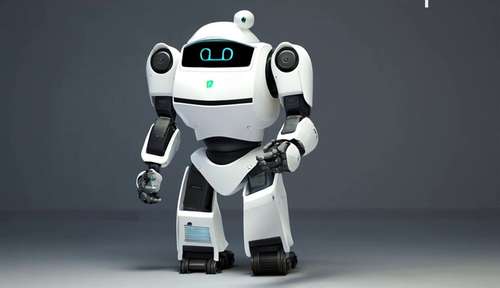}
\includegraphics[width=0.1\linewidth]{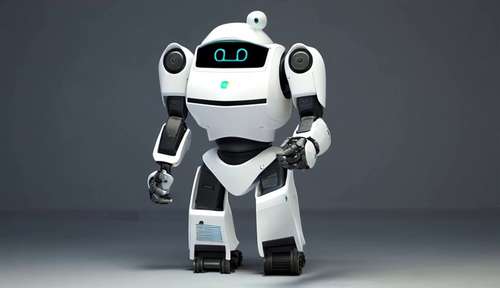}
\includegraphics[width=0.1\linewidth]{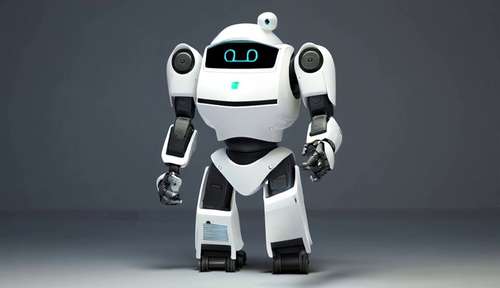}
\hspace{2mm}
\includegraphics[width=0.1\linewidth]{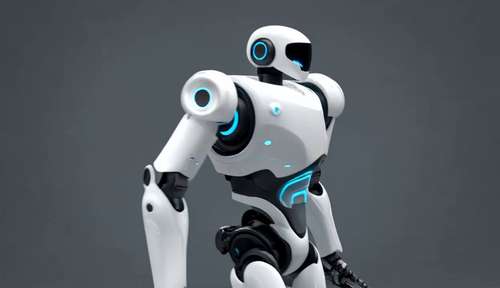}
\includegraphics[width=0.1\linewidth]{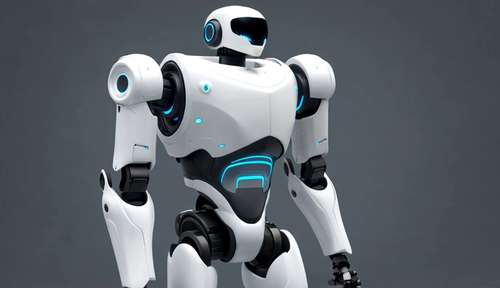}
\includegraphics[width=0.1\linewidth]{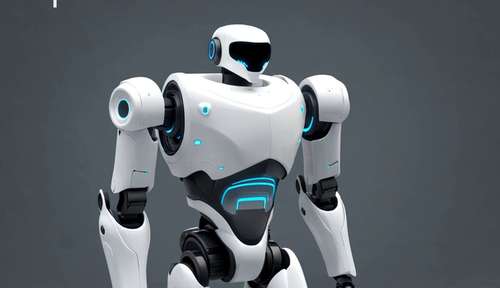}
\includegraphics[width=0.1\linewidth]{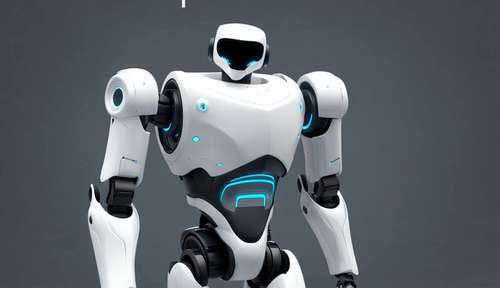}
\\[1.5ex]
\begin{minipage}[c]{2cm}\vspace*{0pt}\vfill\raggedright\textit{\small rocket launching 3rd second}\vfill\end{minipage}
\includegraphics[width=0.1\linewidth]{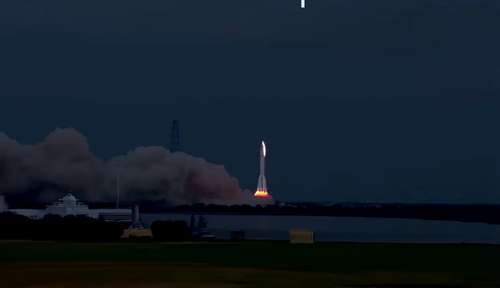}
\includegraphics[width=0.1\linewidth]{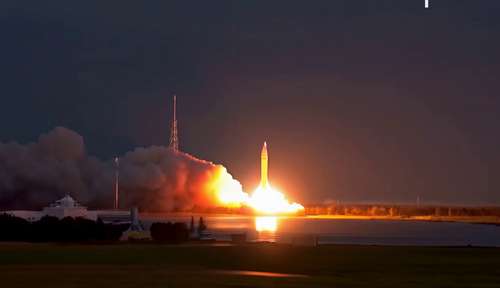}
\includegraphics[width=0.1\linewidth]{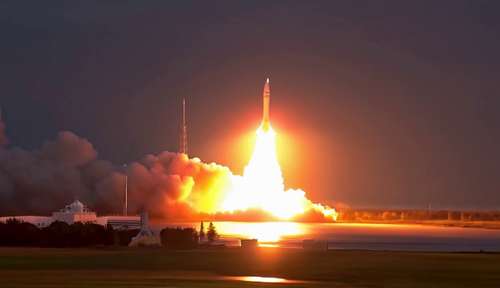}
\includegraphics[width=0.1\linewidth]{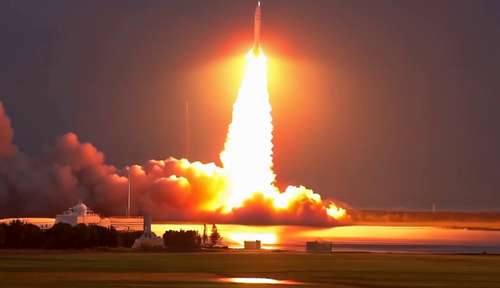}
\hspace{2mm}
\includegraphics[width=0.1\linewidth]{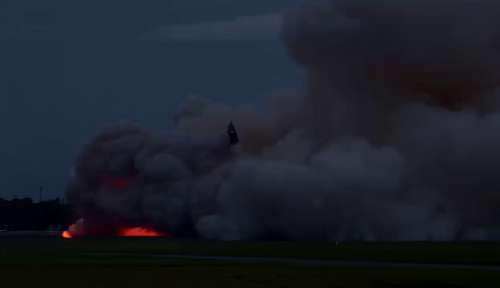}
\includegraphics[width=0.1\linewidth]{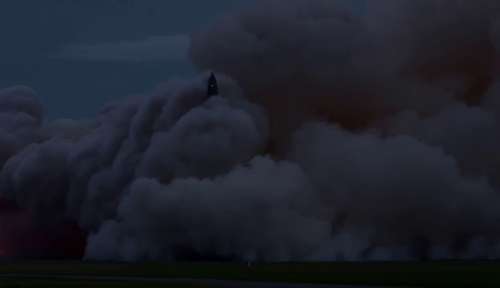}
\includegraphics[width=0.1\linewidth]{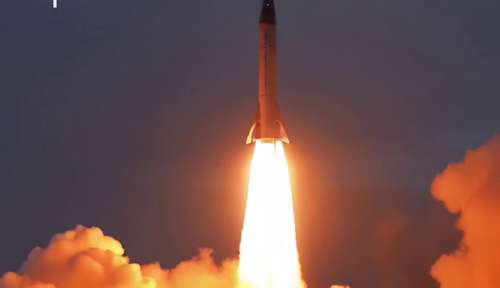}
\includegraphics[width=0.1\linewidth]{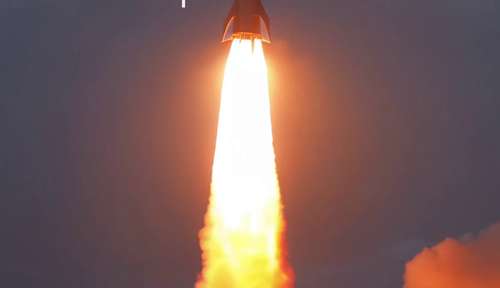}
\\[1.5ex]
\begin{minipage}[c]{2cm}\vspace*{0pt}\vfill\raggedright\textit{\small surfer balancing 4th second}\vfill\end{minipage}
\includegraphics[width=0.1\linewidth]{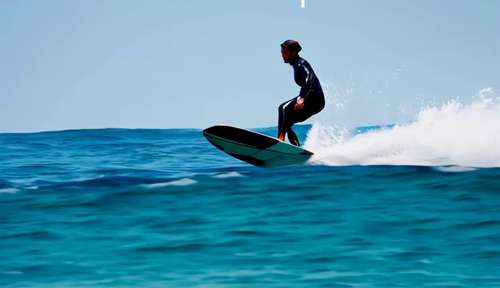}
\includegraphics[width=0.1\linewidth]{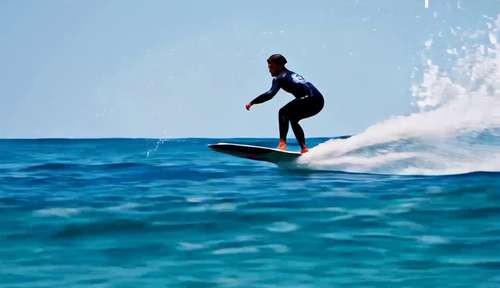}
\includegraphics[width=0.1\linewidth]{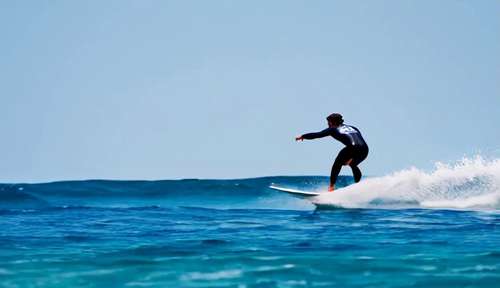}
\includegraphics[width=0.1\linewidth]{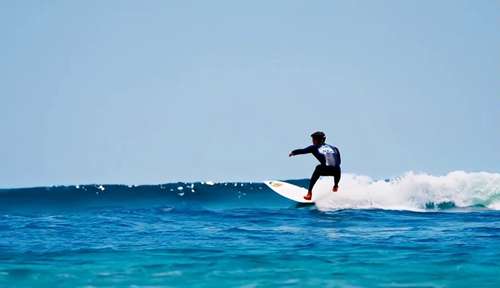}
\hspace{2mm}
\includegraphics[width=0.1\linewidth]{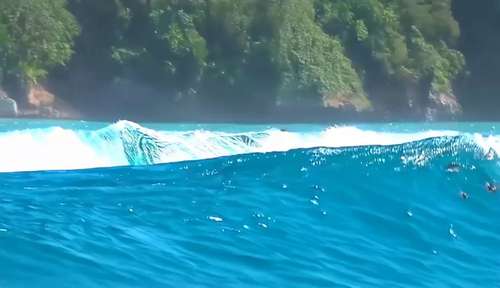}
\includegraphics[width=0.1\linewidth]{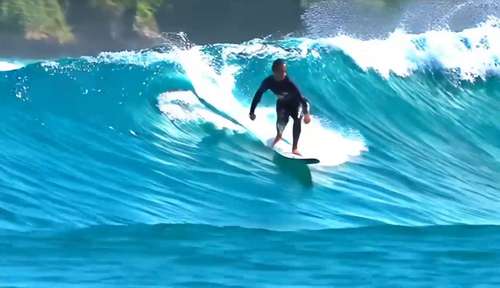}
\includegraphics[width=0.1\linewidth]{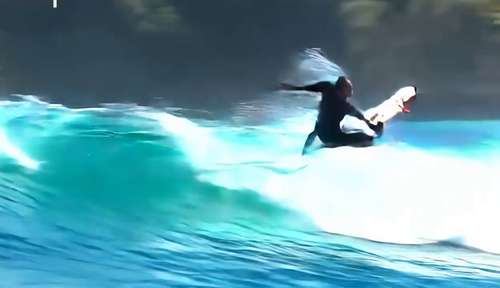}
\includegraphics[width=0.1\linewidth]{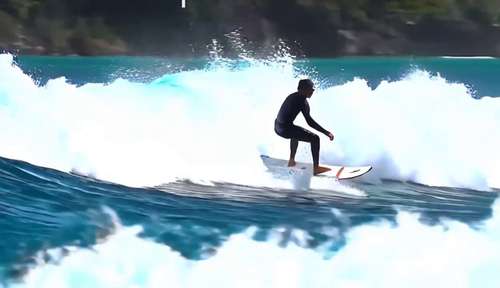}
\\[1.5ex]
\begin{minipage}[c]{2cm}\vspace*{0pt}\vfill\raggedright\textit{\small swimmer diving 4th second}\vfill\end{minipage}
\includegraphics[width=0.1\linewidth]{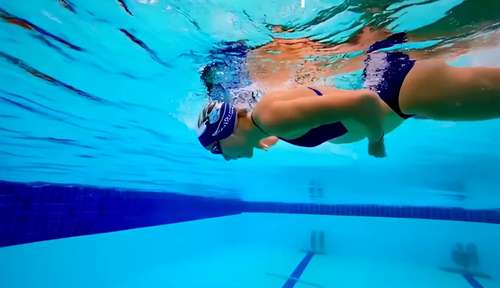}
\includegraphics[width=0.1\linewidth]{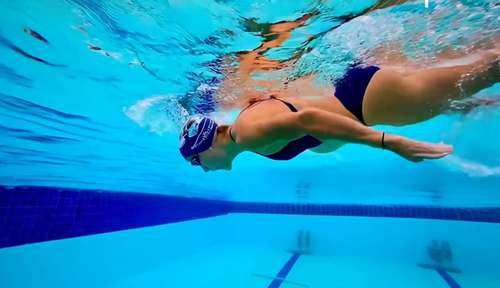}
\includegraphics[width=0.1\linewidth]{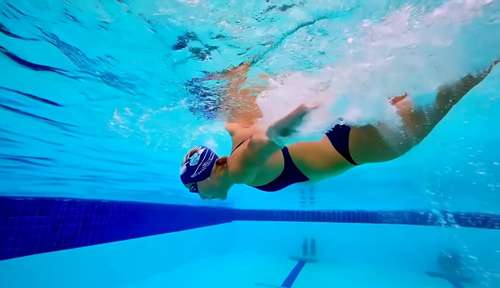}
\includegraphics[width=0.1\linewidth]{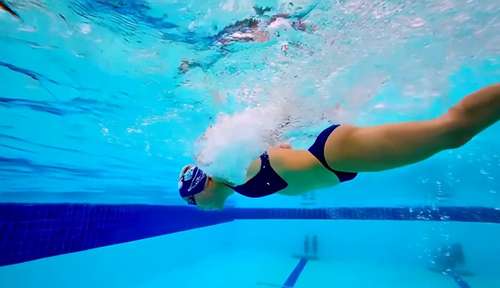}
\hspace{2mm}
\includegraphics[width=0.1\linewidth]{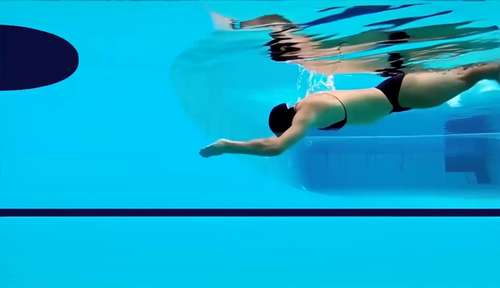}
\includegraphics[width=0.1\linewidth]{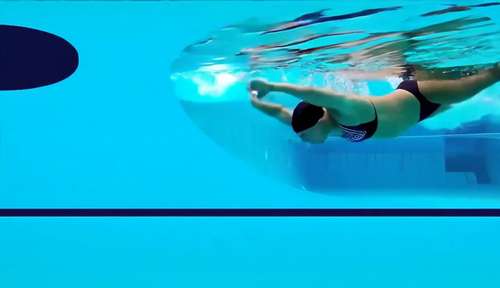}
\includegraphics[width=0.1\linewidth]{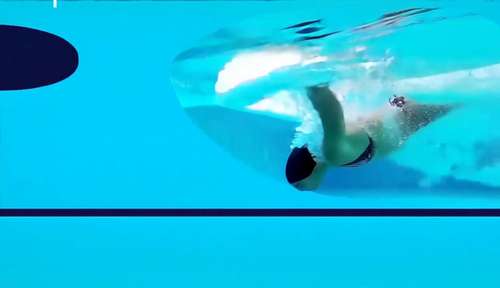}
\includegraphics[width=0.1\linewidth]{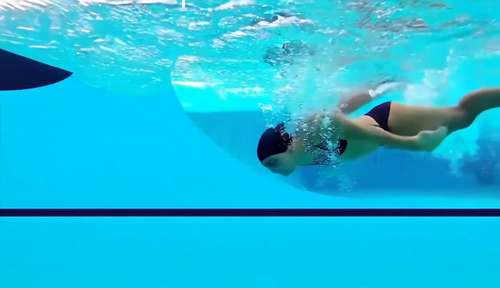}
\\[1.5ex]
\begin{minipage}[c]{2cm}\vspace*{0pt}\vfill\raggedright\textit{\small wolf snarling 4th second}\vfill\end{minipage}
\includegraphics[width=0.1\linewidth]{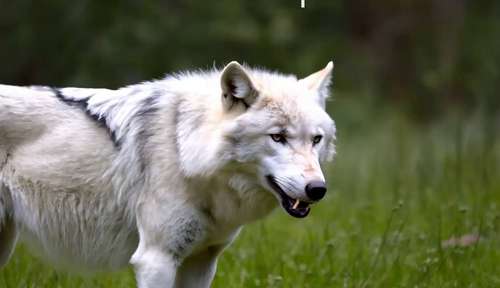}
\includegraphics[width=0.1\linewidth]{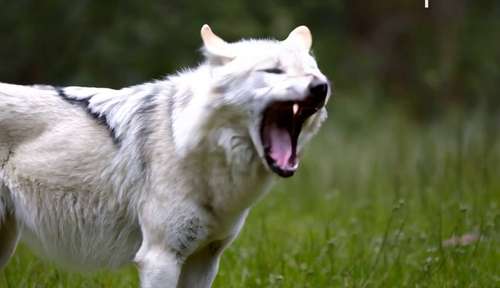}
\includegraphics[width=0.1\linewidth]{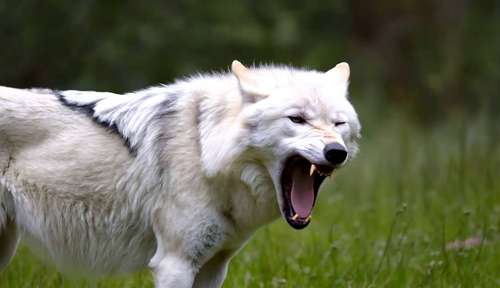}
\includegraphics[width=0.1\linewidth]{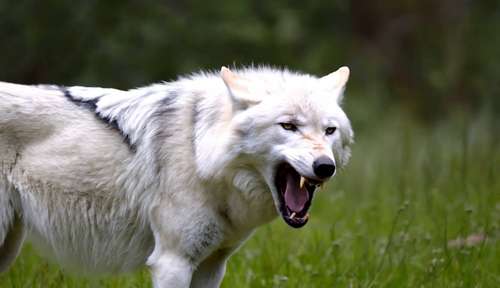}
\hspace{2mm}
\includegraphics[width=0.1\linewidth]{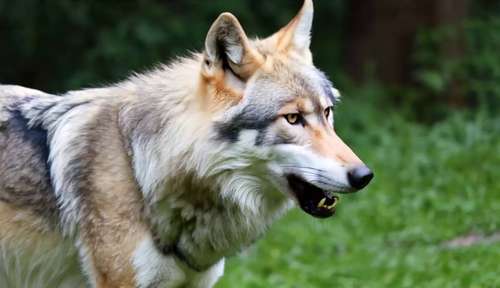}
\includegraphics[width=0.1\linewidth]{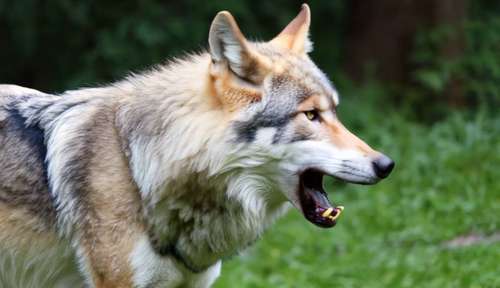}
\includegraphics[width=0.1\linewidth]{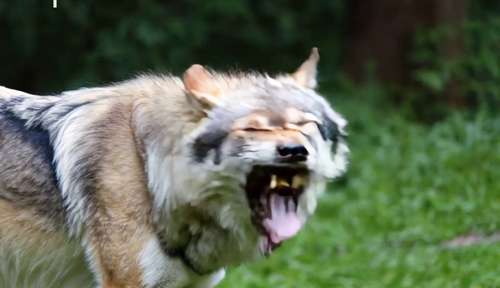}
\includegraphics[width=0.1\linewidth]{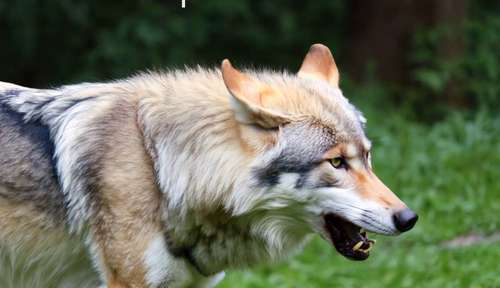}
\\[1.5ex]
\end{minipage}
\caption{Movement Text vs Ours.}
\label{fig:movement2_text_vs_ours}
\end{figure*}

\begin{figure*}[t]
\begin{minipage}{\textwidth}
\makebox[2cm][l]{}\makebox[0.40\linewidth]{\small Text}\hspace{2mm}\makebox[0.40\linewidth]{\small Ours}\\[1.5ex]
\parbox[t]{2cm}{\raggedright\textit{\small elephant ex1}}
\includegraphics[width=0.1\linewidth]{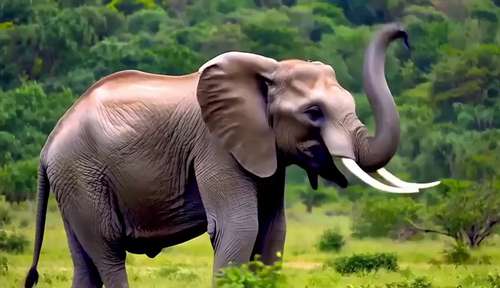}
\includegraphics[width=0.1\linewidth]{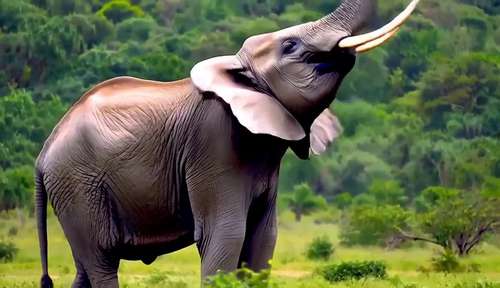}
\includegraphics[width=0.1\linewidth]{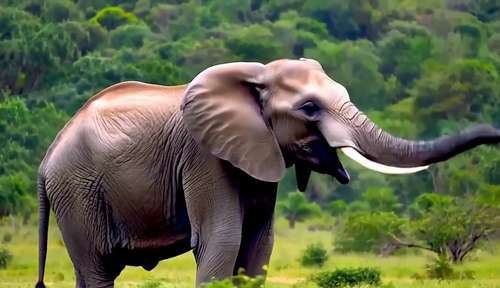}
\includegraphics[width=0.1\linewidth]{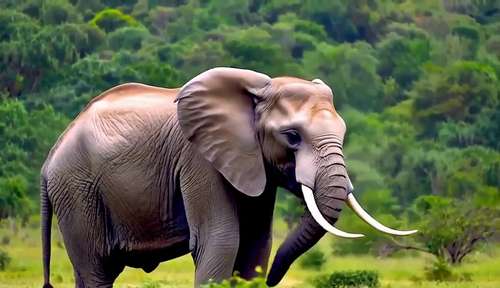}
\hspace{2mm}
\includegraphics[width=0.1\linewidth]{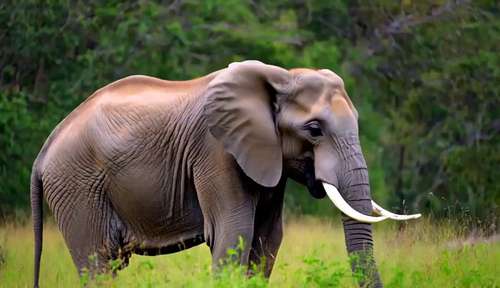}
\includegraphics[width=0.1\linewidth]{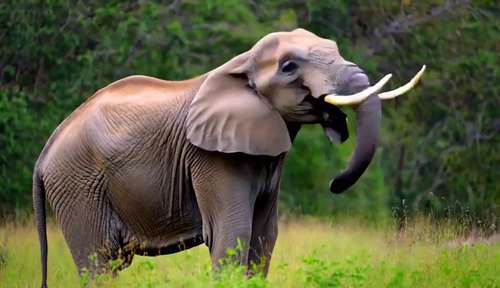}
\includegraphics[width=0.1\linewidth]{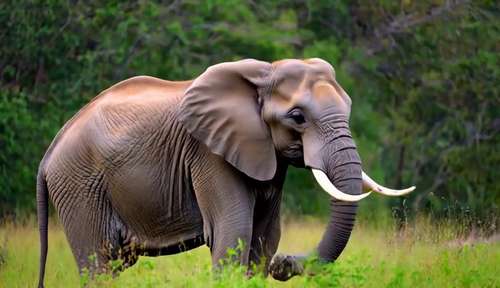}
\includegraphics[width=0.1\linewidth]{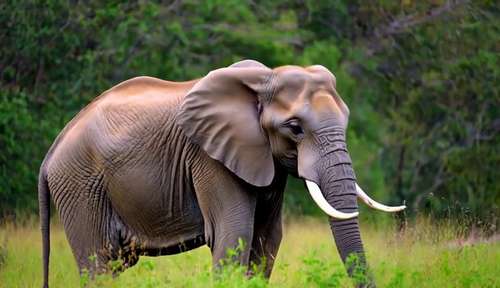}
\\[1.5ex]
\parbox[t]{2cm}{\raggedright\textit{\small elephant ex2}}
\includegraphics[width=0.1\linewidth]{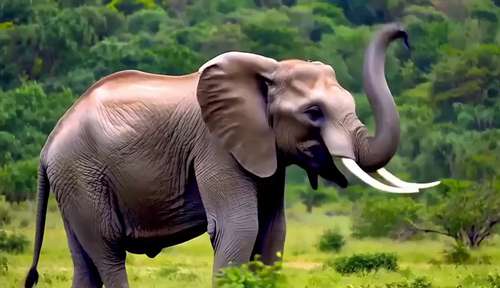}
\includegraphics[width=0.1\linewidth]{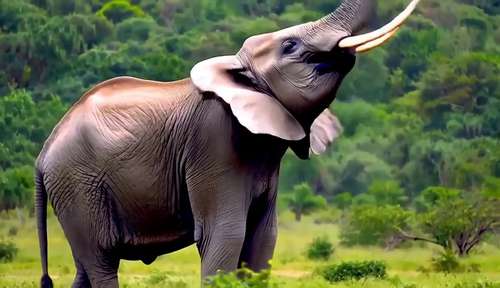}
\includegraphics[width=0.1\linewidth]{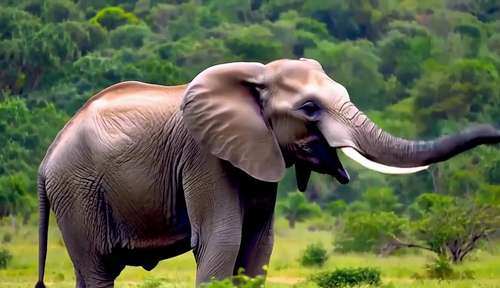}
\includegraphics[width=0.1\linewidth]{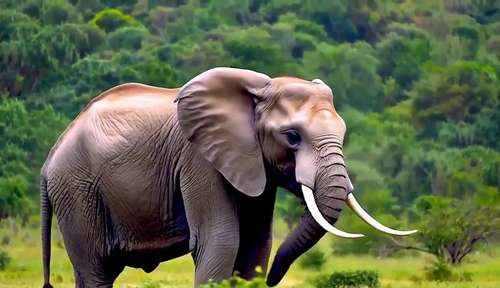}
\hspace{2mm}
\includegraphics[width=0.1\linewidth]{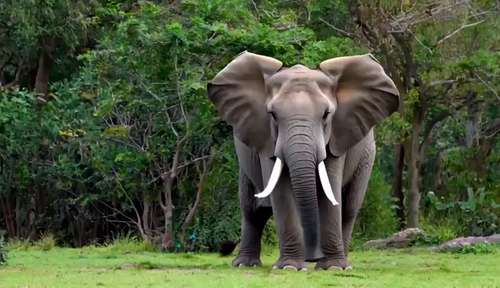}
\includegraphics[width=0.1\linewidth]{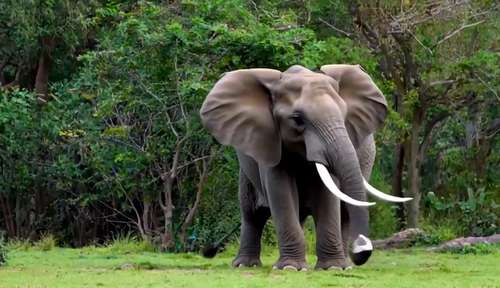}
\includegraphics[width=0.1\linewidth]{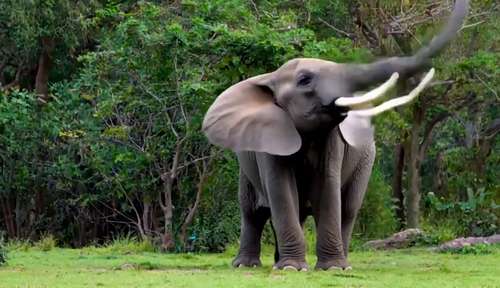}
\includegraphics[width=0.1\linewidth]{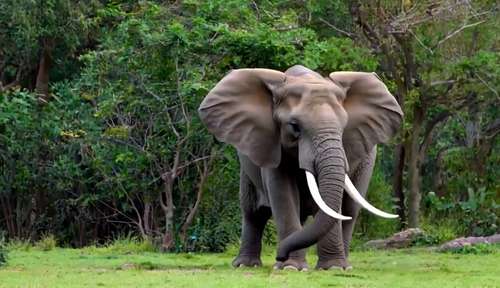}
\\[1.5ex]
\parbox[t]{2cm}{\raggedright\textit{\small lightning ex1}}
\includegraphics[width=0.1\linewidth]{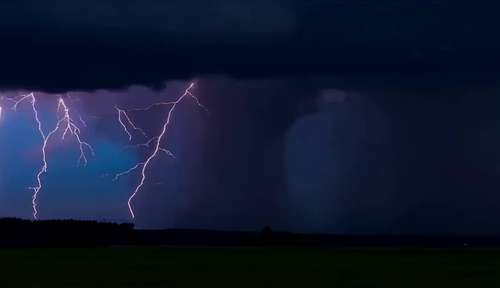}
\includegraphics[width=0.1\linewidth]{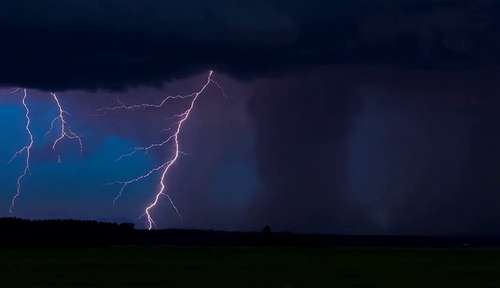}
\includegraphics[width=0.1\linewidth]{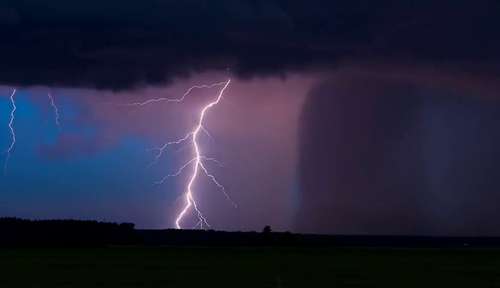}
\includegraphics[width=0.1\linewidth]{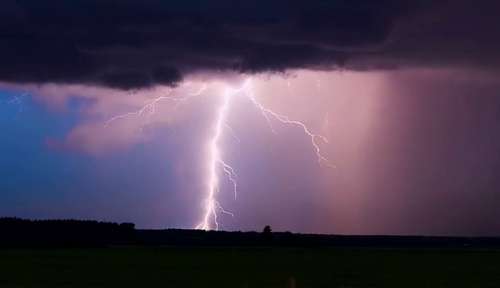}
\hspace{2mm}
\includegraphics[width=0.1\linewidth]{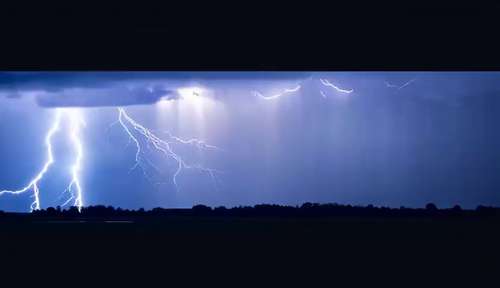}
\includegraphics[width=0.1\linewidth]{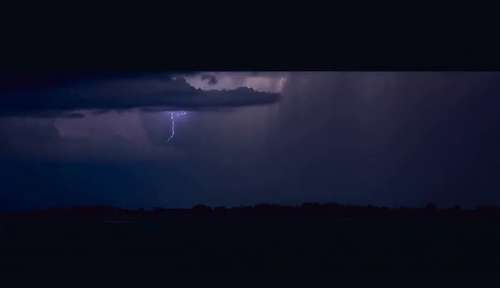}
\includegraphics[width=0.1\linewidth]{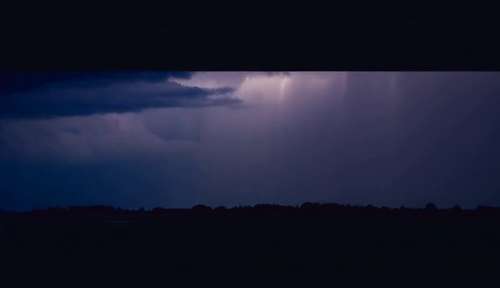}
\includegraphics[width=0.1\linewidth]{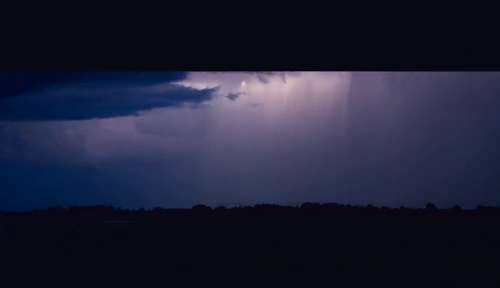}
\\[1.5ex]
\parbox[t]{2cm}{\raggedright\textit{\small lightning ex2}}
\includegraphics[width=0.1\linewidth]{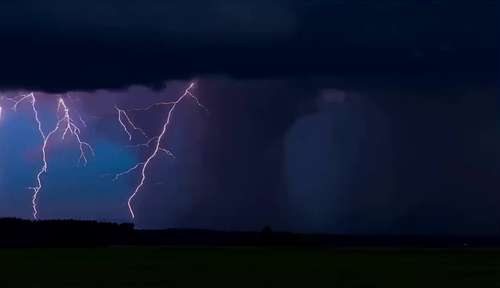}
\includegraphics[width=0.1\linewidth]{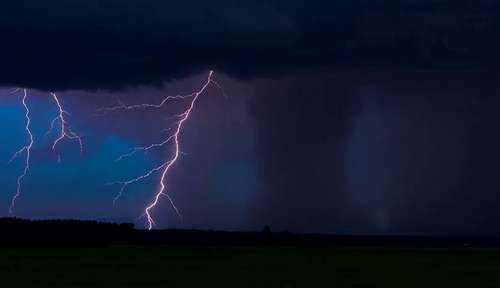}
\includegraphics[width=0.1\linewidth]{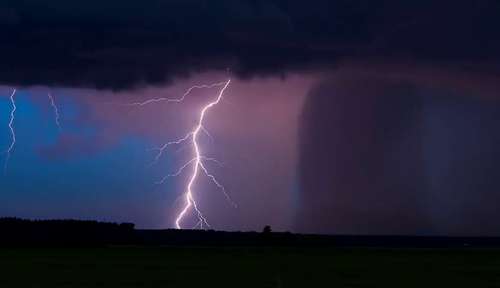}
\includegraphics[width=0.1\linewidth]{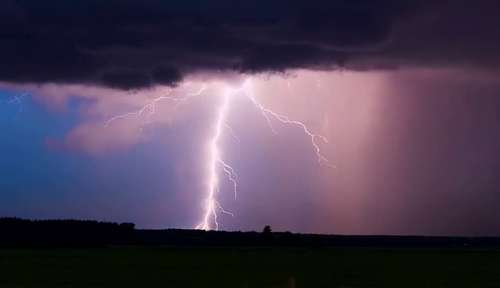}
\hspace{2mm}
\includegraphics[width=0.1\linewidth]{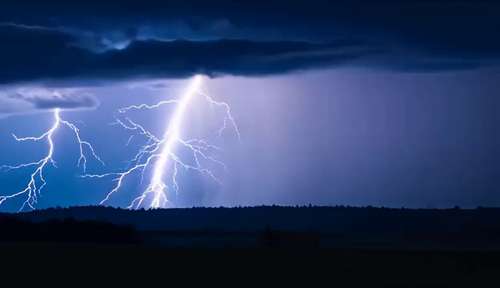}
\includegraphics[width=0.1\linewidth]{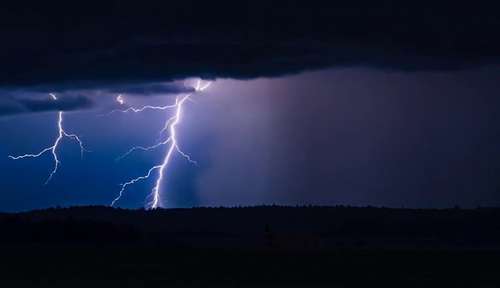}
\includegraphics[width=0.1\linewidth]{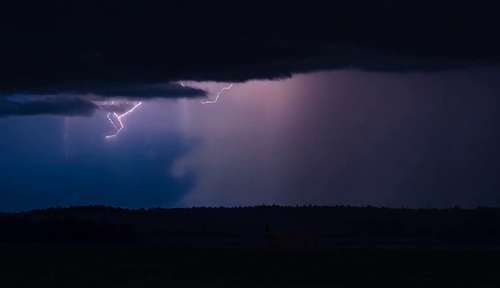}
\includegraphics[width=0.1\linewidth]{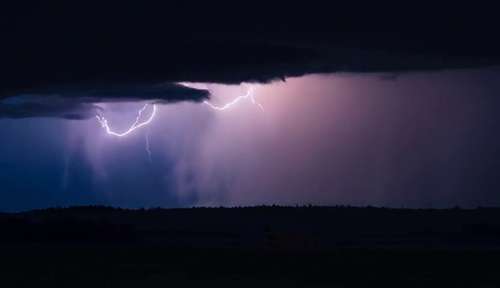}
\\[1.5ex]
\parbox[t]{2cm}{\raggedright\textit{\small rooster ex1}}
\includegraphics[width=0.1\linewidth]{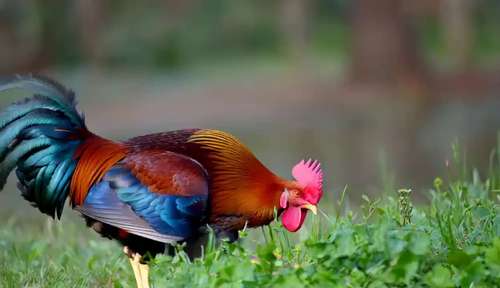}
\includegraphics[width=0.1\linewidth]{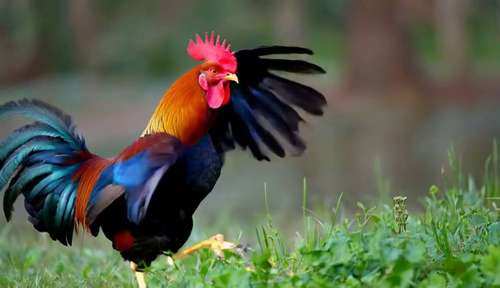}
\includegraphics[width=0.1\linewidth]{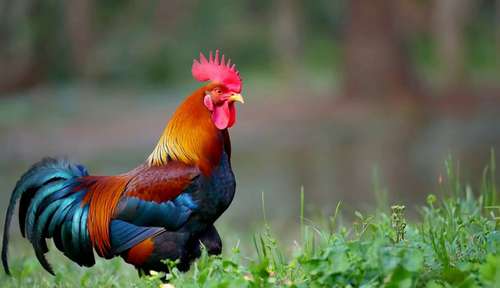}
\includegraphics[width=0.1\linewidth]{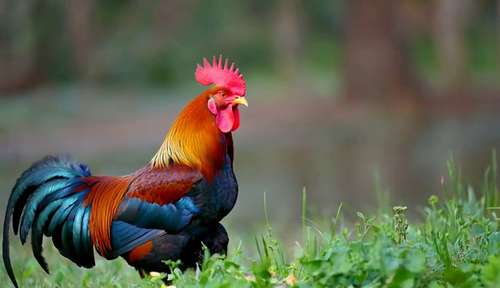}
\hspace{2mm}
\includegraphics[width=0.1\linewidth]{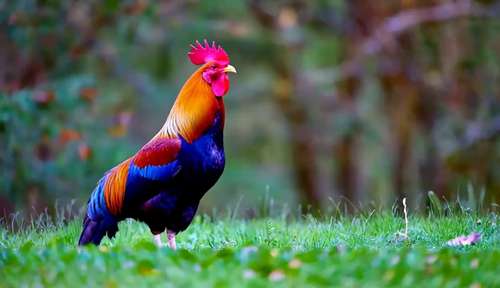}
\includegraphics[width=0.1\linewidth]{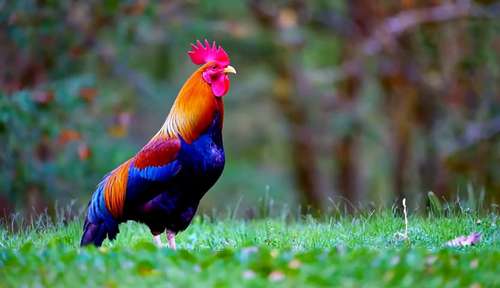}
\includegraphics[width=0.1\linewidth]{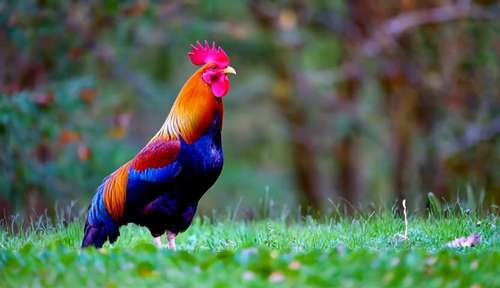}
\includegraphics[width=0.1\linewidth]{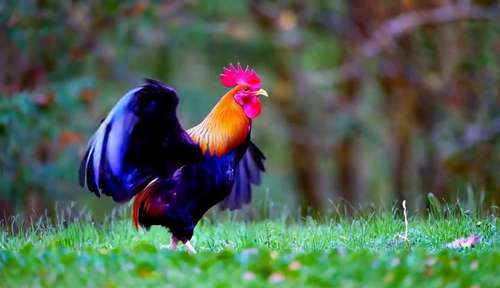}
\\[1.5ex]
\parbox[t]{2cm}{\raggedright\textit{\small rooster ex2}}
\includegraphics[width=0.1\linewidth]{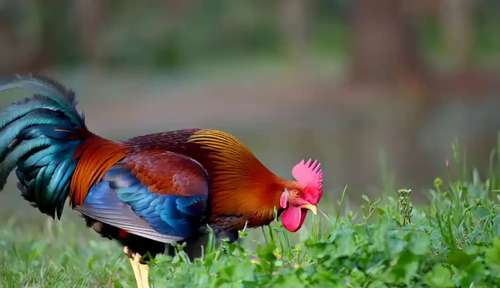}
\includegraphics[width=0.1\linewidth]{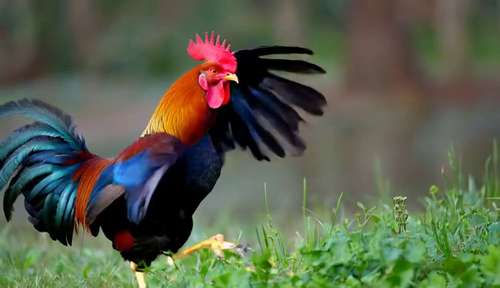}
\includegraphics[width=0.1\linewidth]{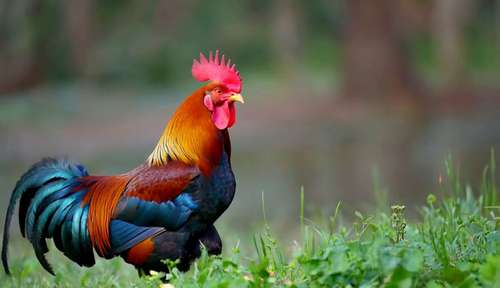}
\includegraphics[width=0.1\linewidth]{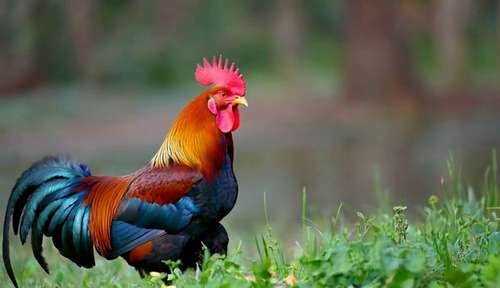}
\hspace{2mm}
\includegraphics[width=0.1\linewidth]{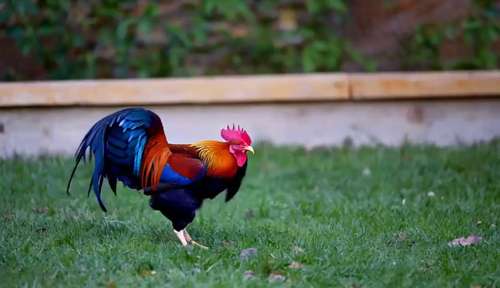}
\includegraphics[width=0.1\linewidth]{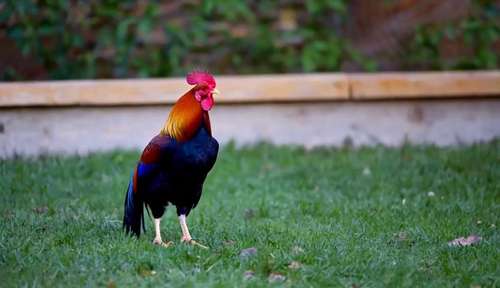}
\includegraphics[width=0.1\linewidth]{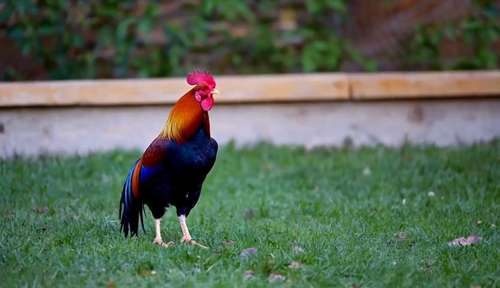}
\includegraphics[width=0.1\linewidth]{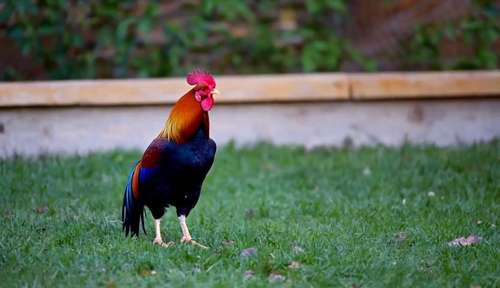}
\\[1.5ex]
\end{minipage}
\caption{Audio-video Alignment Text vs Ours.}
\label{fig:audio_video_alignment_text_vs_ours}
\end{figure*}

\end{document}